Takaaki Fujita, Florentin Smarandache

# A Dynamic Survey of Fuzzy, Intuitionistic Fuzzy, Neutrosophic, Plithogenic, and Extensional Sets

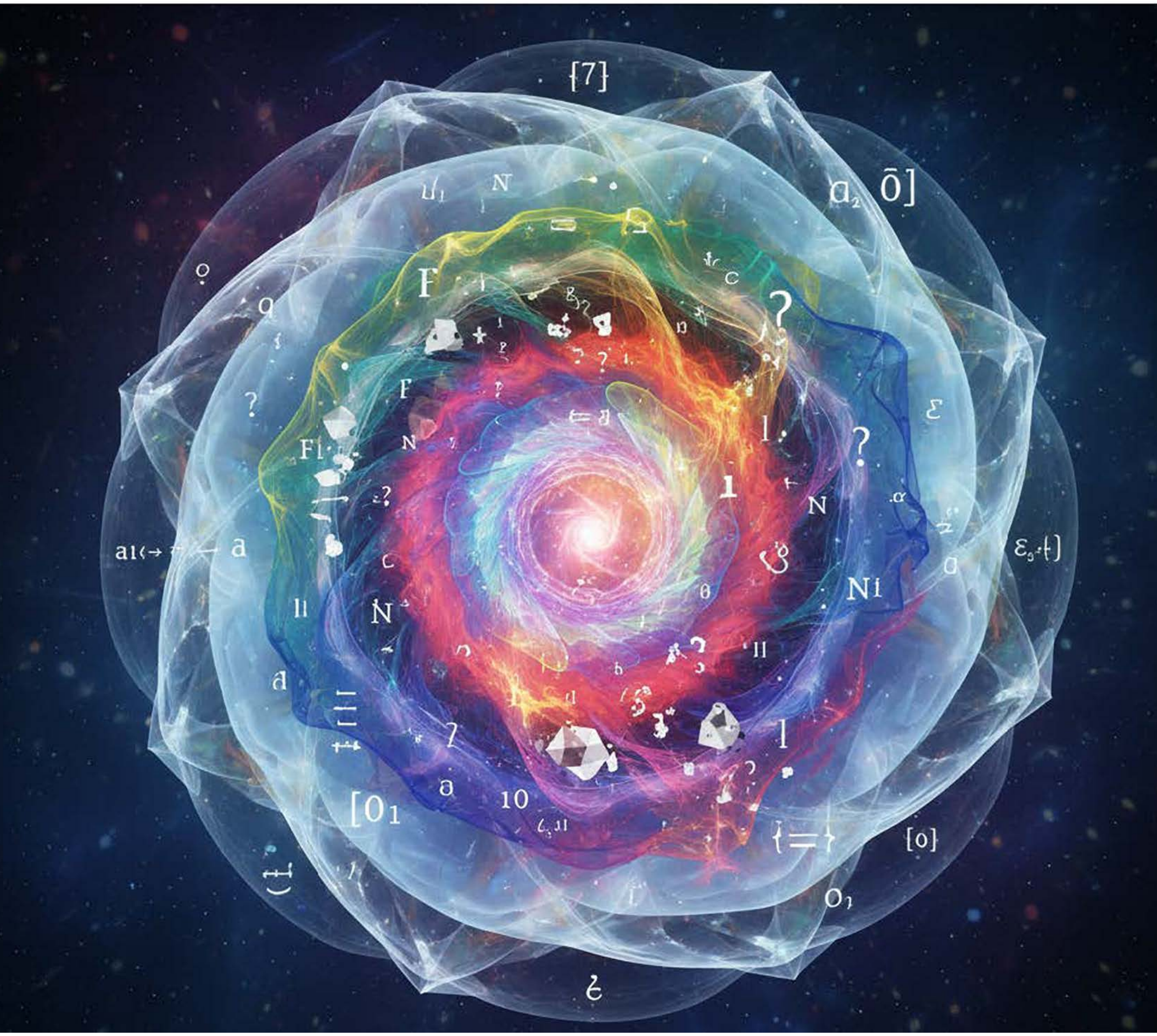

**Takaaki Fujita, Florentin Smarandache**

# A Dynamic Survey of Fuzzy, Intuitionistic Fuzzy, Neutrosophic, Plithogenic, and Extensional Sets

## Authors

**Takaaki Fujita**

Independent Researcher, Tokyo, Japan.
Email: Takaaki.fujita060@gmail.com

**Florentin Smarandache**

University of New Mexico, Gallup Campus, NM 87301, USA.
Email: fsmarandache@gmail.com

## Abstract

Real-world phenomena frequently involve vagueness, partial truth, and incomplete information. To capture such uncertainty in a mathematically rigorous manner, numerous generalized set-theoretic frameworks have been introduced, including Fuzzy Sets [1], Intuitionistic Fuzzy Sets [2], Neutrosophic Sets [3, 4], Vague Sets [5], Hesitant Fuzzy Sets [6], Picture Fuzzy Sets [7], Quadripartitioned Neutro-sophic Sets [8], PentaPartitioned Neutrosophic Sets [9], Plithogenic Sets [10], HyperFuzzy Sets [11], and HyperNeutrosophic Sets [12].

Within these frameworks, a vast number of concepts have been proposed and studied, especially in the contexts of Fuzzy, Intuitionistic Fuzzy, Neutrosophic, and Plithogenic Sets. This extensive body of work highlights both the importance of these theories and the breadth of their application domains. Consequently, many ideas, constructions, and structural patterns recur across these four families of uncertainty-oriented models.

In this book, we present a comprehensive, large-scale survey of Fuzzy, Intuitionistic Fuzzy, Neutro-sophic, and Plithogenic Sets. Our aim is to offer r eaders a s ystematic overview o f e xisting develop-ments and, through this unified exposition, to foster new insights, further conceptual extensions, and additional applications across a wide range of disciplines.

# Table of Contents

# Chapter 1

# Introduction

## 1.1 Uncertain Set

Real-world phenomena often exhibit vagueness, partial truth, and incomplete information. To capture such uncertainty in a mathematically rigorous way, many generalized set-theoretic frameworks have been introduced, including Fuzzy Sets [1], Intuitionistic Fuzzy Sets [2], Neutrosophic Sets [3,4], Vague Sets [5], Hesitant Fuzzy Sets [6], Picture Fuzzy Sets [7], Quadripartitioned Neutrosophic Sets [8], PentaPartitioned Neutrosophic Sets [9], Plithogenic Sets [10], HyperFuzzy Sets [11], and HyperNeutrosophic Sets [12]. Applications of Fuzzy Sets and their extensions—discussed in later sections—have been widely explored in fields such as decision science, chemistry, control systems, and machine learning [13]. Depending on the nature of the application and the number of uncertainty parameters required to characterize the underlying phenomena, an appropriate class of generalized sets is selected to model the problem effectively.

In the classical fuzzy setting, each element $x$ in the universe $X$ is associated with a single membership degree $\mu(x) \in [0, 1]$, which expresses to what extent $x$ belongs to the fuzzy set under consideration [1]. For an Intuitionistic Fuzzy Set, every element $x$ is described by a pair $(\mu(x), \nu(x))$ of membership and non-membership functions $\mu, \nu : X \to [0, 1]$ satisfying

$$0 \leq \mu(x) + \nu(x) \leq 1,$$

[2,14].

A Neutrosophic Set refines this description by assigning to each element $x$ a triple

$$(T(x), I(x), F(x)),$$

where $T(x)$, $I(x)$, and $F(x)$ denote, respectively, the degrees of truth, indeterminacy, and falsity, typically taking values in $[0, 1]$. In contrast to the intuitionistic fuzzy case, these three values are not required to sum to 1, which allows one to encode incomplete, inconsistent, or even redundant information in a flexible way [14, 15]. [1] Neutrosophy highlights the central role of neutrality and indeterminacy, giving rise to neutrosophic set, logic, probability, statistics, measure, integral, and related formalisms. These frameworks now find broad practical applications across numerous scientific and applied domains [13, 17].

[1] Intuitionistic Fuzzy Sets neglect the role of indeterminacy, whereas Neutrosophic Fuzzy Operators assign indeterminacy the same significance as truth-membership and falsehood-nonmembership [14, 16]. Moreover, it is widely recognized that the neutrosophic set generalizes the intuitionistic fuzzy set, the inconsistent intuitionistic fuzzy set (including picture fuzzy and ternary fuzzy sets), the Pythagorean fuzzy set, the spherical fuzzy set, and the $q$-rung orthopair fuzzy set; similarly, neutrosophication generalizes regret theory, grey system theory, and three-way decision theory [16].

Plithogenic Sets further generalize these notions by representing each element through its attribute values, together with the corresponding degrees of appurtenance, and by introducing a contradiction (or dissimilarity) function between distinct attribute values [10, 18, 19]. This additional structure supports context-sensitive aggregation of heterogeneous and possibly conflicting evaluations, thereby refining the classical fuzzy, intuitionistic fuzzy, and neutrosophic models (e.g. [17, 20]). For convenience, Table 1.1 summarizes the main data attached to each element in several well-known set extensions (notation harmonized for this book).

Table 1.1: Representative set extensions and the canonical information stored per element.

| **Set Type** | **Canonical data attached to each element** |
|---|---|
| Fuzzy Set | Membership mapping $\mu : X \to [0,1]$. |
| Intuitionistic Fuzzy Set | Membership $\mu$ and non-membership $\nu$ with $\mu(x) + \nu(x) \leq 1$; the gap $1 - \mu(x) - \nu(x)$ models hesitation. |
| Neutrosophic Set | Triple $(T, I, F)$ with $T, I, F \in [0,1]$ (truth, indeterminacy, falsity) considered as mutually independent coordinates. |
| Plithogenic Set | Tuple $(P, v, Pv, \mathrm{pdf}, \mathrm{pCF})$ where $\mathrm{pdf} : P \times Pv \to [0,1]^s$ encodes $s$-dimensional appurtenance and $\mathrm{pCF} : Pv \times Pv \to [0,1]^t$ is a symmetric contradiction map in $[0,1]^t$. |

Within the plithogenic framework, one can recover *plithogenic fuzzy*, *plithogenic intuitionistic fuzzy*, and *plithogenic neutrosophic* models by choosing suitable dimensions $s$ (for appurtenance) and $t$ (for contradiction) [21–25]. In particular, scalar-contradiction cases with $t = 1$ yield natural extensions of the classical fuzzy, intuitionistic fuzzy, and neutrosophic paradigms; if, in addition, the contradiction function pCF is set identically to zero, one exactly recovers the corresponding non-plithogenic models (cf. [17]). These three frequently used plithogenic variants are summarized in Table 1.2.

Table 1.2: Plithogenic scalar-contradiction variants ($t = 1$) and their classical limits.

| **Variant** | $s$ | $t$ | **Appurtenance vector (semantics)** | **Limit when** $\mathrm{pCF} \equiv 0$ |
|---|---|---|---|---|
| Plithogenic fuzzy | 1 | 1 | $\mu \in [0,1]$ (single membership degree) | Classical fuzzy set |
| Plithogenic intuitionistic fuzzy | 2 | 1 | $(\mu, \nu) \in [0,1]^2$ with $1 - \mu - \nu \geq 0$ | Intuitionistic fuzzy model |
| Plithogenic neutrosophic | 3 | 1 | $(T, I, F) \in [0,1]^3$ (truth, indeterminacy, falsity) | Neutrosophic model |

## 1.2 Applied Area: Fuzzy, Intuitionistic Fuzzy, Neutrosophic, and Plithogenic

Fuzzy, Intuitionistic Fuzzy, Neutrosophic, and Plithogenic Sets play an essential role in modern science due to their mathematical depth, practical applicability, and capacity to model uncertainty effectively (cf. [13, 17]). Because of these properties, they have been widely studied and applied in numerous domains, including algebra, graph theory, hypergraph theory, probability, and decision-making. Tables 1.3, 1.4, and 1.5 provide a summarized overview of the *extensions of classical concepts* under the fuzzy, intuitionistic fuzzy, neutrosophic, and plithogenic frameworks.

From this perspective, it becomes clear that an exceptionally wide range of fields has explored both the applications and the underlying mathematical structures of Fuzzy, Intuitionistic Fuzzy, Neutrosophic, and Plithogenic Sets. These developments highlight not only the theoretical significance of such models but also their substantial contributions to real-world problem solving and practical decision-making.

Table 1.3: Parallel extensions of classical concepts in fuzzy, intuitionistic fuzzy, neutrosophic, and plithogenic frameworks.

| Classical Concept | Fuzzy | Intuitionistic Fuzzy | Neutrosophic | Plithogenic |
|---|---|---|---|---|
| Set [26] | Fuzzy Set [1] | Intuitionistic Fuzzy Set [2] | Neutrosophic Set [27] | Plithogenic Set [10] |
| Relation | Fuzzy Relation [28] | Intuitionistic Fuzzy Relation [29] | Neutrosophic Relation [30] | Plithogenic Relation (cf. [31]) |
| Function / Mapping | Fuzzy Function [32] | Intuitionistic Fuzzy Function [33] | Neutrosophic Function [34, 35] | Plithogenic Function [36] |
| Graph [37] | Fuzzy Graph [38] | Intuitionistic Fuzzy Graph [39] | Neutrosophic Graph [4, 40] | Plithogenic Graph [17] |
| Hypergraph [41, 42] | Fuzzy Hypergraph [43, 44] | Intuitionistic Fuzzy Hypergraph [45] | Neutrosophic Hypergraph [46] | Plithogenic Hypergraph [47] |
| SuperHypergraph [48–50] | Fuzzy SuperHypergraph [51, 52] | Intuitionistic Fuzzy SuperHypergraph [53–55] | Neutrosophic SuperHypergraph [56, 57] | Plithogenic SuperHypergraph [58–60] |
| Matrix / Linear Algebra [61] | Fuzzy Matrix [62] / Linear Algebra | Intuitionistic Fuzzy Matrix [63, 64] / Linear Algebra | Neutrosophic Matrix [63, 65] / Linear Algebra | Plithogenic Matrix [66, 67] / Linear Algebra |
| Algebra (Group/Ring/...) | Fuzzy Algebra [68] (e.g., Fuzzy Group) | Intuitionistic Fuzzy Algebra [69, 70] | Neutrosophic Algebra [71, 72] | Plithogenic Algebra [15, 73] |
| HyperAlgebra [74] (HyperGroup [75]/HyperRing [76]/...) | Fuzzy HyperAlgebra [77] (e.g., Fuzzy HyperGroup [75]) | Intuitionistic Fuzzy HyperAlgebra [78, 79] | Neutrosophic HyperAlgebra [80, 81] | Plithogenic HyperAlgebra |
| Topology [82] | Fuzzy Topology [83, 84] | Intuitionistic Fuzzy Topology [85] | Neutrosophic Topology [86, 87] | Plithogenic Topology [88] |
| Measure / Probability [89] | Fuzzy Probability [90, 91] | Intuitionistic Fuzzy Probability [92, 93] | Neutrosophic Probability [94, 95] | Plithogenic Probability [96, 97] |
| Logic [98] | Fuzzy Logic [1] | Intuitionistic Fuzzy Logic [2] | Neutrosophic Logic [99] | Plithogenic Logic [100] |
| Optimization [101] / Decision [102] | Fuzzy Decision-Making [103, 104] | Intuitionistic Fuzzy Decision-Making [105, 106] | Neutrosophic Decision-Making [107, 108] | Plithogenic Decision-Making [109] |
| Clustering / Classification | Fuzzy Clustering [110, 111] | Intuitionistic Fuzzy Clustering [112, 113] | Neutrosophic Clustering [114, 115] | Plithogenic Clustering |
| Numbers [116] | Fuzzy Numbers [117] | Intuitionistic Fuzzy Numbers [118] | Neutrosophic Numbers [119] | Plithogenic Numbers [120] |

## 1.3 Our Contribution in This Book

A vast number of concepts have been proposed and studied within the frameworks of Fuzzy, Intuitionistic Fuzzy, Neutrosophic, and Plithogenic Sets, reflecting the importance of these theories and the diversity of their application domains (cf. [273]). Because of this richness, many ideas and structural patterns appear repeatedly across these four families of uncertainty-oriented models.

In this book, we undertake a comprehensive and large-scale survey of Fuzzy, Intuitionistic Fuzzy, Neutrosophic, and Plithogenic Sets. Our aim is to provide readers with an organized overview of existing developments and, through this survey, to encourage new insights, novel conceptual extensions, and further applications in a wide range of disciplines.

Table 1.4: Part 2 — Additional concepts across classical, fuzzy, intuitionistic fuzzy, neutrosophic, and plithogenic frameworks.

| Classical Concept | Fuzzy | Intuitionistic Fuzzy | Neutrosophic | Plithogenic |
|---|---|---|---|---|
| Metric Space [121] | Fuzzy Metric Space [122] | Intuitionistic Fuzzy Metric Space [123] | Neutrosophic Metric Space [124, 125] | Plithogenic Metric Space [126] |
| Measure Space [127] | Fuzzy Measure Space [128] | Intuitionistic Fuzzy Measure Space [129, 130] | Neutrosophic Measure Space [131, 132] | Plithogenic Measure Space |
| Stochastic Process [133, 134] | Fuzzy Stochastic Process [135] | Intuitionistic Fuzzy Stochastic Process [136] | Neutrosophic Stochastic Process [137, 138] | Plithogenic Stochastic Process [139] |
| Markov Chain [140] | Fuzzy Markov Chain [141] | Intuitionistic Fuzzy Markov Chain [93] | Neutrosophic Markov Chain [142–144] | Plithogenic Markov Chain [145] |
| Dynamical System | Fuzzy Dynamical System [146, 147] | Intuitionistic Fuzzy Dynamical System | Neutrosophic Dynamical System [148] | Plithogenic Dynamical System |
| Differential Equation [149] | Fuzzy Differential Equation [150, 151] | Intuitionistic Fuzzy Differential Equation [152, 153] | Neutrosophic Differential Equation [154, 155] | Plithogenic Differential Equation |
| Optimization Problem [156] | Fuzzy Optimization [157] | Intuitionistic Fuzzy Optimization [158] | Neutrosophic Optimization [159] | Plithogenic Optimization |
| Control System [160] | Fuzzy Control System [161] | Intuitionistic Fuzzy Control System [162] | Neutrosophic Control System [163, 164] | Plithogenic Control System [165] |
| Automaton [166] | Fuzzy Automaton [167] | Intuitionistic Fuzzy Automaton [168] | Neutrosophic Automaton [169] | Plithogenic Automaton [170] |
| Lattice / Order [171] | Fuzzy Lattice [172] | Intuitionistic Fuzzy Lattice [69] | Neutrosophic Lattice [173] | Plithogenic Lattice [174] |
| Category [175] | Fuzzy Category [176] | Intuitionistic Fuzzy Category [177] | Neutrosophic Category [178] | Plithogenic Category |
| Time Series [179] | Fuzzy Time Series [180, 181] | Intuitionistic Fuzzy Time Series [182, 183] | Neutrosophic Time Series [184–186] | Plithogenic Time Series |
| Ontology [187] | Fuzzy Ontology [188] | Intuitionistic Fuzzy Ontology [189] | Neutrosophic Ontology [190, 191] | Plithogenic Ontology |

Table 1.5: Part 3 — Further concepts across classical, fuzzy, intuitionistic fuzzy, neutrosophic, and plithogenic frameworks.

| Classical Concept | Fuzzy | Intuitionistic Fuzzy | Neutrosophic | Plithogenic |
|---|---|---|---|---|
| Preference Relation [192] | Fuzzy Preference Relation [193, 194] | Intuitionistic Fuzzy Preference Relation [195, 196] | Neutrosophic Preference Relation [197–199] | Plithogenic Preference Relation |
| Aggregation Operator [200] | Fuzzy Aggregation Operator [201] | Intuitionistic Fuzzy Aggregation Operator [202, 203] | Neutrosophic Aggregation Operator [204, 205] | Plithogenic Aggregation Operator [18, 206] |
| Entropy / Information Measure [207] | Fuzzy Entropy / Information Measure [208] | Intuitionistic Fuzzy Entropy / Information Measure [209] | Neutrosophic Entropy / Information Measure [210] | Plithogenic Entropy / Information Measure [211] |
| Similarity Measure [212] | Fuzzy Similarity Measure [213] | Intuitionistic Fuzzy Similarity Measure [214] | Neutrosophic Similarity Measure [215, 216] | Plithogenic Similarity Measure [217] |
| Game (Game Theory) [218] | Fuzzy Game [219] | Intuitionistic Fuzzy Game [220] | Neutrosophic Game [221] | Plithogenic Game |
| Neural Network [222] | Fuzzy Neural Network [223, 224] | Intuitionistic Fuzzy Neural Network [225, 226] | Neutrosophic Neural Network [227–230] | Plithogenic Neural Network (cf. [231]) |
| Regression Model [232] | Fuzzy Regression Model [233, 234] | Intuitionistic Fuzzy Regression Model [235, 236] | Neutrosophic Regression Model [237–239] | Plithogenic Regression Model |
| Database [240] / Knowledge Base | Fuzzy Database [241, 242] / Knowledge Base [243, 244] | Intuitionistic Fuzzy Database [245] / Knowledge Base | Neutrosophic Database [246–249] / Knowledge Base | Plithogenic Database / Knowledge Base |
| Rule-Based System | Fuzzy Rule-Based System [250] | Intuitionistic Fuzzy Rule-Based System [251] | Neutrosophic Rule-Based System [252, 253] | Plithogenic Rule-Based System |
| Matroid [254] | Fuzzy Matroid [255, 256] | Intuitionistic Fuzzy Matroid [257] | Neutrosophic Matroid [258, 259] | Plithogenic Matroid [260] |
| Machine Learning [261] | Fuzzy Machine Learning Model [262, 263] | Intuitionistic Fuzzy Machine Learning Model [264, 265] | Neutrosophic Machine Learning Model [266, 267] | Plithogenic Machine Learning Model |
| HyperStructure [268, 269] | Fuzzy HyperStructure [270] | Intuitionistic Fuzzy HyperStructure [79] | Neutrosophic HyperStructure [271, 272] | Plithogenic HyperStructure [174] |

# Chapter 2

# Preliminaries

This section gathers the background notions and notation required for the main results. Unless explicitly stated otherwise, every set that appears is assumed to be finite.

## 2.1 Fuzzy Set

A Fuzzy Set assigns each element a single membership degree in $[0,1]$ [1,38,274]. The definitions and concrete examples are presented below.

**Definition 2.1.1** (Fuzzy Set)**.** [1] Let $X$ be a nonempty set. A *fuzzy set* $A$ on $X$ is characterized by its membership function

$$\mu_A : X \to [0,1].$$

That is, the fuzzy set $A$ is defined as

$$A = \{(x, \mu_A(x)) \mid x \in X\},$$

where $\mu_A(x)$ represents the degree to which the element $x \in X$ belongs to the set $A$.

A brief concrete example of this concept is provided below.

**Example 2.1.2** (Comfortable room temperature)**.** Let the universe be real temperatures $X = \mathbb{R}$ in degrees Celsius and define the fuzzy set $A =$ "Comfortable temperature" by the triangular membership

$$\mu_A(t) = \begin{cases} 0, & t \le 16, \\ \dfrac{t-16}{22-16} = \dfrac{t-16}{6}, & 16 < t < 22, \\ \dfrac{28-t}{28-22} = \dfrac{28-t}{6}, & 22 \le t < 28, \\ 0, & t \ge 28. \end{cases}$$

Concrete evaluations (numerically explicit):

$$\mu_A(18) = \frac{18-16}{6} = \frac{2}{6} = \frac{1}{3} \approx 0.3333, \qquad \mu_A(22) = 1, \qquad \mu_A(27) = \frac{28-27}{6} = \frac{1}{6} \approx 0.1667.$$

Hence $t = 22°$C is fully comfortable, $18°$C is moderately comfortable, and $27°$C is only slightly comfortable.

**Example 2.1.3** (Premium customer by monthly spend)**.** Let $X = \mathbb{R}_{\geq 0}$ denote monthly customer spending (USD). Define the fuzzy set $P =$ "Premium customer" by the trapezoidal membership with breakpoints $a = 200$, $b = 400$, $c = 1200$, $d = 1600$:

$$\mu_P(x) = \begin{cases} 0, & x \leq a, \\ \dfrac{x-a}{b-a} = \dfrac{x-200}{200}, & a < x < b, \\ 1, & b \leq x \leq c, \\ \dfrac{d-x}{d-c} = \dfrac{1600-x}{400}, & c < x < d, \\ 0, & x \geq d. \end{cases}$$

Concrete evaluations (step-by-step):

$$\mu_P(300) = \frac{300-200}{200} = \frac{100}{200} = 0.5,$$
$$\mu_P(900) = 1 \quad \text{(on the plateau)},$$
$$\mu_P(1400) = \frac{1600-1400}{400} = \frac{200}{400} = 0.5.$$

Thus a \$300 spender is a premium customer to degree 0.5, \$900 is fully premium, and \$1400 declines to 0.5 as spending moves into the upper taper.

## 2.2 Intuitionistic fuzzy set

An intuitionistic fuzzy set assigns each element membership and nonmembership degrees [2,275,276]. The definitions and concrete examples are presented below.

**Definition 2.2.1** (Intuitionistic fuzzy set)**.** [277] Let $E$ be a nonempty set. An *intuitionistic fuzzy set* (IFS) $A$ on $E$ is given by

$$A = \{\langle x, \mu_A(x), \nu_A(x)\rangle : x \in E\},$$

where

$$\mu_A, \nu_A : E \longrightarrow [0,1]$$

are, respectively, the membership and non–membership functions, and for every $x \in E$ one has

$$0 \;\leq\; \mu_A(x) + \nu_A(x) \;\leq\; 1.$$

The quantity

$$\pi_A(x) := 1 - \mu_A(x) - \nu_A(x)$$

represents the hesitation degree at $x$.

The usual fuzzy set notion is recovered in the special case $\nu_A(x) = 1 - \mu_A(x)$ for all $x \in E$, that is, when $\pi_A(x) = 0$ for every $x$.

**Example 2.2.2** (Medical diagnosis: intuitionistic fuzzy "high risk" class)**.** Let $E = \{p_1, p_2, p_3, p_4\}$ be a set of patients and consider the intuitionistic fuzzy concept

$$A = \text{"patient is at high cardiovascular risk"}.$$

We specify $A$ by giving, for each patient $p_i$, the membership degree $\mu_A(p_i)$ and the non–membership degree $\nu_A(p_i)$, with $\mu_A(p_i) + \nu_A(p_i) \leq 1$.

For instance, suppose the cardiologist assesses

| $x$ | $p_1$ | $p_2$ | $p_3$ | $p_4$ |
|---|---|---|---|---|
| $\mu_A(x)$ | 0.85 | 0.60 | 0.30 | 0.10 |
| $\nu_A(x)$ | 0.05 | 0.20 | 0.40 | 0.70 |

Then, for each $p_i$ we have

$$\mu_A(p_i) + \nu_A(p_i) \in \{0.90, 0.80, 0.70, 0.80\} \leq 1,$$

so these values define an intuitionistic fuzzy set. Here $p_1$ has a high membership and very low non–membership (clearly high risk), while $p_4$ has low membership and high non–membership (clearly not high risk). The remaining patients represent intermediate, uncertain cases with a nonzero "hesitation margin" $1 - \mu_A(p_i) - \nu_A(p_i)$.

**Example 2.2.3** (Customer satisfaction: intuitionistic fuzzy "satisfied" class)**.** Let

$$E = \{\text{Service A}, \text{Service B}, \text{Service C}\}$$

denote three online services offered by a company. Consider the intuitionistic fuzzy notion

$$B = \text{"users are satisfied with the service"}.$$

The intuitionistic fuzzy set $B$ is given by

$$B = \big\{\langle x, \mu_B(x), \nu_B(x)\rangle : x \in E\big\},$$

where $\mu_B(x)$ is the degree of satisfaction and $\nu_B(x)$ is the degree of dissatisfaction.

Assume that a survey yields the following aggregated assessments:

| $x$ | Service A | Service B | Service C |
|---|---|---|---|
| $\mu_B(x)$ | 0.70 | 0.40 | 0.20 |
| $\nu_B(x)$ | 0.10 | 0.30 | 0.60 |

Then

$$\mu_B(x) + \nu_B(x) = \begin{cases} 0.80 & \text{for Service A,} \\ 0.70 & \text{for Service B,} \\ 0.80 & \text{for Service C,} \end{cases}$$

all of which are $\leq 1$, so $B$ is an intuitionistic fuzzy set. Service A is mostly satisfactory, Service C is mostly unsatisfactory, and Service B lies in between. The remaining part $1 - \mu_B(x) - \nu_B(x)$ for each service measures the hesitation or lack of information in the survey responses.

## 2.3 Neutrosophic Set

A Neutrosophic Set assigns to each element three independent membership degrees— truth ($T$), indeterminacy ($I$), and falsity ($F$)—each taking values in $[0, 1]$, thereby enabling flexible modeling of uncertainty [4, 14, 15, 278]. Neutrosophic Sets are widely recognized as generalizations of Fuzzy Sets and Intuitionistic Fuzzy Sets, and they offer a highly adaptable framework by explicitly accommodating the component $I$. The definitions and concrete examples are presented below.

**Definition 2.3.1** (Neutrosophic Set)**.** [27, 279] Let $X$ be a non-empty set. A *Neutrosophic Set (NS)* $A$ on $X$ is characterized by three membership functions:

$$T_A : X \to [0, 1], \quad I_A : X \to [0, 1], \quad F_A : X \to [0, 1],$$

where for each $x \in X$, the values $T_A(x)$, $I_A(x)$, and $F_A(x)$ represent the degrees of truth, indeterminacy, and falsity, respectively. These values satisfy the following condition:

$$0 \leq T_A(x) + I_A(x) + F_A(x) \leq 3.$$

A brief concrete example of this concept is provided below.

**Example 2.3.2** (Medical diagnosis under conflicting evidence: "Patient has influenza")**.** Medical diagnosis is the systematic process of identifying diseases or conditions from patient history, examinations, tests, and reasoning by clinicians (cf. [280, 281]).

Let the universe be $X = \{\text{patients}\}$. For $x \in X$, suppose we observe: fever $T_C(x)$ in °C, antigen test score $a(x) \in [0, 1]$, and cough severity $c(x) \in [0, 1]$.

Define the neutrosophic membership of the set $A$ = "has influenza" by

$$T_A(x) = \min\left\{1,\ 0.5\,a(x)\ +\ 0.3\ \max\!\left(0, \frac{T_C(x) - 37}{2}\right)\ +\ 0.2\,c(x)\right\},$$

$$F_A(x) = \min\left\{1,\ 0.6\,(1 - a(x))\ +\ 0.4\ \max\!\left(0, \frac{37 - T_C(x)}{2}\right)\right\},$$

$$I_A(x) = \Big(1 - |2a(x) - 1|\Big) \cdot \Big(1 - \min\{1,\ |T_C(x) - 37|/1.5\}\Big).$$

Numerical instance. Take $T_C = 38.2$, $a = 0.7$, $c = 0.6$ for a patient $x^*$. Then

$$\begin{aligned} T_A(x^*) &= \min\{1,\ 0.5 \cdot 0.7 + 0.3 \cdot (38.2 - 37)/2 + 0.2 \cdot 0.6\} \\ &= \min\{1,\ 0.35 + 0.3 \cdot 0.6 + 0.12\} \\ &= \min\{1,\ 0.35 + 0.18 + 0.12\} = \min\{1,\ 0.65\} = 0.65, \end{aligned}$$

$$\begin{aligned} F_A(x^*) &= \min\{1,\ 0.6 \cdot (1 - 0.7) + 0.4 \cdot \max(0, (37 - 38.2)/2)\} \\ &= \min\{1,\ 0.6 \cdot 0.3 + 0\} = \min\{1,\ 0.18\} = 0.18, \end{aligned}$$

$$\begin{aligned} I_A(x^*) &= \big(1 - |1.4 - 1|\big) \cdot \Big(1 - \min\{1,\ 1.2/1.5\}\Big) \\ &= (1 - 0.4) \cdot (1 - 0.8) = 0.6 \cdot 0.2 = 0.12. \end{aligned}$$

Hence $(T_A, I_A, F_A)(x^*) = (0.65,\ 0.12,\ 0.18)$ and $T_A + I_A + F_A = 0.95 \le 3$ as required.

**Example 2.3.3** (Logistics ETA assessment: "Shipment arrives on time")**.** Let $X = \{\text{shipments}\}$. For $x \in X$, let $r(x) \in [0, 1]$ be the carrier on-time reliability, $\mu(x) > 0$ the predicted remaining transit time (days), $s(x) > 0$ the remaining slack until the promised date (days), and $u(x) \in [0, 1]$ an external-uncertainty score (e.g., weather/customs).

Define membership for $B$ = "arrives on time" by

$$\begin{aligned} g(x) &:= \frac{s(x)}{s(x) + \mu(x)} \in (0, 1), \\ T_B(x) &= \min\{1,\ 0.6\,r(x) + 0.4\,g(x)\}, \\ F_B(x) &= \min\{1,\ 0.6\,(1 - r(x)) + 0.4\,(1 - g(x))\}, \\ I_B(x) &= u(x). \end{aligned}$$

Numerical instance. Let $r = 0.85$, $\mu = 1.8$, $s = 2.0$, $u = 0.25$ for a shipment $x^\dagger$. Then

$$g(x^\dagger) = \frac{2.0}{2.0 + 1.8} = \frac{2.0}{3.8} \approx 0.5263,$$

$$T_B(x^\dagger) = \min\{1,\ 0.6 \cdot 0.85 + 0.4 \cdot 0.5263\} = \min\{1,\ 0.51 + 0.2105\} = 0.7205,$$

$$F_B(x^\dagger) = \min\{1,\ 0.6 \cdot 0.15 + 0.4 \cdot (1 - 0.5263)\} = \min\{1,\ 0.09 + 0.1895\} = 0.2795,$$

$$I_B(x^\dagger) = 0.25.$$

Thus $(T_B, I_B, F_B)(x^\dagger) = (0.7205,\ 0.25,\ 0.2795)$ and $T_B + I_B + F_B = 1.2500 \le 3$, satisfying the neutrosophic bounds.

## 2.4 Rough Set

A Rough Set approximates a subset using lower and upper bounds based on equivalence classes, capturing certainty and uncertainty in membership [282–285]. The definitions and concrete examples are presented below.

**Definition 2.4.1** (Rough Set Approximation)**.** [286] Let $X$ be a non-empty universe of discourse, and let $R \subseteq X \times X$ be an equivalence relation (or indiscernibility relation) on $X$. The equivalence relation $R$ partitions $X$ into disjoint equivalence classes, denoted by $[x]_R$ for $x \in X$, where:

$$[x]_R = \{y \in X \mid (x, y) \in R\}.$$

For any subset $U \subseteq X$, the *lower approximation* $\underline{U}$ and the *upper approximation* $\overline{U}$ of $U$ are defined as follows:

1. *Lower Approximation* $\underline{U}$:
$$\underline{U} = \{x \in X \mid [x]_R \subseteq U\}.$$
The lower approximation $\underline{U}$ includes all elements of $X$ whose equivalence classes are entirely contained within $U$. These are the elements that *definitely* belong to $U$.
2. *Upper Approximation* $\overline{U}$:
$$\overline{U} = \{x \in X \mid [x]_R \cap U \neq \emptyset\}.$$
The upper approximation $\overline{U}$ contains all elements of $X$ whose equivalence classes have a non-empty intersection with $U$. These are the elements that *possibly* belong to $U$.

The pair $(\underline{U}, \overline{U})$ forms the *rough set* representation of $U$, satisfying the relationship:

$$\underline{U} \subseteq U \subseteq \overline{U}.$$

A brief concrete example of this concept is provided below.

**Example 2.4.2** (Email spam filtering with rough approximations)**.** Email spam filtering automatically detects, classifies, and separates unsolicited or malicious messages from legitimate emails using rules and machine learning (cf. [287, 288]).

Consider a mailbox with ten emails $X = \{e_1, e_2, \ldots, e_{10}\}$. Define an indiscernibility (equivalence) relation $R$ that groups emails by two coarse features only: (sender domain category $\in$ {known, unknown} and subject contains the keyword "free" $\in$ {yes, no}). This yields the following $R$-equivalence classes (blocks):

$$B_1 = \{e_1, e_2\} \quad \text{(unknown domain, “free” in subject)},$$
$$B_2 = \{e_3, e_4, e_5\} \quad \text{(unknown domain, no “free”)},$$
$$B_3 = \{e_6, e_7\} \quad \text{(known domain, “free”)},$$
$$B_4 = \{e_8, e_9, e_{10}\} \quad \text{(known domain, no “free”)}.$$

Let the target concept be $U = \{\text{emails that are actually spam}\} = \{e_1, e_2, e_3, e_6, e_7\}$. By definition of rough sets (with respect to $R$):

1) Lower approximation

$$\underline{U} = \{x \in X : [x]_R \subseteq U\} = B_1 \cup B_3 = \{e_1, e_2, e_6, e_7\}.$$

Explanation: $B_1 \subseteq U$ and $B_3 \subseteq U$; hence all elements of these blocks are *certainly* spam.

2) Upper approximation

$$\overline{U} = \{x \in X : [x]_R \cap U \neq \varnothing\} = B_1 \cup B_2 \cup B_3 = \{e_1, e_2, e_3, e_4, e_5, e_6, e_7\}.$$

Explanation: $B_2$ intersects $U$ (it contains $e_3$), so all of $B_2$ are *possibly* spam. Block $B_4$ does not intersect $U$.

3) Boundary, positive, and negative regions

$$\mathrm{BND}_R(U) = \overline{U} \setminus \underline{U} = \{e_3, e_4, e_5\}, \qquad \mathrm{POS}_R(U) = \underline{U} = \{e_1, e_2, e_6, e_7\},$$

$$\mathrm{NEG}_R(U) = X \setminus \overline{U} = \{e_8, e_9, e_{10}\}.$$

4) Numerical indices (with explicit values)

$$|\underline{U}| = 4, \quad |\overline{U}| = 7, \quad |X| = 10.$$

Pawlak accuracy of approximation:

$$\alpha_R(U) = \frac{|\underline{U}|}{|\overline{U}|} = \frac{4}{7} \approx 0.5714.$$

Coverage (certainty rate in the universe):

$$\kappa_R(U) = \frac{|\underline{U}|}{|X|} = \frac{4}{10} = 0.4.$$

Interpretation. Emails in $B_1$ and $B_3$ are *certainly* spam under these coarse features; emails in $B_2$ are ambiguous (boundary); emails in $B_4$ are *certainly* non-spam.

**Example 2.4.3** (Factory quality control with sensor-based indiscernibility)**.** Factory quality control monitors production processes and outputs, inspecting samples, detecting defects, and ensuring products meet safety and performance standards (cf. [289]).

A factory produces twelve items $X = \{p_1, \ldots, p_{12}\}$. Two coarse sensors are used: surface scratch flag $\in \{0, 1\}$ and thickness bin $\in \{\text{thin}, \text{thick}\}$. Items are indiscernible if they share the same ordered pair (scratch, thickness). This induces the $R$-equivalence classes

$$\begin{aligned} E_1 &= \{p_1, p_2, p_3\} \quad (1, \text{thin}), \\ E_2 &= \{p_4, p_5\} \quad (1, \text{thick}), \\ E_3 &= \{p_6, p_7, p_8\} \quad (0, \text{thin}), \\ E_4 &= \{p_9, p_{10}, p_{11}, p_{12}\} \quad (0, \text{thick}). \end{aligned}$$

Let the true concept be $U = \{\text{defective items}\} = \{p_1, p_2, p_3, p_4, p_9\}$ (obtained after a detailed downstream inspection).

1) Lower approximation

$$\underline{U} = \{x \in X : [x]_R \subseteq U\} = E_1 = \{p_1, p_2, p_3\}.$$

Explanation: all of $E_1$ are truly defective; other blocks contain a mix.

2) Upper approximation

$$\overline{U} = \{x \in X : [x]_R \cap U \neq \varnothing\} = E_1 \cup E_2 \cup E_4 = \{p_1, p_2, p_3, p_4, p_5, p_9, p_{10}, p_{11}, p_{12}\}.$$

3) Boundary, positive, and negative regions

$$\mathrm{BND}_R(U) = \overline{U} \setminus \underline{U} = \{p_4, p_5, p_9, p_{10}, p_{11}, p_{12}\}, \qquad \mathrm{POS}_R(U) = \underline{U} = \{p_1, p_2, p_3\},$$

$$\mathrm{NEG}_R(U) = X \setminus \overline{U} = \{p_6, p_7, p_8\}.$$

4) Numerical indices (computed explicitly)

$$|\underline{U}| = 3, \quad |\overline{U}| = 9, \quad |X| = 12.$$

Pawlak accuracy of approximation:

$$\alpha_R(U) = \frac{|\underline{U}|}{|\overline{U}|} = \frac{3}{9} = \frac{1}{3} \approx 0.3333.$$

Coverage (certainty rate in the universe):

$$\kappa_R(U) = \frac{|\underline{U}|}{|X|} = \frac{3}{12} = \frac{1}{4} = 0.25.$$

Interpretation. Items in $E_1$ are *certainly* defective under the coarse sensors; $E_2$ and $E_4$ are ambiguous (boundary); $E_3$ is *certainly* non-defective. Rough approximations separate what can be guaranteed (lower), what is only possible (upper), and what is impossible (negative) using only the coarse sensor information.

Table 2.1 lists the extended rough–set families.

Table 2.1: Concise comparison of extended rough–set families

| **Concept** | **One–line summary** | **Refs.** |
|---|---|---|
| HyperRough Sets | Rough approximations over hyperrelations/hypergraphs (multiway neighborhoods). | [290, 291] |
| SuperHyperRough Sets | Hierarchical (super/hyper) powerset rough models with multilayer approximations. | [292–295] |
| Dominance-Based Rough Sets (DRSA) | Approximations induced by dominance (preference) relations for MCDA. | [296–298] |
| Decision-Theoretic Rough Sets (DTRS) | Bayes risk thresholds yield three-way decisions (accept/defer/reject). | [299–301] |
| Rough Multisets | Rough approximations extended to multisets with element multiplicities. | [302, 303] |
| Composite Rough Sets | Unified rough models combining multiple relations/approximation operators. | [304–306] |

## 2.5 Soft set

A Soft Set is a parameterized family of subsets used to handle uncertainty, introduced by Molodtsov in 1999 for decision-making problems [307,308]. The definitions and concrete examples are presented below.

**Definition 2.5.1** (Soft Set [308])**.** Let $U$ be a universe set and $E$ be a set of parameters. Let $A \subseteq E$ and denote by $\mathcal{P}(U)$ the power set of $U$. A pair $(F, A)$ is called a *soft set* over $U$ if

$$F : A \to \mathcal{P}(U).$$

For each parameter $\epsilon \in A$, the set $F(\epsilon)$ is called the $\epsilon$-approximation of the soft set $(F, A)$. In other words, a soft set over $U$ is a parameterized family of subsets of $U$.

A brief concrete example of this concept is provided below.

**Example 2.5.2** (Apartment Selection (Tokyo Rental Case))**.** Apartment selection evaluates multiple rental options using criteria like location, rent, size, amenities, and suitability for residents and lifestyles preferences (cf. [309]).

Let the universe $U = \{A_1, A_2, A_3, A_4\}$ denote four available apartments. Let the parameter set be

$$E = \{\text{near_station},\ \text{pet_friendly},\ \text{under_¥120,000},\ \text{twoLDK_or_more},\ \text{built_after_2015}\},$$

and take $A = E$. Define the soft set $(F, A)$ over $U$ by listing, for each parameter $\epsilon \in A$, the subset $F(\epsilon) \subseteq U$ of apartments satisfying $\epsilon$:

$$\begin{aligned} F(\text{near_station}) &= \{A_1, A_3, A_4\}, \\ F(\text{pet_friendly}) &= \{A_2, A_3\}, \\ F(\text{under_¥120,000}) &= \{A_1, A_2\}, \\ F(\text{twoLDK_or_more}) &= \{A_1, A_4\}, \\ F(\text{built_after_2015}) &= \{A_3, A_4\}. \end{aligned}$$

Interpretation. The mapping encodes, for each practical requirement, which apartments meet it. For a renter who requires "near_station" and "pet_friendly", the feasible candidates are

$$F(\text{near_station}) \cap F(\text{pet_friendly}) = \{A_3\}.$$

**Example 2.5.3** (Laptop Purchase under Practical Preferences)**.** Laptop purchase is selecting a notebook computer balancing performance, portability, battery life, budget, brand support, future needs, and upgrades (cf. [310]).

Let the universe $U = \{L_1, L_2, L_3, L_4\}$ denote four laptop models under consideration. Let the parameter set be

$$E = \{\text{lightweight}(\leq 1.2 \text{ kg}),\ \text{long_battery}(\geq 10 \text{ h}),\ \text{budget}(\leq ¥100{,}000),\ \text{ram16GB},\ \text{screen14in}\},$$

and take $A = E$. Define the soft set $(F, A)$ over $U$ by

$$\begin{aligned} F(\text{lightweight}) &= \{L_1, L_3\}, \\ F(\text{long_battery}) &= \{L_1, L_2, L_4\}, \\ F(\text{budget}) &= \{L_2, L_3\}, \\ F(\text{ram16GB}) &= \{L_1, L_4\}, \\ F(\text{screen14in}) &= \{L_1, L_3, L_4\}. \end{aligned}$$

Interpretation. Each parameter captures a user preference, and $F(\epsilon)$ lists laptops meeting it. A buyer who insists on "budget" and "screen14in" filters to

$$F(\text{budget}) \cap F(\text{screen14in}) = \{L_3\}.$$

If the buyer also prefers "long_battery", then

$$F(\text{budget}) \cap F(\text{screen14in}) \cap F(\text{long_battery}) = \varnothing,$$

signaling that no single item simultaneously satisfies all three preferences and that trade-offs are required.

Related concepts of the Soft Set include HyperSoft Set [311,312], IndetermSoft Set [313–315], SuperHyperSoft Set [316,317], TreeSoft Set [318–321], ForestSoft Set [322–325], Bipolar Soft Set [326,327], and Double-Framed Soft Set [328,329], all of which extend the classical Soft Set framework in different directions.

The definitions of the HyperSoft Set and SuperHyperSoft Set are given as follows. Note that table 2.2 presents a concise comparison of the Soft Set, Hypersoft Set, and SuperHyperSoft Set.

Table 2.2: Concise comparison of Soft Set, Hypersoft Set, and SuperHyperSoft Set. Here $\mathcal{P}(U)$ denotes the power set of $U$.

| | **Soft Set** | **Hypersoft Set** | **SuperHyperSoft Set** |
|---|---|---|---|
| Universe | $U$ | $U$ | $U$ |
| Parameter domain | $A \subseteq E$ (single attribute) | $\mathcal{C} = \mathcal{A}_1 \times \cdots \times \mathcal{A}_m$ (fixed $m$ attributes) | $\mathcal{C} = \mathcal{P}(A_1) \times \cdots \times \mathcal{P}(A_n)$ (subset–valued per attribute) |
| Input (query key) | $\epsilon \in A$ | $\gamma = (\gamma_1, \dots, \gamma_m)$ with $\gamma_i \in \mathcal{A}_i$ | $\alpha = (\alpha_1, \dots, \alpha_n)$ with $\alpha_i \subseteq A_i$ |
| Mapping | $F : A \to \mathcal{P}(U)$ | $G : \mathcal{C} \to \mathcal{P}(U)$ | $F : \mathcal{C} \to \mathcal{P}(U)$ |
| Granularity | Single parameter value | Exact $m$-tuple of values | Sets of admissible values per attribute ("any of these") |
| Expressiveness | Low (one attribute at a time) | Medium (multi-attribute conjunction) | High (multi-attribute with set-level choices) |
| Reductions | — | $m$=1 ⇒ Soft Set | All $\alpha_i$ singletons ⇒ Hypersoft; $n$=1 and singleton ⇒ Soft |
| Typical query | "cars with color=red" | "laptops with (CPU $=$ i7, RAM $=$ 16, SSD $=$ 512)" | "trips with season$\in$ {Spring, Autumn}, budget$\in$ {Low, Mid}, type$\in$ {Solo, Business}" |

**Definition 2.5.4** (Hypersoft Set)**.** [312] Let $U$ be a universal set, and let $\mathcal{A}_1, \mathcal{A}_2, \dots, \mathcal{A}_m$ be attribute domains. Define $\mathcal{C} = \mathcal{A}_1 \times \mathcal{A}_2 \times \cdots \times \mathcal{A}_m$, the Cartesian product of these domains. A hypersoft set over $U$ is a pair $(G, \mathcal{C})$, where $G : \mathcal{C} \to \mathcal{P}(U)$. The hypersoft set is expressed as:

$$(G, \mathcal{C}) = \{(\gamma, G(\gamma)) \mid \gamma \in \mathcal{C}, G(\gamma) \in \mathcal{P}(U)\}.$$

For an $m$-tuple $\gamma = (\gamma_1, \gamma_2, \dots, \gamma_m) \in \mathcal{C}$, where $\gamma_i \in \mathcal{A}_i$ for $i = 1, 2, \dots, m$, $G(\gamma)$ represents the subset of $U$ corresponding to the combination of attribute values $\gamma_1, \gamma_2, \dots, \gamma_m$.

**Example 2.5.5** (Hypersoft Set — Laptop selection by multi-attribute profile)**.** Let the universe of laptops be

$$U = \{L_1, L_2, L_3, L_4, L_5, L_6\}.$$

Assign each item its (CPU, RAM in GB, SSD in GB):

$$L_1 : (\text{i5}, 8, 256), \quad L_2 : (\text{i5}, 16, 512), \quad L_3 : (\text{i7}, 8, 512),$$
$$L_4 : (\text{i7}, 16, 256), \quad L_5 : (\text{i7}, 16, 512), \quad L_6 : (\text{i5}, 8, 512).$$

Let the attribute domains be

$$\mathcal{A}_1 = \{\text{i5}, \text{i7}\}, \quad \mathcal{A}_2 = \{8, 16\}, \quad \mathcal{A}_3 = \{256, 512\},$$

and let $\mathcal{C} = \mathcal{A}_1 \times \mathcal{A}_2 \times \mathcal{A}_3$. Define the hypersoft mapping $G : \mathcal{C} \to \mathcal{P}(U)$ by

$$G(\gamma_1, \gamma_2, \gamma_3) = \{\, L \in U : \ (\text{CPU}(L), \text{RAM}(L), \text{SSD}(L)) = (\gamma_1, \gamma_2, \gamma_3) \,\}.$$

Then, concretely,

$$\begin{aligned} G(\text{i7}, 16, 512) &= \{L_5\}, \\ G(\text{i5}, 8, 512) &= \{L_6\}, \\ G(\text{i5}, 16, 256) &= \varnothing \quad \text{(no laptop matches)}. \end{aligned}$$

This hypersoft set captures the real-life query "laptops with a given CPU–RAM–SSD triple," returning exactly the subset of products in $U$ that satisfy the chosen attribute combination.

**Definition 2.5.6** (SuperHyperSoft Set)**.** [330] Let $U$ be a universal set, and let $\mathcal{P}(U)$ denote the power set of $U$. Consider $n$ distinct attributes $a_1, a_2, \ldots, a_n$, where $n \geq 1$. Each attribute $a_i$ is associated with a set of attribute values $A_i$, satisfying the property $A_i \cap A_j = \emptyset$ for all $i \neq j$.

Define $\mathcal{P}(A_i)$ as the power set of $A_i$ for each $i = 1, 2, \ldots, n$. Then, the Cartesian product of the power sets of attribute values is given by:

$$\mathcal{C} = \mathcal{P}(A_1) \times \mathcal{P}(A_2) \times \cdots \times \mathcal{P}(A_n).$$

A SuperHyperSoft Set over $U$ is a pair $(F, \mathcal{C})$, where:

$$F : \mathcal{C} \to \mathcal{P}(U),$$

and $F$ maps each element $(\alpha_1, \alpha_2, \ldots, \alpha_n) \in \mathcal{C}$ (with $\alpha_i \in \mathcal{P}(A_i)$) to a subset $F(\alpha_1, \alpha_2, \ldots, \alpha_n) \subseteq U$. Mathematically, the SuperHyperSoft Set is represented as:

$$(F, \mathcal{C}) = \{(\gamma, F(\gamma)) \mid \gamma \in \mathcal{C}, F(\gamma) \in \mathcal{P}(U)\}.$$

Here, $\gamma = (\alpha_1, \alpha_2, \ldots, \alpha_n) \in \mathcal{C}$, where $\alpha_i \in \mathcal{P}(A_i)$ for $i = 1, 2, \ldots, n$, and $F(\gamma)$ corresponds to the subset of $U$ defined by the combined attribute values $\alpha_1, \alpha_2, \ldots, \alpha_n$.

**Example 2.5.7** (SuperHyperSoft Set — Travel package filtering by subset-valued criteria)**.** A travel package bundles transport, accommodation, activities, and services into one pre-arranged offer at a combined, often discounted price overall (cf. [331]).

Let the universe of travel packages be

$$U = \{p_1, p_2, p_3, p_4, p_5\}.$$

Attribute value sets:

$$A_1 = \{\text{Spring}, \text{Summer}, \text{Autumn}, \text{Winter}\}, \quad A_2 = \{\text{Low}, \text{Mid}, \text{High}\}, \quad A_3 = \{\text{Solo}, \text{Family}, \text{Business}\}.$$

Each package carries one concrete value per attribute:

$$\begin{aligned} &p_1 : (\text{Summer}, \text{High}, \text{Family}), \quad p_2 : (\text{Winter}, \text{Mid}, \text{Family}), \\ &p_3 : (\text{Autumn}, \text{Mid}, \text{Solo}), \quad p_4 : (\text{Autumn}, \text{High}, \text{Business}), \quad p_5 : (\text{Spring}, \text{Low}, \text{Solo}). \end{aligned}$$

Form the SuperHyperSoft domain

$$\mathcal{C} = \mathcal{P}(A_1) \times \mathcal{P}(A_2) \times \mathcal{P}(A_3),$$

and define $F : \mathcal{C} \to \mathcal{P}(U)$ by, for $\gamma = (\alpha_1, \alpha_2, \alpha_3)$,

$$F(\alpha_1, \alpha_2, \alpha_3) = \{\, p \in U : \ \text{Season}(p) \in \alpha_1,\ \text{Budget}(p) \in \alpha_2,\ \text{Type}(p) \in \alpha_3 \,\}.$$

Concrete queries with subset-valued criteria:

$$\begin{aligned}
&\alpha_1 = \{\text{Summer}, \text{Autumn}\},\ \alpha_2 = \{\text{High}\},\ \alpha_3 = \{\text{Family}, \text{Business}\} : && F(\alpha_1, \alpha_2, \alpha_3) = \{p_1, p_4\};\\
&\alpha_1 = \{\text{Winter}\},\ \alpha_2 = \{\text{Mid}\},\ \alpha_3 = \{\text{Family}\} : && F(\alpha_1, \alpha_2, \alpha_3) = \{p_2\};\\
&\alpha_1 = \{\text{Spring}, \text{Autumn}\},\ \alpha_2 = \{\text{Low}, \text{Mid}\},\ \alpha_3 = \{\text{Solo}\} : && F(\alpha_1, \alpha_2, \alpha_3) = \{p_3, p_5\}.
\end{aligned}$$

Allowing subsets $\alpha_i \subseteq A_i$ in the query domain enables practical filters such as "any of these seasons," "budget up to Mid," and "Solo or Business travelers," producing the matching subset of packages in $U$.

# Chapter 3

# Dynamic Reviews and Results of Uncertain Sets

This chapter surveys the relationships among Plithogenic Sets, Fuzzy Sets, Intuitionistic Fuzzy Sets, and Neutrosophic Sets. Plithogenic Sets are known to generalize Fuzzy Sets, Intuitionistic Fuzzy Sets, and Neutrosophic Sets by incorporating attribute-based membership and contradiction management. Table 3.1and3.2 presents an overview of the uncertain-set families discussed in this chapter.

## 3.1 Plithogenic Sets of Uncertain Values

The Plithogenic Set is a mathematical framework designed to integrate multi-valued degrees of appurtenance and contradiction, making it particularly effective for addressing complex decision-making scenarios [10,18]. Numerous studies have explored the properties and applications of Plithogenic Sets, as highlighted in works such as [332–335].

**Definition 3.1.1** (Plithogenic Set). [18, 100] Let $S$ be a universal set, and $P \subseteq S$. A *Plithogenic Set $PS$* is defined as:

$$PS = (P, v, Pv, pdf, pCF)$$

where:

- $v$ is an attribute.
- $Pv$ is the range of possible values for the attribute $v$.
- $pdf : P \times Pv \to [0,1]^s$ is the *Degree of Appurtenance Function (DAF)* [1]
- $pCF : Pv \times Pv \to [0,1]^t$ is the *Degree of Contradiction Function (DCF).*

[1] It is important to note that the definition of the Degree of Appurtenance Function varies across different papers. Some studies define this concept using the power set, while others simplify it by avoiding the use of the power set [336]. The author has consistently defined the Classical Plithogenic Set without employing the power set.

Table 3.1: Overview of the uncertain-set families in this chapter (Part I).

| Uncertain-set family | Brief overview |
|---|---|
| Plithogenic Set | Attribute-based framework with appurtenance degrees and a contradiction function, unifying fuzzy, intuitionistic, neutrosophic and related sets. |
| $m$-Polar Plithogenic Set | Splits each attribute value into $m$ semantic poles (e.g. positive / neutral / negative) with pole-level contradiction. |
| Complex Plithogenic Set | Uses complex-valued degrees; modulus measures strength, argument encodes extra features such as time, context, or direction. |
| SuperHyperPlithogenic Set | Plithogenic structure on hyper- and superhyper-objects, with nested degrees on superhyper-elements in higher-order settings. |
| Plithogenic Linguistic Set | Employs linguistic labels ("low", "medium", "high", etc.) as values with plithogenic aggregation and label-wise contradiction. |
| q-rung orthopair Plithogenic Set | Combines q-rung orthopair bounds on truth/falsity with plithogenic attributes and contradiction-sensitive operators. |
| Type-$n$ Plithogenic Set | Family indexed by $n$ organizing multiple membership components and collecting multi-valued plithogenic variants. |
| Iterative MultiPlithogenic Set | Iterated, multi-level plithogenic construction for hierarchical or staged decision and evaluation processes. |
| Interval-Valued Plithogenic Set | Represents appurtenance by intervals to capture measurement imprecision and disagreement between sources or experts. |
| Plithogenic OffSet | Allows over- and under-membership, modeling paradoxical, inconsistent, or extreme information in a plithogenic way. |
| Plithogenic Cubic Set | Combines a crisp component with an interval-valued one, both treated plithogenically, for two-layer uncertainty. |
| Plithogenic Soft / HyperSoft / SuperHyperSoft Set | Soft-like parameter, hyper-parameter, and superhyper-parameter families equipped with plithogenic degrees and contradiction. |
| Hesitant Plithogenic Set | Assigns a finite set of plausible plithogenic degrees to each element–attribute pair, explicitly modeling hesitation. |
| Spherical Plithogenic Set | Uses spherical-type constraints on positive, neutral, and negative components inside a plithogenic structure. |

These functions satisfy the following axioms for all $a, b \in Pv$:

1. *Reflexivity of Contradiction Function*:

$$pCF(a, a) = 0$$

2. *Symmetry of Contradiction Function*:

$$pCF(a, b) = pCF(b, a)$$

We now present concrete examples of the concept.

**Example 3.1.2** (Apartment Rental under Contradictory Criteria)**.** Apartment rental is a housing arrangement where tenants pay regular rent to use residential units owned by landlords for living (cf. [337]).

Table 3.2: Overview of the uncertain-set families in this chapter (Part II).

| Uncertain-set family | Brief overview |
|---|---|
| T-Spherical Plithogenic Set | T-spherical (or q-rung spherical) refinement with powered-sum bounds on truth, indeterminacy, and falsity. |
| Plithogenic Rough Set | Rough lower/upper approximations constructed from plithogenic membership and contradiction-aware similarity. |
| TreePlithogenic Set | Plithogenic sets on tree-structured universes or attributes, where aggregation follows the tree hierarchy. |
| ForestPlithogenic Set | Extension of TreePlithogenic sets to several disjoint hierarchies (a forest) in one plithogenic model. |
| Plithogenic Soft Expert Set | Soft expert system in which each expert's opinions form a plithogenic soft set with attribute-wise contradiction. |
| Dynamic Plithogenic Set | Time- or state-dependent plithogenic degrees and contradiction, evolving across scenarios or time steps. |
| Probabilistic Plithogenic Set | Integrates probability distributions with plithogenic degrees or attributes, mixing probabilistic and plithogenic uncertainty. |
| Nonstandard Plithogenic Set | Uses hyperreal, near-$[0,1]$ membership vectors with infinitesimal over/under-membership under plithogenic contradiction. |
| Linear Diophantine Plithogenic Set | Plithogenic degrees constrained by linear Diophantine relations on parameters, preserving attribute-based contradiction. |
| Subset-Valued Plithogenic Set | Assigns each item–attribute pair a subset of membership vectors in $[0,1]^s$, aggregated via contradiction-weighted plithogenic operators. |
| Triangular Plithogenic Set | Plithogenic extension of triangular fuzzy / intuitionistic / neutrosophic numbers with triangular membership shapes. |
| Trapezoidal Plithogenic Set | Plithogenic extension of trapezoidal fuzzy, intuitionistic fuzzy, and neutrosophic numbers using trapezoidal profiles. |
| Refined Plithogenic Set | Splits components (e.g. truth, indeterminacy, falsity) into several subcomponents within one plithogenic object. |
| Picture Plithogenic Set | Plithogenic analogue of picture fuzzy sets with acceptance, neutrality, rejection, and refusal under contradiction. |

Let the universe be three candidate apartments $P = \{A_1, A_2, A_3\}$. Consider one attribute $v =$ rental_criterion with value-set

$$Pv = \{\text{low_rent},\ \text{near_station},\ \text{large_space}\}.$$

We take a plithogenic Degree of Appurtenance Function (DAF) with $s = 3$ components (truth $T$, indeterminacy $I$, falsity $F$) and a scalar Degree of Contradiction Function (DCF) $pCF : Pv \times Pv \to [0,1]$ $(t = 1)$. The contradiction matrix (symmetric, zeros on the diagonal) is

| $pCF$ | low_rent | near_station | large_space |
|---|---|---|---|
| low_rent | 0 | 0.30 | 0.70 |
| near_station | 0.30 | 0 | 0.50 |
| large_space | 0.70 | 0.50 | 0 |

and the DAF $pdf : P \times Pv \to [0,1]^3$ (entries are $(T, I, F)$) is specified by

$$\begin{aligned}
pdf(A_1, \text{low_rent}) &= (0.85, 0.05, 0.10),\\
pdf(A_1, \text{near_station}) &= (0.90, 0.05, 0.10),\\
pdf(A_1, \text{large_space}) &= (0.40, 0.20, 0.60),\\
pdf(A_2, \text{low_rent}) &= (0.60, 0.15, 0.35),\\
pdf(A_2, \text{near_station}) &= (0.50, 0.20, 0.50),\\
pdf(A_2, \text{large_space}) &= (0.75, 0.10, 0.25),\\
pdf(A_3, \text{low_rent}) &= (0.45, 0.25, 0.55),\\
pdf(A_3, \text{near_station}) &= (0.80, 0.10, 0.20),\\
pdf(A_3, \text{large_space}) &= (0.55, 0.15, 0.45).
\end{aligned}$$

Interpretation. The decision maker may select a dominant value (e.g., near_station); the contradictions $pCF$ quantify how much other values conflict with the dominant one during plithogenic aggregation.

**Example 3.1.3** (Personalized Diet Planning (Clinical Setting))**.** Personalized diet planning designs nutrition plans tailored to an individual's health goals, preferences, restrictions, and lifestyle using data-driven adjustments continuously (cf. [338, 339]).

Let the universe be three patients $P = \{p_1, p_2, p_3\}$. Consider the attribute $v$ = diet_style with value-set

$$Pv = \{\text{ketogenic},\ \text{low_fat},\ \text{vegetarian}\}.$$

Use a scalar DCF with the following matrix:

| $pCF$ | ketogenic | low_fat | vegetarian |
|---|---|---|---|
| ketogenic | 0 | 0.60 | 0.90 |
| low_fat | 0.60 | 0 | 0.20 |
| vegetarian | 0.90 | 0.20 | 0 |

and define the DAF (triples $(T, I, F)$) by

$$\begin{aligned}
pdf(p_1, \text{ketogenic}) &= (0.70, 0.10, 0.30),\\
pdf(p_1, \text{low_fat}) &= (0.40, 0.20, 0.60),\\
pdf(p_1, \text{vegetarian}) &= (0.30, 0.20, 0.70),\\
pdf(p_2, \text{ketogenic}) &= (0.20, 0.20, 0.80),\\
pdf(p_2, \text{low_fat}) &= (0.65, 0.15, 0.35),\\
pdf(p_2, \text{vegetarian}) &= (0.55, 0.15, 0.45),\\
pdf(p_3, \text{ketogenic}) &= (0.50, 0.25, 0.50),\\
pdf(p_3, \text{low_fat}) &= (0.45, 0.20, 0.55),\\
pdf(p_3, \text{vegetarian}) &= (0.75, 0.10, 0.25).
\end{aligned}$$

Interpretation. Strong contradictions (e.g., ketogenic vs vegetarian at 0.90) reflect dietary principles that rarely co-exist; the $(T, I, F)$ entries encode how well each patient aligns with each style under medical, lifestyle, and ethical constraints.

**Example 3.1.4** (Sustainable Car Choice (Powertrain Trade-offs))**.** A sustainable car choice prioritizes low emissions, high fuel efficiency, lifecycle impact, and ethics while still meeting everyday mobility needs (cf. [340]). Let the universe be three cars $P = \{C_1, C_2, C_3\}$. Consider the attribute $v$ = powertrain with value-set

$$Pv = \{\text{electric},\ \text{hybrid},\ \text{gasoline}\}.$$

Take the scalar DCF

| $pCF$ | electric | hybrid | gasoline |
|---|---|---|---|
| electric | 0 | 0.30 | 1.00 |
| hybrid | 0.30 | 0 | 0.60 |
| gasoline | 1.00 | 0.60 | 0 |

and the DAF (triples $(T, I, F)$) as

$$\begin{aligned}
pdf(C_1, \text{electric}) &= (0.90, 0.05, 0.10),\\
pdf(C_1, \text{hybrid}) &= (0.70, 0.10, 0.30),\\
pdf(C_1, \text{gasoline}) &= (0.10, 0.20, 0.90),\\
pdf(C_2, \text{electric}) &= (0.40, 0.20, 0.60),\\
pdf(C_2, \text{hybrid}) &= (0.80, 0.10, 0.20),\\
pdf(C_2, \text{gasoline}) &= (0.35, 0.20, 0.65),\\
pdf(C_3, \text{electric}) &= (0.20, 0.15, 0.80),\\
pdf(C_3, \text{hybrid}) &= (0.50, 0.20, 0.50),\\
pdf(C_3, \text{gasoline}) &= (0.85, 0.05, 0.15).
\end{aligned}$$

Interpretation. The contradiction $pCF(\text{electric}, \text{gasoline}) = 1.00$ models fully opposed values (zero compatibility) in environmental terms, while electric vs hybrid has only mild contradiction 0.30. The $(T, I, F)$ vectors encode each car's appurtenance to the plithogenic set of "sustainable choices" under these value interactions.

Tables 3.3 and 3.4 present examples of sets that can be generalized by the Plithogenic Set framework (cf. [341]). Various concepts for handling uncertainty are being continuously defined and studied in the scientific community.

Table 3.3: A catalogue of Plithogenic Set families by number of components $s$.

| $s$ | $t$ | Representative type(s) |
|---|---|---|
| 1 | 0 | Fuzzy Set [1,44]; N-Set [342]; Shadowed Set [343–345] |
| 2 | 0 | Intuitionistic Fuzzy Set [2,346]; Vague Set [5,347]; Bipolar Fuzzy Set [348]; Intuitionistic Evidence Set [349–351]; Variable Fuzzy Set [352–354]; Paraconsistent Fuzzy Set [355,356]; Bifuzzy Set [357,358] |
| 3 | 0 | Neutrosophic Set[(a)] [278,279]; Hesitant Fuzzy Set [6,359]; Tripolar Fuzzy Set [360–362]; Three-way Fuzzy Set [363,364]; Picture Fuzzy Set [7,365]; Spherical Fuzzy Set [366,367]; Inconsistent Intuitionistic Fuzzy Set [368,369]; Ternary Fuzzy Set [334,370]; Neutrosophic Fuzzy Set [371,372]; (Kleene three-valued logic [373,374];) Neutrosophic Vague Set [375,376] |
| 4 | 0 | Quadripartitioned Neutrosophic Set [8, 377]; Double-Valued Neutrosophic Set [378,379]; Dual hesitant fuzzy sets [380,381]; Ambiguous Set[(b)] [382–384]; Local-Neutrosophic Set [385]; Support Neutrosophic Set [386]; (Belnap four-valued logic [387,388];) Turiyam Neutrosophic Set[(c)] [389–392] |
| 5 | 0 | Pentapartitioned Neutrosophic Set [393–395]; Triple-valued Neutrosophic Set [396–398] |
| 6 | 0 | Hexapartitioned Neutrosophic Set; Quadruple-Valued Neutrosophic Set [397,399] |
| 7 | 0 | Heptapartitioned Neutrosophic Set; Quintuple-Valued Neutrosophic Set [397,400, 401] |
| 8 | 0 | Octapartitioned Neutrosophic Set [402] |
| 9 | 0 | Nonapartitioned Neutrosophic Set [402] |
| $n$ | 0 | $n$-Refined Fuzzy Set [403,404]; ($n$-valued (Łukasiewicz) logic [405];) Multi-valued (Fuzzy) Sets [406]; MultiFuzzy Set [407] |
| $2n$ | 0 | $n$-Refined Intuitionistic Fuzzy Set [404]; Multi-Intuitionistic Fuzzy Set [407] |
| $3n$ | 0 | $n$-Refined Neutrosophic Set [404]; Multi-Neutrosophic Set [407,408] |
| 1 | 1 | Plithogenic Fuzzy Set [23,409,410] |
| 2 | 1 | Plithogenic Intuitionistic Fuzzy Set [411] |
| 3 | 1 | Plithogenic Neutrosophic Set [24,25,412] |
| 4 | 1 | Plithogenic Quadripartitioned Neutrosophic Set |
| 5 | 1 | Plithogenic Pentapartitioned Neutrosophic Set |
| 6 | 1 | Plithogenic Hexapartitioned Neutrosophic Set |
| 7 | 1 | Plithogenic Heptapartitioned Neutrosophic Set |
| 8 | 1 | Plithogenic Octapartitioned Neutrosophic Set |
| 9 | 1 | Plithogenic Nonapartitioned Neutrosophic Set |
| 1 | 2 | Double-valued Plithogenic Fuzzy Set [413] |
| 2 | 2 | Double-valued Plithogenic Intuitionistic Fuzzy Set [413] |
| 3 | 2 | Double-valued Plithogenic Neutrosophic Set [413] |

(a) It is widely recognized that the neutrosophic set generalizes the intuitionistic fuzzy set, the inconsistent intuitionistic fuzzy set (including picture fuzzy and ternary fuzzy sets), the Pythagorean fuzzy set, the spherical fuzzy set, and the $q$-rung orthopair fuzzy set; similarly, neutrosophication generalizes regret theory, grey system theory, and three-way decision theory [16].
(b) Ambiguous Sets are known to form a subclass of Quadripartitioned Neutrosophic Sets [8,377] as well as of Double-Valued Neutrosophic Sets [384].
(c) Turiyam Neutrosophic Sets are known to constitute a subclass of the existing Quadripartitioned Neutrosophic Sets [414].

## 3.2 m-Polar Plithogenic Set

The $m$-polar (multipolar) structure is used in various contexts (e.g. [415]) and can also be applied within Fuzzy, Neutrosophic, and Plithogenic frameworks. An $m$-polar plithogenic set models elements with m contrasting poles, appurtenance vectors, and value/pole contradiction-controlled aggregation for complex decision contexts. The $m$-polar plithogenic set generalizes the Multipolar Fuzzy Set [416, 417], the Multipolar Intuitionistic Fuzzy Set [418, 419], and the Multipolar Neutrosophic Set [420–423].

Table 3.4: Overview of Plithogenic Sets by the contradiction dimension $t$ in the Degree of Contradiction Function $pCF : Pv \times Pv \to [0,1]^t$.

| $t$ | DCF space $[0,1]^t$ | Interpretation of $pCF$ | Reduction / Containment | Typical use or example |
|---|---|---|---|---|
| 0 | $[0,1]^0 \equiv \{\mathbf{0}\}$ | No contradiction modeling (or a constant-zero DCF). Aggregation relies only on the Degree of Appurtenance Function (DAF) *pdf*. | Recovers classical multi-valued families: fuzzy ($s$=1), intuitionistic/vague ($s$=2), neutrosophic ($s$=3), etc. | When attribute values do not meaningfully conflict; e.g., ranking items using only performance scores without trade-off penalties. |
| 1 | $[0,1]$ | Single scalar contradiction per pair of values; typically used to attenuate *pdf* relative to a dominant value or context. | The "classical" plithogenic set. Operations weight/penalize memberships via $(1 - pCF)$ or related schemes. | One principal tension axis (e.g., laptop choice: *lightweight* vs *performance*; purchasing: *price* vs *quality*). |
| 2 | $[0,1]^2$ | Two-dimensional contradiction vector (e.g., *structural* vs *contextual* conflict). Requires a vector-to-scalar reducer (weighted sum, norm, or lexicographic rule) during aggregation. | Double-valued plithogenic variants; strictly more expressive than $t$=1 and reduces to $t$=1 by projecting/ignoring one component. | Modeling dual antagonisms (e.g., treatment planning: *efficacy* vs *side-effects* and *cost*; networking: *latency* vs *energy*). |

**Definition 3.2.1** (***m***-Polar Plithogenic Set)**.** Let $S$ be a universe and $P \subseteq S$ a domain of interest. Let $v$ be a (fixed) attribute with value set $P_v$. Let $\Pi = \{\pi_1, \ldots, \pi_m\}$ be a set of $m$ poles (for instance, $m = 2$ for "positive/negative", $m = 3$ for "positive/neutral/negative"). Fix integers $s \geq 1$ and $t \geq 0$.

An *m-polar plithogenic set* (abbrev. $m$PPS) on $P$ with respect to $v$ is a tuple

$$\mathrm{mPPS} := \big(P,\ v,\ P_v,\ \Pi,\ \mathrm{pdf}^{\Pi},\ \mathrm{pCF}_v,\ \mathrm{pCF}_{\Pi},\ \mathrm{Agg}\big),$$

where

- the *m-polar plithogenic degree of appurtenance function*

  $$\mathrm{pdf}^{\Pi} :\ P \times P_v \longrightarrow \big([0,1]^s\big)^m$$

  assigns to each pair $(x, \alpha) \in P \times P_v$ an $m$-tuple of $s$-component vectors,

  $$\mathrm{pdf}^{\Pi}(x,\alpha)\ =\ \big(\mu_1(x,\alpha), \ldots, \mu_m(x,\alpha)\big), \qquad \mu_r(x,\alpha) \in [0,1]^s\ (r = 1, \ldots, m),$$

  where each $\mu_r(x,\alpha)$ encodes the $s$ plithogenic components (e.g. $s = 1$ fuzzy grade, $s = 2$ intuitionistic pair, $s = 3$ neutrosophic triple) at pole $\pi_r$ for the value $\alpha$;

- $\mathrm{pCF}_v : P_v \times P_v \to [0,1]^t$ is the *value-level Degree of Contradiction Function (DCF)*, symmetric and reflexive:

  $$\mathrm{pCF}_v(\alpha,\alpha) = \mathbf{0}, \qquad \mathrm{pCF}_v(\alpha,\beta) = \mathrm{pCF}_v(\beta,\alpha) \quad (\forall\, \alpha, \beta \in P_v);$$

- $\mathrm{pCF}_{\Pi} : \Pi \times \Pi \to [0,1]^t$ is the *pole-level DCF*, also symmetric and reflexive:

  $$\mathrm{pCF}_{\Pi}(\pi,\pi) = \mathbf{0}, \qquad \mathrm{pCF}_{\Pi}(\pi,\pi') = \mathrm{pCF}_{\Pi}(\pi',\pi) \quad (\forall\, \pi, \pi' \in \Pi);$$

- Agg is a fixed *contradiction-aware reduction operator* that, for any chosen *dominant context* $(d, \delta) \in P_v \times \Pi$, produces an $s$-component aggregated degree

$$\mathrm{pdf}^{(d,\delta)}(x, \alpha) \;:=\; \mathrm{Agg}\Big(\mathrm{pdf}^{\Pi}(x, \alpha),\; \mathrm{pCF}_v(\alpha, d),\; \big(\mathrm{pCF}_{\Pi}(\pi_r, \delta)\big)_{r=1}^{m}\Big) \in [0,1]^s.$$

A canonical choice is the componentwise weighted mean based on *compatibility weights*

$$w_r(\alpha \mid d, \delta) \;:=\; \big(1 - \Phi_v(\mathrm{pCF}_v(\alpha, d))\big)\,\big(1 - \Phi_{\Pi}(\mathrm{pCF}_{\Pi}(\pi_r, \delta))\big),$$

where $\Phi_v, \Phi_{\Pi} : [0,1]^t \to [0,1]$ are monotone fusions of the $t$-dimensional contradiction vectors. Writing

$$\mu_r(x, \alpha) = \big(\mu_{r,1}(x, \alpha), \ldots, \mu_{r,s}(x, \alpha)\big),$$

the aggregated degree is then given componentwise by

$$\mathrm{pdf}_j^{(d,\delta)}(x, \alpha) := \frac{\displaystyle\sum_{r=1}^{m} w_r(\alpha \mid d, \delta)\, \mu_{r,j}(x, \alpha)}{\displaystyle\sum_{r=1}^{m} w_r(\alpha \mid d, \delta)}, \qquad j = 1, \ldots, s, \tag{3.1}$$

with the usual convention that $0/0 := 0$.

The triple $\big(\mathrm{pdf}^{\Pi}, \mathrm{pCF}_v, \mathrm{pCF}_{\Pi}\big)$ describes how each element is evaluated at multiple poles and how contradictions act both between attribute values and between poles. The operator Agg specifies how a dominant context $(d, \delta)$ attenuates or amplifies polar components to obtain a single $s$-component plithogenic degree when required.

**Remark 3.2.2.** If $s = 1$, then $\mathrm{pdf}^{\Pi}(x, \alpha) \in ([0,1]^1)^m \cong [0,1]^m$ and each pole carries a single fuzzy-grade component. For $s = 2$ and $s = 3$ one recovers, respectively, intuitionistic-type and neutrosophic-type $m$-polar structures within the same plithogenic framework.

Tables 3.5 and 3.6 present a selection of concepts that can be generalized within the framework of the Multipolar Plithogenic Set.

We now present concrete examples of the concept.

**Example 3.2.3** (Hiring under Multi-Polar Evaluation ($m = 3$))**.** Let the universe of candidates be $P = \{c_1, c_2, c_3\}$. Consider the attribute $v$ = experience_level with value set

$$P_v = \{\text{junior, mid, senior}\}.$$

Take the pole set

$$\Pi = \{\pi_1, \pi_2, \pi_3\} = \{\text{pro}, \text{neutral}, \text{contra}\}$$

so that $m = 3$. We work in the special case $s = 1$, hence we identify $([0,1]^1)^3 \cong [0,1]^3$ and write each appurtenance vector as a triple of real numbers in $[0,1]$.

The value–level DCF $\mathrm{pCF}_v : P_v \times P_v \to [0,1]$ and pole–level DCF $\mathrm{pCF}_{\Pi} : \Pi \times \Pi \to [0,1]$ are given (in matrix form) by which are symmetric and satisfy $\mathrm{pCF}_v(\alpha, \alpha) = 0$, $\mathrm{pCF}_\Pi(\pi, \pi) = 0$.

| $\mathrm{pCF}_v$ | junior | mid | senior |
|---|---|---|---|
| junior | 0 | 0.30 | 0.80 |
| mid | 0.30 | 0 | 0.40 |
| senior | 0.80 | 0.40 | 0 |

Table 3.5: Cases for $m \in \{1, 2, 3, ...\}$ (polar level), $t = 0$ (no DCF), and $s \in \{1, 2, 3, 4, 5\}$. Here $m$=1 = classical, $m$=2 = bipolar, $m$=3 = tripolar.

| $m$ | $t$ | $s$ | Representative name | Brief note (poles / comment) |
|---|---|---|---|---|
| 1 | 0 | 1 | Fuzzy Set | Single pole $\{\pi_+\}$. |
| 1 | 0 | 2 | Intuitionistic Fuzzy Set | Single pole; stores $(\mu, \nu)$. |
| 1 | 0 | 3 | Neutrosophic Set | Single pole; stores $(T, I, F)$. |
| 2 | 0 | 1 | Bipolar Fuzzy Set [348, 424] | Poles $\{\pi_+, \pi_-\}$; positive vs. negative degrees. |
| 2 | 0 | 2 | Bipolar Intuitionistic Fuzzy Set [425–427] (Bipolar vague sets [428, 429]) | Bipolar analogue of intuitionistic fuzzy (incl. bipolar vague). |
| 2 | 0 | 3 | Bipolar Neutrosophic Set [430–432] | Bipolar analogue of neutrosophic. |
| 2 | 0 | 3 | Bipolar Picture Fuzzy Set [433, 434]/Bipolar Hesitant Fuzzy Set [435, 436] | Bipolar analogue of picture / hesitant fuzzy. |
| 2 | 0 | 3 | Bipolar Spherical Fuzzy Set [437] | Bipolar analogue of spherical fuzzy. |
| 2 | 0 | 4 | Bipolar Quadripartitioned Neutrosophic Set [438–440] | Bipolar analogue of quadripartitioned neutrosophic. |
| 2 | 0 | 5 | Bipolar Pentapartitioned Neutrosophic Set [441, 442] | Bipolar analogue of pentapartitioned neutrosophic. |
| 3 | 0 | 1 | Tripolar Fuzzy Set [360–362, 443] | Poles $\{\pi_+, \pi_0, \pi_-\}$; positive/neutral/negative. |
| 3 | 0 | 2 | Tripolar Intuitionistic Fuzzy Set | Tripolar intuitionistic model. |
| 3 | 0 | 3 | Tripolar Neutrosophic Set [444, 445] | Tripolar neutrosophic model. |
| $m$ | 0 | 1 | $m$-polar Fuzzy Set [417, 446, 447] | Generic $m$ poles; fuzzy case. |
| $m$ | 0 | 2 | $m$-polar Intuitionistic Fuzzy Set [418, 419] | Generic $m$ poles; intuitionistic case. |
| $m$ | 0 | 3 | $m$-polar Neutrosophic Set [420, 421, 448] | Generic $m$ poles; neutrosophic case. |
| $m$ | 0 | 3 | $m$-polar Picture Fuzzy Set [449, 450]/$m$-polar Hesitant Fuzzy Set [451, 452] | Generic $m$ poles; picture / hesitant fuzzy case. |
| $m$ | 0 | 3 | $m$-polar Spherical Fuzzy Set [453, 454] | Generic $m$ poles; spherical fuzzy case. |
| $m$ | 0 | 4 | $m$-polar Quadripartitioned Neutrosophic Set [455, 456] | Generic $m$ poles; quadripartitioned neutrosophic case. |

| $\mathrm{pCF}_\Pi$ | pro | neutral | contra |
|---|---|---|---|
| pro | 0 | 0.20 | 1.00 |
| neutral | 0.20 | 0 | 0.20 |
| contra | 1.00 | 0.20 | 0 |

Define the $m$–polar appurtenance map

$$\mathrm{pdf}^\Pi : P \times P_v \longrightarrow ([0, 1]^1)^3 \cong [0, 1]^3$$

Table 3.6: Cases for $m$ (polar level), $t = 1$ (single DCF), and $s \in \{1, 2, 3, 4, 5\}$. Here $m$=1 = classical, $m$=2 = bipolar, $m$=3 = tripolar.

| $m$ | $t$ | $s$ | Representative name | Brief note (poles / DCF) |
|---|---|---|---|---|
| 1 | 1 | 1 | Plithogenic Fuzzy Set | Single pole; one value-level DCF modulates $\mu$. |
| 1 | 1 | 2 | Plithogenic Intuitionistic Fuzzy Set | Single pole; DCF attenuates $(\mu, \nu)$. |
| 1 | 1 | 3 | Plithogenic Neutrosophic Set | Single pole; DCF acts over $(T, I, F)$. |
| 2 | 1 | 1 | Bipolar Plithogenic Fuzzy Set | Bipolar poles; one value-level DCF; optional pole-level handling. |
| 2 | 1 | 2 | Bipolar Plithogenic Intuitionistic Set | Bipolar, DCF-aware intuitionistic. |
| 2 | 1 | 3 | Bipolar Plithogenic Neutrosophic Set | Bipolar, DCF-aware neutrosophic. |
| 3 | 1 | 1 | Tripolar Plithogenic Fuzzy Set | Tripolar poles; one value-level DCF; optional pole interactions. |
| 3 | 1 | 2 | Tripolar Plithogenic Intuitionistic Set | Tripolar, DCF-aware intuitionistic. |
| 3 | 1 | 3 | Tripolar Plithogenic Neutrosophic Set | Tripolar, DCF-aware neutrosophic. |

whose entries are written as $\mathrm{pdf}^{\Pi}(x, \alpha) = (\mu_{\mathrm{pro}}, \mu_{\mathrm{neutral}}, \mu_{\mathrm{contra}})$. For the three candidates we set

$$\begin{aligned} c_1 :\quad & \mathrm{pdf}^{\Pi}(c_1, \mathrm{junior}) = (0.60, 0.30, 0.10), \\ & \mathrm{pdf}^{\Pi}(c_1, \mathrm{mid}) = (0.70, 0.20, 0.10), \\ & \mathrm{pdf}^{\Pi}(c_1, \mathrm{senior}) = (0.40, 0.20, 0.40); \\ c_2 :\quad & \mathrm{pdf}^{\Pi}(c_2, \mathrm{junior}) = (0.30, 0.20, 0.50), \\ & \mathrm{pdf}^{\Pi}(c_2, \mathrm{mid}) = (0.50, 0.30, 0.20), \\ & \mathrm{pdf}^{\Pi}(c_2, \mathrm{senior}) = (0.80, 0.10, 0.10); \\ c_3 :\quad & \mathrm{pdf}^{\Pi}(c_3, \mathrm{junior}) = (0.20, 0.30, 0.50), \\ & \mathrm{pdf}^{\Pi}(c_3, \mathrm{mid}) = (0.40, 0.30, 0.30), \\ & \mathrm{pdf}^{\Pi}(c_3, \mathrm{senior}) = (0.60, 0.20, 0.20). \end{aligned}$$

Fix the dominant context

$$(d, \delta) = (\mathrm{senior}, \mathrm{pro}),$$

meaning that the decision maker is primarily interested in the value senior and the pole pro. For each $(\alpha, \pi_r) \in P_v \times \Pi$ we use the compatibility weights

$$w_r(\alpha \mid d, \delta) := \big(1 - \mathrm{pCF}_v(\alpha, d)\big)\,\big(1 - \mathrm{pCF}_{\Pi}(\pi_r, \delta)\big), \quad r = 1, 2, 3,$$

and aggregate by the normalized weighted mean

$$\mathrm{pdf}^{(d,\delta)}(x, \alpha) := \frac{\displaystyle\sum_{r=1}^{3} w_r(\alpha \mid d, \delta)\, \mu_r(x, \alpha)}{\displaystyle\sum_{r=1}^{3} w_r(\alpha \mid d, \delta)} \in [0, 1],$$

with the convention $0/0 := 0$.

We illustrate the computation for candidate $c_1$.

(1) Case $\alpha = \text{mid}$. We have

$$1 - \mathrm{pCF}_v(\text{mid}, \text{senior}) = 1 - 0.40 = 0.60,$$

and, against $\delta = \text{pro}$,

$$1 - \mathrm{pCF}_\Pi(\pi_1, \delta) = 1 - 0 = 1, \quad 1 - \mathrm{pCF}_\Pi(\pi_2, \delta) = 1 - 0.20 = 0.8, \quad 1 - \mathrm{pCF}_\Pi(\pi_3, \delta) = 1 - 1.00 = 0.$$

Thus the weight vector is

$$w(\text{mid} \mid d, \delta) = \big(0.60,\ 0.60 \times 0.8,\ 0.60 \times 0\big) = (0.60,\ 0.48,\ 0),$$

and

$$\sum_{r=1}^{3} w_r(\text{mid} \mid d, \delta) = 0.60 + 0.48 + 0 = 1.08.$$

Using $\mathrm{pdf}^{\Pi}(c_1, \text{mid}) = (0.70, 0.20, 0.10)$, we obtain

$$\begin{aligned}\mathrm{pdf}^{(d,\delta)}(c_1, \text{mid}) &= \frac{0.60 \cdot 0.70 + 0.48 \cdot 0.20 + 0 \cdot 0.10}{1.08} \\ &= \frac{0.42 + 0.096}{1.08} = \frac{0.516}{1.08} \approx 0.478.\end{aligned}$$

(2) Case $\alpha = \text{senior}$. Now

$$1 - \mathrm{pCF}_v(\text{senior}, \text{senior}) = 1 - 0 = 1,$$

so the weights are determined purely by the poles:

$$w(\text{senior} \mid d, \delta) = \big(1,\ 0.8,\ 0\big), \quad \sum_{r=1}^{3} w_r(\text{senior} \mid d, \delta) = 1 + 0.8 + 0 = 1.8.$$

Using $\mathrm{pdf}^{\Pi}(c_1, \text{senior}) = (0.40, 0.20, 0.40)$,

$$\begin{aligned}\mathrm{pdf}^{(d,\delta)}(c_1, \text{senior}) &= \frac{1 \cdot 0.40 + 0.8 \cdot 0.20 + 0 \cdot 0.40}{1.8} \\ &= \frac{0.40 + 0.16}{1.8} = \frac{0.56}{1.8} \approx 0.311.\end{aligned}$$

These computations make explicit how contradictions at both the value level ($\mathrm{pCF}_v$) and the pole level ($\mathrm{pCF}_\Pi$) jointly attenuate or emphasize the components of $\mathrm{pdf}^{\Pi}$ in order to produce a single plithogenic score $\mathrm{pdf}^{(d,\delta)}(x, \alpha)$ for each candidate–value pair.

**Example 3.2.4** (Vendor Selection with Four Poles ($m = 4$))**.** Vendor selection is the critical process where organizations evaluate suppliers and choose those meeting cost, quality, reliability, and compliance requirements (cf. [457]).

Let the universe of vendors be $P = \{v_1, v_2\}$. Consider the attribute $v = \text{logistics_objective}$ with value set

$$P_v = \{\text{low_cost, fast_delivery, high_reliability}\}.$$

We use four poles

$$\Pi = \{\text{benefit}, \text{neutral}, \text{risk}, \text{compliance}\},$$

so $m = 4$. Again we take $s = 1$, so $\text{pdf}^{\Pi} : P \times P_v \to [0,1]^4$.

The value–level DCF $\text{pCF}_v$ and pole–level DCF $\text{pCF}_{\Pi}$ are

| $\text{pCF}_v$ | low_cost | fast_delivery | high_reliability |
|---|---|---|---|
| low_cost | 0 | 0.50 | 0.60 |
| fast_delivery | 0.50 | 0 | 0.20 |
| high_reliability | 0.60 | 0.20 | 0 |

| $\text{pCF}_{\Pi}$ | benefit | neutral | risk | compliance |
|---|---|---|---|---|
| benefit | 0 | 0.20 | 0.90 | 0.10 |
| neutral | 0.20 | 0 | 0.20 | 0.20 |
| risk | 0.90 | 0.20 | 0 | 0.80 |
| compliance | 0.10 | 0.20 | 0.80 | 0 |

Both matrices are symmetric with zero diagonal entries, so they are valid DCFs.

For vendor $v_1$, the 4–polar appurtenance vectors $\text{pdf}^{\Pi}(v_1, \alpha) = (\mu_{\text{benefit}}, \mu_{\text{neutral}}, \mu_{\text{risk}}, \mu_{\text{compliance}})$ are chosen as

$$\text{pdf}^{\Pi}(v_1, \text{low_cost}) = (0.65, 0.15, 0.30, 0.40),$$

$$\text{pdf}^{\Pi}(v_1, \text{fast_delivery}) = (0.70, 0.10, 0.20, 0.50),$$

$$\text{pdf}^{\Pi}(v_1, \text{high_reliability}) = (0.60, 0.20, 0.10, 0.90).$$

Fix the dominant context

$$(d, \delta) = (\text{high_reliability}, \text{benefit}).$$

For each $(\alpha, \pi_r) \in P_v \times \Pi$ the weights are again

$$w_r(\alpha \mid d, \delta) := \big(1 - \text{pCF}_v(\alpha, d)\big)\,\big(1 - \text{pCF}_{\Pi}(\pi_r, \delta)\big), \quad r = 1, \ldots, 4,$$

and the scalarized membership is given by

$$\text{pdf}^{(d,\delta)}(x, \alpha) := \frac{\sum_{r=1}^{4} w_r(\alpha \mid d, \delta)\, \mu_r(x, \alpha)}{\sum_{r=1}^{4} w_r(\alpha \mid d, \delta)} \in [0,1].$$

First note that, for the fixed pole $\delta = \text{benefit}$,

$$1 - \text{pCF}_{\Pi}(\cdot, \delta) = \big(1,\ 0.8,\ 0.1,\ 0.9\big)$$

corresponding to $\{\text{benefit}, \text{neutral}, \text{risk}, \text{compliance}\}$.

(1) Case $\alpha = \text{fast_delivery}$. Here

$$1 - \text{pCF}_v(\text{fast_delivery}, \text{high_reliability}) = 1 - 0.20 = 0.80.$$

Thus

$$w(\text{fast_delivery} \mid d, \delta) = 0.80 \cdot (1,\ 0.8,\ 0.1,\ 0.9) = (0.80,\ 0.64,\ 0.08,\ 0.72),$$

and

$$\sum_{r=1}^{4} w_r(\text{fast_delivery} \mid d, \delta) = 0.80 + 0.64 + 0.08 + 0.72 = 2.24.$$

With $\mathrm{pdf}^{\Pi}(v_1, \text{fast_delivery}) = (0.70, 0.10, 0.20, 0.50)$, the numerator is

$$\begin{aligned} \text{num} &= 0.80 \cdot 0.70 + 0.64 \cdot 0.10 + 0.08 \cdot 0.20 + 0.72 \cdot 0.50 \\ &= 0.56 + 0.064 + 0.016 + 0.36 = 1.000, \end{aligned}$$

so

$$\mathrm{pdf}^{(d,\delta)}(v_1, \text{fast_delivery}) = \frac{1.000}{2.24} \approx 0.4464.$$

(2) Case $\alpha = \text{high_reliability}$. Now

$$1 - \mathrm{pCF}_v(\text{high_reliability}, \text{high_reliability}) = 1 - 0 = 1,$$

so the weights coincide with the pole–compatibility vector:

$$w(\text{high_reliability} \mid d, \delta) = (1,\ 0.8,\ 0.1,\ 0.9), \quad \sum_{r=1}^{4} w_r(\text{high_reliability} \mid d, \delta) = 2.8.$$

With $\mathrm{pdf}^{\Pi}(v_1, \text{high_reliability}) = (0.60, 0.20, 0.10, 0.90)$,

$$\begin{aligned} \text{num} &= 1 \cdot 0.60 + 0.8 \cdot 0.20 + 0.1 \cdot 0.10 + 0.9 \cdot 0.90 \\ &= 0.60 + 0.16 + 0.01 + 0.81 = 1.58, \end{aligned}$$

and hence

$$\mathrm{pdf}^{(d,\delta)}(v_1, \text{high_reliability}) = \frac{1.58}{2.8} \approx 0.5643.$$

These values show how the four–pole structure, together with the contradiction degrees $\mathrm{pCF}_v$ and $\mathrm{pCF}_{\Pi}$, allows the decision maker to emphasize "benefit" in a high_reliability–oriented context, while still accounting for neutrality, risk, and compliance through the weights $w_r(\alpha \mid d, \delta)$.

**Example 3.2.5** (Smart-City Intersection Design ($m = 3$)). A smart-city intersection integrates sensors, adaptive signals, and connected vehicles to optimize traffic flow, safety, emissions, and pedestrian experience citywide (cf. [458]).

Let $P = \{\mathrm{A}, \mathrm{B}\}$ be two competing design options for a city intersection. Consider the attribute $v = \text{objective}$ with value set

$$P_v = \{\text{safety, throughput, cost}\}.$$

We take the three poles

$$\Pi = \{\text{pro}, \text{neutral}, \text{contra}\},$$

so that $m = 3$, and again work with $s = 1$, so that $\mathrm{pdf}^{\Pi}(x, \alpha) \in [0, 1]^3$.

The value–level contradiction degrees $\mathrm{pCF}_v$ and pole–level contradiction degrees $\mathrm{pCF}_{\Pi}$ are defined by

| $\mathrm{pCF}_v$ | safety | throughput | cost |
|---|---|---|---|
| safety | 0 | 0.50 | 0.30 |
| throughput | 0.50 | 0 | 0.40 |
| cost | 0.30 | 0.40 | 0 |

| $\mathrm{pCF}_\Pi$ | pro | neutral | contra |
|---|---|---|---|
| pro | 0 | 0.20 | 1.00 |
| neutral | 0.20 | 0 | 0.20 |
| contra | 1.00 | 0.20 | 0 |

which are symmetric with zero diagonal entries, hence valid DCFs.

For each design option $X \in \{\mathrm{A}, \mathrm{B}\}$ and each $\alpha \in P_v$, the triple

$$\mathrm{pdf}^\Pi(X, \alpha) = (\mu_{\mathrm{pro}}(X, \alpha), \mu_{\mathrm{neutral}}(X, \alpha), \mu_{\mathrm{contra}}(X, \alpha))$$

encodes the 3–polar evaluation of $X$ with respect to objective $\alpha$. We choose

$$\begin{aligned}
\mathrm{A}: \quad & \mathrm{pdf}^\Pi(\mathrm{A}, \mathrm{safety}) = (0.85, 0.10, 0.05), \\
& \mathrm{pdf}^\Pi(\mathrm{A}, \mathrm{throughput}) = (0.60, 0.20, 0.20), \\
& \mathrm{pdf}^\Pi(\mathrm{A}, \mathrm{cost}) = (0.55, 0.25, 0.20); \\
\mathrm{B}: \quad & \mathrm{pdf}^\Pi(\mathrm{B}, \mathrm{safety}) = (0.70, 0.20, 0.10), \\
& \mathrm{pdf}^\Pi(\mathrm{B}, \mathrm{throughput}) = (0.80, 0.10, 0.10), \\
& \mathrm{pdf}^\Pi(\mathrm{B}, \mathrm{cost}) = (0.40, 0.30, 0.30).
\end{aligned}$$

We again use the weighted mean reduction

$$\mathrm{pdf}^{(d,\delta)}(x, \alpha) := \frac{\sum_{r=1}^{3} w_r(\alpha \mid d, \delta)\, \mu_r(x, \alpha)}{\sum_{r=1}^{3} w_r(\alpha \mid d, \delta)},$$

where

$$w_r(\alpha \mid d, \delta) := \big(1 - \mathrm{pCF}_v(\alpha, d)\big)\, \big(1 - \mathrm{pCF}_\Pi(\pi_r, \delta)\big).$$

We fix the dominant context

$$(d, \delta) = (\mathrm{safety}, \mathrm{pro}),$$

representing a "safety-first, pro" viewpoint. Then

$$1 - \mathrm{pCF}_\Pi(\cdot, \delta) = \big(1,\ 0.8,\ 0\big)$$

for the poles $\{\mathrm{pro}, \mathrm{neutral}, \mathrm{contra}\}$.

(1) Design B, $\alpha = \mathrm{throughput}$. We have

$$1 - \mathrm{pCF}_v(\mathrm{throughput}, \mathrm{safety}) = 1 - 0.50 = 0.50,$$

so

$$w(\mathrm{throughput} \mid d, \delta) = 0.50 \cdot (1,\ 0.8,\ 0) = (0.50,\ 0.40,\ 0),$$

with

$$\sum_{r=1}^{3} w_r(\mathrm{throughput} \mid d, \delta) = 0.50 + 0.40 + 0 = 0.90.$$

Using $\mathrm{pdf}^\Pi(\mathrm{B}, \mathrm{throughput}) = (0.80, 0.10, 0.10)$,

$$\begin{aligned}
\mathrm{pdf}^{(d,\delta)}(\mathrm{B}, \mathrm{throughput}) &= \frac{0.50 \cdot 0.80 + 0.40 \cdot 0.10 + 0 \cdot 0.10}{0.90} \\
&= \frac{0.40 + 0.04}{0.90} = \frac{0.44}{0.90} \approx 0.489.
\end{aligned}$$

(2) Design A, $\alpha =$ safety. Here

$$1 - \mathrm{pCF}_v(\mathrm{safety}, \mathrm{safety}) = 1 - 0 = 1,$$

so the weights are simply

$$w(\mathrm{safety} \mid d, \delta) = (1,\ 0.8,\ 0), \qquad \sum_{r=1}^{3} w_r(\mathrm{safety} \mid d, \delta) = 1.8.$$

Using $\mathrm{pdf}^{\Pi}(\mathrm{A}, \mathrm{safety}) = (0.85, 0.10, 0.05)$,

$$\begin{aligned} \mathrm{pdf}^{(d,\delta)}(\mathrm{A}, \mathrm{safety}) &= \frac{1 \cdot 0.85 + 0.8 \cdot 0.10 + 0 \cdot 0.05}{1.8} \\ &= \frac{0.85 + 0.08}{1.8} = \frac{0.93}{1.8} \approx 0.517. \end{aligned}$$

These scalarized scores $\mathrm{pdf}^{(d,\delta)}(\mathrm{B}, \mathrm{throughput})$ and $\mathrm{pdf}^{(d,\delta)}(\mathrm{A}, \mathrm{safety})$ explicitly show how a "safety-first, pro" context downweights contradictory poles and objectives, while still preserving the full multi-polar information encoded in $\mathrm{pdf}^{\Pi}$. The $m$–polar plithogenic machinery thus provides a mathematically precise, contradiction–aware comparison of intersection designs in terms of safety, throughput, and cost.

## 3.3 Complex Plithogenic Set (CPS)

A complex plithogenic set uses complex-valued memberships and contradiction functions to aggregate attributes under context-sensitive multidimensional uncertainty and conflict-aware reasoning [459].

**Definition 3.3.1** (Complex Plithogenic Set (CPS))**.** [459] Let $P$ be a nonempty set (the universe under study), $v$ be an attribute with a nonempty finite set of values $P_v$, and let $s \in \mathbb{N}$ and $t \in \mathbb{N}_0$ denote, respectively, the number of appurtenance components and of contradiction channels.

**(i) Complex degrees of appurtenance.** Let $\mathbb{D} := \{z \in \mathbb{C} : |z| \le 1\}$. A *complex degree of appurtenance function (CDAF)* is

$$\mathrm{cpdf}:\ P \times P_v \longrightarrow \mathbb{D}^s, \qquad \mathrm{cpdf}(x, \alpha) = \big(\gamma_1(x, \alpha), \ldots, \gamma_s(x, \alpha)\big),$$

with each component represented in amplitude–phase form $\gamma_i(x, \alpha) = p_i(x, \alpha)\, e^{\mathrm{i}\varphi_i(x,\alpha)}$ where $p_i(x, \alpha) \in [0, 1]$ and $\varphi_i(x, \alpha) \in [0, 2\pi)$.

**(ii) Degrees of contradiction (DCF).** For each $j = 1, \ldots, t$ let

$$\mathrm{pCF}_j:\ P_v \times P_v \longrightarrow [0, 1]$$

be *symmetric* and *reflexive*, i.e., $\mathrm{pCF}_j(a, a) = 0$ and $\mathrm{pCF}_j(a, b) = \mathrm{pCF}_j(b, a)$ for all $a, b \in P_v$.

**(iii) Fusion of contradictions and compatibility weights.** Fix a *contradiction fusion* $\Phi : [0, 1]^t \to [0, 1]$ that is symmetric, monotone (componentwise), and satisfies $\Phi(0, \ldots, 0) = 0$. (If $t = 0$, set $\Phi(\cdot) \equiv 0$.) Given a *dominant value* $\delta \in P_v$, define the compatibility weight

$$w(a \mid \delta) \ :=\ 1 - \Phi\big(\mathrm{pCF}_1(a, \delta), \ldots, \mathrm{pCF}_t(a, \delta)\big)\ \in [0, 1] \qquad (a \in P_v).$$

**(iv) Complex plithogenic aggregation (relative to $\delta$).** For each $x \in P$, define the *$\delta$–relative complex plithogenic degree* $\Gamma^{(\delta)}(x) \in \mathbb{D}^s$ componentwise by the weighted complex mean

$$\Gamma_i^{(\delta)}(x) := \begin{cases} \dfrac{\sum_{a\in P_v} w(a \mid \delta)\, \gamma_i(x,a)}{\sum_{a\in P_v} w(a \mid \delta)}, & \text{if } \sum_{a\in P_v} w(a \mid \delta) > 0, \\ 0, & \text{otherwise,} \end{cases} \qquad i = 1, \dots, s.$$

The *Complex Plithogenic Set (CPS)* associated with $(P, v, P_v)$, $s$, $t$, cpdf, $\{\mathrm{pCF}_j\}$, and $\Phi$ is the tuple

$$\mathrm{CPS} := \big(P,\ v,\ P_v,\ s,\ t,\ \mathrm{cpdf},\ \{\mathrm{pCF}_j\}_{j=1}^t,\ \Phi,\ \Gamma^{(\delta)}\big),$$

where $\Gamma^{(\delta)}$ supplies the contradiction-aware complex membership (as an $s$–vector) relative to the dominant value $\delta \in P_v$.

Table 3.7 presents the summary of the Complex Plithogenic Set (CPS).

Table 3.7: Summary of Complex Plithogenic Set (CPS) cases for $t \in \{0, 1\}$ and $s \in \{1, 2, 3\}$. Here $\Gamma^{(\delta)}$ is the contradiction-aware complex membership (Def. 3.3.1), $\gamma_i \in \mathbb{D}$ are CDAF components, and $w(a \mid \delta) = 1 - \Phi(\mathrm{pCF}_1(a, \delta), \dots, \mathrm{pCF}_t(a, \delta))$. For $t = 1$ we take $\Phi(z) = z$; for $t = 0$ we set $\Phi \equiv 0$.

| $s$ | $t$ | Specification (weights and aggregation) / Interpretation |
|---|---|---|
| *No contradiction channel* ($t = 0$): $\Phi \equiv 0$ so $w(a \mid \delta) = 1$ for all $a \in P_v$ (independent of $\delta$). | | |
| 1 | 0 | $\Gamma_1^{(\delta)}(x) = \dfrac{1}{\lvert P_v\rvert} \sum_{a\in P_v} \gamma_1(x, a)$ (uniform complex mean). Single-component CPS (complex fuzzy set [460–462]–like). |
| 2 | 0 | $\Gamma^{(\delta)}(x) \in \mathbb{D}^2$ with each component the uniform complex mean over $P_v$. Two-component CPS (complex intuitionistic fuzzy set [463–465] and complex vague set [466, 467]–like). |
| 3 | 0 | $\Gamma^{(\delta)}(x) \in \mathbb{D}^3$ with componentwise uniform complex means. Three-component CPS (complex neutrosophic set [468–470],Complex picture fuzzy set [471, 472], Complex Hesitant fuzzy [473–475], and Complex Spherical fuzzy [476, 477]-like). |
| 4 | 0 | $\Gamma^{(\delta)}(x) \in \mathbb{D}^4$ with componentwise uniform complex means. Four-component CPS (complex quadripartitioned neutrosophic set [478, 479]-like). |
| 5 | 0 | $\Gamma^{(\delta)}(x) \in \mathbb{D}^5$ with componentwise uniform complex means. Five-component CPS (complex pentapartitioned neutrosophic set [480, 481]-like). |
| *Single contradiction channel* ($t = 1$): $\Phi(z) = z$, so $w(a \mid \delta) = 1 - \mathrm{pCF}(a, \delta)$. | | |
| 1 | 1 | $\Gamma_1^{(\delta)}(x) = \dfrac{\sum_{a\in P_v} \big(1 - \mathrm{pCF}(a, \delta)\big)\, \gamma_1(x, a)}{\sum_{a\in P_v} \big(1 - \mathrm{pCF}(a, \delta)\big)}$. Single-component CPS with DCF-based reweighting (complex plithogenic fuzzy–like). |
| 2 | 1 | Same weighting; $\Gamma^{(\delta)}(x) \in \mathbb{D}^2$ computed componentwise by the above weighted complex mean. |
| 3 | 1 | Same weighting; $\Gamma^{(\delta)}(x) \in \mathbb{D}^3$ computed componentwise by the above weighted complex mean. |

*Notes.* (i) If all phases vanish ($\varphi_i \equiv 0$), the table reduces to the real-valued plithogenic counterparts. (ii) For $t = 0$, $\Gamma^{(\delta)}$ does not depend on $\delta$; for $t = 1$, larger $\mathrm{pCF}(a, \delta)$ downweights the contribution of value $a$.

We now present concrete examples of the concept.

**Example 3.3.2** (CPS in Smart-Grid EV Charging (tariff-aware dispatch))**.** Smart-grid EV charging coordinates electric vehicle charging with grid signals, optimizing energy use, reducing costs, balancing demand and renewables integration (cf. [482]).

Consider EV charging decisions for one car $x = \mathrm{EV}_1$. Universe $P = \{\mathrm{EV}_1\}$; attribute $v$ = tariff period with values $P_v = \{\text{offpeak}, \text{shoulder}, \text{peak}\}$. Take $s = 2$ complex components (grid benefit, stability) and $t = 2$ contradiction channels (price, transformer load). Let the dominant value be $\delta = \text{peak}$. For the DCFs (symmetric, reflexive) we only need the pairs $(a, \delta)$:

$$\begin{aligned}
&\mathrm{pCF}_1(\text{offpeak}, \text{peak}) = 0.8, \quad \mathrm{pCF}_1(\text{shoulder}, \text{peak}) = 0.4, \quad \mathrm{pCF}_1(\text{peak}, \text{peak}) = 0,\\
&\mathrm{pCF}_2(\text{offpeak}, \text{peak}) = 0.5, \quad \mathrm{pCF}_2(\text{shoulder}, \text{peak}) = 0.2, \quad \mathrm{pCF}_2(\text{peak}, \text{peak}) = 0.
\end{aligned}$$

Fuse contradictions by the mean $\Phi(c_1, c_2) = \frac{c_1+c_2}{2}$, hence the compatibility weights

$$w(a \mid \delta) = 1 - \Phi\big(\mathrm{pCF}_1(a, \delta), \mathrm{pCF}_2(a, \delta)\big) \Rightarrow \begin{cases} w(\text{offpeak} \mid \text{peak}) = 1 - \frac{0.8+0.5}{2} = 0.35,\\ w(\text{shoulder} \mid \text{peak}) = 1 - \frac{0.4+0.2}{2} = 0.70,\\ w(\text{peak} \mid \text{peak}) = 1. \end{cases}$$

Complex degrees (amplitude–phase): for component 1 (grid benefit) and 2 (stability),

$$\begin{aligned}
&\gamma^{(1)}(\mathrm{EV}_1, \text{offpeak}) = 0.60\, e^{\mathrm{i}\, 0^\circ} = (0.600 + 0.000\, \mathrm{i}),\\
&\gamma^{(1)}(\mathrm{EV}_1, \text{shoulder}) = 0.80\, e^{\mathrm{i}\, 30^\circ} = (0.800 \cos 30^\circ + 0.800 \sin 30^\circ\, \mathrm{i}) = (0.693 + 0.400\, \mathrm{i}),\\
&\gamma^{(1)}(\mathrm{EV}_1, \text{peak}) = 0.40\, e^{\mathrm{i}\, 60^\circ} = (0.200 + 0.346\, \mathrm{i});
\end{aligned}$$

$$\begin{aligned}
&\gamma^{(2)}(\mathrm{EV}_1, \text{offpeak}) = 0.50\, e^{\mathrm{i}\, 90^\circ} = (0.000 + 0.500\, \mathrm{i}),\\
&\gamma^{(2)}(\mathrm{EV}_1, \text{shoulder}) = 0.70\, e^{\mathrm{i}\, 60^\circ} = (0.350 + 0.606\, \mathrm{i}),\\
&\gamma^{(2)}(\mathrm{EV}_1, \text{peak}) = 0.30\, e^{\mathrm{i}\, 90^\circ} = (0.000 + 0.300\, \mathrm{i}).
\end{aligned}$$

Weighted complex mean (denominator $D = 0.35 + 0.70 + 1 = 2.05$):

$$\begin{aligned}
\mathrm{Num}^{(1)} &= 0.35(0.600 + 0.000\mathrm{i}) + 0.70(0.693 + 0.400\mathrm{i}) + 1(0.200 + 0.346\mathrm{i})\\
&= (0.8950 + 0.6264\, \mathrm{i}),\\
\Gamma_1^{(\text{peak})}(\mathrm{EV}_1) &= \frac{\mathrm{Num}^{(1)}}{D} = (0.2373 + 0.4778\, \mathrm{i}) = 0.5329\, e^{\mathrm{i}\, 34.99^\circ};\\
\mathrm{Num}^{(2)} &= 0.35(0.000 + 0.500\mathrm{i}) + 0.70(0.350 + 0.606\mathrm{i}) + 1(0.000 + 0.300\mathrm{i})\\
&= (0.2450 + 0.8994\, \mathrm{i}),\\
\Gamma_2^{(\text{peak})}(\mathrm{EV}_1) &= \frac{\mathrm{Num}^{(2)}}{D} = (0.1195 + 0.4384\, \mathrm{i}) = 0.4547\, e^{\mathrm{i}\, 74.83^\circ}.
\end{aligned}$$

Thus the CPS yields a contradiction-aware complex membership vector

$$\Gamma^{(\text{peak})}(\mathrm{EV}_1) = (0.5329 e^{\mathrm{i}35.0^\circ},\ 0.4547 e^{\mathrm{i}74.8^\circ})$$

.

**Example 3.3.3** (CPS in Medical Imaging Triage (CT/MRI/Ultrasound fusion))**.** Medical image triage prioritizes radiology or diagnostic images for review, flagging urgent abnormalities using automated or AI-assisted analysis to clinicians [483].

Patient $x = \mathrm{A}$, attribute $v$ = imaging modality with $P_v = \{\mathrm{CT}, \mathrm{US}, \mathrm{MRI}\}$. Take one complex component ($s = 1$) representing *diagnostic confidence*, and $t = 2$ DCF channels: radiation risk vs. diagnostic yield. Dominant value $\delta = \mathrm{MRI}$. Let

$$\begin{aligned}
&\mathrm{pCF}_1(\mathrm{CT}, \mathrm{MRI}) = 0.9, \quad \mathrm{pCF}_1(\mathrm{US}, \mathrm{MRI}) = 0.0,\\
&\mathrm{pCF}_2(\mathrm{CT}, \mathrm{MRI}) = 0.2, \quad \mathrm{pCF}_2(\mathrm{US}, \mathrm{MRI}) = 0.6,
\end{aligned} \qquad \Phi(c_1, c_2) = \tfrac{c_1+c_2}{2}.$$

Weights:

$$\begin{aligned}
&w(\mathrm{CT} \mid \mathrm{MRI}) = 1 - \tfrac{0.9+0.2}{2} = 0.45, \quad w(\mathrm{US} \mid \mathrm{MRI}) = 1 - \tfrac{0.0+0.6}{2} = 0.70,\\
&w(\mathrm{MRI} \mid \mathrm{MRI}) = 1, \qquad D = 0.45 + 0.70 + 1 = 2.15.
\end{aligned}$$

CDAF (amplitude–phase in degrees):

$$\gamma(\mathrm{A},\mathrm{CT}) = 0.65\,e^{\mathrm{i}\,10^\circ} = (0.640 + 0.113\,\mathrm{i}), \quad \gamma(\mathrm{A},\mathrm{US}) = 0.50\,e^{\mathrm{i}\,70^\circ} = (0.171 + 0.470\,\mathrm{i}),$$

$$\gamma(\mathrm{A},\mathrm{MRI}) = 0.85\,e^{\mathrm{i}\,20^\circ} = (0.798 + 0.291\,\mathrm{i}).$$

Weighted sum and aggregation:

$$\begin{aligned} \mathrm{Num} &= 0.45(0.640 + 0.113\mathrm{i}) + 0.70(0.171 + 0.470\mathrm{i}) + 1(0.798 + 0.291\mathrm{i}) \\ &= (1.2066 + 0.6700\,\mathrm{i}), \\ \Gamma^{(\mathrm{MRI})}(\mathrm{A}) = \frac{\mathrm{Num}}{D} &= (0.5612 + 0.3118\,\mathrm{i}) = 0.6420\,e^{\mathrm{i}\,29.06^\circ}. \end{aligned}$$

Hence MRI-dominant fusion produces a high complex confidence with phase $\approx 29^\circ$ (moderate uncertainty skew).

**Example 3.3.4** (CPS in E-Commerce Multi-Tier Product Positioning)**.** Product $x$ = LaptopA, attribute $v$ = tier with $P_v = \{\text{budget}, \text{balanced}, \text{premium}\}$. Use $s = 2$ complex components: (1) user sentiment, (2) delivery satisfaction. Let $t = 2$ DCF channels (price gap, feature gap), dominant tier $\delta$ = balanced. Contradictions (only pairs to $\delta$):

$$\begin{aligned} \mathrm{pCF}_1(\text{budget}, \text{balanced}) &= 0.4, & \mathrm{pCF}_1(\text{premium}, \text{balanced}) &= 0.5, \\ \mathrm{pCF}_2(\text{budget}, \text{balanced}) &= 0.6, & \mathrm{pCF}_2(\text{premium}, \text{balanced}) &= 0.2, \end{aligned} \qquad \Phi(c_1, c_2) = \tfrac{c_1 + c_2}{2}.$$

Weights:

$$w(\text{budget} \mid \text{balanced}) = 1 - \tfrac{0.4+0.6}{2} = 0.50,$$

$$w(\text{premium} \mid \text{balanced}) = 1 - \tfrac{0.5+0.2}{2} = 0.65,$$

$$w(\text{balanced} \mid \text{balanced}) = 1, \; D = 2.15.$$

CDAFs (amplitude–phase; rectangular shown to three decimals):

$$\begin{aligned} \gamma^{(1)}(\text{LaptopA}, \text{budget}) &= 0.55\,e^{\mathrm{i}15^\circ} = (0.532 + 0.142\,\mathrm{i}), \\ \gamma^{(1)}(\text{LaptopA}, \text{balanced}) &= 0.75\,e^{\mathrm{i}25^\circ} = (0.680 + 0.317\,\mathrm{i}), \\ \gamma^{(1)}(\text{LaptopA}, \text{premium}) &= 0.70\,e^{\mathrm{i}35^\circ} = (0.574 + 0.402\,\mathrm{i}); \end{aligned}$$

$$\begin{aligned} \gamma^{(2)}(\text{LaptopA}, \text{budget}) &= 0.60\,e^{\mathrm{i}80^\circ} = (0.104 + 0.591\,\mathrm{i}), \\ \gamma^{(2)}(\text{LaptopA}, \text{balanced}) &= 0.65\,e^{\mathrm{i}50^\circ} = (0.418 + 0.498\,\mathrm{i}), \\ \gamma^{(2)}(\text{LaptopA}, \text{premium}) &= 0.55\,e^{\mathrm{i}40^\circ} = (0.421 + 0.354\,\mathrm{i}). \end{aligned}$$

Weighted complex means (piecewise numerators shown; then divide by $D = 2.15$):

$$\begin{aligned} \mathrm{Num}^{(1)} &= 0.50(0.532 + 0.142\mathrm{i}) + 1(0.680 + 0.317\mathrm{i}) + 0.65(0.574 + 0.402\mathrm{i}) \\ &= (1.3181 + 0.6491\,\mathrm{i}), \\ \Gamma_1^{(\text{balanced})}(\text{LaptopA}) &= (1.3181 + 0.6491\,\mathrm{i})/2.15 = (0.6131 + 0.3019\,\mathrm{i}) \\ &= 0.6834\,e^{\mathrm{i}\,26.22^\circ}; \\ \mathrm{Num}^{(2)} &= 0.50(0.104 + 0.591\mathrm{i}) + 1(0.418 + 0.498\mathrm{i}) + 0.65(0.421 + 0.354\mathrm{i}) \\ &= (0.7438 + 1.0232\,\mathrm{i}), \\ \Gamma_2^{(\text{balanced})}(\text{LaptopA}) &= (0.7438 + 1.0232\,\mathrm{i})/2.15 = (0.3459 + 0.4759\,\mathrm{i}) \\ &= 0.5883\,e^{\mathrm{i}\,53.99^\circ}. \end{aligned}$$

So the CPS reports a balanced-tier, contradiction-aware complex appraisal vector $\Gamma^{(\text{balanced})}(\text{LaptopA}) = (0.6834e^{\mathrm{i}26.22^\circ}, 0.5883e^{\mathrm{i}53.99^\circ})$.

## 3.4 SuperHyperPlithogenic Set

A $(m, n)$-SuperhyperPlithogenic Set assigns $m$-level parameter-subsets and their attribute values to $n$-level fuzzy-contradiction degree-sets, capturing multi-faceted membership and inter-attribute conflicts and uncertainty patterns [484–486]. The definition of the $(m, n)$-SuperhyperPlithogenic Set is presented below.

**Definition 3.4.1** ($(m, n)$-SuperhyperPlithogenic Set)**.** [485] Let $X$ be a nonempty set and let $V = \{v_1, \ldots, v_k\}$ be a finite set of attributes. For each $v \in V$, let $P_v$ be the set of its possible values. Fix positive integers $m, n$ and positive dimensions $s, t$. Define the $m$-th nested powerset [81, 487] of $X$ by

$$\mathcal{P}^0(X) = X, \quad \mathcal{P}^r(X) = \mathcal{P}\big(\mathcal{P}^{r-1}(X)\big) \quad (r \geq 1),$$

and similarly $\mathcal{P}^n\big([0,1]^s\big)$ for the $s$-dimensional unit cube. An $(m, n)$-*SuperhyperPlithogenic Set* over $X$ is the quintuple

$$SHP^{(m,n)} = \Big(\mathcal{P}^m(X),\ V,\ \{P_v\}_{v\in V},\ \{\tilde{pdf}_v^{(m,n)}\}_{v\in V},\ pCF^{(m,n)}\Big),$$

where

(i) $\mathcal{P}^m(X)$ is the domain of "super-elements" of level $m$.

(ii) For each $v \in V$, $P_v$ is the finite set of its values.

(iii) The *Hyper Degree of Appurtenance Function*

$$\tilde{pdf}_v^{(m,n)} : \mathcal{P}^m(X) \times P_v \longrightarrow \mathcal{P}^n\big([0,1]^s\big)$$

assigns to each $(A, a)$ with $A \in \mathcal{P}^m(X)$ and $a \in P_v$ a nonempty subset $\tilde{pdf}_v^{(m,n)}(A, a) \subseteq [0, 1]^s$ representing all possible membership-degree vectors of dimension $s$.

(iv) The *Degree of Contradiction Function*

$$pCF^{(m,n)} : \Big(\bigcup_{v\in V} P_v\Big) \times \Big(\bigcup_{v\in V} P_v\Big) \longrightarrow [0,1]^t$$

satisfies for all $a, b$:

(a) $pCF^{(m,n)}(a, a) = 0$ (reflexivity),

(b) $pCF^{(m,n)}(a, b) = pCF^{(m,n)}(b, a)$ (symmetry).

Tables 3.8 and 3.9 present the associated concepts related to $(m, n)$-SuperhyperPlithogenic Sets.

We now present concrete examples of the concept.

**Example 3.4.2** ((m,n)=(2,1): Hospital "team-of-teams" triage with hyper–neutrosophic memberships)**.** Let $X = \{\mathrm{ER}, \mathrm{ICU}, \mathrm{OR}, \mathrm{RAD}\}$ be hospital units. Work on the superlevel $\mathcal{P}^2(X)$; consider the super–element

$$A = \big\{\{\mathrm{ER}, \mathrm{ICU}\},\ \{\mathrm{OR}, \mathrm{RAD}\}\big\} \in \mathcal{P}^2(X).$$

Use a single attribute $v$ ="clinical risk bracket" with values $P_v = \{\mathrm{low}, \mathrm{moderate}, \mathrm{high}\}$. Choose $s = 3$ (neutrosophic $(T, I, F)$), $n = 1$ (each membership is a finite subset of $[0, 1]^3$), and $t = 1$.

Table 3.8: Specializations of $(m, n)$-SuperhyperPlithogenic Sets

| **Parameter choice** | **Domain level** | **Membership / contradiction structure** | **Resulting model** |
|---|---|---|---|
| $(0, 0)$-SuperhyperPlithogenic Set | $\mathcal{P}^0(X) = X$ (no super-level) | Single plithogenic DAF $pdf : X \times Pv \to [0, 1]^s$ and DCF $pCF : Pv \times Pv \to [0, 1]^t$ | Plithogenic Set |
| $(0, n)$-SuperhyperPlithogenic Set | $\mathcal{P}^0(X) = X$ (no super-level) | DAF takes values in $\mathcal{P}^n([0, 1]^s)$: plithogenic memberships are hyper/nested | HyperPlithogenic Set (cf. [488–490]) |
| $(m, 0)$-SuperhyperPlithogenic Set | $\mathcal{P}^m(X)$ (super / hyper-level on the universe) | Usual plithogenic DAF/DCF on super-elements (no hyper–membership nesting) | SuperPlithogenic Set (cf. [491]) |

Table 3.9: HyperPlithogenic Set specializations by number of DAF components

| **Model** | **DAF dim.** $s$ | **Membership / contradiction description** | **Resulting structure** |
|---|---|---|---|
| Hyper–Plithogenic Fuzzy Set | $s = 1$ | Each super/hyper–element has a single plithogenic membership in $[0, 1]$, modulated by a value–level DCF over attribute values. | HyperFuzzy (plithogenic) Set [492–495] |
| Hyper–Plithogenic Intuitionistic Fuzzy Set | $s = 2$ | Each super/hyper–element carries a pair (membership, non–membership) in $[0, 1]^2$, aggregated via the plithogenic contradiction function. | Hyper–Intuitionistic Fuzzy Set |
| Hyper–Plithogenic Neutrosophic Set | $s = 3$ | Each super/hyper–element stores $(T, I, F) \in [0, 1]^3$; the plithogenic DCF weights them according to value–contradiction. | Hyper–Neutrosophic Set [291, 496–499] |

Hyper Degree of Appurtenance (HDAF) for $A$:

$$\tilde{pdf}^{(2,1)}(A, \text{low}) = \{(0.30, 0.50, 0.40),\ (0.40, 0.40, 0.50)\},$$
$$\tilde{pdf}^{(2,1)}(A, \text{moderate}) = \{(0.50, 0.30, 0.40),\ (0.60, 0.30, 0.30)\},$$
$$\tilde{pdf}^{(2,1)}(A, \text{high}) = \{(0.80, 0.10, 0.10),\ (0.70, 0.20, 0.20)\}.$$

Centroids (componentwise means) per value:

$$\bar{u}(\text{low}) = (0.35, 0.45, 0.45), \quad \bar{u}(\text{moderate}) = (0.55, 0.30, 0.35), \quad \bar{u}(\text{high}) = (0.75, 0.15, 0.15).$$

Contradiction (symmetric, reflexive, only off–diagonals shown):

$$pCF^{(2,1)}(\text{low}, \text{high}) = 0.7, \quad pCF^{(2,1)}(\text{moderate}, \text{high}) = 0.3, \quad pCF^{(2,1)}(\text{low}, \text{moderate}) = 0.4.$$

For a "dominant" value $\delta = \text{high}$, set compatibility weights $w(a \mid \delta) = 1 - pCF^{(2,1)}(a, \delta)$:

$$w(\text{low} \mid \text{high}) = 0.3, \quad w(\text{moderate} \mid \text{high}) = 0.7, \quad w(\text{high} \mid \text{high}) = 1, \quad D = 0.3 + 0.7 + 1 = 2.$$

A plithogenic reduction (weighted mean of centroids) yields

$$\mu^{(\text{high})}(A) = \frac{0.3\,\bar{u}(\text{low}) + 0.7\,\bar{u}(\text{moderate}) + 1\,\bar{u}(\text{high})}{2}.$$

Coordinatewise calculations:

$$T : \frac{0.3 \cdot 0.35 + 0.7 \cdot 0.55 + 1 \cdot 0.75}{2} = \frac{0.105 + 0.385 + 0.75}{2} = \frac{1.240}{2} = 0.620,$$
$$I : \frac{0.3 \cdot 0.45 + 0.7 \cdot 0.30 + 1 \cdot 0.15}{2} = \frac{0.135 + 0.210 + 0.150}{2} = \frac{0.495}{2} = 0.2475,$$
$$F : \frac{0.3 \cdot 0.45 + 0.7 \cdot 0.35 + 1 \cdot 0.15}{2} = \frac{0.135 + 0.245 + 0.150}{2} = \frac{0.530}{2} = 0.265.$$

Therefore, under "high–risk" dominance, the super–team $A$ has reduced hyper–neutrosophic degree $\mu^{(\text{high})}(A) = (0.620,\ 0.2475,\ 0.265)$.

**Example 3.4.3** ((m,n)=(1,2): Urban multi–modal trip bundle with nested (scenario–clustered) feasibility)**.** A multi-modal trip combines multiple transport modes, such as walking, buses, trains, and bikes, within one integrated journey for travelers [500].

Let $X$ be trip segments (bus links, rail legs, walk connectors). On $\mathcal{P}^1(X) = \mathcal{P}(X)$ choose the super–element

$$A = \{\text{BusSeg}_1,\ \text{RailSeg}_3\}.$$

Use the attribute $v$ ="congestion level" with $P_v = \{\text{free}, \text{moderate}, \text{heavy}\}$. Take $s = 1$ (scalar feasibility in $[0, 1]$), $n = 2$ (a set of *sets* of feasibilities), and $t = 1$.

HDAF as scenario clusters (each inner set is a scenario; values are segment–level feasibilities aggregated to $A$):

$$\tilde{pdf}^{(1,2)}(A, \text{free}) = \big\{\{0.90, 0.80\},\ \{0.85, 0.75\}\big\},$$
$$\tilde{pdf}^{(1,2)}(A, \text{moderate}) = \big\{\{0.70, 0.60\},\ \{0.75, 0.55\}\big\},$$
$$\tilde{pdf}^{(1,2)}(A, \text{heavy}) = \big\{\{0.50, 0.30\},\ \{0.45, 0.35\}\big\}.$$

Inner means (per scenario), then outer mean (per value):

$$\bar{u}(\text{free}) = \tfrac{1}{2}\big(\tfrac{0.90+0.80}{2} + \tfrac{0.85+0.75}{2}\big) = \tfrac{1}{2}(0.85 + 0.80) = 0.825,$$
$$\bar{u}(\text{moderate}) = \tfrac{1}{2}\big(\tfrac{0.70+0.60}{2} + \tfrac{0.75+0.55}{2}\big) = \tfrac{1}{2}(0.65 + 0.65) = 0.65,$$
$$\bar{u}(\text{heavy}) = \tfrac{1}{2}\big(\tfrac{0.50+0.30}{2} + \tfrac{0.45+0.35}{2}\big) = \tfrac{1}{2}(0.40 + 0.40) = 0.40.$$

Contradiction (symmetric, reflexive):

$$pCF^{(1,2)}(\text{free}, \text{heavy}) = 0.8, \quad pCF^{(1,2)}(\text{moderate}, \text{heavy}) = 0.5, \quad pCF^{(1,2)}(\text{free}, \text{moderate}) = 0.4.$$

With dominant $\delta$ = heavy, weights $w(a \mid \delta) = 1 - pCF^{(1,2)}(a, \delta)$:

$$w(\text{free} \mid \text{heavy}) = 0.2, \quad w(\text{moderate} \mid \text{heavy}) = 0.5, \quad w(\text{heavy} \mid \text{heavy}) = 1, \quad D = 0.2{+}0.5{+}1 = 1.7.$$

Reduced feasibility:

$$\mu^{(\text{heavy})}(A) = \frac{0.2 \cdot 0.825 + 0.5 \cdot 0.65 + 1 \cdot 0.40}{1.7} = \frac{0.165 + 0.325 + 0.400}{1.7} = \frac{0.890}{1.7} \approx 0.5235.$$

Thus, for rush–hour dominance, the super–trip $A$ remains feasible at $\approx 0.524$ after contradiction–aware nesting.

**Example 3.4.4** ((m,n)=(2,2): Multi–tier supplier portfolio with clustered neutrosophic appraisals)**.** A multi-tier supplier portfolio organizes vendors across primary, secondary, and tertiary levels to diversify risk, ensure resilience, and optimize costs (cf. [501]).

Let $X$ be base–level suppliers. On the domain $\mathcal{P}^2(X)$, consider

$$A = \big\{\{\text{Tier1A}, \text{Tier1B}\},\ \{\text{Tier2C}\}\big\}.$$

Attribute $v$ ="sourcing strategy" with $P_v = \{\text{single}, \text{dual}, \text{multi}\}$. Choose $s = 3$ (neutrosophic $(T, I, F)$), $n = 2$ (clustered scenario sets), $t = 1$.

Clustered HDAF for $A$ (each inner brace is a scenario cluster; vectors are $(T, I, F)$):

$$\tilde{pdf}^{(2,2)}(A, \text{single}) = \big\{\{(0.70, 0.20, 0.30), (0.60, 0.30, 0.30)\},\ \{(0.75, 0.20, 0.25)\}\big\},$$
$$\tilde{pdf}^{(2,2)}(A, \text{dual}) = \big\{\{(0.80, 0.15, 0.15), (0.78, 0.15, 0.12)\},\ \{(0.75, 0.18, 0.18)\}\big\},$$
$$\tilde{pdf}^{(2,2)}(A, \text{multi}) = \big\{\{(0.72, 0.20, 0.18), (0.70, 0.22, 0.18)\},\ \{(0.68, 0.25, 0.20)\}\big\}.$$

Inner (scenario) centroids, then outer (value) centroids:

Single:
$$c_1 = \tfrac{1}{2}\big((0.70, 0.20, 0.30) + (0.60, 0.30, 0.30)\big) = (0.65, 0.25, 0.30),$$
$$c_2 = (0.75, 0.20, 0.25), \quad \bar{u}(\text{single}) = \tfrac{1}{2}(c_1 + c_2) = (0.70, 0.225, 0.275).$$

Dual:
$$c_1 = \tfrac{1}{2}\big((0.80, 0.15, 0.15) + (0.78, 0.15, 0.12)\big) = (0.79, 0.15, 0.135),$$
$$c_2 = (0.75, 0.18, 0.18), \quad \bar{u}(\text{dual}) = \tfrac{1}{2}(c_1 + c_2) = (0.77, 0.165, 0.1575).$$

Multi:
$$c_1 = \tfrac{1}{2}\big((0.72, 0.20, 0.18) + (0.70, 0.22, 0.18)\big) = (0.71, 0.21, 0.18),$$
$$c_2 = (0.68, 0.25, 0.20), \quad \bar{u}(\text{multi}) = \tfrac{1}{2}(c_1 + c_2) = (0.695, 0.23, 0.19).$$

Contradiction (symmetric, reflexive):

$$pCF^{(2,2)}(\text{single}, \text{dual}) = 0.4, \quad pCF^{(2,2)}(\text{multi}, \text{dual}) = 0.3, \quad pCF^{(2,2)}(\text{single}, \text{multi}) = 0.7.$$

With dominant $\delta = \text{dual}$, weights $w(a \mid \delta) = 1 - pCF^{(2,2)}(a, \delta)$:

$$w(\text{single} \mid \text{dual}) = 0.6, \quad w(\text{dual} \mid \text{dual}) = 1, \quad w(\text{multi} \mid \text{dual}) = 0.7, \quad D = 0.6 + 1 + 0.7 = 2.3.$$

Reduced neutrosophic degree (componentwise):

$$\mu^{(\text{dual})}(A) = \frac{0.6\,\bar{u}(\text{single}) + 1\,\bar{u}(\text{dual}) + 0.7\,\bar{u}(\text{multi})}{2.3}.$$

Calculations:

$$T: \frac{0.6 \cdot 0.70 + 1 \cdot 0.77 + 0.7 \cdot 0.695}{2.3} = \frac{0.420 + 0.770 + 0.4865}{2.3} = \frac{1.6765}{2.3} \approx 0.7298,$$
$$I: \frac{0.6 \cdot 0.225 + 1 \cdot 0.165 + 0.7 \cdot 0.230}{2.3} = \frac{0.135 + 0.165 + 0.161}{2.3} = \frac{0.461}{2.3} \approx 0.2004,$$
$$F: \frac{0.6 \cdot 0.275 + 1 \cdot 0.1575 + 0.7 \cdot 0.190}{2.3} = \frac{0.165 + 0.1575 + 0.133}{2.3} = \frac{0.4555}{2.3} \approx 0.1980.$$

Hence, under "dual–sourcing" dominance, the super–portfolio $A$ attains

$$\mu^{(\text{dual})}(A) \approx (0.7298,\ 0.2004,\ 0.1980)$$

.

## 3.5 Plithogenic Linguistic Set

A plithogenic linguistic set assigns linguistic-term memberships and contradiction-aware weights to objects, aggregating hesitant or single-term evaluations under a dominant context. The plithogenic linguistic set is closely related to the Hesitant Fuzzy Linguistic Set [502–504] and the Linguistic Neutrosophic Set [505–509].

**Definition 3.5.1** (Plithogenic Linguistic Set (single-term and hesitant variants))**.** Let $P$ be a nonempty finite universe of objects and let $S = \{s_0, s_1, \dots, s_g\}$ be a finite *linguistic term set* endowed with its natural total order $s_0 \prec s_1 \prec \cdots \prec s_g$. Write $\iota : S \to \{0, 1, \dots, g\}$ for the (order-preserving) index map $\iota(s_i) = i$, and define the *normalized linguistic distance*

$$d_S(s_i, s_j) \ := \ \frac{|\iota(s_i) - \iota(s_j)|}{g} \ \in [0, 1].$$

**(A) Linguistic contradiction (on terms and on hesitant sets).**

(A1) (*Term-level DCF*) A *linguistic degree of contradiction* is any map $pCF : S \times S \to [0,1]$ that is symmetric and reflexive: $pCF(a,a) = 0$ and $pCF(a,b) = pCF(b,a)$. A canonical choice is $pCF(a,b) := d_S(a,b)$.

(A2) (*Lift to hesitant sets*) For nonempty $H, K \subseteq S$ define the *lifted* contradiction

$$\widehat{pCF}(H,K) \;:=\; \max\Big\{ \max_{a\in H}\min_{b\in K} pCF(a,b),\; \max_{b\in K}\min_{a\in H} pCF(a,b) \Big\} \in [0,1],$$

and set $\widehat{pCF}(\varnothing,\varnothing) := 0$ and $\widehat{pCF}(H,\varnothing) = \widehat{pCF}(\varnothing,K) := 1$ otherwise.

**(B) Value domain and linguistic appurtenance.** Fix $s \in \mathbb{N}$ (the number of appurtenance components). We consider two admissible value domains:

$$P_v \in \big\{\, S \;\; (\textit{single-term}\ \text{variant}), \quad \mathcal{P}^\star(S) := \mathcal{P}(S) \setminus \{\varnothing\} \;\; (\textit{hesitant}\ \text{variant}) \,\big\}.$$

A *linguistic degree of appurtenance function (LDAF)* is

$$pdf : \; P \times P_v \;\longrightarrow\; [0,1]^s, \qquad pdf(x,\alpha) = \big(\mu_1(x,\alpha), \dots, \mu_s(x,\alpha)\big),$$

with each component $\mu_i$ nonnegative and bounded by 1.

**(C) Fusion of contradiction channels (optional).** For $t \in \mathbb{N}_0$ let $\{pCF_j\}_{j=1}^t$ be $t$ (term-level) contradiction maps as in (A1) and let $\Phi : [0,1]^t \to [0,1]$ be a symmetric, coordinatewise-monotone *fusion* with $\Phi(0,\dots,0) = 0$. (If $t = 0$, we set $\Phi(\cdot) \equiv 0$.)

**(D) Compatibility weights and plithogenic aggregation.** Fix a *dominant linguistic value* $\delta \in P_v$. Define a compatibility weight $w(\cdot \mid \delta) \in [0,1]$ by

$$\text{single-term:} \quad w(a \mid \delta) \;:=\; 1 - \Phi\big(pCF_1(a,\delta), \dots, pCF_t(a,\delta)\big) \qquad (a \in S),$$

$$\text{hesitant:} \quad w(H \mid \delta) \;:=\; 1 - \Phi\big(\widehat{pCF}_1(H,\delta), \dots, \widehat{pCF}_t(H,\delta)\big) \qquad (H \in \mathcal{P}^\star(S)).$$

For $x \in P$, the *$\delta$–relative plithogenic linguistic degree* $\Gamma^{(\delta)}(x) \in [0,1]^s$ is defined componentwise by the weighted mean

$$\Gamma_i^{(\delta)}(x) \;:=\; \begin{cases} \dfrac{\sum\limits_{\alpha\in P_v} w(\alpha \mid \delta)\,\mu_i(x,\alpha)}{\sum\limits_{\alpha\in P_v} w(\alpha \mid \delta)}, & \text{if } \sum_{\alpha\in P_v} w(\alpha \mid \delta) > 0, \\ 0, & \text{otherwise,} \end{cases} \qquad i = 1,\dots,s.$$

The *Plithogenic Linguistic Set (PLS)* associated with $(P, S, P_v)$, $s$, $t$, $pdf$, $\{pCF_j\}$, and $\Phi$ is the tuple

$$\mathrm{PLS} \;:=\; \big(P,\; v,\; S,\; P_v,\; s,\; t,\; pdf,\; \{pCF_j\}_{j=1}^t,\; \Phi,\; \Gamma^{(\delta)}\big),$$

where $v$ designates the (linguistic) attribute and $\Gamma^{(\delta)}$ yields the contradiction-aware aggregated linguistic membership relative to the dominant value $\delta$.

Table 3.10 summarizes the reductions of the Plithogenic Linguistic Set.

We now present concrete examples of the concept.

Table 3.10: Reductions of the Plithogenic Linguistic Set

| Parameter setting | Reduction / meaning |
|---|---|
| $t = 0$ (no contradiction channel) | $w(\alpha \mid \delta) \equiv 1$; $\Gamma^{(\delta)}$ becomes the simple unweighted average over $P_v$ (independent of $\delta$). |
| $P_v = S$, $s = 1$, $t = 1$, $\Phi(z) = z$ | Standard plithogenic linguistic weighting: $\Gamma^{(\delta)}(x)$ is a contradiction-based weighted mean w.r.t. the dominant term $\delta$. |
| $P_v = \mathcal{P}^{\star}(S)$, $t = 0$ | Hesitant fuzzy linguistic case [436, 510]: nonempty subsets of $S$ act as hesitant term sets, aggregated by unweighted means. |

**Example 3.5.2** (Procurement: Supplier quality under a "Very Good" policy (single–term PLS))**.** Let the linguistic term set be

$$S = \{s_0 = \text{Very Poor},\ s_1 = \text{Poor},\ s_2 = \text{Fair},\ s_3 = \text{Good},\ s_4 = \text{Very Good},\ s_5 = \text{Excellent}\},$$

with index map $\iota(s_i) = i$ and $g = 5$. Define the linguistic distance $d_S(s_i, s_j) = \frac{|\iota(s_i) - \iota(s_j)|}{g}$. Take $t = 1$ and set $pCF(a, b) = d_S(a, b)$. The dominant linguistic value is $\delta = s_4$ ("Very Good"), hence the compatibility weights are $w(s_i \mid \delta) = 1 - d_S(s_i, s_4)$:

$$\big(w(s_0), w(s_1), w(s_2), w(s_3), w(s_4), w(s_5)\big) = (0.2,\ 0.4,\ 0.6,\ 0.8,\ 1.0,\ 0.8),$$

and $\sum_i w(s_i \mid \delta) = 3.8$.

For supplier $A$ (the universe $P = \{A\}$, single–term value domain $P_v = S$), suppose the linguistic appurtenance (for $s = 1$) is

$$\big(\mu(A, s_0), \mu(A, s_1), \mu(A, s_2), \mu(A, s_3), \mu(A, s_4), \mu(A, s_5)\big) = (0.02,\ 0.08,\ 0.20,\ 0.40,\ 0.22,\ 0.08).$$

The $\delta$–relative plithogenic linguistic degree is

$$\Gamma^{(\delta)}(A) = \frac{\sum_{i=0}^{5} w(s_i \mid \delta)\, \mu(A, s_i)}{\sum_{i=0}^{5} w(s_i \mid \delta)}.$$

Compute the numerator explicitly:

$$\begin{aligned} &0.2 \cdot 0.02 + 0.4 \cdot 0.08 + 0.6 \cdot 0.20 + 0.8 \cdot 0.40 + 1.0 \cdot 0.22 + 0.8 \cdot 0.08 \\ &= 0.004 + 0.032 + 0.120 + 0.320 + 0.220 + 0.064 \\ &= 0.760. \end{aligned}$$

Hence

$$\Gamma^{(\delta)}(A) = \frac{0.760}{3.8} = 0.2000.$$

Thus, under a "Very Good" policy, supplier $A$'s contradiction–aware aggregated quality is 0.20.

**Example 3.5.3** (Restaurant service: hesitant evaluations under a "Very Good" target (hesitant PLS))**.** Use the same $S$, $\iota$, $g = 5$, $d_S$, and $pCF = d_S$ as above. Work in the hesitant variant with value domain a finite subfamily of nonempty term sets

$$P_v = \{H_1, H_2, H_3\}, \quad H_1 = \{s_3, s_4\},\ H_2 = \{s_2, s_3\},\ H_3 = \{s_4, s_5\}.$$

Let the dominant value be $\delta = \{s_4\}$. The lifted contradiction from Definition (A2) gives

$$\widehat{pCF}(H, \delta) = \max_{a \in H} d_S(a, s_4).$$

Therefore the compatibility weights are

$$w(H_1 \mid \delta) = 1 - \max\{d_S(s_3, s_4), d_S(s_4, s_4)\} = 1 - \max\{0.2, 0\} = 0.8,$$

$$w(H_2 \mid \delta) = 1 - \max\{d_S(s_2, s_4), d_S(s_3, s_4)\} = 1 - \max\{0.4, 0.2\} = 0.6,$$

$$w(H_3 \mid \delta) = 1 - \max\{d_S(s_4, s_4), d_S(s_5, s_4)\} = 1 - \max\{0, 0.2\} = 0.8,$$

so $\sum w = 0.8 + 0.6 + 0.8 = 2.2$.

For restaurant $R$ (universe $P = \{R\}$) on the attribute "service speed", take (with $s = 1$)

$$\mu(R, H_1) = 0.60, \qquad \mu(R, H_2) = 0.25, \qquad \mu(R, H_3) = 0.35.$$

Then

$$\Gamma^{(\delta)}(R) = \frac{0.8 \cdot 0.60 + 0.6 \cdot 0.25 + 0.8 \cdot 0.35}{0.8 + 0.6 + 0.8} = \frac{0.480 + 0.150 + 0.280}{2.2} = \frac{0.910}{2.2} \approx 0.4136.$$

Thus, against a "Very Good" target, the hesitant plithogenic aggregation yields an overall service–speed score $\approx 0.414$.

**Example 3.5.4** (IT helpdesk: contradiction–aware $(T, I, F)$ aggregation (multi–component PLS))**.** An IT helpdesk provides users with technical support, troubleshooting, system guidance, and issue resolution to maintain smooth daily computer and network operations (cf. [511]).

Keep the same $S$, $\iota$, $g = 5$, $d_S$, $pCF = d_S$, and take the dominant value $\delta = s_4$. Let $s = 3$ encode $(T, I, F)$ = (truth, indeterminacy, falsity) components of satisfaction. For a resolved ticket $X$ on attribute "user satisfaction", specify single–term appurtenances for each $s_i \in S$:

| term $s_i$ | $s_0$ | $s_1$ | $s_2$ | $s_3$ | $s_4$ | $s_5$ |
|---|---|---|---|---|---|---|
| $T = \mu_1(X, s_i)$ | 0.05 | 0.10 | 0.30 | 0.55 | 0.70 | 0.60 |
| $I = \mu_2(X, s_i)$ | 0.10 | 0.15 | 0.20 | 0.20 | 0.15 | 0.15 |
| $F = \mu_3(X, s_i)$ | 0.80 | 0.70 | 0.50 | 0.30 | 0.20 | 0.25 |

With $\delta = s_4$, the weights are

$$\big(w(s_0), w(s_1), w(s_2), w(s_3), w(s_4), w(s_5)\big) = (0.2,\ 0.4,\ 0.6,\ 0.8,\ 1.0,\ 0.8), \qquad \sum w = 3.8.$$

Aggregate componentwise:

$$\Gamma_T^{(\delta)}(X) = \frac{0.2 \cdot 0.05 + 0.4 \cdot 0.10 + 0.6 \cdot 0.30 + 0.8 \cdot 0.55 + 1.0 \cdot 0.70 + 0.8 \cdot 0.60}{3.8}$$

$$= \frac{0.01 + 0.04 + 0.18 + 0.44 + 0.70 + 0.48}{3.8} = \frac{1.85}{3.8} \approx 0.4868,$$

$$\Gamma_I^{(\delta)}(X) = \frac{0.2 \cdot 0.10 + 0.4 \cdot 0.15 + 0.6 \cdot 0.20 + 0.8 \cdot 0.20 + 1.0 \cdot 0.15 + 0.8 \cdot 0.15}{3.8}$$

$$= \frac{0.02 + 0.06 + 0.12 + 0.16 + 0.15 + 0.12}{3.8} = \frac{0.63}{3.8} \approx 0.1658,$$

$$\Gamma_F^{(\delta)}(X) = \frac{0.2 \cdot 0.80 + 0.4 \cdot 0.70 + 0.6 \cdot 0.50 + 0.8 \cdot 0.30 + 1.0 \cdot 0.20 + 0.8 \cdot 0.25}{3.8}$$

$$= \frac{0.16 + 0.28 + 0.30 + 0.24 + 0.20 + 0.20}{3.8} = \frac{1.38}{3.8} \approx 0.3632.$$

Therefore, under a "Very Good" dominant context, the contradiction–aware aggregated triplet is

$$\Gamma^{(\delta)}(X) \approx (T, I, F) = (0.4868,\ 0.1658,\ 0.3632).$$

## 3.6 $q$-rung orthopair Plithogenic sets

A $q$-Rung $n$-Tuple Plithogenic Set models elements with multi-attribute memberships and contradictions, generalizing q-rung orthopair fuzzy sets [512, 513].

**Notation 3.6.1.** *Fix an integer $n \geq 2$ and $q \in \mathbb{N}$ with $q \geq 1$. Define the q-orthant constraint set in $[0,1]^n$ by*

$$\mathbb{O}_{q,n} := \Big\{ (m_1, \ldots, m_n) \in [0,1]^n : \sum_{i=1}^{n} m_i^q \leq n-1 \Big\}.$$

**Definition 3.6.2** (q-Rung $n$-Tuple Plithogenic Set (general form))**.** Let $S$ be a universe and let $P \subseteq S$ be nonempty. Fix an attribute $v$ with a nonempty value set $Pv$. A *q-rung n-tuple plithogenic set* is a quintuple

$$\mathrm{PS}_{q,n} = (P,\, v,\, Pv,\, \mathrm{pdf}_{q,n},\, \mathrm{pCF}),$$

where

- $\mathrm{pdf}_{q,n} : P \times Pv \to [0,1]^n$ is the *degree of appurtenance function* that returns an $n$-tuple

  $$\mathrm{pdf}_{q,n}(x,a) = \big(m_{a,1}(x), \ldots, m_{a,n}(x)\big),$$

  and satisfies the $q$-rung constraint

  $$\sum_{i=1}^{n} \big(m_{a,i}(x)\big)^q \leq n-1 \qquad \text{for all } (x,a) \in P \times Pv;$$

  equivalently, $\mathrm{pdf}_{q,n}(x,a) \in \mathbb{O}_{q,n}$ for all $(x,a)$;
- $\mathrm{pCF} : Pv \times Pv \to [0,1]$ is a *degree of contradiction* satisfying $\mathrm{pCF}(a,a) = 0$ and $\mathrm{pCF}(a,b) = \mathrm{pCF}(b,a)$ for all $a, b \in Pv$.

When $n = 2$ we recover the orthopair case; for general $n$, we obtain an *orthovector* of appurtenance degrees constrained by the $q$-sum bound $n-1$.

Table 3.11 provides an overview of the naming map for $q$-rung $n$-tuple plithogenic sets $\mathrm{PS}_{q,n}$.

We now present concrete examples of the concept.

**Example 3.6.3** (Medical triage (Fermatean orthopair, $q = 3$, $n = 2$; attribute = symptom severity))**.** Let the value set be $Pv = \{s_0 = \text{none}, s_1 = \text{mild}, s_2 = \text{moderate}, s_3 = \text{severe}\}$, indexed by $\iota(s_i) = i$. Define the plithogenic contradiction by the normalized distance

$$\mathrm{pCF}(s_i, s_j) = \frac{|\iota(s_i) - \iota(s_j)|}{3} \in [0,1],$$

and fix the dominant value $\delta = s_3$ (severe). Then the compatibility weights are

$$w(s_i \mid \delta) = 1 - \mathrm{pCF}(s_i, \delta) = \Big(1 - \tfrac{3}{3},\, 1 - \tfrac{2}{3},\, 1 - \tfrac{1}{3},\, 1 - 0\Big) = (0, \tfrac{1}{3}, \tfrac{2}{3}, 1),$$

with $\sum_i w(s_i \mid \delta) = 2$.

For a patient $x$, suppose the Fermatean (orthopair) degrees for “has the disease” vs “does not have the disease” are

| $a$ | $s_0$ | $s_1$ | $s_2$ | $s_3$ |
|---|---|---|---|---|
| $m_a(x) = \mu(x,a)$ | 0.05 | 0.35 | 0.70 | 0.88 |
| $n_a(x) = \nu(x,a)$ | 0.98 | 0.90 | 0.80 | 0.38 |

Table 3.11: Naming map for $q$-rung $n$-tuple plithogenic sets $\mathrm{PS}_{q,n}$

| $n$ **(tuple size)** | $q = 2$ | $q = 3$ |
|---|---|---|
| $n = 2$<br><br>$q$-rung orthopair fuzzy sets [520, 521] in $\mathrm{PS}_{q,2}$ | Pythagorean fuzzy set [514–516] embedded in $\mathrm{PS}_{2,2}$ | Fermatean fuzzy set [517–519] embedded in $\mathrm{PS}_{3,2}$ |
| $n = 3$<br><br>$q$-rung orthopair neutrosophic sets [536, 537], $q$-rung orthopair Picture Fuzzy Set [538–540], and $q$-rung orthopair Hesitant Fuzzy Set [541, 542] embedded in $\mathrm{PS}_{q,3}$ | Pythagorean neutrosophic set [334, 522, 523], Pythagorean hesitant fuzzy set [524, 525], and Pythagorean picture fuzzy set [526–528] embedded in $\mathrm{PS}_{2,3}$ | Fermatean neutrosophic set [529–531], Fermatean Picture Fuzzy set [532, 533], and Fermatean Hesitant Fuzzy set [534, 535] embedded in $\mathrm{PS}_{3,3}$ |

*Notes.* (1) "Pythagorean" $\Leftrightarrow q = 2$; "Fermatean" $\Leftrightarrow q = 3$. (2) $n = 2$ corresponds to the orthopair fuzzy case; $n = 3$ to the orthopair neutrosophic case.

and each pair satisfies the $q$-rung constraint $\mu^3 + \nu^3 \leq 1$ (e.g., for $s_3$: $0.88^3 + 0.38^3 = 0.681472 + 0.054872 = 0.736344 \leq 1$).

Plithogenic aggregation (weighted by $w$) gives

$$\mu^{(\delta)}(x) = \frac{0 \cdot 0.05 + \frac{1}{3} \cdot 0.35 + \frac{2}{3} \cdot 0.70 + 1 \cdot 0.88}{2} = \frac{0 + 0.116666\ldots + 0.466666\ldots + 0.88}{2} = 0.731666\ldots,$$

$$\nu^{(\delta)}(x) = \frac{0 \cdot 0.98 + \frac{1}{3} \cdot 0.90 + \frac{2}{3} \cdot 0.80 + 1 \cdot 0.38}{2} = \frac{0 + 0.30 + 0.533333\ldots + 0.38}{2} = 0.606666\ldots.$$

Constraint check after aggregation:

$$(\mu^{(\delta)})^3 + (\nu^{(\delta)})^3 \approx 0.731666^3 + 0.606667^3 \approx 0.392 + 0.223 = 0.615 < 1.$$

Thus the contradiction–aware Fermatean orthopair for triage under the "severe" context is $\left(\mu^{(\delta)}, \nu^{(\delta)}\right) \approx (0.732,\ 0.607)$.

**Example 3.6.4** (Credit risk (Pythagorean neutrosophic orthovector, $q = 2$, $n = 3$; attribute = income stability)). Let $Pv = \{a_0 = \text{low}, a_1 = \text{medium}, a_2 = \text{high}\}$ with $\iota(a_i) = i$. Set

$$\mathrm{pCF}(a_i, a_j) = \frac{|\iota(a_i) - \iota(a_j)|}{2}, \qquad w(a \mid \delta) = 1 - \mathrm{pCF}(a, \delta), \quad \delta = a_2.$$

Hence $w(a_0 \mid \delta) = 0$, $w(a_1 \mid \delta) = \frac{1}{2}$, $w(a_2 \mid \delta) = 1$ and $\sum w = 1.5$.

For an applicant $x$, let the Pythagorean neutrosophic triple $(T, I, F)$ (creditworthy, indeterminate, not creditworthy) be

| $a$ | $a_0$ | $a_1$ | $a_2$ |
|---|---|---|---|
| $T = \mu_1(x, a)$ | 0.20 | 0.55 | 0.85 |
| $I = \mu_2(x, a)$ | 0.20 | 0.25 | 0.10 |
| $F = \mu_3(x, a)$ | 0.90 | 0.55 | 0.35 |

Each satisfies the $q$-rung constraint $\sum_{i=1}^{3} \mu_i^2 \leq n - 1 = 2$ (e.g., for $a_2$: $0.85^2 + 0.10^2 + 0.35^2 = 0.7225 + 0.01 + 0.1225 = 0.855 \leq 2$).

Aggregate componentwise:

$$T^{(\delta)}(x) = \frac{0 \cdot 0.20 + \frac{1}{2} \cdot 0.55 + 1 \cdot 0.85}{1.5} = \frac{0.275 + 0.85}{1.5} = \frac{1.125}{1.5} = 0.75,$$

$$I^{(\delta)}(x) = \frac{0 \cdot 0.20 + \frac{1}{2} \cdot 0.25 + 1 \cdot 0.10}{1.5} = \frac{0.125 + 0.10}{1.5} = \frac{0.225}{1.5} = 0.15,$$

$$F^{(\delta)}(x) = \frac{0 \cdot 0.90 + \frac{1}{2} \cdot 0.55 + 1 \cdot 0.35}{1.5} = \frac{0.275 + 0.35}{1.5} = \frac{0.625}{1.5} = 0.416666\ldots.$$

Constraint check after aggregation:

$$(0.75)^2 + (0.15)^2 + (0.416666\ldots)^2 = 0.5625 + 0.0225 + 0.1736\ldots \approx 0.7586 < 2.$$

Thus the contradiction–aware Pythagorean neutrosophic assessment under the "high income stability" context is

$$\bigl(T^{(\delta)}, I^{(\delta)}, F^{(\delta)}\bigr) = (0.75,\ 0.15,\ 0.4167).$$

**Example 3.6.5** (Consumer electronics reliability (orthopair, $q = 4$, $n = 2$; attribute = warranty length))**.** Consumer electronics are everyday electronic devices that individuals buy for personal use, including phones, laptops, televisions, audio equipment, and wearables (cf. [543]).

Let $Pv = \{b_0 = 1 \text{ yr},\ b_1 = 2 \text{ yr},\ b_2 = 3 \text{ yr}\}$ with $\iota(b_i) = i$ and

$$\mathrm{pCF}(b_i, b_j) = \frac{|\iota(b_i) - \iota(b_j)|}{2}, \qquad \delta = b_2.$$

Then $w(b_0 \mid \delta) = 0$, $w(b_1 \mid \delta) = \frac{1}{2}$, $w(b_2 \mid \delta) = 1$ and $\sum w = 1.5$.

For a laptop model $x$, consider the $q$-rung ($q = 4$) orthopair for "reliable" vs "not reliable":

| $b$ | $b_0$ | $b_1$ | $b_2$ |
|---|---|---|---|
| $\mu(x, b)$ | 0.20 | 0.65 | 0.90 |
| $\nu(x, b)$ | 0.98 | 0.60 | 0.30 |

Each pair respects $\mu^4 + \nu^4 \le 1$ (e.g., for $b_2$: $0.90^4 + 0.30^4 = 0.6561 + 0.0081 = 0.6642 \le 1$).

Aggregate:

$$\mu^{(\delta)}(x) = \frac{0 \cdot 0.20 + \frac{1}{2} \cdot 0.65 + 1 \cdot 0.90}{1.5} = \frac{0.325 + 0.90}{1.5} = \frac{1.225}{1.5} = 0.816666\ldots,$$

$$\nu^{(\delta)}(x) = \frac{0 \cdot 0.98 + \frac{1}{2} \cdot 0.60 + 1 \cdot 0.30}{1.5} = \frac{0.30 + 0.30}{1.5} = \frac{0.60}{1.5} = 0.4.$$

Constraint check after aggregation:

$$(\mu^{(\delta)})^4 + (\nu^{(\delta)})^4 \approx 0.816666^4 + 0.4^4 \approx 0.4449 + 0.0256 = 0.4705 < 1.$$

Therefore, under the "3-year warranty" dominant context, the contradiction–aware $q$-rung ($q = 4$) orthopair is $\bigl(\mu^{(\delta)}, \nu^{(\delta)}\bigr) \approx (0.8167,\ 0.4000)$.

## 3.7 Type-$n$ Plithogenic Set

A type-$n$ plithogenic set recursively nests plithogenic memberships and contradictions across n levels to model hierarchical uncertain attributes and relations [544].

**Definition 3.7.1** (Type-$n$ Plithogenic Set)**.** [544] Let $S$ be a universal set, and let $P \subseteq S$. Let $v$ be an attribute taking values in a set $Pv$. Let $s \ge 1$ and $t \ge 1$ be fixed integers (the dimensions for the Degree of Appurtenance Function and the Degree of Contradiction Function, respectively). A *Type-n Plithogenic Set* of dimension $(s, t)$, denoted by

$$PS^{(n,s,t)} \;=\; \bigl(P,\ v,\ Pv,\ pdf,\ pCF\bigr),$$

is defined recursively as follows:

Table 3.12: Naming for Type-$n$ Plithogenic Sets when $t = 0$ (no explicit contradiction dimension)

| **Appurtenance dim.** $s$ | $t$ | **Standard name for** $PS^{(n,s,t)}$ |
|---|---|---|
| 1 | 0 | Type-$n$ Fuzzy Set [545–547] |
| 2 | 0 | Type-$n$ Intuitionistic Fuzzy Set |
| 3 | 0 | Type-$n$ Neutrosophic Set [496, 544] |
| 4 | 0 | Type-$n$ Quadripartitioned Neutrosophic Set |
| 5 | 0 | Type-$n$ Pentapartitioned Neutrosophic Set |

*Note.* Here $t = 0$ indicates that no Degree of Contradiction Function is carried at the top level. This is a degenerate ($t = 0$) variant of Definition 3.7.1, interpreted as "no explicit contradiction channel".

1. *Base Case ($n = 1$):* A *Type-1 Plithogenic Set* is simply a *classical Plithogenic Set* (cf. [10]), i.e.

$$pdf : P \times Pv \to [0,1]^s, \quad pCF : Pv \times Pv \to [0,1]^t,$$

subject to the axioms:

(a) *Reflexivity of Contradiction*:

$$pCF(a, a) = 0, \quad \forall a \in Pv,$$

(b) *Symmetry of Contradiction*:

$$pCF(a, b) = pCF(b, a), \quad \forall a, b \in Pv.$$

(c) Hence, a Type-1 Plithogenic Set is

$$PS^{(1,s,t)} = \big(P, v, Pv, pdf, pCF\big),$$

where $pdf$ and $pCF$ map into $[0,1]^s$ and $[0,1]^t$, respectively.

2. *Recursive Case ($n > 1$):* A *Type-n Plithogenic Set* is given by

$$pdf : P \times Pv \longrightarrow \mathcal{M}_{n-1}^{(s,t)}([0,1]), \quad pCF : Pv \times Pv \longrightarrow \mathcal{M}_{n-1}^{(s,t)}([0,1]),$$

where each $\mathcal{M}_k^{(s,t)}([0,1])$ denotes the set of all *Type-k Plithogenic Sets* (of dimension $(s,t)$) on the unit interval $[0,1]$ for their degree of appurtenance and contradiction values.[2]

Concretely, if $(n-1) = 1$, then the recursion bottoms out in a classical Plithogenic Set; if $(n-1) > 1$, the recursion continues. Thus, a Type-$n$ Plithogenic Set is an *($n$-level) hierarchical structure* where each membership entry is itself a Type-$(n-1)$ Plithogenic object.

Tables 3.12 and 3.13 present the associated concepts of Type-$n$ Plithogenic Sets.

We now present concrete examples of the concept.

**Example 3.7.2** (Healthcare treatment planning — Type-2 Plithogenic Set ($n = 2$, neutrosophic $s = 3$, one contradiction channel $t = 1$)**).** Healthcare treatment planning designs patient-specific interventions by integrating diagnosis, risks, goals, and resources to optimize clinical outcomes and minimize complications (cf. [569]).

Let the universe $P$ contain a patient $x$. The top-level attribute is *therapy intensity* with values $Pv = \{L = \text{low}, M = \text{medium}, H = \text{high}\}$, indexed by $\iota(L) = 0$, $\iota(M) = 1$, $\iota(H) = 2$. Define the top-level contradiction by normalized distance

$$\mathrm{pCF}(a, b) = \frac{|\iota(a) - \iota(b)|}{2} \in [0,1], \qquad w(a \mid \delta) = 1 - \mathrm{pCF}(a, \delta), \quad \delta = H.$$

[2]More precisely, at each recursion level, the range $[0,1]^s$ for the DAF becomes replaced by a set of dimension-$s$ *plithogenic* membership objects, each of which is itself a Type-$(n-1)$ Plithogenic Set. Likewise, the range $[0,1]^t$ for the DCF is replaced by Type-$(n-1)$ plithogenic contradiction objects.

Table 3.13: Concrete instantiations for $n = 2$ and $n = 3$ with $t = 0$

| $n$ | $s$ | $t$ | **Standard name** |
|---|---|---|---|
| 2 | 1 | 0 | Type-2 Fuzzy Set [548–550] |
| 2 | 1 | 0 | Type-2 Shadowed Set [551, 552] |
| 2 | 2 | 0 | Type-2 Intuitionistic Fuzzy Set [553, 554] |
| 2 | 2 | 0 | Type-2 Vague Set [555, 556] |
| 2 | 3 | 0 | Type-2 Neutrosophic Set [191, 557] |
| 2 | 3 | 0 | Type-2 Picture Fuzzy Set [558, 559] |
| 2 | 3 | 0 | Type-2 Hesitant Fuzzy Set [560, 561] |
| 2 | 3 | 0 | Type-2 Spherical Fuzzy Set [562] |
| 2 | 4 | 0 | Type-2 Quadripartitioned Neutrosophic Set |
| 2 | 5 | 0 | Type-2 Pentapartitioned Neutrosophic Set |
| 3 | 1 | 0 | Type-3 Fuzzy Set [161, 262, 563, 564] |
| 3 | 2 | 0 | Type-3 Intuitionistic Fuzzy Set [565–567] |
| 3 | 3 | 0 | Type-3 Neutrosophic Set [568] |
| 3 | 4 | 0 | Type-3 Quadripartitioned Neutrosophic Set |
| 3 | 5 | 0 | Type-3 Pentapartitioned Neutrosophic Set |

*Reading.* Increasing $n$ raises the *type level* (depth of iteration of plithogenic-valued memberships). Choosing $s \in \{1, 2, 3, 4, 5\}$ selects the membership tuple shape: fuzzy ($s = 1$), intuitionistic fuzzy ($s = 2$), neutrosophic ($s = 3$), ...; setting $t = 0$ suppresses the contradiction dimension at the top level.

Hence $w(L \mid H) = 0,\ w(M \mid H) = \frac{1}{2},\ w(H \mid H) = 1$ and $\sum w = 1.5$.

At each top value $a \in \{L, M, H\}$, the membership object is a *Type-1 plithogenic* (neutrosophic) triple $(T, I, F) \in [0, 1]^3$ *aggregated* from a micro-attribute *adherence* with values $Q = \{q_0 = \text{poor}, q_1 = \text{fair}, q_2 = \text{good}\}$, $\jmath(q_i) = i$, micro-DCF $\widehat{\text{pCF}}(q_i, q_j) = |i - j|/2$, and micro-weights $w(q \mid q_2) = 1 - \widehat{\text{pCF}}(q, q_2)$, i.e. $(0, \frac{1}{2}, 1)$ with sum 1.5.

Micro-level data (neutrosophic $(T, I, F)$) for $x$:

| intensity $a$ | $q_0$ (poor) | $q_1$ (fair) | $q_2$ (good) |
|---|---|---|---|
| $L$ | $(0.35, 0.25, 0.60)$ | $(0.55, 0.20, 0.45)$ | $(0.70, 0.15, 0.35)$ |
| $M$ | $(0.45, 0.25, 0.55)$ | $(0.65, 0.18, 0.40)$ | $(0.82, 0.12, 0.28)$ |
| $H$ | $(0.52, 0.20, 0.48)$ | $(0.78, 0.14, 0.26)$ | $(0.90, 0.10, 0.18)$ |

Micro-aggregation to obtain the *Type-1* neutrosophic triple at each $a$ (weights $0, \frac{1}{2}, 1$; sum 1.5):

$$(T, I, F)_L = \Big(\tfrac{0 \cdot 0.35 + \frac{1}{2} \cdot 0.55 + 1 \cdot 0.70}{1.5},\ \tfrac{0 \cdot 0.25 + \frac{1}{2} \cdot 0.20 + 1 \cdot 0.15}{1.5},\ \tfrac{0 \cdot 0.60 + \frac{1}{2} \cdot 0.45 + 1 \cdot 0.35}{1.5}\Big)$$
$$= (0.6500,\ 0.1667,\ 0.3833),$$
$$(T, I, F)_M = (0.7633,\ 0.1400,\ 0.3200), \qquad (T, I, F)_H = (0.8600,\ 0.1133,\ 0.2067).$$

Top-level Type-2 aggregation (weights $0, \frac{1}{2}, 1$; sum 1.5) relative to $\delta = H$:

$$T^{(\delta)}(x) = \frac{0 \cdot 0.6500 + \frac{1}{2} \cdot 0.7633 + 1 \cdot 0.8600}{1.5} = 0.8278,$$
$$I^{(\delta)}(x) = \frac{0 \cdot 0.1667 + \frac{1}{2} \cdot 0.1400 + 1 \cdot 0.1133}{1.5} = 0.1222,$$
$$F^{(\delta)}(x) = \frac{0 \cdot 0.3833 + \frac{1}{2} \cdot 0.3200 + 1 \cdot 0.2067}{1.5} = 0.2444.$$

Thus, under the dominant context "high intensity", the *Type-2* neutrosophic evaluation is $(T, I, F) \approx (0.828, 0.122, 0.244)$.

**Example 3.7.3** (Global logistics reliability — Type-3 Plithogenic Set ($n = 3$, fuzzy $s = 1$, $t = 1$)). Global logistics reliability measures how consistently international supply chains deliver goods on time, intact, and as contracted across worldwide markets (cf. [570]).

We assess a vendor $x$ along three nested levels.

*Level 3 (top):* attribute *vendor scope* $V = \{v_0 = \text{local}, v_1 = \text{regional}, v_2 = \text{global}\}$, indices $\iota(v_i) = i$, DCF $\text{pCF}(v_i, v_j) = |i - j|/2$, dominant $\delta_V = v_2$, weights $w_V = (0, \frac{1}{2}, 1)$, sum 1.5.

*Level 2 (mid):* attribute *delivery mode* $D = \{d_0 = \text{road}, d_1 = \text{air}, d_2 = \text{sea}\}$, indices $0, 1, 2$, DCF $|i - j|/2$, dominant $\delta_D = d_1$, weights $w_D = (\frac{1}{2}, 1, 0)$, sum 1.5.

*Level 1 (bottom):* attribute *weather* $W = \{w_0 = \text{clear}, w_1 = \text{rain}, w_2 = \text{storm}\}$, indices $0, 1, 2$, DCF $|i - j|/2$, dominant $\delta_W = w_0$, weights $w_W = (1, \frac{1}{2}, 0)$, sum 1.5.

Bottom-level fuzzy reliability scores $\mu \in [0, 1]$ (by delivery mode and weather):

| $d$ | $\mu$(clear) | $\mu$(rain) | $\mu$(storm) |
|---|---|---|---|
| road | 0.80 | 0.60 | 0.25 |
| air | 0.90 | 0.75 | 0.30 |
| sea | 0.85 | 0.55 | 0.20 |

Level 1 aggregation (weights $1, \frac{1}{2}, 0$; sum 1.5):

$$\mu_{\text{road}} = \frac{1{\cdot}0.80 + \frac{1}{2}{\cdot}0.60 + 0{\cdot}0.25}{1.5} = 0.7333,$$

$$\mu_{\text{air}} = \frac{1{\cdot}0.90 + \frac{1}{2}{\cdot}0.75 + 0{\cdot}0.30}{1.5} = 0.8500,$$

$$\mu_{\text{sea}} = \frac{1{\cdot}0.85 + \frac{1}{2}{\cdot}0.55 + 0{\cdot}0.20}{1.5} = 0.7500.$$

Level 2 aggregation per vendor scope (dominant $d_1$; weights $\frac{1}{2}, 1, 0$; sum 1.5). For illustration, suppose local/regional/global slightly differ in these level-1 results:

| scope | $\mu_{\text{road}}$ | $\mu_{\text{air}}$ | $\mu_{\text{sea}}$ |
|---|---|---|---|
| local ($v_0$) | 0.7333 | 0.8500 | 0.7500 |
| regional ($v_1$) | 0.7000 | 0.8800 | 0.7200 |
| global ($v_2$) | 0.7600 | 0.9200 | 0.7800 |

Then

$$\mu_{v_0} = \frac{\frac{1}{2}{\cdot}0.7333 + 1{\cdot}0.8500 + 0{\cdot}0.7500}{1.5} = 0.8111,$$

$$\mu_{v_1} = \frac{\frac{1}{2}{\cdot}0.7000 + 1{\cdot}0.8800}{1.5} = 0.8200,$$

$$\mu_{v_2} = \frac{\frac{1}{2}{\cdot}0.7600 + 1{\cdot}0.9200}{1.5} = 0.8667.$$

Level 3 aggregation (dominant $v_2$; weights $0, \frac{1}{2}, 1$; sum 1.5):

$$\mu^{(\delta_V)}(x) = \frac{0 \cdot 0.8111 + \frac{1}{2} \cdot 0.8200 + 1 \cdot 0.8667}{1.5} = \frac{0.4100 + 0.8667}{1.5} = 0.8511.$$

Thus, the *Type-3* fuzzy plithogenic reliability under the "global scope" context is $\mu \approx 0.851$.

**Example 3.7.4** (University admission suitability — Type-2 Plithogenic Set ($n = 2$, intuitionistic fuzzy $s = 2$, $t = 1$))**.** University admission suitability evaluates how well an applicant's academic record, abilities, and background match a university's entry requirements and expectations (cf. [571]).

Top attribute *program intensity* $Pv = \{L = \text{light}, B = \text{balanced}, I = \text{intensive}\}$, indices $0, 1, 2$, DCF $\text{pCF}(a, b) = |i - j|/2$, dominant $\delta = I$, weights $w(L\,|\,I) = 0,\ w(B\,|\,I) = \frac{1}{2},\ w(I\,|\,I) = 1$, sum 1.5.

At each $a \in Pv$, the membership object is *Type-1 intuitionistic fuzzy* $(\mu, \nu)$ aggregated from micro-attribute *study habits* $Q = \{q_0 = \text{poor}, q_1 = \text{average}, q_2 = \text{strong}\}$, with micro-DCF $|i - j|/2$, dominant $q_2$, micro-weights $(0, \frac{1}{2}, 1)$, sum 1.5. All pairs satisfy $\mu + \nu \leq 1$.

Micro-level data and aggregation:

| $a$ | $q_0$ | $q_1$ | $q_2$ |
|---|---|---|---|
| $L$ | $(0.30, 0.60)$ | $(0.55, 0.35)$ | $(0.70, 0.20)$ |
| $B$ | $(0.40, 0.55)$ | $(0.65, 0.30)$ | $(0.80, 0.15)$ |
| $I$ | $(0.45, 0.50)$ | $(0.72, 0.22)$ | $(0.90, 0.08)$ |

Level-1 aggregation (weights $0, \frac{1}{2}, 1$; sum 1.5):

$$(\mu, \nu)_L = \left(\tfrac{0\cdot 0.30 + \frac{1}{2}\cdot 0.55 + 1\cdot 0.70}{1.5},\ \tfrac{0\cdot 0.60 + \frac{1}{2}\cdot 0.35 + 1\cdot 0.20}{1.5}\right) = (0.6500,\ 0.2500),$$
$$(\mu, \nu)_B = (0.7500,\ 0.2000), \qquad (\mu, \nu)_I = (0.8400,\ 0.1267).$$

(Check: $0.65{+}0.25 = 0.90 \leq 1$, $0.75{+}0.20 = 0.95 \leq 1$, $0.84{+}0.1267 \approx 0.9667 \leq 1$.)

Top-level Type-2 aggregation (weights $0, \frac{1}{2}, 1$; sum 1.5):

$$\mu^{(\delta)} = \frac{0 \cdot 0.65 + \frac{1}{2} \cdot 0.75 + 1 \cdot 0.84}{1.5} = \frac{0.375 + 0.84}{1.5} = 0.8100,$$

$$\nu^{(\delta)} = \frac{0 \cdot 0.25 + \frac{1}{2} \cdot 0.20 + 1 \cdot 0.1267}{1.5} = \frac{0.10 + 0.1267}{1.5} = 0.1511.$$

Hence, under the "intensive program" context, the *Type-2* intuitionistic fuzzy assessment is $(\mu, \nu) \approx (0.810,\ 0.151)$ with $\mu + \nu \approx 0.961 \leq 1$.

## 3.8 Iterative MultiPlithogenic Set

A MultiPlithogenic Set models elements with multi-component degrees and contradiction among attribute values for rich, flexible uncertainty representation in applications. An Iterative MultiPlithogenic Set recursively assigns plithogenic-valued memberships, layering appurtenance and contradiction across levels for hierarchical, multi-attribute, robust uncertainty modeling [407].

**Notation 3.8.1.** *Fix a nonempty universe $P$ (of objects to be evaluated), a single attribute $v$ with a finite value set $P_v$, and two finite index sets:*

$$\mathcal{I} = \{1, \ldots, k\} \quad \textit{for DAF components}, \qquad \mathcal{J} = \{1, \ldots, \ell\} \quad \textit{for DCF components.}$$

*Let $\{L_i\}_{i \in \mathcal{I}}$ and $\{L'_j\}_{j \in \mathcal{J}}$ be complete lattices (typically $L_i = L'_j = [0, 1]$).*

**Definition 3.8.2** (Iterative MultiPlithogenic codomains)**.** [407] Define the level-0 plithogenic fiber as

$$\mathfrak{L}^{(0)} := \prod_{i\in\mathcal{I}} L_i.$$

Recursively, for $n \geq 1$ set

$$\mathfrak{L}^{(n)}(P, P_v) := \Big\{ \Phi : P \times P_v \longrightarrow \mathfrak{L}^{(n-1)}(P, P_v) \Big\}.$$

(Thus an element of $\mathfrak{L}^{(n)}$ is a *plithogenic-valued* mapping on $P \times P_v$ whose values are level-$(n-1)$ objects.)

**Definition 3.8.3** (Iterative MultiPlithogenic Set (order $n$))**.** [407] Let $n \in \mathbb{N}$, $n \geq 1$. An *Iterative MultiPlithogenic Set of order* $n$ on $P$ is a tuple

$$\mathrm{IMPS}^{(n)} := \Big(P,\ v,\ P_v,\ \{\,\mathrm{PDF}_i^{(n)}\,\}_{i\in\mathcal{I}},\ \{\,\mathrm{pCF}_j\,\}_{j\in\mathcal{J}}\Big),$$

where

- for each $i \in \mathcal{I}$, the $i$-th *iterative Degree of Appurtenance Function* (DAF) is

$$\mathrm{PDF}_i^{(n)} :\ P \times P_v \ \longrightarrow\ \mathfrak{L}^{(n-1)}(P, P_v);$$

- for each $j \in \mathcal{J}$, the *Degree of Contradiction Function* (DCF)

$$\mathrm{pCF}_j :\ P_v \times P_v \ \longrightarrow\ L'_j$$

  is symmetric and reflexive: $\mathrm{pCF}_j(\alpha, \alpha) = 0$, $\quad \mathrm{pCF}_j(\alpha, \beta) = \mathrm{pCF}_j(\beta, \alpha)$ for all $\alpha, \beta \in P_v$.

When $n = 1$ we get the classical (non-iterative) case $\mathrm{PDF}_i^{(1)} : P \times P_v \to \mathfrak{L}^{(0)} = \prod_{i\in\mathcal{I}} L_i$, i.e., ordinary MultiPlithogenic DAFs.

Table 3.14 presents the taxonomy of Iterative MultiPlithogenic Sets by order and instance.

We now present concrete examples of the proposed concept.

**Example 3.8.4** (Smart-home HVAC mode selection — IMPS of order 2 with two fuzzy components)**.** We evaluate a dwelling $x \in P$ under attribute *HVAC mode* with values $P_v = \{E = \text{Eco}, N = \text{Normal}, T = \text{Turbo}\}$. Define $\iota(E) = 0,\ \iota(N) = 1,\ \iota(T) = 2$ and the term-level contradiction

$$\mathrm{pCF}(a, b) = \frac{|\iota(a) - \iota(b)|}{2} \in [0, 1].$$

Fix the dominant context $\delta = T$ so that the compatibility weights are

$$w(E\,|\,T) = 0, \qquad w(N\,|\,T) = \tfrac{1}{2}, \qquad w(T\,|\,T) = 1, \qquad \sum w = 1.5.$$

This is an *Iterative MultiPlithogenic Set of order* 2 with two DAF components: (i) *cost-saving* $c \in [0, 1]$ (higher is cheaper), and (ii) *comfort* $u \in [0, 1]$ (higher is more comfortable). At level 1 (the inner plithogenic objects), each mode $a \in P_v$ aggregates over the micro-attribute *outdoor temperature* $Q = \{\text{Mild}, \text{Warm}, \text{Hot}\}$, with indices $0, 1, 2$, micro-DCF $|i - j|/2$, dominant micro-context "Hot," and micro-weights $w_Q(\text{Mild}) = 0$, $w_Q(\text{Warm}) = \frac{1}{2}$, $w_Q(\text{Hot}) = 1$ (sum 1.5).

Table 3.14: Taxonomy of Iterative MultiPlithogenic Sets by order and instance

| **Order** | **General family** | **Fuzzy instance** | **Intuitionistic fuzzy instance** | **Neutrosophic instance** |
|---|---|---|---|---|
| 1 | MultiPlithogenic Set | MultiFuzzy Set [407, 572, 573] | MultiIntuitionistic Fuzzy Set [574–576] (cf. MultiVague Set [577, 578]) | MultiNeutrosophic Set [408, 579–581] (cf.Multi-Hesitant Fuzzy Set [582–584] and Multi-Picture Fuzzy Set [585]) |
| $n \geq 2$ | Iterative MultiPlithogenic Set (order $n$) | Iterative MultiFuzzy Set (order $n$) [407] | Iterative MultiIntuitionistic Fuzzy Set (order $n$) [407] | Iterative MultiNeutrosophic Set (order $n$) [407] |

*DAF codomain and typical component shapes.* At order 1: $\mathrm{PDF}_i^{(1)} : P \times P_v \to \mathfrak{L}^{(0)} = \prod_{i \in \mathcal{I}} L_i$. Typical choices:

$$\begin{aligned} &\text{MultiFuzzy:} && L_i = [0,1] \ \Rightarrow\ (\mu_i)_{i\in\mathcal{I}} \in [0,1]^k, \\ &\text{MultiIntuitionistic Fuzzy:} && L_i = [0,1]^2 \ \Rightarrow\ (\mu_i, \nu_i) \text{ with } \mu_i + \nu_i \leq 1, \\ &\text{MultiNeutrosophic:} && L_i = [0,1]^3 \ \Rightarrow\ (T_i, I_i, F_i) \in [0,1]^3. \end{aligned}$$

$$\text{At order } n \geq 2\text{: } \mathrm{PDF}_i^{(n)} : P \times P_v \to \mathfrak{L}^{(n-1)}(P, P_v).$$

Level-1 data for $x$:

| mode $a$ | Mild | Warm | Hot |
|---|---|---|---|
| $E:\ (c,u)$ | $(0.85, 0.60)$ | $(0.75, 0.55)$ | $(0.65, 0.50)$ |
| $N:\ (c,u)$ | $(0.70, 0.75)$ | $(0.60, 0.80)$ | $(0.50, 0.85)$ |
| $T:\ (c,u)$ | $(0.55, 0.88)$ | $(0.45, 0.92)$ | $(0.35, 0.95)$ |

Level-1 aggregation (weights $0, \frac{1}{2}, 1$; denominator 1.5):

$$\begin{aligned} (c,u)_E &= \left( \tfrac{0\cdot 0.85 + \frac{1}{2}\cdot 0.75 + 1\cdot 0.65}{1.5}, \ \tfrac{0\cdot 0.60 + \frac{1}{2}\cdot 0.55 + 1\cdot 0.50}{1.5} \right) = (0.6833,\ 0.5167), \\ (c,u)_N &= (0.5333,\ 0.8333), \qquad (c,u)_T = (0.3833,\ 0.9400). \end{aligned}$$

Top-level (order 2) aggregation relative to $\delta = T$ (weights $0, \frac{1}{2}, 1$; denominator 1.5):

$$\begin{aligned} c^{(\delta)}(x) &= \tfrac{0\cdot 0.6833 + \frac{1}{2}\cdot 0.5333 + 1\cdot 0.3833}{1.5} = \tfrac{0.2667 + 0.3833}{1.5} = 0.4333, \\ u^{(\delta)}(x) &= \tfrac{0\cdot 0.5167 + \frac{1}{2}\cdot 0.8333 + 1\cdot 0.9400}{1.5} = \tfrac{0.4167 + 0.9400}{1.5} = 0.9044. \end{aligned}$$

Hence, under the dominant "Turbo" context, the order-2 multi-fuzzy evaluation is $(c,u) \approx (0.4333,\ 0.9044)$.

**Example 3.8.5** (Fraud-alert triage — IMPS of order 2 with two fuzzy components)**.** Let $P_v = \{I = \text{Ignore}, M = \text{Monitor}, B = \text{Block}\}$ with $\iota(I) = 0, \iota(M) = 1, \iota(B) = 2$, $\mathrm{pCF}(a,b) = |\iota(a) - \iota(b)|/2$, dominant $\delta = B$ giving $w(I \,|\, B) = 0,\ w(M \,|\, B) = \frac{1}{2},\ w(B \,|\, B) = 1$ (sum 1.5).

Two components are recorded at the inner level: precision $p \in [0,1]$ and recall $r \in [0,1]$. Each action $a$ aggregates over micro-attribute *evidence severity* $Q = \{\ell = \text{low}, m = \text{mid}, h = \text{high}\}$ with indices $0, 1, 2$, micro-DCF $|i - j|/2$, dominant $h$, and micro-weights $(0, \frac{1}{2}, 1)$ (sum 1.5).

Level-1 data:

| $a$ | $(p,r)_\ell$ | $(p,r)_m$ | $(p,r)_h$ |
|---|---|---|---|
| $I$ | $(0.95, 0.20)$ | $(0.85, 0.35)$ | $(0.70, 0.50)$ |
| $M$ | $(0.90, 0.50)$ | $(0.82, 0.70)$ | $(0.75, 0.85)$ |
| $B$ | $(0.80, 0.60)$ | $(0.72, 0.80)$ | $(0.65, 0.95)$ |

Level-1 aggregation (weights $0, \frac{1}{2}, 1$; denominator 1.5):

$$(p,r)_I = \left(\frac{0\cdot 0.95+\frac{1}{2}\cdot 0.85+1\cdot 0.70}{1.5},\ \frac{0\cdot 0.20+\frac{1}{2}\cdot 0.35+1\cdot 0.50}{1.5}\right) = (0.7500,\ 0.4500),$$
$$(p,r)_M = (0.7733,\ 0.8000), \qquad (p,r)_B = (0.6733,\ 0.9000).$$

Top-level (order 2) aggregation relative to $\delta = B$ (weights $0, \frac{1}{2}, 1$; denominator 1.5):

$$p^{(\delta)}(x) = \frac{0\cdot 0.7500+\frac{1}{2}\cdot 0.7733+1\cdot 0.6733}{1.5} = \frac{0.3867+0.6733}{1.5} = 0.7066,$$
$$r^{(\delta)}(x) = \frac{0\cdot 0.4500+\frac{1}{2}\cdot 0.8000+1\cdot 0.9000}{1.5} = \frac{0.4000+0.9000}{1.5} = 0.8667.$$

Thus, in the dominant “Block” context, the order-2 multi-fuzzy assessment is $(p,r) \approx (0.7066,\ 0.8667)$.

**Example 3.8.6** (Public transit planning — IMPS of order 2 with two neutrosophic components)**.** Public transit planning designs, schedules, and optimizes buses, trains, and routes to move people efficiently, affordably, safely, and sustainably citywide (cf. [586, 587]).

We choose service frequency $P_v = \{L = \text{low},\ M = \text{medium},\ H = \text{high}\}$, $\iota(L) = 0, \iota(M) = 1, \iota(H) = 2$, $\mathrm{pCF}(a,b) = |\iota(a) - \iota(b)|/2$, dominant $\delta = H$, so $w(L \mid H) = 0,\ w(M \mid H) = \frac{1}{2},\ w(H \mid H) = 1$ (sum 1.5).

This order-2 *MultiNeutrosophic* instance has two components, each a neutrosophic triple $(T, I, F)$: component 1 = *reliability*, component 2 = *low crowding (comfort)*. At level 1, each frequency $a$ aggregates by the time-of-day micro-attribute $Q = \{\text{OP} = \text{off-peak},\ \text{PK} = \text{peak},\ \text{LN} = \text{late-night}\}$, with indices $0, 1, 2$, micro-DCF $|i - j|/2$, dominant micro-context PK, micro-weights $w_Q(\text{OP}) = \frac{1}{2},\ w_Q(\text{PK}) = 1,\ w_Q(\text{LN}) = \frac{1}{2}$ (sum 2.0).

Level-1 data for $x$:

Reliability $(T, I, F)$

| $a$ | OP | PK | LN |
|---|---|---|---|
| $L$ | (0.70, 0.20, 0.30) | (0.55, 0.25, 0.45) | (0.65, 0.22, 0.35) |
| $M$ | (0.78, 0.18, 0.25) | (0.68, 0.20, 0.35) | (0.72, 0.18, 0.28) |
| $H$ | (0.82, 0.15, 0.22) | (0.80, 0.14, 0.20) | (0.78, 0.16, 0.22) |

Low-crowding $(T, I, F)$

| $a$ | OP | PK | LN |
|---|---|---|---|
| $L$ | (0.80, 0.15, 0.20) | (0.40, 0.30, 0.60) | (0.75, 0.18, 0.25) |
| $M$ | (0.75, 0.18, 0.25) | (0.55, 0.25, 0.45) | (0.70, 0.20, 0.30) |
| $H$ | (0.70, 0.22, 0.30) | (0.50, 0.28, 0.50) | (0.65, 0.24, 0.35) |

Level-1 aggregation (weights $\frac{1}{2}, 1, \frac{1}{2}$; denominator 2.0):

Reliability

$$(T,I,F)_L = \left(\frac{0.5\cdot 0.70+1\cdot 0.55+0.5\cdot 0.65}{2},\ \frac{0.5\cdot 0.20+1\cdot 0.25+0.5\cdot 0.22}{2},\ \frac{0.5\cdot 0.30+1\cdot 0.45+0.5\cdot 0.35}{2}\right)$$
$$= (0.6125,\ 0.2300,\ 0.3875),$$
$$(T,I,F)_M = (0.7150,\ 0.1900,\ 0.3075), \qquad (T,I,F)_H = (0.8000,\ 0.1475,\ 0.2100).$$

Low-crowding

$$(T,I,F)_L = \left(\frac{0.5\cdot0.80+1\cdot0.40+0.5\cdot0.75}{2},\ \frac{0.5\cdot0.15+1\cdot0.30+0.5\cdot0.18}{2},\ \frac{0.5\cdot0.20+1\cdot0.60+0.5\cdot0.25}{2}\right)$$
$$= (0.5875,\ 0.2325,\ 0.4125),$$
$$(T,I,F)_M = (0.6375,\ 0.2200,\ 0.3625), \qquad (T,I,F)_H = (0.5875,\ 0.2550,\ 0.3958).$$

Top-level (order 2) aggregation relative to $\delta = H$ (weights $0, \frac{1}{2}, 1$; denominator 1.5):

Reliability

$$T^{(\delta)} = \frac{0\cdot0.6125+\frac{1}{2}\cdot0.7150+1\cdot0.8000}{1.5} = \frac{0.3575+0.8000}{1.5} = 0.7717,$$
$$I^{(\delta)} = \frac{0\cdot0.2300+\frac{1}{2}\cdot0.1900+1\cdot0.1475}{1.5} = \frac{0.0950+0.1475}{1.5} = 0.1617,$$
$$F^{(\delta)} = \frac{0\cdot0.3875+\frac{1}{2}\cdot0.3075+1\cdot0.2100}{1.5} = \frac{0.1538+0.2100}{1.5} = 0.2425.$$

Low-crowding

$$T^{(\delta)} = \frac{0\cdot0.5875+\frac{1}{2}\cdot0.6375+1\cdot0.5875}{1.5} = \frac{0.3188+0.5875}{1.5} = 0.6042,$$
$$I^{(\delta)} = \frac{0\cdot0.2325+\frac{1}{2}\cdot0.2200+1\cdot0.2550}{1.5} = \frac{0.1100+0.2550}{1.5} = 0.2433,$$
$$F^{(\delta)} = \frac{0\cdot0.4125+\frac{1}{2}\cdot0.3625+1\cdot0.3958}{1.5} = \frac{0.1813+0.3958}{1.5} = 0.3958.$$

Therefore, under the dominant "high frequency" context, the order-2 multi-neutrosophic outputs are: Reliability $(T,I,F) \approx (0.7717,\ 0.1617,\ 0.2425)$ and Low-crowding $(T,I,F) \approx (0.6042,\ 0.2433,\ 0.3958)$.

## 3.9 Interval-Valued Plithogenic Set

An Interval-Valued Plithogenic Set assigns each object interval-valued memberships and contradiction degrees, handling uncertainty with bounded ranges instead of single numeric values [588].

**Definition 3.9.1** (Interval–Valued Plithogenic Set)**.** [588] Let $S$ be a universal set and let $P \subseteq S$ be a nonempty subset. Let $v$ be an attribute whose set of possible values is $Pv$. Fix two positive integers $s \geq 1$ and $t \geq 1$. Denote by

$$\mathcal{I}([0,1]) \ := \ \{\, [\ell,u] \subseteq [0,1] \mid 0 \leq \ell \leq u \leq 1 \,\}$$

the family of all closed subintervals of $[0,1]$.

An *Interval–Valued Plithogenic Set* (IVPS) of dimension $(s,t)$ is a tuple

$$PS_{\mathrm{IV}}^{(s,t)} \ = \ \big(P,\ v,\ Pv,\ pdf_{\mathrm{IV}},\ pCF_{\mathrm{IV}}\big),$$

where

1. the *interval–valued degree of appurtenance function* (interval–valued DAF)

$$pdf_{\mathrm{IV}} : P \times Pv \longrightarrow \big(\mathcal{I}([0,1])\big)^s$$

assigns to every pair $(x,a) \in P \times Pv$ an $s$–tuple of membership–intervals

$$pdf_{\mathrm{IV}}(x,a) \ = \ \big([\alpha_1^L(x,a), \alpha_1^U(x,a)], \ldots, [\alpha_s^L(x,a), \alpha_s^U(x,a)]\big),$$

where $0 \leq \alpha_i^L(x,a) \leq \alpha_i^U(x,a) \leq 1$ for all $i = 1, \ldots, s$;

2. the *interval–valued degree of contradiction function* (interval–valued DCF)

$$pCF_{\mathrm{IV}} : Pv \times Pv \longrightarrow \big(\mathcal{I}([0,1])\big)^t$$

assigns to every pair $(a,b) \in Pv \times Pv$ a $t$–tuple of contradiction–intervals

$$pCF_{\mathrm{IV}}(a,b) \;=\; \big([\beta_1^L(a,b), \beta_1^U(a,b)], \dots, [\beta_t^L(a,b), \beta_t^U(a,b)]\big),$$

where $0 \le \beta_j^L(a,b) \le \beta_j^U(a,b) \le 1$ for all $j = 1,\dots,t$, and which satisfies the usual plithogenic axioms:

- (Reflexivity) for all $a \in Pv$,

$$pCF_{\mathrm{IV}}(a,a) \;=\; \big([0,0], \dots, [0,0]\big);$$

- (Symmetry) for all $a, b \in Pv$,

$$pCF_{\mathrm{IV}}(a,b) \;=\; pCF_{\mathrm{IV}}(b,a).$$

If, for every $(x,a)$ and every $(a,b)$, the intervals are degenerate, i.e.

$$[\alpha_i^L(x,a), \alpha_i^U(x,a)] = [\alpha_i(x,a), \alpha_i(x,a)], \qquad [\beta_j^L(a,b), \beta_j^U(a,b)] = [\beta_j(a,b), \beta_j(a,b)],$$

then $PS_{\mathrm{IV}}^{(s,t)}$ reduces to the classical (crisp–valued) plithogenic set of dimension $(s,t)$.

The contents of Table 3.15 describe how typical interval-valued fuzzy and neutrosophic families arise as special cases of the Interval-Valued Plithogenic Set (IVPS).

Table 3.15: Typical interval–valued fuzzy/neutrosophic families captured as special cases of the Interval–Valued Plithogenic Set (IVPS) by choosing $s$ equal to the number of interval components and $t = 0$ (no DCF) or $t \ge 1$ (DCF–aware).

| **Interval–valued set** | $s$ | $t$ | **Generalization by IVPS** |
|---|---|---|---|
| Interval–Valued Fuzzy Set [589–591] | 1 | 0/1 | Single interval membership is $pdf_{\mathrm{IV}}(\cdot,\cdot) \in \mathcal{I}([0,1])$; contradiction optional. |
| Interval–Valued Intuitionistic Fuzzy Set [592–595] (Interval-valued vague sets [596, 597]) | 2 | 0/1 | Two interval components (membership, nonmembership) become $s$=2 plithogenic intervals. |
| Interval valued pythagorean fuzzy sets [598, 599] | 2 | 0/1 | Two interval components become $s$=2 plithogenic intervals. |
| Interval–Valued Neutrosophic Set [600–603] | 3 | 0/1 | Three interval components (truth, indeterminacy, falsity) become $s$=3 plithogenic intervals. |
| Interval-valued picture fuzzy sets [604–606] | 3 | 0/1 | Three interval components become $s$=3 plithogenic intervals. |
| Interval-valued hesitant fuzzy sets [607, 608] | 3 | 0/1 | Three interval components become $s$=3 plithogenic intervals. |
| Interval-valued spherical fuzzy sets [609, 610] | 3 | 0/1 | Three interval components become $s$=3 plithogenic intervals. |
| Interval–Valued Quadripartitioned Neutrosophic Set [611, 612] | 4 | 0/1 | Four interval components embedded as $s$=4 in IVPS. |
| Interval–Valued Pentapartitioned Neutrosophic Set [9, 613, 614] | 5 | 0/1 | Five interval components embedded as $s$=5 in IVPS. |

We now present concrete examples of the proposed concept.

**Example 3.9.2** (E–commerce shipping choice — Interval–Valued Fuzzy Plithogenic (s=1, t=1))**.** E-commerce shipping choice is selecting delivery options balancing speed, cost, reliability, tracking, environmental impact, customer convenience, and overall price tradeoffs (cf. [615]).

Consider a product $x \in P$ to be recommended with respect to attribute $v =$ "shipping option" and $Pv = \{E = \text{Economy}, S = \text{Standard}, X = \text{Express}\}$. Let the interval–valued DAF (suitability) for $x$ be

$$pdf_{\text{IV}}(x, E) = [0.50, 0.70], \quad pdf_{\text{IV}}(x, S) = [0.60, 0.85], \quad pdf_{\text{IV}}(x, X) = [0.40, 0.65].$$

Assume an interval–valued DCF (larger means more contradictory) given by

$$pCF_{\text{IV}}(E, S) = [0.20, 0.30], \quad pCF_{\text{IV}}(X, S) = [0.30, 0.50],$$

$$pCF_{\text{IV}}(E, X) = [0.60, 0.80], \quad pCF_{\text{IV}}(a, a) = [0, 0].$$

Fix the dominant value $\delta = S$. Define compatibility intervals $w(a \mid \delta) := 1 - pCF_{\text{IV}}(a, \delta)$:

$$w(E \mid S) = [0.70, 0.80], \quad w(S \mid S) = [1, 1], \quad w(X \mid S) = [0.50, 0.70],$$

so the total weight interval is

$$W := \sum w = [0.70 + 1 + 0.50,\ 0.80 + 1 + 0.70] = [2.20,\ 2.50].$$

Using interval arithmetic with non–negative bounds: $[\alpha, \beta] \cdot [\gamma, \delta] = [\alpha\gamma, \beta\delta]$ and $[A, B] + [C, D] = [A + C, B + D]$, the numerator interval is

$$\begin{aligned} N &= w(E \mid S)\, pdf_{\text{IV}}(x, E) + w(S \mid S)\, pdf_{\text{IV}}(x, S) + w(X \mid S)\, pdf_{\text{IV}}(x, X) \\ &= [0.70, 0.80] \cdot [0.50, 0.70] + [1, 1] \cdot [0.60, 0.85] + [0.50, 0.70] \cdot [0.40, 0.65] \\ &= [0.35, 0.56] + [0.60, 0.85] + [0.20, 0.455] = [1.15, 1.865]. \end{aligned}$$

For division by a positive interval $[c, d]$ we use $[A, B]/[c, d] = [A/d, B/c]$. Hence the aggregated (Standard–relative) suitability interval is

$$\Gamma^{(S)}(x) = \frac{N}{W} = \left[\frac{1.15}{2.50}, \frac{1.865}{2.20}\right] = [0.4600,\ 0.8477].$$

Interpretation: under the dominant "Standard" context, $x$ has overall IV–fuzzy suitability in $[0.4600, 0.8477]$.

**Example 3.9.3** (Hiring decision (candidate $y$) — Interval–Valued Intuitionistic Plithogenic (s=2, t=1))**.** Hiring decision selects candidates based on skills, experience, cultural fit, and organizational needs, balancing fairness, cost, and long-term potential carefully (cf. [616]).

Let $v =$ "contract type" with $Pv = \{I = \text{Intern}, F = \text{Fixed-term}, P = \text{Permanent}\}$. For candidate $y$, take interval–valued intuitionistic appurtenances $(\mu, \nu)$:

$$\begin{aligned} pdf_{\text{IV}}(y, I) &= (\,[0.55, 0.75],\ [0.10, 0.25]\,), \\ pdf_{\text{IV}}(y, F) &= (\,[0.65, 0.80],\ [0.10, 0.20]\,), \\ pdf_{\text{IV}}(y, P) &= (\,[0.50, 0.65],\ [0.20, 0.35]\,), \end{aligned}$$

(each pair chosen so that $\max \mu + \max \nu \le 1$). Let the DCF intervals be

$$pCF_{\text{IV}}(I, P) = [0.50, 0.70], \quad pCF_{\text{IV}}(F, P) = [0.20, 0.30], \quad pCF_{\text{IV}}(a, a) = [0, 0],$$

and set the dominant value $\delta = P$. Compatibility intervals:

$$w(I \mid P) = [0.30, 0.50], \quad w(F \mid P) = [0.70, 0.80], \quad w(P \mid P) = [1, 1], \quad W = [2.00, 2.30].$$

Aggregating $\mu$ (membership) by interval arithmetic:

$$\begin{aligned} N_\mu &= [0.30, 0.50] \cdot [0.55, 0.75] + [0.70, 0.80] \cdot [0.65, 0.80] + [1, 1] \cdot [0.50, 0.65] \\ &= [0.165, 0.375] + [0.455, 0.640] + [0.50, 0.65] \\ &= [1.120, 1.665], \end{aligned}$$

$$\mu^{(P)}(y) = \frac{N_\mu}{W} = \Big[\frac{1.120}{2.30}, \frac{1.665}{2.00}\Big] = [0.4870,\ 0.8325].$$

Aggregating $\nu$ (nonmembership):

$$\begin{aligned} N_\nu &= [0.30, 0.50]\cdot[0.10, 0.25] + [0.70, 0.80]\cdot[0.10, 0.20] + [1, 1]\cdot[0.20, 0.35] \\ &= [0.030, 0.125] + [0.070, 0.160] + [0.20, 0.35] = [0.300, 0.635], \end{aligned}$$

$$\nu^{(P)}(y) = \frac{N_\nu}{W} = \Big[\frac{0.300}{2.30}, \frac{0.635}{2.00}\Big] = [0.1304,\ 0.3175].$$

Thus, relative to the dominant "Permanent" context, the candidate's intuitionistic IV assessment is

$$(\mu, \nu)^{(P)}(y) \in [0.4870, 0.8325] \times [0.1304, 0.3175].$$

**Example 3.9.4** (Precision agriculture irrigation — Interval–Valued Neutrosophic Plithogenic (s=3, t=1))**.** Agriculture irrigation supplies controlled water to crops through canals, sprinklers, or drip systems, improving yields, reducing drought risk and erosion (cf. [617]).

Let $v$ = "irrigation strategy" with $Pv = \{C = \text{Conservative},\ B = \text{Balanced},\ A = \text{Aggressive}\}$. For a plot $z$, take IV–neutrosophic triples $(T, I, F)$:

$$\begin{aligned} pdf_{\mathrm{IV}}(z, C) &= (\,[0.60, 0.80],\ [0.10, 0.25],\ [0.10, 0.20]\,), \\ pdf_{\mathrm{IV}}(z, B) &= (\,[0.70, 0.90],\ [0.05, 0.15],\ [0.05, 0.10]\,), \\ pdf_{\mathrm{IV}}(z, A) &= (\,[0.40, 0.60],\ [0.20, 0.30],\ [0.25, 0.40]\,). \end{aligned}$$

Assume DCF intervals to the target (balanced) policy:

$$pCF_{\mathrm{IV}}(C, B) = [0.10, 0.20], \quad pCF_{\mathrm{IV}}(A, B) = [0.30, 0.50], \quad pCF_{\mathrm{IV}}(B, B) = [0, 0].$$

Dominant value $\delta = B$. Compatibility intervals and total weight:

$$w(C\,|\,B) = [0.80, 0.90], \quad w(B\,|\,B) = [1, 1], \quad w(A\,|\,B) = [0.50, 0.70], \quad W = [2.30, 2.60].$$

*Truth* component:

$$\begin{aligned} N_T &= [0.80, 0.90]\cdot[0.60, 0.80] \ + \ [1, 1]\cdot[0.70, 0.90] \ + \ [0.50, 0.70]\cdot[0.40, 0.60] \\ &= [0.48, 0.72] + [0.70, 0.90] + [0.20, 0.42] = [1.38, 2.04], \\ T^{(B)}(z) &= \frac{N_T}{W} = \Big[\frac{1.38}{2.60}, \frac{2.04}{2.30}\Big] = [0.5308,\ 0.8869]. \end{aligned}$$

*Indeterminacy* component:

$$\begin{aligned} N_I &= [0.80, 0.90]\cdot[0.10, 0.25] + [1, 1]\cdot[0.05, 0.15] + [0.50, 0.70]\cdot[0.20, 0.30] \\ &= [0.08, 0.225] + [0.05, 0.15] + [0.10, 0.21] = [0.23, 0.585], \\ I^{(B)}(z) &= \frac{N_I}{W} = \Big[\frac{0.23}{2.60}, \frac{0.585}{2.30}\Big] = [0.0885,\ 0.2543]. \end{aligned}$$

*Falsity* component:

$$\begin{aligned} N_F &= [0.80, 0.90]\cdot[0.10, 0.20] + [1, 1]\cdot[0.05, 0.10] + [0.50, 0.70]\cdot[0.25, 0.40] \\ &= [0.08, 0.18] + [0.05, 0.10] + [0.125, 0.28] = [0.255, 0.56], \\ F^{(B)}(z) &= \frac{N_F}{W} = \Big[\frac{0.255}{2.60}, \frac{0.56}{2.30}\Big] = [0.0981,\ 0.2435]. \end{aligned}$$

Therefore, relative to the dominant "Balanced" policy, the plot's IV–neutrosophic irrigation assessment is

$$(T, I, F)^{(B)}(z) \in [0.5308, 0.8869] \times [0.0885, 0.2543] \times [0.0981, 0.2435].$$

## 3.10 Plithogenic OffSet

A plithogenic offset allows membership and contradiction degrees beyond $[0, 1]$, modeling negative and over-membership under attribute-based contradictions in complex systems [618–621].

**Definition 3.10.1** (Plithogenic Offset). [260,622] Let $S$ be a universal set, and $P \subseteq S$. A *Plithogenic Offset* $PS_{\text{offset}}$ is defined as:

$$PS_{\text{offset}} = (P, v, Pv, pdf, pCF),$$

where:

- $v$ is an attribute.
- $Pv$ is the set of possible values for the attribute $v$.
- $pdf : P \times Pv \to [\Psi_v, \Omega_v]^s$ is the *Degree of Appurtenance Function (DAF)*, where $\Psi_v < 0$ and $\Omega_v > 1$.
- $pCF : Pv \times Pv \to [\Psi_v, \Omega_v]^t$ is the *Degree of Contradiction Function (DCF)*.

In a Plithogenic Offset, membership degrees $pdf(x, a)$ range from below 0 (negative membership) to above 1 (over-membership), providing flexibility to model diverse membership states.

Table 3.16 presents the comparison of plithogenic sets, oversets, undersets, and offsets based on the range of their degree of appurtenance functions.

| Concept | Membership range of DAF *pdf* | Brief description |
|---|---|---|
| Plithogenic Set | $pdf : P \times Pv \to [0, 1]^s$ | Standard plithogenic model: truth / indeterminacy / falsity components in $[0, 1]$, with a contradiction function $pCF : Pv \times Pv \to [0, 1]^t$ controlling attribute–based aggregation. |
| Plithogenic OverSet | $pdf : P \times Pv \to (1, \Omega_v]^s$ with $\Omega_v > 1$ | Over–membership only: each component strictly larger than 1, up to an upper bound $\Omega_v$. Models redundancy, over–satisfaction, or "super–belonging" under plithogenic contradictions. |
| Plithogenic UnderSet | $pdf : P \times Pv \to [\Psi_v, 0)^s$ with $\Psi_v < 0$ | Negative–membership only: each component strictly below 0, down to a lower bound $\Psi_v$. Represents active opposition, counter–evidence, or anti–belonging in a plithogenic way. |
| Plithogenic OffSet | $pdf : P \times Pv \to [\Psi_v, \Omega_v]^s$ with $\Psi_v < 0 < 1 < \Omega_v$ | Full offset scale: a single plithogenic structure that simultaneously admits under–membership ($< 0$), classical membership ($[0, 1]$), and over–membership ($> 1$), together with a (possibly extended) contradiction function $pCF : Pv \times Pv \to [\Psi_v, \Omega_v]^t$. |

Table 3.16: Comparison of plithogenic set, overset, underset, and offset via the range of the degree of appurtenance function.

Table 3.17: Special cases of the plithogenic offset $PS_{\text{off}}(s,t)$ for $t = 1$.

| $s$ | $t$ | **Offset Type** |
|---|---|---|
| 1 | 1 | Plithogenic fuzzy offset (cf. [623,624]) (/ Plithogenic Grey sets (cf. [625–627]) ) |
| 2 | 1 | Plithogenic intuitionistic fuzzy offset (cf. [623,624]) |
| 3 | 1 | Plithogenic neutrosophic offset (cf. [623,628,629]) |
| 4 | 1 | Plithogenic quadripartitioned neutrosophic offset |
| 5 | 1 | Plithogenic pentapartitioned neutrosophic offset |
| 6 | 1 | Plithogenic hexapartitioned neutrosophic offset |
| 7 | 1 | Plithogenic heptapartitioned neutrosophic offset |
| 8 | 1 | Plithogenic octapartitioned neutrosophic offset |
| 9 | 1 | Plithogenic nonapartitioned neutrosophic offset |

The following table 3.17 summarizes important special cases of the plithogenic offset $PS_{\text{off}}(s,t)$ when $t = 1$. (For $t = 0$, each case reduces to the corresponding classical offset.)

We now present concrete examples of the proposed concept.

**Example 3.10.2** (Bank credit limit upgrade with plithogenic fuzzy offsets (fuzzy case, $s$=1, $t$=1))**.** Bank credit limit is the maximum amount a customer may borrow on a credit facility, set by risk assessment policies (cf. [630]).

Let $P = \{\text{cust}\}$ be a customer under review. The attribute is $v$ = "recommended credit limit tier" with $Pv = \{\text{Low}, \text{Medium}, \text{High}\}$. Because we allow offsets, the DAF can take values below 0 and above 1:

$$pdf(\text{cust}, \text{Low}) = -0.10, \quad pdf(\text{cust}, \text{Medium}) = 0.80, \quad pdf(\text{cust}, \text{High}) = 1.40.$$

Use a contradiction map $pCF : Pv \times Pv \to [0,1]$ (we keep $[0,1]$ for simplicity):

$$pCF(\text{Low}, \text{High}) = 0.85, \quad pCF(\text{Medium}, \text{High}) = 0.30, \quad pCF(\text{Low}, \text{Medium}) = 0.50, \quad pCF(a,a) = 0.$$

Fix the dominant context $\delta = \text{High}$ and set weights $w(a \mid \delta) = 1 - pCF(a, \delta)$. Then

$$w(\text{Low} \mid \text{High}) = 0.15, \quad w(\text{Medium} \mid \text{High}) = 0.70, \quad w(\text{High} \mid \text{High}) = 1.$$

The plithogenic aggregate relative to $\delta$ is the weighted mean

$$\Gamma^{(\text{High})}(\text{cust}) = \frac{0.15 \cdot (-0.10) + 0.70 \cdot 0.80 + 1 \cdot 1.40}{0.15 + 0.70 + 1} = \frac{-0.015 + 0.56 + 1.40}{1.85} = \frac{1.945}{1.85} \approx 1.0514.$$

Interpretation. Although there is weak *negative* evidence for "Low", strong over-membership for "High" and moderate support for "Medium" yield an *over-membership* aggregate slightly above 1, supporting an upgrade beyond the standard "High" threshold.

**Example 3.10.3** (Hospital triage with plithogenic intuitionistic fuzzy offsets (IF-offset, $s$=2, $t$=1))**.** Let $P = \{\text{pt}\}$ be a patient and $v$ = "triage category" with $Pv = \{\text{Immediate}, \text{Urgent}, \text{Routine}\}$. Use a two-component DAF $pdf(\cdot,\cdot) = (\mu, \nu)$ allowing offsets:

$$pdf(\text{pt}, \text{Immediate}) = (-0.10,\ 1.05),$$
$$pdf(\text{pt}, \text{Urgent}) = (1.10,\ 0.00),$$
$$pdf(\text{pt}, \text{Routine}) = (0.20,\ 0.95).$$

Here $\mu$ is the membership and $\nu$ the non-membership; both may be $< 0$ or $> 1$. Set contradictions

$$pCF(\text{Immediate}, \text{Urgent}) = 0.20, \quad pCF(\text{Routine}, \text{Urgent}) = 0.60, \quad pCF(a,a) = 0,$$

and choose $\delta = \text{Urgent}$. Weights $w = 1 - pCF$ give

$$w(\text{Immediate} \mid \text{Urgent}) = 0.80,$$

$$w(\text{Urgent} \mid \text{Urgent}) = 1,$$

$$w(\text{Routine} \mid \text{Urgent}) = 0.40,$$

$$W = 0.80 + 1 + 0.40 = 2.20.$$

Aggregate membership and non-membership separately:

$$\Gamma_{\mu}^{(\delta)}(\text{pt}) = \frac{0.80(-0.10) + 1(1.10) + 0.40(0.20)}{2.20} = \frac{-0.08 + 1.10 + 0.08}{2.20} = \frac{1.10}{2.20} = 0.50,$$

$$\Gamma_{\nu}^{(\delta)}(\text{pt}) = \frac{0.80(1.05) + 1(0.00) + 0.40(0.95)}{2.20} = \frac{0.84 + 0 + 0.38}{2.20} = \frac{1.22}{2.20} \approx 0.5545.$$

Interpretation. The raw offsets include an over-membership 1.10 for “Urgent” and a negative membership for “Immediate”. After contradiction-aware aggregation toward $\delta$ = Urgent, the final intuitionistic pair is $(\mu, \nu) \approx (0.50, 0.5545)$, reflecting balanced support with slightly higher aggregated hesitation/contradiction against urgency.

**Example 3.10.4** (Power-grid dispatch with plithogenic neutrosophic offsets (NS-offset, $s$=3, $t$=1))**.** Power-grid dispatch coordinates generation units in real time to balance supply and demand while respecting network constraints and reliability requirements (cf. [631]).

Consider $P = \{\text{hour}_\star\}$ (a specific operating hour) and $v$ = “primary energy source” with $Pv = \{\text{Coal}, \text{Gas}, \text{Solar}\}$. Use neutrosophic triples $pdf = (T, I, F)$ with offsets allowed:

$$pdf(\text{hour}_\star, \text{Coal}) = (-0.20,\ 0.10,\ 1.15),$$
$$pdf(\text{hour}_\star, \text{Gas}) = (0.80,\ 0.20,\ 0.20),$$
$$pdf(\text{hour}_\star, \text{Solar}) = (1.30,\ -0.05,\ 0.00).$$

Contradictions to a solar-dominant context $\delta$ = Solar are

$$pCF(\text{Coal}, \text{Solar}) = 0.85, \qquad pCF(\text{Gas}, \text{Solar}) = 0.40, \qquad pCF(a, a) = 0.$$

Weights $w = 1 - pCF$ yield

$$w(\text{Coal} \mid \text{Solar}) = 0.15, \quad w(\text{Gas} \mid \text{Solar}) = 0.60, \quad w(\text{Solar} \mid \text{Solar}) = 1, \quad W = 1.75.$$

Aggregate each neutrosophic component:

$$\Gamma_{T}^{(\delta)} = \frac{0.15(-0.20) + 0.60(0.80) + 1(1.30)}{1.75} = \frac{-0.03 + 0.48 + 1.30}{1.75} = \frac{1.75}{1.75} = 1.00,$$

$$\Gamma_{I}^{(\delta)} = \frac{0.15(0.10) + 0.60(0.20) + 1(-0.05)}{1.75} = \frac{0.015 + 0.12 - 0.05}{1.75} = \frac{0.085}{1.75} \approx 0.0486,$$

$$\Gamma_{F}^{(\delta)} = \frac{0.15(1.15) + 0.60(0.20) + 1(0.00)}{1.75} = \frac{0.1725 + 0.12 + 0}{1.75} = \frac{0.2925}{1.75} \approx 0.1671.$$

Interpretation. Offsets encode strong policy/subsidy preference for Solar ($T$=1.30) and counter-evidence for Coal ($T = -0.20$, $F$=1.15). The contradiction-aware neutrosophic aggregate under a solar-dominant context is $(T, I, F) \approx (1.00, 0.0486, 0.1671)$, i.e., decisive endorsement of Solar with very low indeterminacy and modest falsity.

## 3.11 Plithogenic Cubic Set

A Plithogenic Cubic Set assigns each element interval and crisp membership data, weighted by attribute-wise contradiction, unifying fuzzy, intuitionistic fuzzy, and neutrosophic cubic models [217, 484, 632–636].

**Definition 3.11.1** (Plithogenic Cubic Set)**.** [217, 632, 633] Let $A$ be a nonempty universe. Let an *attribute* $v$ take values in a set $R$ ("attribute value set"), equipped with a *degree of contradiction* function $C : R \times R \to [0,1]$ satisfying (i) $C(r,r) = 0$ and (ii) $C(r_1,r_2) = C(r_2,r_1)$. Write $\mathcal{I}([0,1]) = \{ [\alpha,\beta] \subseteq [0,1] \mid 0 \le \alpha \le \beta \le 1 \}$.

A *Plithogenic Cubic Set (PCS)* on $A$ is a collection of triples

$$\mathcal{P} = \{ \langle a,\ I(a),\ c(a) \rangle \mid a \in A \},$$

where $I(\cdot)$ is an *interval–valued plithogenic* appurtenance object and $c(\cdot)$ is its *point–valued* counterpart of the same type, chosen from one of the following canonical instances.

**(1) Plithogenic Fuzzy Cubic Set (PFCS).**

$$\Gamma = \langle X, \mu \rangle, \qquad X : A \to \mathcal{I}([0,1]),\ X(a) = [X^-(a), X^+(a)], \qquad \mu : A \to [0,1],$$

with $0 \le X^-(a) \le X^+(a) \le 1$ for all $a \in A$. Here $X$ is an *interval–valued plithogenic fuzzy set* (its interval endpoints arise from plithogenic appurtenance aggregation driven by $C$ and, optionally, a dominant value in $R$), while $\mu$ is an ordinary fuzzy membership.

**(2) Plithogenic Intuitionistic Fuzzy Cubic Set (PIFCS).**

$$\Psi = \langle Y, \delta \rangle, \qquad Y : A \to \mathcal{I}([0,1]) \times \mathcal{I}([0,1]),$$

$$Y(a) = \big([\mu^-(a), \mu^+(a)],\ [\nu^-(a), \nu^+(a)]\big), \qquad \delta(a) = (\mu(a), \nu(a)) \in [0,1]^2,$$

subject to $0 \le \mu^- \le \mu^+ \le 1$, $0 \le \nu^- \le \nu^+ \le 1$, and the intuitionistic constraint $\mu^+(a) + \nu^+(a) \le 1$ for all $a \in A$. Here $Y$ is an *interval–valued plithogenic intuitionistic fuzzy set* and $\delta$ is an intuitionistic fuzzy membership pair.

**(3) Plithogenic Neutrosophic Cubic Set (PNCS).**

$$\Omega = \langle Z, \lambda \rangle, \qquad Z : A \to \mathcal{I}([0,1])^3, \quad Z(a) = \big([T^-(a), T^+(a)],\ [I^-(a), I^+(a)],\ [F^-(a), F^+(a)]\big),$$

$$\lambda(a) = (T(a), I(a), F(a)) \in [0,1]^3,$$

with $0 \le T^- \le T^+ \le 1$, $0 \le I^- \le I^+ \le 1$, $0 \le F^- \le F^+ \le 1$ for all $a \in A$. Here $Z$ is an *interval–valued plithogenic neutrosophic set* and $\lambda$ is a (single–valued) neutrosophic triple.

We now present concrete examples of the proposed concept.

**Example 3.11.2** (E–commerce targeting with a Plithogenic Fuzzy Cubic Set (PFCS))**.** Let $A = \{\text{item}_\star\}$ be a specific product. The attribute is $v$ = "customer segment" with value set $R = \{\text{Budget}, \text{Standard}, \text{Premium}\}$. Let the contradiction $C : R \times R \to [0,1]$ be symmetric with

$$C(\text{Budget}, \text{Premium}) = 0.80, \quad C(\text{Standard}, \text{Premium}) = 0.30,$$

$$C(\text{Budget}, \text{Standard}) = 0.40, \quad C(r,r) = 0.$$

Fix the dominant segment $r^* = \text{Premium}$ and set compatibility weights $w(r \mid r^*) = 1 - C(r, r^*)$, hence

$$w(\text{Budget} \mid r^*) = 0.20, \quad w(\text{Standard} \mid r^*) = 0.70,$$

$$w(\text{Premium} \mid r^*) = 1.00, \quad W = 0.20 + 0.70 + 1.00 = 1.90.$$

Table 3.18: Classical cubic set families as special cases of the Plithogenic Cubic Set

| Classical cubic set | Interval (set–valued) component | Point (crisp) component | Generalization inside Plithogenic Cubic |
|---|---|---|---|
| Fuzzy cubic set | $X(a) = [X^-(a), X^+(a)] \subseteq [0,1]$ | $\mu(a) \in [0,1]$ | Obtain PFCS by letting the interval endpoints arise from plithogenic aggregation over attribute values and contradiction $C(\cdot,\cdot)$; the fuzzy cubic set is the case with no (or trivial) contradiction and a single attribute value. |
| Intuitionistic fuzzy cubic set | $Y(a) = ([\mu^-(a), \mu^+(a)], [\nu^-(a), \nu^+(a)])$ with $\mu^+(a) + \nu^+(a) \leq 1$ | $(\mu(a), \nu(a))$ with $\mu(a) + \nu(a) \leq 1$ | Obtain PIFCS by generating both interval pairs through a plithogenic DAF driven by an attribute $v$ and a DCF $C$; the classical intuitionistic fuzzy cubic set is recovered when $C \equiv 0$ and there is only one attribute value. |
| Neutrosophic cubic set [637–642] | $Z(a) = ([T^-, T^+], [I^-, I^+], [F^-, F^+]) \subseteq [0,1]^3$ | $(T(a), I(a), F(a)) \in [0,1]^3$ | Obtain PNCS by letting each neutrosophic interval come from plithogenic fusion against a dominant value and contradiction $C$; the ordinary neutrosophic cubic set is the specialization with $C \equiv 0$ (or a single value of $v$). |

Experts supply segment–wise fuzzy intervals and crisp memberships for $\text{item}_\star$:

| $r$ | $[X^-(r), X^+(r)]$ | $\mu(r)$ |
|---|---|---|
| Budget | $[0.30, 0.55]$ | 0.50 |
| Standard | $[0.60, 0.80]$ | 0.70 |
| Premium | $[0.85, 0.95]$ | 0.90 |

Aggregate interval endpoints and crisp membership by a weighted mean:

$$X^-(\text{item}_\star) = \frac{0.20 \cdot 0.30 + 0.70 \cdot 0.60 + 1.00 \cdot 0.85}{1.90} = \frac{0.06 + 0.42 + 0.85}{1.90} = \frac{1.33}{1.90} = 0.7000,$$

$$X^+(\text{item}_\star) = \frac{0.20 \cdot 0.55 + 0.70 \cdot 0.80 + 1.00 \cdot 0.95}{1.90} = \frac{0.11 + 0.56 + 0.95}{1.90} = \frac{1.62}{1.90} \approx 0.8526,$$

$$\mu(\text{item}_\star) = \frac{0.20 \cdot 0.50 + 0.70 \cdot 0.70 + 1.00 \cdot 0.90}{1.90} = \frac{0.10 + 0.49 + 0.90}{1.90} = \frac{1.49}{1.90} \approx 0.7842.$$

Therefore the PFCS triple is

$$\langle \text{item}_\star,\ [0.7000,\ 0.8526],\ 0.7842 \rangle.$$

Interpretation: contradiction–aware weighting toward the dominant "Premium" segment tightens the interval near high affinity and raises the crisp membership.

**Example 3.11.3** (Scholarship selection with a Plithogenic Intuitionistic Fuzzy Cubic Set (PIFCS))**.** Scholarship selection evaluates applicants' academic merit, research potential, financial need, and community impact to allocate limited educational available funding fairly (cf. [643]).

Let $A = \{\text{applicant}_\star\}$ and $v =$ "evidence source" with $R = \{\text{Grades, Research, Community}\}$. Choose $r^* = \text{Research}$. Let the symmetric contradiction be

$$C(\text{Grades, Research}) = 0.25, \quad C(\text{Community, Research}) = 0.40, \quad C(r, r) = 0,$$

so that $w(\text{Grades} \mid r^*) = 0.75$, $w(\text{Community} \mid r^*) = 0.60$, $w(\text{Research} \mid r^*) = 1.00$, and $W = 2.35$. Each source provides an intuitionistic interval pair and a crisp pair $(\mu, \nu)$:

| $r$ | $[\mu^-(r), \mu^+(r)]$ | $[\nu^-(r), \nu^+(r)]$ | $(\mu(r), \nu(r))$ |
|---|---|---|---|
| Grades | $[0.65, 0.80]$ | $[0.10, 0.20]$ | $(0.72, 0.18)$ |
| Research | $[0.75, 0.92]$ | $[0.03, 0.10]$ | $(0.85, 0.08)$ |
| Community | $[0.50, 0.70]$ | $[0.10, 0.20]$ | $(0.60, 0.18)$ |

Contradiction–weighted aggregation (componentwise) yields

$$\mu^-(\text{app}_\star) = \frac{0.75 \cdot 0.65 + 1.00 \cdot 0.75 + 0.60 \cdot 0.50}{2.35} = \frac{0.4875 + 0.75 + 0.30}{2.35} = \frac{1.5375}{2.35} \approx 0.6543,$$

$$\mu^+(\text{app}_\star) = \frac{0.75 \cdot 0.80 + 1.00 \cdot 0.92 + 0.60 \cdot 0.70}{2.35} = \frac{0.60 + 0.92 + 0.42}{2.35} = \frac{1.94}{2.35} \approx 0.8255,$$

$$\nu^-(\text{app}_\star) = \frac{0.75 \cdot 0.10 + 1.00 \cdot 0.03 + 0.60 \cdot 0.10}{2.35} = \frac{0.075 + 0.03 + 0.06}{2.35} = \frac{0.165}{2.35} \approx 0.0702,$$

$$\nu^+(\text{app}_\star) = \frac{0.75 \cdot 0.20 + 1.00 \cdot 0.10 + 0.60 \cdot 0.20}{2.35} = \frac{0.15 + 0.10 + 0.12}{2.35} = \frac{0.37}{2.35} \approx 0.1574,$$

which respects the intuitionistic bound $\mu^+ + \nu^+ \le 1$. For the crisp pair,

$$\mu(\text{app}_\star) = \frac{0.75 \cdot 0.72 + 1.00 \cdot 0.85 + 0.60 \cdot 0.60}{2.35} = \frac{0.54 + 0.85 + 0.36}{2.35} = \frac{1.75}{2.35} \approx 0.7447,$$

$$\nu(\text{app}_\star) = \frac{0.75 \cdot 0.18 + 1.00 \cdot 0.08 + 0.60 \cdot 0.18}{2.35} = \frac{0.135 + 0.08 + 0.108}{2.35} = \frac{0.323}{2.35} \approx 0.1372.$$

Hence the PIFCS triple is

$$\Big\langle \text{applicant}_\star,\ ([\mu^-, \mu^+], [\nu^-, \nu^+]) = ([0.6543, 0.8255], [0.0702, 0.1574]),\ (\mu, \nu) = (0.7447, 0.1372) \Big\rangle.$$

Interpretation: emphasizing research reduces disagreement and concentrates the interval toward eligibility while keeping nonmembership low.

**Example 3.11.4** (Urban bike-station siting with a Plithogenic Neutrosophic Cubic Set (PNCS))**.** Let $A = \{\text{site}_\star\}$ and $v =$ "planning criterion" with $R = \{\text{Demand, Safety, Cost}\}$. Choose $r^* = \text{Demand}$. Define a symmetric contradiction

$$C(\text{Safety, Demand}) = 0.30, \quad C(\text{Cost, Demand}) = 0.60, \quad C(r, r) = 0,$$

so that $w(\text{Safety} \mid r^*) = 0.70$, $w(\text{Cost} \mid r^*) = 0.40$, $w(\text{Demand} \mid r^*) = 1.00$, with $W = 2.10$. Criterion–wise neutrosophic intervals and crisp triples $(T, I, F)$ are:

| $r$ | $[T^-(r), T^+(r)]$ | $[I^-(r), I^+(r)]$ | $[F^-(r), F^+(r)]$ |
|---|---|---|---|
| Demand | $[0.80, 0.95]$ | $[0.03, 0.10]$ | $[0.02, 0.07]$ |
| Safety | $[0.70, 0.88]$ | $[0.05, 0.12]$ | $[0.07, 0.15]$ |
| Cost | $[0.40, 0.60]$ | $[0.10, 0.20]$ | $[0.30, 0.45]$ |

| $r$ | $(T(r), I(r), F(r))$ |
|---|---|
| Demand | $(0.90, 0.06, 0.04)$ |
| Safety | $(0.80, 0.09, 0.11)$ |
| Cost | $(0.50, 0.15, 0.35)$ |

Aggregate (componentwise) with the weights:

$$T^-(\text{site}_\star) = \frac{1 \cdot 0.80 + 0.70 \cdot 0.70 + 0.40 \cdot 0.40}{2.10} = \frac{0.80 + 0.49 + 0.16}{2.10} = \frac{1.45}{2.10} \approx 0.6905,$$

$$T^{+}(\text{site}_{\star}) = \frac{1 \cdot 0.95 + 0.70 \cdot 0.88 + 0.40 \cdot 0.60}{2.10} = \frac{0.95 + 0.616 + 0.24}{2.10} = \frac{1.806}{2.10} = 0.8600,$$
$$I^{-}(\text{site}_{\star}) = \frac{1 \cdot 0.03 + 0.70 \cdot 0.05 + 0.40 \cdot 0.10}{2.10} = \frac{0.03 + 0.035 + 0.04}{2.10} = \frac{0.105}{2.10} = 0.0500,$$
$$I^{+}(\text{site}_{\star}) = \frac{1 \cdot 0.10 + 0.70 \cdot 0.12 + 0.40 \cdot 0.20}{2.10} = \frac{0.10 + 0.084 + 0.08}{2.10} = \frac{0.264}{2.10} \approx 0.1257,$$
$$F^{-}(\text{site}_{\star}) = \frac{1 \cdot 0.02 + 0.70 \cdot 0.07 + 0.40 \cdot 0.30}{2.10} = \frac{0.02 + 0.049 + 0.12}{2.10} = \frac{0.189}{2.10} = 0.0900,$$
$$F^{+}(\text{site}_{\star}) = \frac{1 \cdot 0.07 + 0.70 \cdot 0.15 + 0.40 \cdot 0.45}{2.10} = \frac{0.07 + 0.105 + 0.18}{2.10} = \frac{0.355}{2.10} \approx 0.1690.$$

For the crisp neutrosophic triple:

$$T(\text{site}_{\star}) = \frac{1 \cdot 0.90 + 0.70 \cdot 0.80 + 0.40 \cdot 0.50}{2.10} = \frac{0.90 + 0.56 + 0.20}{2.10} = \frac{1.66}{2.10} \approx 0.7905,$$
$$I(\text{site}_{\star}) = \frac{1 \cdot 0.06 + 0.70 \cdot 0.09 + 0.40 \cdot 0.15}{2.10} = \frac{0.06 + 0.063 + 0.06}{2.10} = \frac{0.183}{2.10} \approx 0.0871,$$
$$F(\text{site}_{\star}) = \frac{1 \cdot 0.04 + 0.70 \cdot 0.11 + 0.40 \cdot 0.35}{2.10} = \frac{0.04 + 0.077 + 0.14}{2.10} = \frac{0.257}{2.10} \approx 0.1224.$$

Thus the PNCS triple is

$$\big\langle \text{site}_{\star},\ ([0.6905, 0.8600], [0.0500, 0.1257], [0.0900, 0.1690]),\ (0.7905, 0.0871, 0.1224) \big\rangle.$$

Interpretation: prioritizing demand yields high truth and low indeterminacy about suitability; modest falsity persists due to cost and safety trade–offs.

## 3.12 Plithogenic Soft, HyperSoft, and SuperHyperSoft Set

Recall that soft set represents objects by parameterized approximate subsets, enabling flexible, simple modeling of uncertainty in decision-making contexts under vague information [307,644]. HyperSoft set extends soft sets to multi-attribute, multi-valued parameter configurations, capturing higher-order, context-dependent uncertainties in datasets with structured aggregation rules [311,312,645,646]. SuperHyperSoft set iterates powerset-based soft constructions, linking hyper-parameters across levels to encode hierarchical, multilayer, interacting uncertainties over complex object families [330, 647, 648]. Plithogenic SuperHyperSoft set augments superhypersoft structures with contradiction-aware membership functions, modeling conflicting multi-attribute information across hierarchical parameters in realistic systems [649].

**Definition 3.12.1** (Plithogenic SuperHyperSoft Set)**.** Let $U$ be a universe and let $a_1, \dots, a_n$ be $n \geq 1$ distinct attributes. For each $i$, let $A_i$ be a finite, pairwise–disjoint set of values ($A_i \cap A_j = \emptyset$ for $i \neq j$), and write

$$\mathcal{C} \ = \ \mathcal{P}(A_1) \times \cdots \times \mathcal{P}(A_n).$$

Fix a nonempty $Y \subseteq U$. A *Plithogenic SuperHyperSoft Set* (PSHSS) over $U$ is the data

$$\text{PSHSS} \ = \ \big(U,\ Y,\ \{a_i\}_{i=1}^{n},\ \{A_i\}_{i=1}^{n},\ \mathcal{C},\ d,\ c\big),$$

together with a mapping $F : \mathcal{C} \to \mathcal{P}(U)$, where:

- $d : Y \times \mathcal{C} \to \mathcal{P}\big([0,1]^j\big)$ is the (possibly multi–valued) *appurtenance* function, with $j \in \{1,2,3\}$ indicating fuzzy ($j$=1), intuitionistic fuzzy ($j$=2), or neutrosophic ($j$=3) membership codomain;
- for each attribute $a_i$, $c : \mathcal{P}(A_i) \times \mathcal{P}(A_i) \to \mathcal{P}\big([0,1]^j\big)$ is a (possibly multi–valued) *contradiction* function satisfying reflexivity and symmetry:

$$c(\alpha,\alpha) = \{0\}, \qquad c(\alpha,\beta) = c(\beta,\alpha) \quad (\alpha,\beta \subseteq A_i);$$

- the image $F(\gamma) \subseteq U$ for $\gamma = (\alpha_1, \ldots, \alpha_n) \in \mathcal{C}$ is determined plithogenically from $\{d_y(\gamma)\}_{y \in Y}$ together with the contradictions $\{c(\alpha_i, \alpha_i), c(\alpha_i, \alpha_{i'})\}$ via the chosen plithogenic aggregation/composition rules.

Table 3.20 presents the reductions from a PSHSS to Plithogenic HyperSoft and Plithogenic Soft Sets. In addition, Table 3.19 summarizes the relationships between a PSHSS and the Fuzzy SuperHyperSoft Set, Intuitionistic Fuzzy SuperHyperSoft Set, and Neutrosophic SuperHyperSoft Set.

Table 3.19: Plithogenic SuperHyperSoft Set (PSHSS) as a common generalization of Fuzzy, Intuitionistic Fuzzy, Neutrosophic, and partitioned Neutrosophic SuperHyperSoft Sets.

| Model | Codomain of $d$ | $c$ | Specialization of PSHSS |
|---|---|---|---|
| Plithogenic SuperHyperSoft Set (PSHSS) | $\mathcal{P}([0,1]^j)$, $j \in \{1, \ldots, 5\}$ | reflexive, symmetric $c$ | Base model |
| Fuzzy SuperHyperSoft Set [650] | $\mathcal{P}([0,1])$ | $c(\alpha,\beta) \equiv \{0\}$ | $j = 1$, $c \equiv \{0\}$ |
| Intuitionistic Fuzzy SuperHyperSoft Set | $\mathcal{P}([0,1]^2)$ | $c(\alpha,\beta) \equiv \{0\}$ | $j = 2$, $c \equiv \{0\}$ |
| Neutrosophic SuperHyperSoft Set [651–654] | $\mathcal{P}([0,1]^3)$ | $c(\alpha,\beta) \equiv \{0\}$ | $j = 3$, $c \equiv \{0\}$ |
| Quadripartitioned Neutrosophic SuperHyperSoft Set | $\mathcal{P}([0,1]^4)$ | $c(\alpha,\beta) \equiv \{0\}$ | $j = 4$, $c \equiv \{0\}$ |
| Pentapartitioned Neutrosophic SuperHyperSoft Set | $\mathcal{P}([0,1]^5)$ | $c(\alpha,\beta) \equiv \{0\}$ | $j = 5$, $c \equiv \{0\}$ |

A concrete example of this concept is provided below.

**Example 3.12.2** (Hospital bed allocation (neutrosophic PSHSS, $j = 3$))**.** Hospital bed allocation is the process where hospitals assign limited beds to patients to optimize care, reduce delays, manage capacity, and ensure efficient treatment flow (cf. [691]).

Universe $U = \{\mathrm{p}_A, \mathrm{p}_B, \mathrm{p}_C\}$ (patients); focus subset $Y = \{\mathrm{p}_A, \mathrm{p}_B\}$. Attributes and value-sets (pairwise disjoint):

$$a_1 = \text{Severity},\ A_1 = \{\text{Mild}, \text{Moderate}, \text{Severe}\};$$

$$a_2 = \text{Resources},\ A_2 = \{\text{Low}, \text{High}\};$$

$$a_3 = \text{InfectionRisk},\ A_3 = \{\text{Low}, \text{High}\}.$$

Superhyper–parameter domain $\mathcal{C} = \mathcal{P}(A_1) \times \mathcal{P}(A_2) \times \mathcal{P}(A_3)$. Choose configuration

$$\gamma = (\{\text{Moderate}, \text{Severe}\},\ \{\text{High}\},\ \{\text{High}\})$$

and dominant configuration

$$\gamma^* = (\{\text{Severe}\},\ \{\text{High}\},\ \{\text{High}\}).$$

Element-level (symmetric, reflexive) contradictions $c_1, c_2, c_3 : A_i \times A_i \to [0,1]$:

$$\begin{aligned}
&c_1(\text{Mild}, \text{Severe}) = 0.90, \quad c_1(\text{Moderate}, \text{Severe}) = 0.30, \quad c_1(x,x) = 0;\\
&c_2(\text{Low}, \text{High}) = 0.80, \quad c_2(x,x) = 0;\\
&c_3(\text{Low}, \text{High}) = 0.90, \quad c_3(x,x) = 0.
\end{aligned}$$

Lift to subset–level via the max–min Hausdorff–type rule

$$\widehat{c}_i(H,K) := \max\Big\{ \max_{h \in H} \min_{k \in K} c_i(h,k),\ \max_{k \in K} \min_{h \in H} c_i(h,k) \Big\}.$$

Then

$$\widehat{c}_1(\{\text{Mod}, \text{Sev}\}, \{\text{Sev}\}) = \max\{\, c_1(\text{Mod}, \text{Sev}),\ c_1(\text{Sev}, \text{Sev})\,\} = \max\{0.30, 0\} = 0.30,$$

Table 3.20: PSHSS reductions to Plithogenic HyperSoft / Plithogenic Soft Sets and their fuzzy, intuitionistic fuzzy, and neutrosophic specializations.

| Target structure | Domain restriction | Function restriction | Outcome |
|---|---|---|---|
| **Plithogenic HyperSoft Set (PHS [655–659]** | Replace the power–set domain by singletons: $\mathcal{C}' = A_1 \times \cdots \times A_n$ (identify $\{\omega_i\}$ with $\omega_i \in A_i$) | Restrict $d$ to $d\vert_{Y\times\mathcal{C}'} : Y \times \mathcal{C}' \to \mathcal{P}([0,1]^j)$; restrict $c$ from $\mathcal{P}(A_i)\times\mathcal{P}(A_i)$ to $A_i \times A_i$ via $c'(\omega,\omega') := c(\{\omega\},\{\omega'\})$ | Yields the standard PHS on $A_1 \times \cdots \times A_n$ |
| Plithogenic Soft Set (PSS) [206, 660, 661] | Take a single attribute ($n = 1$) and singletons: $\mathcal{C}'' = A_1$ | $d\vert_{Y\times A_1} : Y\times A_1 \to \mathcal{P}([0,1]^j)$; $c\vert_{A_1\times A_1}$ via $c'(\omega,\omega') := c(\{\omega\},\{\omega'\})$ | Reduces to a plithogenic soft set over parameter set $A_1$ |
| (From PSS) fuzzy / IF / neutrosophic soft | Same as above, single attribute, singletons | Choose $j = 1$ (fuzzy [662, 663] and shadowed [664, 665]), $j = 2$ (intuitionistic [604, 666] and vague [667, 668]), $j = 3$ (neutrosophic [669, 670], hesitant Fuzzy [671, 672], picture fuzzy [673, 674], spherical fuzzy [675, 676]), $j = 4$ (quadripartitioned neutrosophic [377, 677, 678] and Double-valued neutrosophic [679, 680]), $j = 5$ (pentapartitioned neutrosophic) [9, 681] and set $pCF \equiv 0$ | Recovers fuzzy soft, IF soft, neutrosophic soft |
| (From PHS) fuzzy / IF / neutrosophic hypersoft | Same as above, $\mathcal{C}' = A_1 \times \cdots \times A_n$ | Choose $j = 1, 2, 3$ respectively and set contradiction to zero or constant | Recovers fuzzy [682, 683]/IF [684, 685]/neutrosophic [422, 686](/hesitant fuzzy [687]/picture fuzzy [688, 689]/spherical fuzzy [690]) hypersoft set |

$$\widehat{c}_2(\{\text{High}\},\{\text{High}\}) = 0, \qquad \widehat{c}_3(\{\text{High}\},\{\text{High}\}) = 0.$$

Compatibility weights per attribute $w_i := 1 - \widehat{c}_i(\alpha_i, \alpha_i^*)$ and global weight

$$w := \prod_{i=1}^{3} w_i = (1 - 0.30)(1 - 0)(1 - 0) = 0.70.$$

Neutrosophic appurtenance (singleton image of $d$) for $\mathrm{p}_A$ at $\gamma$:

$$d_{\mathrm{p}_A}(\gamma) = \{(T, I, F)\} = \{(0.85,\ 0.10,\ 0.08)\}.$$

One admissible plithogenic selection score is

$$\mathrm{score}(\mathrm{p}_A \mid \gamma, \gamma^*) := w\,(T-F)-I = 0.70\,(0.85-0.08)-0.10 = 0.70\cdot 0.77-0.10 = 0.439-0.10 = 0.339.$$

With threshold $\tau = 0$, score $> 0$ so $\mathrm{p}_A \in F(\gamma)$ (assign bed). The same pipeline applies to $\mathrm{p}_B$, possibly yielding exclusion if the score falls below $\tau$ due to higher contradictions or lower truth.

**Example 3.12.3** (Retail assortment planning (fuzzy PSHSS, $j = 1$))**.** Retail assortment planning is the process where retailers determine optimal product mixes to meet customer demand, maximize sales, reduce stockouts, and improve profitability (cf. [692]).

Universe $U$ is the SKU catalog; focus item $s_\star \in Y \subseteq U$. Attributes and values:

$$a_1 = \text{Season},\ A_1 = \{\text{Spring}, \text{Summer}, \text{Autumn}, \text{Winter}\};$$

$$a_2 = \text{PriceBand},\ A_2 = \{\text{Low}, \text{Mid}, \text{High}\};$$
$$a_3 = \text{BrandStyle},\ A_3 = \{\text{Classic}, \text{Sport}, \text{Tech}\}.$$

Choose

$$\gamma = (\{\text{Summer}, \text{Autumn}\},\ \{\text{Mid}, \text{High}\},\ \{\text{Sport}, \text{Tech}\}),$$
$$\gamma^* = (\{\text{Autumn}\},\ \{\text{High}\},\ \{\text{Sport}\}).$$

Singleton contradictions (symmetric, reflexive), nonzero entries:

$$c_1(\text{Summer}, \text{Autumn}) = 0.20, \quad c_2(\text{Mid}, \text{High}) = 0.30, \quad c_3(\text{Tech}, \text{Sport}) = 0.40.$$

Consider four singleton triples inside $\gamma$:

$$\theta_1 = (\text{Autumn}, \text{High}, \text{Sport}),$$
$$\theta_2 = (\text{Summer}, \text{High}, \text{Sport}),$$
$$\theta_3 = (\text{Autumn}, \text{Mid}, \text{Sport}),$$
$$\theta_4 = (\text{Autumn}, \text{High}, \text{Tech}).$$

Per-triple compatibility $w(\theta) = \prod_{i=1}^{3} \big(1 - c_i(\theta_i, \theta_i^*)\big)$ with $\theta^* = (\text{Autumn}, \text{High}, \text{Sport})$ gives

$$w(\theta_1) = 1, \quad w(\theta_2) = (1 - 0.20) \cdot 1 \cdot 1 = 0.80,$$

$$w(\theta_3) = 1 \cdot (1 - 0.30) \cdot 1 = 0.70, \quad w(\theta_4) = 1 \cdot 1 \cdot (1 - 0.40) = 0.60.$$

Let the fuzzy memberships (from $d_{s_\star}$) be

$$\mu(\theta_1) = 0.86,\ \mu(\theta_2) = 0.70,\ \mu(\theta_3) = 0.75,\ \mu(\theta_4) = 0.68.$$

Normalize and aggregate:

$$W = \sum_{k=1}^{4} w(\theta_k) = 1.00 + 0.80 + 0.70 + 0.60 = 3.10,$$

$$\mu_{\text{agg}}(s_\star \mid \gamma, \gamma^*) = \frac{\sum_{k=1}^{4} w(\theta_k)\, \mu(\theta_k)}{W}$$
$$= \frac{1 \cdot 0.86 + 0.8 \cdot 0.70 + 0.7 \cdot 0.75 + 0.6 \cdot 0.68}{3.10} = \frac{2.353}{3.10} \approx 0.7584.$$

Decision: include $s_\star$ in the autumn–premium–sport capsule if $\mu_{\text{agg}} \geq 0.75$ (here borderline but acceptable under a 0.75 cap with managerial override).

**Example 3.12.4** (University timetabling (intuitionistic PSHSS, $j = 2$))**.** University timetabling is the process where institutions schedule courses, rooms, instructors, and times to avoid conflicts, optimize resources, and support students' learning (cf. [693]).

Universe $U$ is the set of candidate time–room slots; focus $y_\star \in Y \subseteq U$. Attributes and values:

$$a_1 = \text{Period},\ A_1 = \{\text{Morning}, \text{Afternoon}\};$$

$$a_2 = \text{RoomType},\ A_2 = \{\text{Lecture}, \text{Lab}\};$$
$$a_3 = \text{Day},\ A_3 = \{\text{Mon}, \text{Tue}\}.$$

Take

$$\gamma = (\{\text{Morning}, \text{Afternoon}\},$$
$$\{\text{Lecture}, \text{Lab}\},\ \{\text{Mon}, \text{Tue}\}),$$
$$\gamma^* = (\{\text{Morning}\},\ \{\text{Lecture}\},\ \{\text{Tue}\}).$$

Singleton contradictions (symmetric, reflexive):

$$c_1(\text{Afternoon}, \text{Morning}) = 0.40,$$

$$c_2(\text{Lab}, \text{Lecture}) = 0.50, \quad c_3(\text{Mon}, \text{Tue}) = 0.30.$$

Select four representative triples inside $\gamma$:

$$\theta_1 = (\text{Morning}, \text{Lecture}, \text{Tue}),\ \theta_2 = (\text{Afternoon}, \text{Lecture}, \text{Tue}),$$

$$\theta_3 = (\text{Morning}, \text{Lecture}, \text{Mon}),\ \theta_4 = (\text{Morning}, \text{Lab}, \text{Tue}).$$

Weights:

$$w(\theta_1) = 1, \quad w(\theta_2) = (1 - 0.40) = 0.60,$$

$$w(\theta_3) = (1 - 0.30) = 0.70, \quad w(\theta_4) = (1 - 0.50) = 0.50; \quad W = 2.80.$$

From $d_{y_\star}$ obtain intuitionistic pairs $(\mu, \nu)$:

$$(\mu, \nu)(\theta_1) = (0.85, 0.10),\ (\mu, \nu)(\theta_2) = (0.62, 0.25),\ (\mu, \nu)(\theta_3) = (0.70, 0.20),\ (\mu, \nu)(\theta_4) = (0.60, 0.22).$$

Weighted aggregation:

$$\mu_{\text{agg}} = \frac{1 \cdot 0.85 + 0.6 \cdot 0.62 + 0.7 \cdot 0.70 + 0.5 \cdot 0.60}{2.80} = \frac{0.85 + 0.372 + 0.49 + 0.30}{2.80} = \frac{2.012}{2.80} \approx 0.7186,$$

$$\nu_{\text{agg}} = \frac{1 \cdot 0.10 + 0.6 \cdot 0.25 + 0.7 \cdot 0.20 + 0.5 \cdot 0.22}{2.80} = \frac{0.10 + 0.15 + 0.14 + 0.11}{2.80} = \frac{0.50}{2.80} \approx 0.1786,$$

which satisfies $\mu_{\text{agg}} + \nu_{\text{agg}} \leq 1$. Decision rule (example): schedule $y_\star$ if $\mu_{\text{agg}} - \nu_{\text{agg}} \geq 0.50$; here $0.7186 - 0.1786 = 0.5400 \geq 0.50$, so $y_\star \in F(\gamma)$ (slot accepted).

## 3.13 Hesitant Plithogenic Set

A Hesitant Plithogenic Set assigns to each object and attribute value a *finite nonempty set* [588] of plithogenic membership vectors in $[0, 1]^s$, together with a contradiction function on the attribute value set. It unifies hesitant fuzzy [6] and hesitant neutrosophic sets as special cases by choosing $s = 1$ and $s = 3$, respectively, and by setting the contradiction degrees to zero.

**Notation 3.13.1** (Finite hesitation families)**.** *Let $X$ be a nonempty set. Denote by*

$$\mathcal{H}(X) \ := \ \big\{ H \subseteq X \ \big| \ H \neq \emptyset,\ H \text{ is finite} \big\}$$

*the family of all* finite nonempty *subsets of $X$. Elements of $\mathcal{H}(X)$ will be called* hesitation sets *over $X$.*

**Definition 3.13.2** (Hesitant Plithogenic Set)**.** Let $S$ be a universal set and let $P \subseteq S$ be a nonempty subset. Let $v$ be an attribute with value set $Pv$. Fix integers $s \geq 1$ and $t \geq 1$.

A *Hesitant Plithogenic Set* (HPS) of dimension $(s, t)$ is a tuple

$$PS_{\mathrm{H}}^{(s,t)} \ = \ \big(P,\ v,\ Pv,\ hpdf,\ pCF\big),$$

where

1. the *hesitant plithogenic degree of appurtenance function* (hesitant DAF)

$$hpdf : P \times Pv \longrightarrow \mathcal{H}\big([0, 1]^s\big)$$

   assigns to each pair $(x, a) \in P \times Pv$ a finite nonempty set

$$hpdf(x, a) \ = \ \big\{\alpha^{(1)}(x, a), \ldots, \alpha^{(m_{x,a})}(x, a)\big\} \subseteq [0, 1]^s,$$

   where $m_{x,a} \in \mathbb{N}$ and each vector $\alpha^{(r)}(x, a) = (\alpha_1^{(r)}(x, a), \ldots, \alpha_s^{(r)}(x, a))$ has components in $[0, 1]$;

2. the *degree of contradiction function* (DCF)

$$pCF : Pv \times Pv \longrightarrow [0,1]^t$$

satisfies the usual plithogenic axioms:

- (Reflexivity) for all $a \in Pv$,

$$pCF(a,a) = (0,\ldots,0) \in [0,1]^t;$$

- (Symmetry) for all $a, b \in Pv$,

$$pCF(a,b) = pCF(b,a).$$

When every hesitation set $hpdf(x,a)$ is a singleton, $PS_{\mathrm{H}}^{(s,t)}$ reduces to the classical (non-hesitant) plithogenic set of dimension $(s,t)$.

Table 3.21 presents a summary of the roles of the parameters $s$ and $t$ in hesitant plithogenic memberships.

Table 3.21: Roles of the parameters $s$ and $t$ in hesitant plithogenic memberships

| Parameter | Interpretation |
|---|---|
| $s = 1$ | Encodes hesitant fuzzy [6, 359]-type memberships (each hesitant element is a list of scalar membership degrees). |
| $s = 2$ | Encodes hesitant intuitionistic fuzzy [694,695]-type pairs (each hesitant element is a list of membership/nonmembership pairs). |
| $s = 3$ | Encodes hesitant neutrosophic triples $(T, I, F)$ [696,697] (each hesitant element is a list of truth, indeterminacy, and falsity triples). |
| $s = 3$ | Encodes hesitant picture fuzzy triples [698, 699]. |
| $s = 3$ | Encodes spherical hesitant fuzzy triples [700, 701]. |
| $s = 4$ | Encodes hesitant Quadripartitioned neutrosophic quadruple. |
| $s = 5$ | Encodes hesitant Pentapartitioned neutrosophic quintuple. |
| $t$ | Counts the number of contradiction components in $pCF(a,b) \in [0,1]^t$. The special case $pCF(a,b) \equiv (0,\ldots,0)$, for all attribute values $a, b$, recovers the usual hesitant structures without contradiction-awareness. |

A concrete example of this concept is provided below.

**Example 3.13.3** (Hesitant Plithogenic Set in Medical Diagnosis Triage)**.** Consider an emergency department that must triage incoming patients according to the severity of a suspected condition.

Let the universe of discourse be the set of patients

$$P = \{x_1, x_2\},$$

where $x_1$ is a young adult patient and $x_2$ is an elderly patient with comorbidities.

Let $v$ be the attribute "clinical severity", with value set

$$Pv = \{\text{mild},\ \text{moderate},\ \text{severe}\}.$$

We work with dimension $(s,t) = (3,1)$ so that each membership vector has the neutrosophic form $(T, I, F) \in [0,1]^3$, and there is a single contradiction component.

The degree of contradiction function $pCF : Pv \times Pv \to [0,1]$ is chosen as

$$pCF(a,b) = \begin{cases} 0, & a = b, \\ 0.4, & \{a,b\} = \{\text{mild}, \text{moderate}\}, \\ 0.6, & \{a,b\} = \{\text{moderate}, \text{severe}\}, \\ 0.9, & \{a,b\} = \{\text{mild}, \text{severe}\}, \end{cases}$$

and extended symmetrically, so that the values reflect how "far apart" the severity labels are.

For each patient $x \in P$ and each attribute value $a \in Pv$, the hesitant plithogenic degree of appurtenance

$$hpdf(x,a) \in \mathcal{H}([0,1]^3)$$

collects several candidate neutrosophic assessments $(T, I, F)$ from different physicians or diagnostic tools.

For the young adult patient $x_1$, suppose we have:

$$hpdf(x_1, \text{mild}) = \{(0.75,\ 0.15,\ 0.10),\ (0.68,\ 0.22,\ 0.10)\},$$

$$hpdf(x_1, \text{moderate}) = \{(0.40,\ 0.30,\ 0.30)\}, \qquad hpdf(x_1, \text{severe}) = \{(0.10,\ 0.20,\ 0.70)\}.$$

Here, for instance, $(0.75, 0.15, 0.10)$ means that one expert believes with truth 0.75 and falsity 0.10 (with indeterminacy 0.15) that the patient is "mild", while another expert provides the slightly different assessment $(0.68, 0.22, 0.10)$; the set of such vectors forms the hesitation set.

For the elderly patient $x_2$, assume:

$$hpdf(x_2, \text{mild}) = \{(0.20,\ 0.25,\ 0.55)\},$$

$$hpdf(x_2, \text{moderate}) = \{(0.55,\ 0.20,\ 0.25),\ (0.48,\ 0.32,\ 0.20)\},$$

$$hpdf(x_2, \text{severe}) = \{(0.35,\ 0.25,\ 0.40)\}.$$

The tuple

$$PS_{\mathrm{H}}^{(3,1)} = (P,\ v,\ Pv,\ hpdf,\ pCF)$$

is then a concrete Hesitant Plithogenic Set. It simultaneously captures:

- multiple competing neutrosophic opinions for each (patient, severity-label) pair; and
- the degrees of contradiction between different labels (mild, moderate, severe), which can be used in plithogenic aggregation rules (for example, giving more weight to labels with smaller contradiction against a chosen dominant label).

A triage policy can subsequently aggregate each hesitation set (e.g. via weighted averaging or ordered selection) while explicitly accounting for the contradiction values $pCF$ in order to prioritize treatments or allocate limited resources.

**Example 3.13.4** (Hesitant Plithogenic Set in Supplier Risk Assessment)**.** Consider a manufacturing company that evaluates suppliers according to their reliability in delivering critical components on time.

Let the universe be the set of candidate suppliers

$$P = \{s_1, s_2, s_3\}.$$

Let $v$ be the attribute "reliability class", with attribute value set

$$Pv = \{\text{high}, \ \text{medium}, \ \text{low}\}.$$

Here we work with dimension $(s, t) = (1, 1)$, so each membership vector is a single fuzzy-type grade $\alpha \in [0, 1]$, and there is a single contradiction component.

The contradiction function $pCF : Pv \times Pv \to [0, 1]$ is given by

$$pCF(a, b) = \begin{cases} 0, & a = b, \\ 0.6, & \{a, b\} = \{\text{high}, \text{medium}\}, \\ 0.7, & \{a, b\} = \{\text{medium}, \text{low}\}, \\ 0.95, & \{a, b\} = \{\text{high}, \text{low}\}, \end{cases}$$

extended symmetrically. Thus "high" and "low" reliability classes are almost maximally contradictory.

The hesitant plithogenic degree of appurtenance

$$hpdf : P \times Pv \longrightarrow \mathcal{H}([0, 1])$$

collects multiple fuzzy membership scores obtained from different departments (e.g. logistics, finance, and quality control).

For supplier $s_1$ (long-term partner with mostly on-time deliveries) we may have:

$$hpdf(s_1, \text{high}) = \{0.82, \ 0.88, \ 0.91\}, \qquad hpdf(s_1, \text{medium}) = \{0.10, \ 0.15\}, \qquad hpdf(s_1, \text{low}) = \{0.02\}.$$

Here, for example, 0.82 and 0.88 could come from historical on-time rates as assessed by two different analysts, while 0.91 comes from a recent predictive model.

For supplier $s_2$ (new supplier with limited track record), suppose:

$$hpdf(s_2, \text{high}) = \{0.40, \ 0.55\}, \qquad hpdf(s_2, \text{medium}) = \{0.45, \ 0.50, \ 0.60\}, \qquad hpdf(s_2, \text{low}) = \{0.15, \ 0.25\}.$$

The hesitation sets express the fact that different stakeholders hold somewhat conflicting views about $s_2$'s reliability.

For supplier $s_3$ (supplier with recurring delays), assume:

$$hpdf(s_3, \text{high}) = \{0.05\}, \qquad hpdf(s_3, \text{medium}) = \{0.25, \ 0.30\}, \qquad hpdf(s_3, \text{low}) = \{0.70, \ 0.80\}.$$

Thus the tuple

$$PS_{\mathrm{H}}^{(1,1)} = \big(P, \ v, \ Pv, \ hpdf, \ pCF\big)$$

is a Hesitant Plithogenic Set that models supplier reliability assessment.

A decision maker can choose a dominant attribute value (for example, "high" reliability) and perform plithogenic aggregation of each hesitation set $hpdf(s_i, a)$ by:

- first combining the hesitant fuzzy memberships inside each $hpdf(s_i, a)$ (e.g. by average, median, or optimistic/pessimistic selection); and
- then contradiction-weighting the resulting aggregated scores using $pCF(\text{high}, a)$ so that memberships coming from values more contradictory to "high" (such as "low") are penalized more strongly.

This provides a flexible, contradiction-aware mechanism to rank or screen suppliers under heterogeneous and hesitant expert opinions.

For completeness, we recall the standard hesitant fuzzy and hesitant neutrosophic sets in a form compatible with Definition 3.13.2.

**Definition 3.13.5** (Hesitant Fuzzy Set)**.** [359] Let $P$ be a nonempty universe. A *Hesitant Fuzzy Set* (HFS) on $P$ is a pair

$$H_{\mathrm{F}} = (P,\ h_{\mathrm{F}}),$$

where

$$h_{\mathrm{F}} : P \longrightarrow \mathcal{H}([0,1])$$

assigns to each $x \in P$ a finite nonempty set $h_{\mathrm{F}}(x) \subseteq [0,1]$ of possible membership degrees of $x$ in the set.

**Definition 3.13.6** (Hesitant Neutrosophic Set)**.** [697] Let $P$ be a nonempty universe. A *Hesitant Neutrosophic Set* (HNS) on $P$ is a pair

$$H_{\mathrm{N}} = (P,\ h_{\mathrm{N}}),$$

where

$$h_{\mathrm{N}} : P \longrightarrow \mathcal{H}\big([0,1]^3\big)$$

assigns to each $x \in P$ a finite nonempty set $h_{\mathrm{N}}(x) \subseteq [0,1]^3$ of neutrosophic triples

$$(T, I, F) \in [0,1]^3,$$

interpreted as hesitant degrees of truth, indeterminacy, and falsity for $x$.

We now show formally that the Hesitant Plithogenic Set simultaneously generalizes Hesitant Fuzzy Sets and Hesitant Neutrosophic Sets by suitable choices of $s$, $t$, and the attribute value set.

**Theorem 3.13.7** (Hesitant Plithogenic Set generalizes Hesitant Fuzzy Set)**.** *Fix a nonempty universe $P$. Consider the class of all Hesitant Fuzzy Sets on $P$,*

$$\mathsf{HFS}(P) \ := \ \big\{\, H_{\mathrm{F}} = (P, h_{\mathrm{F}}) \ \big|\ h_{\mathrm{F}} : P \to \mathcal{H}([0,1]) \,\big\},$$

*and the subclass of Hesitant Plithogenic Sets*

$$\mathsf{HPS}_{\mathrm{F}}(P) \ := \ \Big\{\, PS_{\mathrm{H}}^{(1,1)} = (P, v, Pv, hpdf, pCF) \ \Big|\ Pv = \{a_0\},\ pCF(a,b) \equiv 0 \,\Big\}.$$

*Then there exists a bijection*

$$\Phi_{\mathrm{F}} : \mathsf{HFS}(P) \longrightarrow \mathsf{HPS}_{\mathrm{F}}(P),$$

*so every Hesitant Fuzzy Set can be identified with a unique Hesitant Plithogenic Set with $s = 1$, $t = 1$, a single attribute value, and zero contradiction.*

*Proof.* Step 1. Construction of the map $\Phi_{\mathrm{F}} : \mathsf{HFS}(P) \to \mathsf{HPS}_{\mathrm{F}}(P)$.

Let $H_{\mathrm{F}} = (P, h_{\mathrm{F}}) \in \mathsf{HFS}(P)$ be arbitrary. Fix an attribute name $v$ and a singleton value set $Pv = \{a_0\}$. Define

$$pCF(a,b) := 0 \in [0,1] \qquad \text{for all } a, b \in Pv.$$

Clearly $pCF$ is reflexive and symmetric.

Define $hpdf : P \times Pv \to \mathcal{H}([0,1]^1)$ by

$$hpdf(x, a_0) \ := \ \big\{ (\alpha) \in [0,1]^1 \ \big| \ \alpha \in h_{\mathrm{F}}(x) \big\}, \qquad x \in P. \tag{3.2}$$

Since $h_{\mathrm{F}}(x)$ is finite and nonempty, the set on the right-hand side is also finite and nonempty, so $hpdf(x, a_0) \in \mathcal{H}([0,1]^1)$. Thus

$$PS_{\mathrm{H}}^{(1,1)}(H_{\mathrm{F}}) \ := \ \big(P,\ v,\ Pv,\ hpdf,\ pCF\big)$$

is a well-defined Hesitant Plithogenic Set of dimension $(s,t) = (1,1)$ with $Pv = \{a_0\}$ and $pCF \equiv 0$. We set

$$\Phi_{\mathrm{F}}(H_{\mathrm{F}}) \ := \ PS_{\mathrm{H}}^{(1,1)}(H_{\mathrm{F}}).$$

Step 2. Construction of the inverse map $\Psi_{\mathrm{F}} : \mathsf{HPS}_{\mathrm{F}}(P) \to \mathsf{HFS}(P)$.

Let

$$PS_{\mathrm{H}}^{(1,1)} = (P, v, \{a_0\}, hpdf, pCF) \in \mathsf{HPS}_{\mathrm{F}}(P)$$

be arbitrary. Define $h_{\mathrm{F}} : P \to \mathcal{H}([0,1])$ by

$$h_{\mathrm{F}}(x) \ := \ \big\{ \alpha \in [0,1] \ \big| \ (\alpha) \in hpdf(x, a_0) \big\}, \qquad x \in P. \tag{3.3}$$

Since each $hpdf(x, a_0)$ is finite and nonempty, the set of its first coordinates $h_{\mathrm{F}}(x)$ is also finite and nonempty, so $h_{\mathrm{F}}(x) \in \mathcal{H}([0,1])$. Hence $H_{\mathrm{F}} := (P, h_{\mathrm{F}})$ is a Hesitant Fuzzy Set on $P$. Define

$$\Psi_{\mathrm{F}}(PS_{\mathrm{H}}^{(1,1)}) \ := \ H_{\mathrm{F}}.$$

Step 3. Verification that $\Psi_{\mathrm{F}}$ and $\Phi_{\mathrm{F}}$ are mutual inverses.

(i) For $H_{\mathrm{F}} = (P, h_{\mathrm{F}}) \in \mathsf{HFS}(P)$ and $x \in P$, combine (3.2) and (3.3) to compute

$$\big(\Psi_{\mathrm{F}} \circ \Phi_{\mathrm{F}}\big)(H_{\mathrm{F}})(x) \ = \ \Psi_{\mathrm{F}}\big(PS_{\mathrm{H}}^{(1,1)}(H_{\mathrm{F}})\big)(x) \ = \ \big\{\alpha \in [0,1] \ \big| \ (\alpha) \in hpdf(x, a_0)\big\}.$$

By (3.2),

$$hpdf(x, a_0) \ = \ \big\{ (\beta) \in [0,1]^1 \ \big| \ \beta \in h_{\mathrm{F}}(x) \big\}.$$

Therefore

$$\big(\Psi_{\mathrm{F}} \circ \Phi_{\mathrm{F}}\big)(H_{\mathrm{F}})(x) \ = \ \Big\{\alpha \in [0,1] \ \Big| \ (\alpha) \in \big\{(\beta) \mid \beta \in h_{\mathrm{F}}(x)\big\}\Big\}.$$

The condition $(\alpha) \in \{(\beta) \mid \beta \in h_{\mathrm{F}}(x)\}$ is equivalent to $\exists \beta \in h_{\mathrm{F}}(x)$ with $(\alpha) = (\beta)$. Equality of 1-dimensional vectors $(\alpha) = (\beta)$ implies $\alpha = \beta$. Thus

$$\big(\Psi_{\mathrm{F}} \circ \Phi_{\mathrm{F}}\big)(H_{\mathrm{F}})(x) \ = \ \big\{\alpha \in [0,1] \ \big| \ \alpha \in h_{\mathrm{F}}(x)\big\} \ = \ h_{\mathrm{F}}(x).$$

Since this holds for all $x \in P$, we obtain

$$\Psi_{\mathrm{F}} \circ \Phi_{\mathrm{F}} = \mathrm{id}_{\mathsf{HFS}(P)}.$$

(ii) Conversely, let $PS_{\mathrm{H}}^{(1,1)} = (P, v, \{a_0\}, hpdf, pCF) \in \mathsf{HPS}_{\mathrm{F}}(P)$ and $x \in P$. Then $\Psi_{\mathrm{F}}(PS_{\mathrm{H}}^{(1,1)}) = (P, h_{\mathrm{F}})$ with $h_{\mathrm{F}}$ given by (3.3). Applying $\Phi_{\mathrm{F}}$ we obtain a new hesitant plithogenic DAF

$$hpdf'(x, a_0) \;:=\; \{\, (\alpha) \in [0,1]^1 \;\big|\; \alpha \in h_{\mathrm{F}}(x) \,\}.$$

Using (3.3), this becomes

$$hpdf'(x, a_0) \;=\; \Big\{ (\alpha) \in [0,1]^1 \;\Big|\; \alpha \in \{\beta \in [0,1] \;\big|\; (\beta) \in hpdf(x, a_0)\} \Big\}.$$

Thus

$$hpdf'(x, a_0) \;=\; \Big\{ (\alpha) \in [0,1]^1 \;\Big|\; \exists \beta \in [0,1] \text{ with } (\beta) \in hpdf(x, a_0) \text{ and } \alpha = \beta \Big\}.$$

But the condition "there exists $\beta$ with $(\beta) \in hpdf(x, a_0)$ and $\alpha = \beta$" is equivalent to "$(\alpha) \in hpdf(x, a_0)$". Hence

$$hpdf'(x, a_0) \;=\; \{\, (\alpha) \in [0,1]^1 \;\big|\; (\alpha) \in hpdf(x, a_0) \,\} \;=\; hpdf(x, a_0).$$

Therefore the DAFs of $PS_{\mathrm{H}}^{(1,1)}$ and $\Phi_{\mathrm{F}}\big(\Psi_{\mathrm{F}}(PS_{\mathrm{H}}^{(1,1)})\big)$ coincide, and the other components $(P, v, Pv, pCF)$ are unchanged by construction. Thus

$$\Phi_{\mathrm{F}} \circ \Psi_{\mathrm{F}} \;=\; \mathrm{id}_{\mathsf{HPS}_{\mathrm{F}}(P)}\,.$$

Step 4. Since $\Psi_{\mathrm{F}} \circ \Phi_{\mathrm{F}} = \mathrm{id}_{\mathsf{HFS}(P)}$ and $\Phi_{\mathrm{F}} \circ \Psi_{\mathrm{F}} = \mathrm{id}_{\mathsf{HPS}_{\mathrm{F}}(P)}$, the maps $\Phi_{\mathrm{F}}$ and $\Psi_{\mathrm{F}}$ are mutual inverses and hence $\Phi_{\mathrm{F}}$ is a bijection. This proves that every Hesitant Fuzzy Set is canonically realized as a Hesitant Plithogenic Set with $(s,t) = (1,1)$, a single attribute value, and zero contradiction. □

**Theorem 3.13.8** (Hesitant Plithogenic Set generalizes Hesitant Neutrosophic Set)**.** *Fix a nonempty universe $P$. Consider the class of all Hesitant Neutrosophic Sets on $P$,*

$$\mathsf{HNS}(P) \;:=\; \{\, H_{\mathrm{N}} = (P, h_{\mathrm{N}}) \;\big|\; h_{\mathrm{N}} : P \to \mathcal{H}([0,1]^3) \,\},$$

*and the subclass of Hesitant Plithogenic Sets*

$$\mathsf{HPS}_{\mathrm{N}}(P) \;:=\; \Big\{ PS_{\mathrm{H}}^{(3,1)} = (P, v, Pv, hpdf, pCF) \;\Big|\; Pv = \{a_0\},\; pCF(a,b) \equiv 0 \Big\}.$$

*Then there exists a bijection*

$$\Phi_{\mathrm{N}} : \mathsf{HNS}(P) \longrightarrow \mathsf{HPS}_{\mathrm{N}}(P),$$

*so every Hesitant Neutrosophic Set can be identified with a unique Hesitant Plithogenic Set with $s = 3$, $t = 1$, a single attribute value, and zero contradiction.*

*Proof.* The proof is parallel to Theorem 3.13.7, but now $s = 3$ and no change of dimension is needed.

Step 1. For $H_{\mathrm{N}} = (P, h_{\mathrm{N}}) \in \mathsf{HNS}(P)$, fix an attribute $v$, set $Pv = \{a_0\}$, and define $pCF(a,b) := 0$ for all $a, b \in Pv$. Define

$$hpdf(x, a_0) \;:=\; h_{\mathrm{N}}(x) \subseteq [0,1]^3, \qquad x \in P.$$

Since $h_{\mathrm{N}}(x)$ is finite and nonempty, this yields $hpdf(x, a_0) \in \mathcal{H}([0,1]^3)$, so

$$PS_{\mathrm{H}}^{(3,1)}(H_{\mathrm{N}}) \;:=\; \big(P,\; v,\; \{a_0\},\; hpdf,\; pCF\big) \in \mathsf{HPS}_{\mathrm{N}}(P).$$

Set $\Phi_{\mathrm{N}}(H_{\mathrm{N}}) := PS_{\mathrm{H}}^{(3,1)}(H_{\mathrm{N}})$.

Step 2. For $PS_{\mathrm{H}}^{(3,1)} = (P, v, \{a_0\}, hpdf, pCF) \in \mathsf{HPS}_{\mathrm{N}}(P)$, define $h_{\mathrm{N}} : P \to \mathcal{H}([0,1]^3)$ simply by

$$h_{\mathrm{N}}(x) := hpdf(x, a_0), \qquad x \in P,$$

and put $\Psi_{\mathrm{N}}(PS_{\mathrm{H}}^{(3,1)}) := (P, h_{\mathrm{N}}) \in \mathsf{HNS}(P)$.

Step 3. For any $H_{\mathrm{N}} = (P, h_{\mathrm{N}})$ and $x \in P$ we have

$$\big(\Psi_{\mathrm{N}} \circ \Phi_{\mathrm{N}}\big)(H_{\mathrm{N}})(x) = \Psi_{\mathrm{N}}\big(PS_{\mathrm{H}}^{(3,1)}(H_{\mathrm{N}})\big)(x) = hpdf(x, a_0) = h_{\mathrm{N}}(x),$$

so $\Psi_{\mathrm{N}} \circ \Phi_{\mathrm{N}} = \mathrm{id}_{\mathsf{HNS}(P)}$. Conversely, for any $PS_{\mathrm{H}}^{(3,1)}$ and $x \in P$ we have

$$\big(\Phi_{\mathrm{N}} \circ \Psi_{\mathrm{N}}\big)(PS_{\mathrm{H}}^{(3,1)})(x, a_0) = h_{\mathrm{N}}(x) = hpdf(x, a_0),$$

so the DAF is preserved and the remaining components $(P, v, Pv, pCF)$ are unchanged. Thus $\Phi_{\mathrm{N}} \circ \Psi_{\mathrm{N}} = \mathrm{id}_{\mathsf{HPS}_{\mathrm{N}}(P)}$.

Step 4. Hence $\Phi_{\mathrm{N}}$ is a bijection between $\mathsf{HNS}(P)$ and $\mathsf{HPS}_{\mathrm{N}}(P)$. This shows that every Hesitant Neutrosophic Set is realized as a particular Hesitant Plithogenic Set with $(s, t) = (3, 1)$, a single attribute value, and trivial contradiction. □

## 3.14 Spherical Plithogenic Sets

Spherical plithogenic sets model attribute-valued elements using neutrosophic triples constrained on a sphere, integrating contradiction degrees for complex uncertainty interactions. In this section we recall the standard notions of spherical fuzzy set and spherical neutrosophic set, then introduce the notion of a spherical plithogenic set, and finally prove that it generalizes both previous concepts.

**Definition 3.14.1** (Spherical Fuzzy Set)**.** [366,367,702] Let $U$ be a nonempty universe. A *spherical fuzzy set* $A$ on $U$ is determined by three membership functions

$$T_A,\ I_A,\ F_A : U \longrightarrow [0,1]$$

such that, for every $x \in U$,

$$0 \ \le\ T_A(x)^2 + I_A(x)^2 + F_A(x)^2 \ \le\ 1. \tag{3.4}$$

We usually write

$$A \ =\ \big\{\ \big(x,\ T_A(x),\ I_A(x),\ F_A(x)\big)\ :\ x \in U\ \big\}.$$

**Definition 3.14.2** (Spherical Neutrosophic Set)**.** [703–705] Let $U$ be a nonempty universe. A *spherical neutrosophic set* $A$ on $U$ is described by three functions

$$T_A,\ I_A,\ F_A : U \longrightarrow [0,3]$$

such that, for every $x \in U$,

$$0 \ \le\ T_A(x)^2 + I_A(x)^2 + F_A(x)^2 \ \le\ 3. \tag{3.5}$$

As before we may write

$$A \ =\ \big\{\ \big(x,\ T_A(x),\ I_A(x),\ F_A(x)\big)\ :\ x \in U\ \big\}.$$

**Example 3.14.3** (Spherical Neutrosophic Set)**.** Let $U = \{x_1, x_2, x_3\}$ be a universe, where $x_1$ = "normal", $x_2$ = "warning", $x_3$ = "critical" state of a machine. Define a spherical neutrosophic set $A$ on $U$ by

$$\begin{aligned} T_A(x_1) = 0.9,\ I_A(x_1) = 0.2,\ F_A(x_1) = 0.1,\\ T_A(x_2) = 0.6,\ I_A(x_2) = 0.6,\ F_A(x_2) = 0.4,\\ T_A(x_3) = 0.3,\ I_A(x_3) = 0.8,\ F_A(x_3) = 0.9. \end{aligned}$$

Then, for each $x \in U$ we have

$$\begin{aligned} T_A(x_1)^2 + I_A(x_1)^2 + F_A(x_1)^2 &= 0.9^2 + 0.2^2 + 0.1^2 = 0.81 + 0.04 + 0.01 = 0.86 \le 3,\\ T_A(x_2)^2 + I_A(x_2)^2 + F_A(x_2)^2 &= 0.6^2 + 0.6^2 + 0.4^2 = 0.36 + 0.36 + 0.16 = 0.88 \le 3,\\ T_A(x_3)^2 + I_A(x_3)^2 + F_A(x_3)^2 &= 0.3^2 + 0.8^2 + 0.9^2 = 0.09 + 0.64 + 0.81 = 1.54 \le 3, \end{aligned}$$

so the constraint (3.5) is satisfied for all $x \in U$, and hence $A$ is a spherical neutrosophic set on $U$.

We now impose a spherical constraint on the three components of $d$ and thus obtain the notion of a spherical plithogenic set.

**Definition 3.14.4** (Spherical Plithogenic Set)**.** Let $U$ be a nonempty universe, let $a$ be a fixed attribute, and let $V$ be a nonempty set of attribute values. Fix a radius $\lambda > 0$ and a contradiction degree function $c : V \times V \to [0, 1]$ with

$$c(v, v) = 0, \qquad c(v_1, v_2) = c(v_2, v_1) \quad \text{for all } v, v_1, v_2 \in V.$$

A *spherical plithogenic set of radius* $\lambda$ on $(U, a, V)$ is a plithogenic set

$$A^{\mathrm{sph}} \;=\; (U, a, V, d_A, c)$$

such that

- the degree of appurtenance function

  $$d_A : U \times V \longrightarrow [0, \lambda]^3$$

  is given by

  $$d_A(x, v) \;=\; \big(T_A(x, v),\, I_A(x, v),\, F_A(x, v)\big),$$

- and for every $(x, v) \in U \times V$ we have the spherical constraint

$$0 \;\le\; T_A(x, v)^2 + I_A(x, v)^2 + F_A(x, v)^2 \;\le\; \lambda^2. \tag{3.6}$$

We call $\big(T_A(x, v), I_A(x, v), F_A(x, v)\big)$ the *spherical plithogenic neutrosophic triple* of $x$ with respect to the attribute value $v$.

**Remark 3.14.5.** Note that if $|V| = 1$ and $c \equiv 0$, a spherical plithogenic set reduces to a single spherical triple assigned to each element $x \in U$, which is precisely the situation of spherical fuzzy and spherical neutrosophic sets, once $\lambda$ is chosen appropriately.

The specialization relationships between Spherical Plithogenic Sets and previous spherical models are summarized in Table 3.22.

Table 3.22: Spherical fuzzy–type models as special cases of a Spherical Plithogenic Set (single attribute $a^*$, $pCF \equiv 0$).

| Model | $pdf(x, a^*)$ | Spherical constraint |
|---|---|---|
| Spherical Fuzzy Set [367, 702] | $(\mu(x), \eta(x), \nu(x))$ | $\mu(x)^2 + \eta(x)^2 + \nu(x)^2 \le 1$ |
| Spherical Neutrosophic Set [703–705] | $(T(x), I(x), F(x))$ | $T(x)^2 + I(x)^2 + F(x)^2 \le 1$ |
| Spherical Picture Fuzzy Set [706, 707] | $(\xi(x), \psi(x), \rho(x), \gamma(x))$ | $\xi(x)^2 + \psi(x)^2 + \rho(x)^2 + \gamma(x)^2 \le 1$ |
| Spherical Hesitant Fuzzy Set [701] | $\big(H_\mu(x), H_\eta(x), H_\nu(x)\big)$ | for all $(\mu, \eta, \nu) \in H_\mu(x) \times H_\eta(x) \times H_\nu(x)$: $\mu^2 + \eta^2 + \nu^2 \le 1$ |

A concrete example of this concept is provided below.

**Example 3.14.6** (Spherical plithogenic air–quality risk assessment)**.** Let $U$ be the set of city districts and let $a$ be the attribute "air–quality health risk." Take the value set

$$V = \{\text{low, medium, high}\}.$$

We consider a spherical plithogenic set of radius $\lambda = 1$,

$$A^{\text{sph}} = (U, a, V, d_A, c),$$

with contradiction degrees

$$c(\text{low}, \text{low}) = c(\text{medium}, \text{medium}) = c(\text{high}, \text{high}) = 0,$$

$$c(\text{low}, \text{medium}) = c(\text{medium}, \text{low}) = 0.5,$$
$$c(\text{medium}, \text{high}) = c(\text{high}, \text{medium}) = 0.6,$$
$$c(\text{low}, \text{high}) = c(\text{high}, \text{low}) = 1.$$

For a specific district $x_0 \in U$, suppose the environmental authority assigns the following spherical plithogenic neutrosophic triples:

$$d_A(x_0, \text{low}) = \big(T_A, I_A, F_A\big)(x_0, \text{low}) = (0.10,\ 0.20,\ 0.95),$$

$$d_A(x_0, \text{medium}) = \big(T_A, I_A, F_A\big)(x_0, \text{medium}) = (0.50,\ 0.30,\ 0.40),$$
$$d_A(x_0, \text{high}) = \big(T_A, I_A, F_A\big)(x_0, \text{high}) = (0.75,\ 0.40,\ 0.10).$$

Each triple satisfies the spherical constraint $T^2 + I^2 + F^2 \leq 1$:

$$0.10^2 + 0.20^2 + 0.95^2 = 0.01 + 0.04 + 0.9025 = 0.9525 \leq 1,$$

$$0.50^2 + 0.30^2 + 0.40^2 = 0.25 + 0.09 + 0.16 = 0.50 \leq 1,$$
$$0.75^2 + 0.40^2 + 0.10^2 = 0.5625 + 0.16 + 0.01 = 0.7325 \leq 1.$$

Here $T_A(x_0, v)$ is the degree to which $x_0$ truly has risk level $v$, $I_A(x_0, v)$ is the degree of indeterminacy (due to sensor noise, seasonal variation, or incomplete data), and $F_A(x_0, v)$ is the degree to which $x_0$ fails to have risk level $v$. The contradiction function $c$ expresses that "low" and "high" risk are maximally contradictory, while "medium" is partially compatible with both.

When aggregating expert opinions or monitoring data, the plithogenic machinery weights the triples using $c$, so that evidence supporting "high" is discounted when combined with evidence supporting "low," and vice versa. Thus $A^{\text{sph}}$ provides a spherical, contradiction–aware representation of uncertain air–quality risk for all districts $x \in U$.

**Example 3.14.7** (Spherical plithogenic evaluation of diabetes severity)**.** Diabetes severity quantifies how advanced the disease is, combining glycemic control, complications, comorbidities, treatment intensity, and long-term risk of outcomes (cf. [708]).

Let $U$ be the set of patients in a clinic and let $a$ be the attribute "diabetes severity." Consider the set

$$V = \{\text{mild, moderate, severe}\},$$

and a spherical plithogenic set of radius $\lambda = 1$,

$$B^{\text{sph}} = (U, a, V, d_B, c),$$

with contradiction degrees

$$c(\text{mild}, \text{mild}) = c(\text{moderate}, \text{moderate}) = c(\text{severe}, \text{severe}) = 0,$$

$$c(\text{mild}, \text{moderate}) = c(\text{moderate}, \text{mild}) = 0.3,$$
$$c(\text{moderate}, \text{severe}) = c(\text{severe}, \text{moderate}) = 0.4,$$
$$c(\text{mild}, \text{severe}) = c(\text{severe}, \text{mild}) = 1.$$

For a particular patient $y \in U$, suppose laboratory tests and physician assessments yield:

$$d_B(y, \text{mild}) = \big(T_B, I_B, F_B\big)(y, \text{mild}) = (0.25,\ 0.40,\ 0.80),$$

$$d_B(y, \text{moderate}) = \big(T_B, I_B, F_B\big)(y, \text{moderate}) = (0.60,\ 0.35,\ 0.30),$$
$$d_B(y, \text{severe}) = \big(T_B, I_B, F_B\big)(y, \text{severe}) = (0.75,\ 0.30,\ 0.15).$$

Again, each triple lies in the unit sphere:

$$0.25^2 + 0.40^2 + 0.80^2 = 0.0625 + 0.16 + 0.64 = 0.8625 \le 1,$$

$$0.60^2 + 0.35^2 + 0.30^2 = 0.36 + 0.1225 + 0.09 = 0.5725 \le 1,$$
$$0.75^2 + 0.30^2 + 0.15^2 = 0.5625 + 0.09 + 0.0225 = 0.675 \le 1.$$

Here $T_B(y, v)$ represents the degree to which the current medical evidence supports severity level $v$ for patient $y$, $I_B(y, v)$ captures uncertainty (e.g., missing tests, conflicting indicators), and $F_B(y, v)$ encodes the degree of rejection of severity level $v$. The contradiction function $c$ encodes that "mild" and "severe" are highly incompatible, whereas "moderate" is closer to each.

In multi–expert decision support, spherical plithogenic aggregation uses $c$ to reconcile conflicting evaluations (e.g., one expert supporting "mild" and another supporting "severe") while preserving the spherical neutrosophic constraint for each $(y, v) \in U \times V$. This yields a medically interpretable, contradiction–aware assessment of diabetes severity for every patient.

We now formalize and prove the generalization property.

**Theorem 3.14.8.** *Let $U$ be a nonempty universe.*

*(i)* (Spherical fuzzy case*) Consider the class $\mathcal{SFS}(U)$ of all spherical fuzzy sets on $U$. Then $\mathcal{SFS}(U)$ is in bijection with the subclass $\mathcal{SPS}_1(U)$ of spherical plithogenic sets of radius $\lambda = 1$ such that*

$$V = \{v_0\} \quad \textit{is a singleton and} \quad c(v_0, v_0) = 0.$$

*(ii)* (Spherical neutrosophic case*) Consider the class $\mathcal{SNS}(U)$ of all spherical neutrosophic sets on $U$. Then $\mathcal{SNS}(U)$ is in bijection with the subclass $\mathcal{SPS}_{\sqrt{3}}(U)$ of spherical plithogenic sets of radius $\lambda = \sqrt{3}$ satisfying*

$$V = \{v_0\} \quad \textit{and} \quad c(v_0, v_0) = 0.$$

*Consequently, the notion of spherical plithogenic set strictly generalizes both spherical fuzzy sets and spherical neutrosophic sets.*

*Proof.* We prove (i) and (ii) separately.

**Proof of (i).** Fix $\lambda = 1$, a singleton value set $V = \{v_0\}$, and let $c(v_0, v_0) = 0$.

*Step 1: From spherical fuzzy to spherical plithogenic.* Let $A \in \mathcal{SFS}(U)$ be a spherical fuzzy set. By definition we have three maps

$$T_A,\ I_A,\ F_A : U \to [0,1]$$

such that for every $x \in U$,

$$0 \ \le\ T_A(x)^2 + I_A(x)^2 + F_A(x)^2 \ \le\ 1.$$

Define a degree of appurtenance function

$$d_A : U \times V \to [0,1]^3$$

by

$$d_A(x, v_0) \ :=\ \big(T_A(x),\, I_A(x),\, F_A(x)\big). \tag{3.7}$$

Since $T_A(x), I_A(x), F_A(x) \in [0,1]$ for all $x$, the codomain condition $d_A(x, v_0) \in [0,1]^3$ is satisfied. Moreover, the spherical inequality

$$T_A(x)^2 + I_A(x)^2 + F_A(x)^2 \le 1$$

is precisely (3.6) with $\lambda = 1$ and $v = v_0$. Therefore,

$$A^{\mathrm{sph}} := (U, a, V, d_A, c)$$

is a spherical plithogenic set of radius 1 with singleton value set and trivial contradiction function. This defines a map

$$\Phi : \mathcal{SFS}(U) \longrightarrow \mathcal{SPS}_1(U), \quad A \longmapsto A^{\mathrm{sph}}.$$

*Step 2: From spherical plithogenic (radius 1, singleton V) to spherical fuzzy.* Conversely, let

$$B^{\mathrm{sph}} = (U, a, V, d_B, c) \in \mathcal{SPS}_1(U)$$

with $V = \{v_0\}$ and $c(v_0, v_0) = 0$. By Definition 3.14.4, for each $x \in U$ there exist real numbers

$$T_B(x, v_0),\ I_B(x, v_0),\ F_B(x, v_0) \in [0,1]$$

such that

$$d_B(x, v_0) \ =\ \big(T_B(x, v_0),\, I_B(x, v_0),\, F_B(x, v_0)\big)$$

and

$$0 \ \le\ T_B(x, v_0)^2 + I_B(x, v_0)^2 + F_B(x, v_0)^2 \ \le\ 1$$

for every $x \in U$. Define

$$T_B^*(x) := T_B(x, v_0), \quad I_B^*(x) := I_B(x, v_0), \quad F_B^*(x) := F_B(x, v_0).$$

Then $T_B^*, I_B^*, F_B^* : U \to [0,1]$ and for each $x \in U$,

$$0 \ \le\ T_B^*(x)^2 + I_B^*(x)^2 + F_B^*(x)^2 \ =\ T_B(x, v_0)^2 + I_B(x, v_0)^2 + F_B(x, v_0)^2 \ \le\ 1.$$

Therefore

$$B := \big\{(x, T_B^*(x), I_B^*(x), F_B^*(x)) : x \in U\big\}$$

is a spherical fuzzy set, i.e. $B \in \mathcal{SFS}(U)$. This defines a map

$$\Psi : \mathcal{SPS}_1(U) \longrightarrow \mathcal{SFS}(U), \quad B^{\mathrm{sph}} \longmapsto B.$$

*Step 3:* $\Phi$ *and* $\Psi$ *are inverse to each other.* Take $A \in \mathcal{SFS}(U)$ and compute $\Psi(\Phi(A))$. By construction, $\Phi(A)$ is the spherical plithogenic set $A^{\mathrm{sph}}$ with

$$d_A(x, v_0) = \big(T_A(x), I_A(x), F_A(x)\big).$$

Applying $\Psi$ recovers the spherical fuzzy triple

$$T_A^*(x) = T_A(x), \quad I_A^*(x) = I_A(x), \quad F_A^*(x) = F_A(x)$$

for all $x \in U$. Hence $\Psi(\Phi(A)) = A$.

Conversely, take $B^{\mathrm{sph}} \in \mathcal{SPS}_1(U)$ and compute $\Phi(\Psi(B^{\mathrm{sph}}))$. The spherical fuzzy set $\Psi(B^{\mathrm{sph}})$ has components

$$T_B^*(x) = T_B(x, v_0), \quad I_B^*(x) = I_B(x, v_0), \quad F_B^*(x) = F_B(x, v_0).$$

Then $\Phi(\Psi(B^{\mathrm{sph}}))$ is the spherical plithogenic set $\widetilde{B}^{\mathrm{sph}}$ with degree function

$$\tilde{d}_B(x, v_0) = \big(T_B^*(x), I_B^*(x), F_B^*(x)\big) = \big(T_B(x, v_0), I_B(x, v_0), F_B(x, v_0)\big) = d_B(x, v_0).$$

Thus

$$\tilde{d}_B = d_B \quad \text{and hence} \quad \Phi(\Psi(B^{\mathrm{sph}})) = B^{\mathrm{sph}}.$$

Therefore $\Phi$ and $\Psi$ are mutually inverse bijections between $\mathcal{SFS}(U)$ and $\mathcal{SPS}_1(U)$, proving (i).

**Proof of (ii).** Now fix $\lambda = \sqrt{3}$, $V = \{v_0\}$, and $c(v_0, v_0) = 0$.

*Step 1: From spherical neutrosophic to spherical plithogenic.* Let $A \in \mathcal{SNS}(U)$ be a spherical neutrosophic set. Then

$$T_A,\ I_A,\ F_A : U \to [0, 3]$$

and for every $x \in U$,

$$0 \ \le\ T_A(x)^2 + I_A(x)^2 + F_A(x)^2 \ \le\ 3. \tag{3.8}$$

Define

$$d_A : U \times V \to [0, \sqrt{3}]^3$$

by

$$d_A(x, v_0) \ :=\ \big(T_A(x),\ I_A(x),\ F_A(x)\big). \tag{3.9}$$

First we check the codomain condition. Fix $x \in U$. From (3.8) we have

$$T_A(x)^2 \le T_A(x)^2 + I_A(x)^2 + F_A(x)^2 \le 3,$$

hence $T_A(x)^2 \le 3$ and therefore

$$0 \le T_A(x) \le \sqrt{3}.$$

The same argument applies to $I_A(x)$ and $F_A(x)$, so

$$0 \le T_A(x), I_A(x), F_A(x) \le \sqrt{3}$$

for every $x \in U$. Thus $d_A(x, v_0) \in [0, \sqrt{3}]^3$.

Next, the spherical constraint (3.6) with $\lambda = \sqrt{3}$ becomes

$$0 \le T_A(x)^2 + I_A(x)^2 + F_A(x)^2 \le (\sqrt{3})^2 = 3,$$

which holds by (3.8). Therefore

$$A^{\mathrm{sph}} := (U, a, V, d_A, c)$$

is a spherical plithogenic set of radius $\sqrt{3}$ with singleton value set and trivial contradiction function. This defines a map

$$\Phi' : \mathcal{SNS}(U) \longrightarrow \mathcal{SPS}_{\sqrt{3}}(U), \quad A \longmapsto A^{\mathrm{sph}}.$$

*Step 2: From spherical plithogenic (radius $\sqrt{3}$, singleton $V$) to spherical neutrosophic.* Conversely, let

$$B^{\mathrm{sph}} = (U, a, V, d_B, c) \in \mathcal{SPS}_{\sqrt{3}}(U)$$

with $V = \{v_0\}$ and $c(v_0, v_0) = 0$. Again, by Definition 3.14.4, we have for each $x \in U$

$$d_B(x, v_0) \;=\; \big(T_B(x, v_0),\, I_B(x, v_0),\, F_B(x, v_0)\big) \in [0, \sqrt{3}]^3$$

and

$$0 \;\le\; T_B(x, v_0)^2 + I_B(x, v_0)^2 + F_B(x, v_0)^2 \;\le\; (\sqrt{3})^2 = 3.$$

Define

$$T_B^{\dagger}(x) := T_B(x, v_0), \quad I_B^{\dagger}(x) := I_B(x, v_0), \quad F_B^{\dagger}(x) := F_B(x, v_0).$$

Then

$$T_B^{\dagger}, I_B^{\dagger}, F_B^{\dagger} : U \to [0, \sqrt{3}] \subseteq [0, 3],$$

and for every $x \in U$,

$$0 \;\le\; T_B^{\dagger}(x)^2 + I_B^{\dagger}(x)^2 + F_B^{\dagger}(x)^2 \;\le\; 3.$$

Thus

$$B := \big\{(x, T_B^{\dagger}(x), I_B^{\dagger}(x), F_B^{\dagger}(x)) : x \in U\big\}$$

is a spherical neutrosophic set, i.e. $B \in \mathcal{SNS}(U)$. This gives a map

$$\Psi' : \mathcal{SPS}_{\sqrt{3}}(U) \longrightarrow \mathcal{SNS}(U), \quad B^{\mathrm{sph}} \longmapsto B.$$

*Step 3: $\Phi'$ and $\Psi'$ are inverse to each other.* The verification is identical in structure to the spherical fuzzy case. For any $A \in \mathcal{SNS}(U)$, the composite $\Psi'(\Phi'(A))$ recovers the original triple $(T_A(x), I_A(x), F_A(x))$ for each $x \in U$, so $\Psi'(\Phi'(A)) = A$. Conversely, for any $B^{\mathrm{sph}} \in \mathcal{SPS}_{\sqrt{3}}(U)$, the composite $\Phi'(\Psi'(B^{\mathrm{sph}}))$ preserves $d_B$ pointwise (via the same coordinate equalities as in the proof of (i)), hence $\Phi'(\Psi'(B^{\mathrm{sph}})) = B^{\mathrm{sph}}$.

Therefore $\Phi'$ and $\Psi'$ are mutually inverse bijections between $\mathcal{SNS}(U)$ and $\mathcal{SPS}_{\sqrt{3}}(U)$, establishing (ii).

Combining (i) and (ii), we conclude that:

- every spherical fuzzy set can be realized as a spherical plithogenic set of radius 1 with a single attribute value and zero contradiction;
- every spherical neutrosophic set can be realized as a spherical plithogenic set of radius $\sqrt{3}$ with a single attribute value and zero contradiction;
- spherical plithogenic sets in general allow multiple attribute values and a nontrivial contradiction degree $c$, and hence strictly extend both frameworks.

This completes the proof. □

## 3.15 T-Spherical Plithogenic Set

A $T$-Spherical Plithogenic Set models uncertainty by assigning triples constrained by a $t$-spherical radius, incorporating attribute-based contradictions to capture richer multi-valued information.

**Definition 3.15.1** (T-Spherical Plithogenic Set)**.** Let $U$ be a nonempty universe, let $a$ be a fixed attribute, and let $V$ be a nonempty set of attribute values. Fix a radius $\lambda > 0$, a real parameter $t \geq 1$, and a contradiction degree function $c : V \times V \to [0, 1]$ with

$$c(v, v) = 0, \qquad c(v_1, v_2) = c(v_2, v_1) \quad \text{for all } v, v_1, v_2 \in V.$$

A *T-spherical plithogenic set of order* $t$ *and radius* $\lambda$ on $(U, a, V)$ is a plithogenic set

$$A^{\mathrm{TSph}} = (U, a, V, d_A, c)$$

such that

- the degree of appurtenance function

$$d_A : U \times V \longrightarrow [0, \lambda]^3$$

  is given by

$$d_A(x, v) = \big(T_A(x, v),\ I_A(x, v),\ F_A(x, v)\big),$$

- and for every $(x, v) \in U \times V$ we have the $t$–spherical constraint

$$0 \leq T_A(x, v)^t + I_A(x, v)^t + F_A(x, v)^t \leq \lambda^t. \tag{3.10}$$

The triple $\big(T_A(x, v), I_A(x, v), F_A(x, v)\big)$ is called the $t$*–spherical plithogenic neutrosophic triple* of $x$ with respect to the attribute value $v$.

Table 3.23 provides a brief overview showing that various T-spherical fuzzy–type models arise as special cases of a T-Spherical Plithogenic Set. Each model is obtained by fixing a single attribute value, setting the contradiction function to zero, and applying the $t$-spherical constraint to its corresponding membership structure.

Table 3.23: T-spherical fuzzy–type models as special cases of a T-Spherical Plithogenic Set (single attribute $a^*$, $pCF \equiv 0$, order $t \geq 1$, radius 1).

| Model | $pdf(x, a^*)$ | $t$-spherical constraint |
|---|---|---|
| T-Spherical Fuzzy Set [709, 710] | $(\mu(x), \eta(x), \nu(x))$ | $\mu(x)^t + \eta(x)^t + \nu(x)^t \leq 1^t$ |
| T-Spherical Neutrosophic Set [711, 712] | $(T(x), I(x), F(x))$ | $T(x)^t + I(x)^t + F(x)^t \leq 1^t$ |
| T-Spherical Picture Fuzzy Set | $(\xi(x), \psi(x), \rho(x), \gamma(x))$ | $\xi(x)^t + \psi(x)^t + \rho(x)^t + \gamma(x)^t \leq 1^t$ |
| T-Spherical Hesitant Fuzzy Set [701, 713] | $\big(H_\mu(x), H_\eta(x), H_\nu(x)\big)$ | for all $(\mu, \eta, \nu) \in H_\mu(x) \times H_\eta(x) \times H_\nu(x)$: $\mu^t + \eta^t + \nu^t \leq 1^t$ |

A concrete example of this concept is provided below.

**Example 3.15.2** (A simple T-Spherical Plithogenic Set)**.** Let $U = \{x_1, x_2\}$ be a set of two alternatives (e.g. two candidate suppliers), and let

$$V = \{v_1, v_2\}$$

be a set of two attribute values (e.g. $v_1$ = "cost-efficient", $v_2$ = "high quality"). Fix $\lambda = 1$ and $t = 3$.

Define the contradiction function $c : V \times V \to [0, 1]$ by

$$c(v_1, v_1) = c(v_2, v_2) = 0, \qquad c(v_1, v_2) = c(v_2, v_1) = 0.4.$$

Define $d_A : U \times V \to [0, 1]^3$ by:

$$d_A(x_1, v_1) = (0.8, 0.3, 0.2), \qquad d_A(x_1, v_2) = (0.5, 0.5, 0.2),$$

$$d_A(x_2, v_1) = (0.4, 0.7, 0.2), \qquad d_A(x_2, v_2) = (0.6, 0.2, 0.6).$$

We check the constraint (3.10) for $t = 3$.

For $(x_1, v_1)$:

$$0.8^3 + 0.3^3 + 0.2^3 \; = \; 0.512 + 0.027 + 0.008 \; = \; 0.547 \; \leq \; 1^3.$$

For $(x_1, v_2)$:

$$0.5^3 + 0.5^3 + 0.2^3 \; = \; 0.125 + 0.125 + 0.008 \; = \; 0.258 \; \leq \; 1^3.$$

For $(x_2, v_1)$:

$$0.4^3 + 0.7^3 + 0.2^3 \; = \; 0.064 + 0.343 + 0.008 \; = \; 0.415 \; \leq \; 1^3.$$

For $(x_2, v_2)$:

$$0.6^3 + 0.2^3 + 0.6^3 \; = \; 0.216 + 0.008 + 0.216 \; = \; 0.440 \; \leq \; 1^3.$$

Thus all triples satisfy $T_A(x, v)^3 + I_A(x, v)^3 + F_A(x, v)^3 \leq 1^3$, so $A^{\mathrm{TSph}} = (U, a, V, d_A, c)$ is a T-spherical plithogenic set of order $t = 3$ and radius $\lambda = 1$.

In this framework, the following theorems hold.

**Theorem 3.15.3** (T-Spherical Plithogenic Set generalizes Spherical Plithogenic Set)**.** *Every spherical plithogenic set of Definition* 3.14.4 *is a T-spherical plithogenic set of order* $t = 2$ *in the sense of Definition* 3.15.1. *Hence the class of T-spherical plithogenic sets strictly generalizes the class of spherical plithogenic sets.*

*Proof.* Let

$$A^{\mathrm{sph}} = (U, a, V, d_A, c)$$

be a spherical plithogenic set of radius $\lambda$ as in Definition 3.14.4. By definition we have

$$d_A(x, v) = \big(T_A(x, v), I_A(x, v), F_A(x, v)\big) \in [0, \lambda]^3$$

and, for all $(x, v) \in U \times V$,

$$0 \le T_A(x, v)^2 + I_A(x, v)^2 + F_A(x, v)^2 \le \lambda^2. \tag{3.11}$$

Now fix $t = 2$. Compare (3.11) with the $t$–spherical constraint (3.10) in Definition 3.15.1:

$$0 \le T_A(x, v)^t + I_A(x, v)^t + F_A(x, v)^t \le \lambda^t.$$

For $t = 2$ these two inequalities coincide exactly, and the remaining data $(U, a, V, d_A, c)$ are unchanged. Therefore $A^{\mathrm{sph}}$ satisfies all requirements of a T-spherical plithogenic set of order $t = 2$ and radius $\lambda$.

Consequently, the mapping

$$\Phi : \{\text{spherical plithogenic sets}\} \longrightarrow \{\text{T-spherical plithogenic sets}\}, \quad \Phi(A^{\mathrm{sph}}) = A^{\mathrm{sph}} \text{ with } t = 2$$

embeds the class of spherical plithogenic sets into the class of T-spherical plithogenic sets. Since for $t > 2$ there exist T-spherical plithogenic sets which do not satisfy the quadratic constraint of Definition 3.14.4, the inclusion is proper, so T-spherical plithogenic sets strictly generalize spherical plithogenic sets. □

## 3.16 Plithogenic Rough Set

A plithogenic rough set approximates a subset using plithogenic relations combining membership and contradiction, generalizing fuzzy and neutrosophic rough models [714].

**Definition 3.16.1** (Plithogenic rough approximation space)**.** [714] Let $U \neq \varnothing$ be a universe and let $R$ be a plithogenic relation on $U$. A plithogenic relation is specified by

$$pdf_R : U \times U \longrightarrow [0, 1]^s, \qquad pCF_R : U \times U \longrightarrow [0, 1]^t,$$

where $pdf_R(x, y)$ is the (vector) plithogenic degree of appurtenance of $y$ to $x$, and $pCF_R(x, y)$ is the (vector) degree of contradiction between the attribute values of $x$ and $y$. Assume there is a fixed aggregation

$$\Phi : [0, 1]^s \times [0, 1]^t \longrightarrow [0, 1]$$

that is monotone in each argument and satisfies $\Phi(\mathbf{a}, \mathbf{0}) = \Phi(\mathbf{0}, \mathbf{b}) = 0$. For convenience write

$$\tilde{R}(x, y) := \Phi\big(pdf_R(x, y), pCF_R(x, y)\big) \in [0, 1].$$

Then $(U, R)$ is called a plithogenic rough approximation space.

**Definition 3.16.2** (Plithogenic lower approximation)**.** [714] Let $A \subseteq U$ and let $(U, R)$ be as above. The plithogenic lower approximation of $A$ with respect to $R$ is the fuzzy set

$$\underline{PL}_R(A) : U \to [0, 1], \qquad \underline{PL}_R(A)(x) := \inf_{y \in A} \max\big(1 - \tilde{R}(x, y), 1 - \tilde{R}(y, x)\big), \quad x \in U.$$

In the scalar case $s = t = 1$ and $\Phi(a, b) = \max\{a, b\}$, this becomes

$$\underline{PL}_R(A)(x) = \inf_{y \in A} \max\big(1 - pdf_R(x, y), 1 - pCF_R(x, y)\big), \quad x \in U.$$

**Definition 3.16.3** (Plithogenic upper approximation)**.** [714] Under the same assumptions, the plithogenic upper approximation of $A$ with respect to $R$ is the fuzzy set

$$\overline{PL}_R(A) : U \to [0,1], \qquad \overline{PL}_R(A)(x) := \sup_{y\in U} \min\big(\tilde{R}(x,y),\, 1-\tilde{R}(y,x)\big), \quad x \in U.$$

In the scalar case $s = t = 1$ and $\Phi(a,b) = \max\{a,b\}$, this reduces to

$$\overline{PL}_R(A)(x) = \sup_{y\in U} \min\big(pdf_R(x,y),\, 1 - pCF_R(x,y)\big), \quad x \in U.$$

**Definition 3.16.4** (Plithogenic Rough Set)**.** Let $A \subseteq U$. The *plithogenic rough set* of $A$ (with respect to $R$) is the ordered pair

$$PL_R(A) := \big(\underline{PL}_R(A),\, \overline{PL}_R(A)\big).$$

Table 3.24 presents the relationships between plithogenic rough sets and their related concepts.

Table 3.24: Examples of rough-type models that are special cases of the plithogenic rough set $PL_R(A) = (\underline{PL}_R(A), \overline{PL}_R(A))$ by choosing the DAF dimension $s$ and (optionally) the DCF dimension $t$.

| Target rough model | Membership type | $s$ | $t$ | Generalization by plithogenic rough set |
|---|---|---|---|---|
| Fuzzy Rough Set [715–717] | single fuzzy degree $\mu(x) \in [0,1]$ | 1 | 0 | obtained by taking $pdf(x,a) \in [0,1]$ and $pCF \equiv 0$; lower/upper are fuzzy-graded approximations |
| Intuitionistic Fuzzy Rough Set [718–720] | intuitionistic pair $(\mu,\nu)$ | 2 | 0 | recovered by letting $pdf(x,a) \in [0,1]^2$ and $pCF \equiv 0$; lower/upper use both membership and nonmembership |
| Vague Rough Set [721–723] | Vague pair $(\mu,\nu)$ | 2 | 0 | recovered by letting $pdf(x,a) \in [0,1]^2$ and $pCF \equiv 0$; |
| Neutrosophic Rough Set [724–727] | neutrosophic triple $(T,I,F)$ | 3 | 0 | recovered by letting $pdf(x,a) \in [0,1]^3$ and $pCF \equiv 0$; lower/upper defined componentwise on $(T,I,F)$ |
| Picture Fuzzy Rough Set [728–730] | picture fuzzy triple | 3 | 0 | recovered by letting $pdf(x,a) \in [0,1]^3$ and $pCF \equiv 0$; |
| Hesitant Fuzzy Rough Set [731–733] | Hesitant fuzzy triple | 3 | 0 | recovered by letting $pdf(x,a) \in [0,1]^3$ and $pCF \equiv 0$; |
| Spherical Fuzzy Rough Set [734–736] | Spherical fuzzy triple | 3 | 0 | recovered by letting $pdf(x,a) \in [0,1]^3$ and $pCF \equiv 0$; |
| Quadripartitioned Neutrosophic Rough Set (cf. [737]) | 4-part neutrosophic/rough components | 4 | 0 | obtained by $pdf(x,a) \in [0,1]^4$; plithogenic framework stores four uncertainty channels per element |
| Pentapartitioned Neutrosophic Rough Set | 5-part neutrosophic/rough components | 5 | 0 | obtained by $pdf(x,a) \in [0,1]^5$; plithogenic rough set directly supports higher partitioned rough information |

A concrete example of this concept is provided below.

**Example 3.16.5** (Hospital triage: "needs ICU" under plithogenic clinical similarity/contradiction)**.** Let the universe be patients $U = \{p_1, p_2, p_3\}$. Target concept $A = \{p_2\}$ (clinically judged "needs ICU"). For a plithogenic relation $R$, take scalar degrees $s = t = 1$ and adopt the aggregation

$$\tilde{R}(x,y) := \Phi\big(pdf_R(x,y),\, pCF_R(x,y)\big) = pdf_R(x,y) \cdot \big(1 - pCF_R(x,y)\big),$$

so that high clinical similarity and low contradiction increase $\tilde{R} \in [0,1]$.

Assume (diagonal values omitted):

| $pdf_R(x,y)$ | $y = p_1$ | $p_2$ | $p_3$ |
|---|---|---|---|
| $x = p_1$ | – | 0.80 | 0.30 |
| $p_2$ | 0.70 | – | 0.40 |
| $p_3$ | 0.50 | 0.60 | – |

| $pCF_R(x, y)$ | $y = p_1$ | $p_2$ | $p_3$ |
|---|---|---|---|
| $x = p_1$ | – | 0.20 | 0.60 |
| $p_2$ | 0.30 | – | 0.40 |
| $p_3$ | 0.50 | 0.40 | – |

Then (again off–diagonal shown)

$$\tilde{R}(x, y) = pdf_R(x, y)\big(1 - pCF_R(x, y)\big) :$$

| $\tilde{R}(x, y)$ | $y = p_1$ | $p_2$ | $p_3$ |
|---|---|---|---|
| $x = p_1$ | – | $0.80 \times 0.80 = 0.64$ | $0.30 \times 0.40 = 0.12$ |
| $p_2$ | $0.70 \times 0.70 = 0.49$ | – | $0.40 \times 0.60 = 0.24$ |
| $p_3$ | $0.50 \times 0.50 = 0.25$ | $0.60 \times 0.60 = 0.36$ | – |

Plithogenic lower/upper approximations (finite $U \Rightarrow \inf = \min,\ \sup = \max$):

$$\underline{PL}_R(A)(x) = \min_{y \in A} \max\big(1 - \tilde{R}(x, y),\ 1 - \tilde{R}(y, x)\big),$$

$$\overline{PL}_R(A)(x) = \max_{y \in U} \min\big(\tilde{R}(x, y),\ 1 - \tilde{R}(y, x)\big).$$

For $x = p_1$ (unlabeled):

$$\underline{PL}_R(A)(p_1) = \max\big(1 - \tilde{R}(p_1, p_2),\ 1 - \tilde{R}(p_2, p_1)\big) = \max(1 - 0.64,\ 1 - 0.49) = \max(0.36, 0.51) = 0.51.$$

$$\overline{PL}_R(A)(p_1) = \max\big\{\min(\tilde{R}(p_1, p_1), 1 - \tilde{R}(p_1, p_1)),$$
$$\min(\tilde{R}(p_1, p_2), 1 - \tilde{R}(p_2, p_1)),\ \min(\tilde{R}(p_1, p_3), 1 - \tilde{R}(p_3, p_1))\big\}.$$

Taking $\tilde{R}(p_1, p_1) = 1$ (self–consistency), the three mins are

$\min(1, 0) = 0, \quad \min(0.64, 1-0.49) = \min(0.64, 0.51) = 0.51, \quad \min(0.12, 1-0.25) = \min(0.12, 0.75) = 0.12,$

so $\overline{PL}_R(A)(p_1) = \max(0, 0.51, 0.12) = 0.51$. Hence $p_1$ lies on the boundary: $\underline{PL} = 0.51,\ \overline{PL} = 0.51$.

For $x = p_3$ (unlabeled):

$$\underline{PL}_R(A)(p_3) = \max(1 - \tilde{R}(p_3, p_2),\ 1 - \tilde{R}(p_2, p_3))$$
$$= \max(1 - 0.36,\ 1 - 0.24) = \max(0.64, 0.76) = 0.76,$$
$$\overline{PL}_R(A)(p_3) = \max\big\{\min(1, 0),$$
$$\min(0.36, 1 - 0.49) = \min(0.36, 0.51) = 0.36,\ \min(0.25, 1 - 0.25) = 0.25\big\} = 0.36.$$

Thus $p_3$ is weakly covered in the upper region (0.36) with a comparatively large lower penalty (0.76), reflecting moderate similarity to the ICU case but notable contradictions.

**Example 3.16.6** (Credit risk screening: "high–risk borrowers")**.** Universe $U = \{a, b, c\}$ (loan applicants); target $A = \{b\}$ ("high–risk" profile from historical labels). Use the same scalar model with $\tilde{R}(x, y) = pdf_R(x, y)\big(1 - pCF_R(x, y)\big)$. Assume

| $pdf_R$ | $y = a$ | $b$ | $c$ |
|---|---|---|---|
| $x = a$ | – | 0.65 | 0.30 |
| $b$ | 0.55 | – | 0.35 |
| $c$ | 0.25 | 0.50 | – |

| $pCF_R$ | $y = a$ | $b$ | $c$ |
|---|---|---|---|
| $x = a$ | – | 0.25 | 0.50 |
| $b$ | 0.40 | – | 0.35 |
| $c$ | 0.30 | 0.45 | – |

Hence

$$\tilde{R}(a, b) = 0.65(1 - 0.25) = 0.4875, \quad \tilde{R}(b, a) = 0.55(1 - 0.40) = 0.33,$$
$$\tilde{R}(c, b) = 0.50(1 - 0.45) = 0.275, \quad \tilde{R}(b, c) = 0.35(1 - 0.35) = 0.2275,$$

and $\tilde{R}(a,c) = 0.30(1-0.50) = 0.15,\ \tilde{R}(c,a) = 0.25(1-0.30) = 0.175.$

Lower/upper for $x = a$:

$$\underline{PL}_R(A)(a) = \max(1-\tilde{R}(a,b),$$

$$1-\tilde{R}(b,a)) = \max(1-0.4875,\ 1-0.33) = \max(0.5125, 0.67) = 0.67,$$

$$\overline{PL}_R(A)(a) = \max\{\min(1,0),$$

$$\min(0.4875, 1-0.33) = \min(0.4875, 0.67) = 0.4875,\ \min(0.15, 1-0.175) = 0.15\} = 0.4875.$$

Applicant $a$ sits close to the boundary (upper $\approx 0.49$) but with a sizable lower penalty (0.67), i.e., some similarity to high–risk patterns yet notable contradictions.

For $x = c$:

$$\underline{PL}_R(A)(c) = \max(1-\tilde{R}(c,b),\ 1-\tilde{R}(b,c)) = \max(1-0.275,\ 1-0.2275) = \max(0.725, 0.7725) = 0.7725,$$

$$\overline{PL}_R(A)(c) = \max\{\min(1,0),\ \min(0.275, 1-0.33) =$$

$$\min(0.275, 0.67) = 0.275,\ \min(0.175, 1-0.15) = 0.175\} = 0.275.$$

Thus $c$ is loosely covered by the upper region (0.275) and far from the lower region, indicating predominantly non–high–risk behavior with mild resemblance to $b$.

**Example 3.16.7** (Manufacturing quality control: "defective batches")**.** Manufacturing quality control monitors production processes, inspects products, and corrects defects to ensure consistent standards, safety, compliance, and customer satisfaction (cf. [738]).

Universe $U = \{B_1, B_2, B_3, B_4\}$ (batches). Target $A = \{B_3, B_4\}$ (confirmed defective). Let $\tilde{R} = \mathit{pdf} \cdot (1 - pCF)$ as before, with $s = t = 1$. Suppose the (off–diagonal) entries summarizing similarity in fault signatures (PDF) and contradictions due to differing process settings (DCF) are

| $pdf_R$ | $B_1$ | $B_2$ | $B_3$ | $B_4$ |
|---|---|---|---|---|
| $B_1$ | – | 0.55 | 0.70 | 0.40 |
| $B_2$ | 0.60 | – | 0.45 | 0.35 |
| $B_3$ | 0.50 | 0.40 | – | 0.80 |
| $B_4$ | 0.45 | 0.30 | 0.75 | – |

| $pCF_R$ | $B_1$ | $B_2$ | $B_3$ | $B_4$ |
|---|---|---|---|---|
| $B_1$ | – | 0.30 | 0.20 | 0.40 |
| $B_2$ | 0.35 | – | 0.25 | 0.30 |
| $B_3$ | 0.25 | 0.35 | – | 0.15 |
| $B_4$ | 0.40 | 0.30 | 0.10 | – |

Compute the needed $\tilde{R}$ values:

$$\tilde{R}(B_1,B_3) = 0.70(1-0.20) = 0.56, \quad \tilde{R}(B_3,B_1) = 0.50(1-0.25) = 0.375,$$

$$\tilde{R}(B_1,B_4) = 0.40(1-0.40) = 0.24, \quad \tilde{R}(B_4,B_1) = 0.45(1-0.40) = 0.27.$$

Lower/upper for $x = B_1$ w.r.t. $A = \{B_3, B_4\}$:

$$\underline{PL}_R(A)(B_1) = \min_{y\in\{B_3,B_4\}} \max\bigl(1-\tilde{R}(B_1,y),\ 1-\tilde{R}(y,B_1)\bigr).$$

For $y = B_3$: $\max(1-0.56,\ 1-0.375) = \max(0.44, 0.625) = 0.625$. For $y = B_4$: $\max(1-0.24,\ 1-0.27) = \max(0.76, 0.73) = 0.76$. Thus $\underline{PL}_R(A)(B_1) = \min\{0.625, 0.76\} = 0.625$.

$$\overline{PL}_R(A)(B_1) = \max_{y\in U} \min\bigl(\tilde{R}(B_1,y),\ 1-\tilde{R}(y,B_1)\bigr).$$

Evaluate four mins (incl. self):

$$\min(1,0) = 0, \quad \min(0.55(1-0.30) = 0.385,\ 1-0.60(1-0.35) = 1-0.39 = 0.61) = 0.385,$$

$$\min(0.56,\ 1-0.375 = 0.625) = 0.56, \quad \min(0.24,\ 1-0.27 = 0.73) = 0.24.$$

Therefore $\overline{PL}_R(A)(B_1) = \max\{0, 0.385, 0.56, 0.24\} = 0.56$. Interpretation: $B_1$ is moderately similar to defective batches (upper 0.56) but fails certainty (lower 0.625), i.e., flagged for re–inspection.

For $x = B_2$ (quick check against $A$):

$$\tilde{R}(B_2, B_3) = 0.45(1 - 0.25) = 0.3375, \quad \tilde{R}(B_3, B_2) = 0.40(1 - 0.35) = 0.26,$$

$$\tilde{R}(B_2, B_4) = 0.35(1 - 0.30) = 0.245, \quad \tilde{R}(B_4, B_2) = 0.30(1 - 0.30) = 0.21.$$

$$\underline{PL}_R(A)(B_2) = \min\{\max(1 - 0.3375, 1 - 0.26),\ \max(1 - 0.245, 1 - 0.21)\}$$

$$= \min\{\max(0.6625, 0.74),\ \max(0.755, 0.79)\} = \min\{0.74, 0.79\} = 0.74,$$

$$\overline{PL}_R(A)(B_2) \geq \max\{\min(0.3375, 1 - 0.26) = 0.3375,\ \min(0.245, 1 - 0.21) = 0.245\} = 0.3375,$$

hence $B_2$ is weakly in the upper region, suggesting a lower likelihood of defect yet non–negligible resemblance to faulty signatures.

## 3.17 Plithogenic soft rough set

A plithogenic soft rough set combines parameterized soft memberships with plithogenic rough approximations, modeling contradiction-aware, granular uncertainty in decision-making contexts (cf. [739]). In this section we couple the plithogenic rough approximation machinery with plithogenic soft information. The resulting structure will be called a *Plithogenic soft rough set*; it will be shown to specialize both to a plithogenic rough set and to a plithogenic soft set.

**Definition 3.17.1** (Plithogenic soft approximation space)**.** Let $U \neq \varnothing$ be a universe and let $E$ be a nonempty set of parameters. A *plithogenic soft approximation space* is a tuple

$$\mathbb{S} \ := \ (U, E, \mu_S, pCF_E, \{R_e\}_{e \in E}, \Phi, \Psi),$$

where:

- $\mu_S : U \times E \to [0,1]^j$ is a plithogenic degree of appurtenance (DAF) assigning to each pair $(x, e) \in U \times E$ a (possibly vector-valued) membership $\mu_S(x, e)$, with $j \in \{1, 2, 3, 4, 5\}$;
- $pCF_E : E \times E \to [0,1]^t$ is a degree of contradiction function between parameters, satisfying

$$pCF_E(e, e) = \mathbf{0}, \qquad pCF_E(e, e') = pCF_E(e', e) \quad (e, e' \in E);$$

- for each $e \in E$, $R_e$ is a plithogenic relation on $U$ given by

$$pdf_{R_e} : U \times U \to [0,1]^s, \qquad pCF_{R_e} : U \times U \to [0,1]^t,$$

  together with a fixed aggregation

$$\Phi : [0,1]^s \times [0,1]^t \longrightarrow [0,1]$$

  which is monotone in each argument and satisfies

$$\Phi(\mathbf{a}, \mathbf{0}) = \Phi(\mathbf{0}, \mathbf{b}) = 0 \quad (\mathbf{a} \in [0,1]^s,\ \mathbf{b} \in [0,1]^t).$$

  As in the plithogenic rough case, we abbreviate

$$\tilde{R}_e(x, y) \ := \ \Phi\big(pdf_{R_e}(x, y),\, pCF_{R_e}(x, y)\big) \in [0,1].$$

- $\Psi : [0,1]^j \times [0,1] \to [0,1]^j$ is a fixed plithogenic aggregation operator which is monotone in each argument, i.e.,

$$\mu_1 \leq \mu_2,\ \lambda_1 \leq \lambda_2 \quad \Longrightarrow \quad \Psi(\mu_1, \lambda_1) \leq \Psi(\mu_2, \lambda_2),$$

  and satisfies the boundary conditions

$$\Psi(\mathbf{0}, \lambda) = \mathbf{0}, \qquad \Psi(\mu, 0) = \mathbf{0} \quad (\mu \in [0,1]^j,\ \lambda \in [0,1]).$$

For each $e \in E$, the pair $(U, R_e)$ is a plithogenic rough approximation space in the sense of the previous subsection, and hence admits plithogenic lower and upper approximations.

**Definition 3.17.2** (Plithogenic soft rough lower and upper approximations)**.** Let $\mathbb{S}$ be a plithogenic soft approximation space as above, and let $A \subseteq U$. For each fixed parameter $e \in E$ and element $x \in U$, define the *local plithogenic rough lower and upper approximations* of $A$ with respect to $R_e$ by

$$\underline{PL}_{R_e}(A)(x) \ := \ \inf_{y \in A} \max\bigl(1 - \tilde{R}_e(x,y),\, 1 - \tilde{R}_e(y,x)\bigr),$$

$$\overline{PL}_{R_e}(A)(x) \ := \ \sup_{y \in U} \min\bigl(\tilde{R}_e(x,y),\, 1 - \tilde{R}_e(y,x)\bigr),$$

exactly as in the plithogenic rough case, but with $R$ replaced by $R_e$.

The *plithogenic soft rough lower approximation* of $A$ is the mapping

$$\underline{PSR}(A) : U \times E \to [0,1]^j,$$

defined by

$$\underline{PSR}(A)(x,e) \ := \ \Psi\bigl(\mu_S(x,e),\, \underline{PL}_{R_e}(A)(x)\bigr), \qquad (x,e) \in U \times E,$$

and the *plithogenic soft rough upper approximation* of $A$ is the mapping

$$\overline{PSR}(A) : U \times E \to [0,1]^j,$$

given by

$$\overline{PSR}(A)(x,e) \ := \ \Psi\bigl(\mu_S(x,e),\, \overline{PL}_{R_e}(A)(x)\bigr), \qquad (x,e) \in U \times E.$$

Thus, for each fixed $e \in E$, the functions $x \mapsto \underline{PSR}(A)(x,e)$ and $x \mapsto \overline{PSR}(A)(x,e)$ describe how strongly $x$ belongs to the plithogenic rough lower/upper approximation of $A$ under parameter $e$, after combining the soft membership $\mu_S(x,e)$ with the rough information derived from $R_e$.

**Definition 3.17.3** (Plithogenic soft rough set)**.** Let $\mathbb{S}$ and $A$ be as above. The ordered pair

$$PSR_{\mathbb{S}}(A) \ := \ \bigl(\underline{PSR}(A),\, \overline{PSR}(A)\bigr)$$

is called the *plithogenic soft rough set* of $A$ over the plithogenic soft approximation space $\mathbb{S}$.

Equivalently, for each $e \in E$ we may regard

$$F_L(e) : U \to [0,1]^j, \quad F_L(e)(x) := \underline{PSR}(A)(x,e),$$

$$F_U(e) : U \to [0,1]^j, \quad F_U(e)(x) := \overline{PSR}(A)(x,e),$$

so that $PSR_{\mathbb{S}}(A)$ naturally corresponds to the pair of plithogenic soft sets $(F_L, F_U)$ on $(U, E)$.

The following Table 3.25 presents how the plithogenic soft rough set serves as a common generalization of several existing soft rough models.

A concrete example is given below.

Table 3.25: Plithogenic soft rough set as a common generalization of several soft rough models.

| Target soft rough model | Codomain of $F_L, F_U$ | $j$ | Specialization of $PSR_{\mathbb{S}}(A)$ |
|---|---|---|---|
| Plithogenic soft rough set | $[0,1]^j$ | $j \in \{1,2,3\}$ | Base model with general $pCF$ on attribute values. |
| Fuzzy soft rough set [740, 741] | $[0,1]$ | $j = 1$ | Take $F_L, F_U : U \to [0,1]$ and set $pCF \equiv 0$. |
| Intuitionistic fuzzy soft rough set [742–744] | $[0,1]^2$ | $j = 2$ | Take $F_L, F_U : U \to [0,1]^2$ and set $pCF \equiv 0$. |
| Neutrosophic soft rough set [745–748] | $[0,1]^3$ | $j = 3$ | Take $F_L, F_U : U \to [0,1]^3$ and set $pCF \equiv 0$. |
| Hesitant fuzzy soft rough set [749–751] | $[0,1]^3$ | $j = 3$ | Represent hesitant information in $[0,1]^3$ and set $pCF \equiv 0$. |
| Picture fuzzy soft rough set [752] | $[0,1]^3$ | $j = 3$ | Use picture fuzzy acceptance/neutrality/rejection in $[0,1]^3$ with $pCF \equiv 0$. |
| Spherical fuzzy soft rough set [709, 710, 753] | $[0,1]^3$ | $j = 3$ | Use spherical fuzzy components in $[0,1]^3$ and set $pCF \equiv 0$. |
| Quadripartitioned Neutrosophic soft rough set | $[0,1]^4$ | $j = 4$ | Take $F_L, F_U : U \to [0,1]^4$ and set $pCF \equiv 0$. |
| Pentapartitioned Neutrosophic soft rough set | $[0,1]^5$ | $j = 5$ | Take $F_L, F_U : U \to [0,1]^5$ and set $pCF \equiv 0$. |

**Example 3.17.4** (Supplier selection under two parameters)**.** Consider a purchasing department that must decide whether to treat a supplier as "strategic partner."

Universe and parameters.

$$U = \{S_1, S_2\}, \quad E = \{e_1, e_2\},$$

where $S_1$ = local supplier, $S_2$ = global supplier, $e_1$ = "on–time delivery," $e_2$ = "eco–compliance." The target concept is

$$A = \{S_2\},$$

meaning "strategic partner candidates."

Soft (plithogenic) parameter–based view. For each pair $(x, e) \in U \times E$, the decision maker gives a fuzzy soft membership $\mu_S(x, e) \in [0,1]$:

| $\mu_S(x,e)$ | $e_1$ (on–time) | $e_2$ (eco) |
|---|---|---|
| $S_1$ | 0.70 | 0.60 |
| $S_2$ | 0.90 | 0.80 |

Plithogenic rough relations. For each parameter $e \in E$ we have a plithogenic relation $R_e$ on $U$ with scalar degree of appurtenance $pdf_{R_e}$ and contradiction $pCF_{R_e}$. We work in the scalar case $s = t = 1$ and define

$$\tilde{R}_e(x,y) := pdf_{R_e}(x,y)\,\big(1 - pCF_{R_e}(x,y)\big) \in [0,1],$$

setting $\tilde{R}_e(x,x) = 1$ by convention.

For on–time delivery $e_1$:

| $pdf_{R_{e_1}}(x,y)$ | $S_1$ | $S_2$ |
|---|---|---|
| $S_1$ | – | 0.80 |
| $S_2$ | 0.75 | – |

| $pCF_{R_{e_1}}(x,y)$ | $S_1$ | $S_2$ |
|---|---|---|
| $S_1$ | – | 0.20 |
| $S_2$ | 0.25 | – |

so

$$\tilde{R}_{e_1}(S_1, S_2) = 0.80(1 - 0.20) = 0.64, \qquad \tilde{R}_{e_1}(S_2, S_1) = 0.75(1 - 0.25) = 0.5625.$$

For eco–compliance $e_2$:

| $pdf_{R_{e_2}}(x,y)$ | $S_1$ | $S_2$ |
|---|---|---|
| $S_1$ | – | 0.60 |
| $S_2$ | 0.70 | – |

| $pCF_{R_{e_2}}(x,y)$ | $S_1$ | $S_2$ |
|---|---|---|
| $S_1$ | – | 0.30 |
| $S_2$ | 0.20 | – |

thus

$$\tilde{R}_{e_2}(S_1,S_2) = 0.60(1-0.30) = 0.42, \qquad \tilde{R}_{e_2}(S_2,S_1) = 0.70(1-0.20) = 0.56.$$

Plithogenic rough lower/upper approximations (per parameter). For $A = \{S_2\}$ and any $x \in U$, the local plithogenic rough lower and upper approximations are

$$\underline{PL}_{R_e}(A)(x) = \max\bigl(1-\tilde{R}_e(x,S_2),\, 1-\tilde{R}_e(S_2,x)\bigr),$$

$$\overline{PL}_{R_e}(A)(x) = \max_{y\in U}\min\bigl(\tilde{R}_e(x,y),\, 1-\tilde{R}_e(y,x)\bigr).$$

For $x = S_1$ and $e_1$:

$$\underline{PL}_{R_{e_1}}(A)(S_1) = \max\bigl(1-0.64,\ 1-0.5625\bigr) = \max(0.36,\ 0.4375) = 0.4375.$$

For the upper approximation,

$$\overline{PL}_{R_{e_1}}(A)(S_1) = \max\Bigl\{\min\bigl(\tilde{R}_{e_1}(S_1,S_1), 1-\tilde{R}_{e_1}(S_1,S_1)\bigr),\ \min\bigl(\tilde{R}_{e_1}(S_1,S_2), 1-\tilde{R}_{e_1}(S_2,S_1)\bigr)\Bigr\}$$

$$= \max\bigl\{\min(1,0),\ \min(0.64, 1-0.5625)\bigr\} = \max\bigl\{0,\ \min(0.64, 0.4375)\bigr\} = \max\{0, 0.4375\} = 0.4375.$$

For $x = S_1$ and $e_2$:

$$\underline{PL}_{R_{e_2}}(A)(S_1) = \max\bigl(1-0.42,\ 1-0.56\bigr) = \max(0.58,\ 0.44) = 0.58,$$

$$\overline{PL}_{R_{e_2}}(A)(S_1) = \max\bigl\{\min(1,0),\ \min(0.42, 1-0.56)\bigr\} = \max\{0,\ \min(0.42, 0.44)\} = 0.42.$$

For the "prototypical" strategic supplier $x = S_2$ we have $\tilde{R}_e(S_2,S_2) = 1$ and thus

$$\underline{PL}_{R_e}(A)(S_2) = \max(1-1, 1-1) = 0$$

for both $e_1, e_2$, while the upper penalties remain moderate, e.g.

$$\overline{PL}_{R_{e_1}}(A)(S_2) = \max\bigl\{\min(1,0),\ \min(0.5625, 1-0.64)\bigr\} = \max\{0,\ \min(0.5625, 0.36)\} = 0.36,$$

$$\overline{PL}_{R_{e_2}}(A)(S_2) = \max\bigl\{\min(1,0),\ \min(0.56, 1-0.42)\bigr\} = \max\{0,\ \min(0.56, 0.58)\} = 0.56.$$

Plithogenic soft rough approximations. Take $j = 1$ and choose the scalarization

$$\Psi(a,b) := a\,(1-b), \qquad a, b \in [0,1],$$

so that the rough "penalty" $b$ is converted into a support factor $(1-b)$ and combined with the soft membership $a$.

The plithogenic soft rough lower and upper approximations of $A$ are

$$\underline{PSR}(A)(x,e) = \Psi\bigl(\mu_S(x,e),\, \underline{PL}_{R_e}(A)(x)\bigr) = \mu_S(x,e)\bigl(1-\underline{PL}_{R_e}(A)(x)\bigr),$$

$$\overline{PSR}(A)(x,e) = \Psi\big(\mu_S(x,e),\, \overline{PL}_{R_e}(A)(x)\big) = \mu_S(x,e)\big(1 - \overline{PL}_{R_e}(A)(x)\big).$$

For $x = S_1$:

$$\underline{PSR}(A)(S_1,e_1) = 0.70\big(1 - 0.4375\big) = 0.70 \times 0.5625 = 0.39375,$$
$$\overline{PSR}(A)(S_1,e_1) = 0.70\big(1 - 0.4375\big) = 0.39375,$$
$$\underline{PSR}(A)(S_1,e_2) = 0.60\big(1 - 0.58\big) = 0.60 \times 0.42 = 0.252,$$
$$\overline{PSR}(A)(S_1,e_2) = 0.60\big(1 - 0.42\big) = 0.60 \times 0.58 = 0.348.$$

For $x = S_2$:

$$\underline{PSR}(A)(S_2,e_1) = 0.90\big(1 - 0\big) = 0.90, \qquad \overline{PSR}(A)(S_2,e_1) = 0.90\big(1 - 0.36\big) = 0.90 \times 0.64 = 0.576,$$
$$\underline{PSR}(A)(S_2,e_2) = 0.80\big(1 - 0\big) = 0.80, \qquad \overline{PSR}(A)(S_2,e_2) = 0.80\big(1 - 0.56\big) = 0.80 \times 0.44 = 0.352.$$

Interpretation. For the "strategic partner" concept $A = \{S_2\}$, the plithogenic soft rough set

$$PSR_{\mathbb{S}}(A) = \big(\underline{PSR}(A), \overline{PSR}(A)\big)$$

yields, parameter by parameter, a pair of graded soft approximations that jointly reflect:

- the soft, parameterized preferences of the purchasing team ($\mu_S$ on on–time delivery and eco–compliance),
- the plithogenic rough neighborhood of the strategic supplier $S_2$ under each parameter (via $R_{e_1}, R_{e_2}$, including contradiction),
- and a combined assessment through $\Psi$.

Here $S_2$ has large lower values (0.90 for $e_1$, 0.80 for $e_2$), meaning it is strongly inside the plithogenic soft rough lower approximation of "strategic partner." Supplier $S_1$ has moderate values, especially under eco–compliance, indicating a borderline candidate that may require contractual improvements or further audits before being upgraded to strategic status.

Next we show that the above notion simultaneously extends the plithogenic rough set and the plithogenic soft set.

**Theorem 3.17.5** (Generalization of plithogenic rough set)**.** *Let $(U, R)$ be a plithogenic rough approximation space as in Definition 2.XX (Plithogenic rough approximation space), and let $A \subseteq U$. Consider the plithogenic soft approximation space*

$$\mathbb{S}_R \;:=\; (U, E, \mu_S, pCF_E, \{R_e\}_{e\in E}, \Phi, \Psi)$$

*obtained as follows:*

- *take a singleton parameter set $E := \{e_0\}$;*

- *define the plithogenic soft membership to be constant*

$$\mu_S(x, e_0) \ := \ 1 \in [0,1]^j, \qquad x \in U,$$

*and the parameter contradiction to be trivial*

$$pCF_E(e_0, e_0) \ := \ \mathbf{0};$$

- *set* $R_{e_0} := R$ *and use the same aggregation* $\Phi$ *as in the definition of* $\tilde{R}(x, y)$*;*
- *choose the plithogenic aggregation* $\Psi : [0,1]^j \times [0,1] \to [0,1]$ *in the scalar case* $j = 1$ *by*

$$\Psi(a, \lambda) \ := \ \lambda, \qquad a, \lambda \in [0,1],$$

*i.e.* $\Psi$ *simply projects to the second argument.*

*Then, for every* $x \in U$*,*

$$\underline{PSR}(A)(x, e_0) \ = \ \underline{PL}_R(A)(x), \qquad \overline{PSR}(A)(x, e_0) \ = \ \overline{PL}_R(A)(x),$$

*so that* $PSR_{\mathbb{S}_R}(A)$ *coincides with the plithogenic rough set* $PL_R(A) = (\underline{PL}_R(A), \overline{PL}_R(A))$*.*

*Proof.* Fix $x \in U$. Since $E = \{e_0\}$, all expressions are evaluated at $e_0$. By definition of $\underline{PSR}(A)$ we have

$$\underline{PSR}(A)(x, e_0) \ = \ \Psi\big(\mu_S(x, e_0), \, \underline{PL}_{R_{e_0}}(A)(x)\big).$$

In $\mathbb{S}_R$ we imposed

$$\mu_S(x, e_0) = 1, \qquad R_{e_0} = R,$$

hence

$$\underline{PSR}(A)(x, e_0) \ = \ \Psi\big(1, \, \underline{PL}_R(A)(x)\big).$$

The chosen $\Psi(a, \lambda) = \lambda$ yields

$$\underline{PSR}(A)(x, e_0) \ = \ \underline{PL}_R(A)(x).$$

The computation for the upper approximation is analogous:

$$\overline{PSR}(A)(x, e_0) \ = \ \Psi\big(\mu_S(x, e_0), \, \overline{PL}_{R_{e_0}}(A)(x)\big) \ = \ \Psi\big(1, \, \overline{PL}_R(A)(x)\big) \ = \ \overline{PL}_R(A)(x).$$

Therefore the pair

$$PSR_{\mathbb{S}_R}(A) \ = \big(\underline{PSR}(A)(\cdot, e_0), \, \overline{PSR}(A)(\cdot, e_0)\big)$$

is exactly the plithogenic rough set $PL_R(A)$, up to the natural identification of the singleton parameter $e_0$. This shows that every plithogenic rough set arises as a special case of a plithogenic soft rough set. □

**Theorem 3.17.6** (Generalization of plithogenic soft set)**.** *Let* $(U, E, \mu_S, pCF_E)$ *be a plithogenic soft set (cf. the reduction to PSS in Table 3.20), where*

$$\mu_S : U \times E \to [0,1]^j, \qquad pCF_E : E \times E \to [0,1]^t,$$

*and let* $A \subseteq U$ *be arbitrary. Consider any family of plithogenic relations* $\{R_e\}_{e \in E}$ *on* $U$ *and define* $\mathbb{S}_S$ *by*

$$\mathbb{S}_S \ := \ (U, E, \mu_S, pCF_E, \{R_e\}_{e \in E}, \Phi, \Psi),$$

*where* $\Phi$ *is as before and* $\Psi$ *is now chosen as the projection to the first argument:*

$$\Psi(\mu, \lambda) \ := \ \mu, \qquad \mu \in [0,1]^j, \ \lambda \in [0,1].$$

*Then, for all* $(x, e) \in U \times E$*,*

$$\underline{PSR}(A)(x, e) \ = \ \mu_S(x, e), \qquad \overline{PSR}(A)(x, e) \ = \ \mu_S(x, e),$$

*so that* $PSR_{\mathbb{S}_S}(A)$ *coincides with the original plithogenic soft set* $(U, E, \mu_S, pCF_E)$ *(viewed as a pair of identical lower/upper plithogenic soft sets).*

*Proof.* Fix $(x,e) \in U \times E$. By definition of $\underline{PSR}(A)$,

$$\underline{PSR}(A)(x,e) \;=\; \Psi\big(\mu_S(x,e),\, \underline{PL}_{R_e}(A)(x)\big).$$

With $\Psi(\mu,\lambda) = \mu$ we obtain

$$\underline{PSR}(A)(x,e) \;=\; \mu_S(x,e),$$

independently of the value of $\underline{PL}_{R_e}(A)(x)$. The same argument applies to the upper approximation:

$$\overline{PSR}(A)(x,e) \;=\; \Psi\big(\mu_S(x,e),\, \overline{PL}_{R_e}(A)(x)\big) \;=\; \mu_S(x,e).$$

Therefore, for each $e \in E$ the maps

$$x \longmapsto \underline{PSR}(A)(x,e), \qquad x \longmapsto \overline{PSR}(A)(x,e)$$

both coincide with the original plithogenic soft membership $x \mapsto \mu_S(x,e)$. Equivalently, the pair of plithogenic soft sets $(F_L, F_U)$ encoded by $PSR_{\mathbb{S}_S}(A)$ satisfies

$$F_L(e) = F_U(e) = \mu_S(\cdot,e) \qquad (e \in E),$$

which is exactly the given plithogenic soft set $(U, E, \mu_S, pCF_E)$. Hence every plithogenic soft set is realized as a specialization of a plithogenic soft rough set. □

## 3.18 Linear Diophantine Plithogenic set

A linear Diophantine plithogenic set uses multi-component memberships and integer coefficients, constrained by linear Diophantine equations, inside a contradiction-aware plithogenic attribute structure for complex uncertainty.

**Definition 3.18.1** (Linear Diophantine plithogenic number)**.** Fix an integer $s \geq 1$ and a constant $C > 0$. An *s–dimensional linear Diophantine plithogenic number* is a pair

$$\Lambda \;:=\; \big(\boldsymbol{\mu}, \boldsymbol{\alpha}\big) \;\in\; [0,1]^s \times [0,1]^s,$$

where

$$\boldsymbol{\mu} = (\mu_1, \ldots, \mu_s), \qquad \boldsymbol{\alpha} = (\alpha_1, \ldots, \alpha_s),$$

satisfy the linear Diophantine–type constraints

$$0 \;\leq\; \sum_{i=1}^{s} \alpha_i\, \mu_i \;\leq\; C, \qquad 0 \;\leq\; \sum_{i=1}^{s} \alpha_i \;\leq\; C. \tag{3.12}$$

The quantity

$$\pi(\Lambda) \;:=\; C - \sum_{i=1}^{s} \alpha_i\, \mu_i$$

plays the role of a (reference–dependent) hesitancy or residual degree.

**Definition 3.18.2** (Linear Diophantine plithogenic set)**.** Let $X$ be a nonempty universe and let $v$ be a plithogenic attribute with set of attribute values $P_v$. Let

$$pCF : P_v \times P_v \longrightarrow [0,1]$$

be the plithogenic degree–of–contradiction function. Fix an integer $s \geq 1$ and a constant $C > 0$.

A *linear Diophantine plithogenic set* (LDPS) on $X$ with respect to $(v, P_v, pCF)$ is a structure

$$\mathsf{PS}_{LD} \;:=\; \big(X,\ v,\ P_v,\ pCF,\ \text{LDpdf},\ C\big),$$

where

$$\mathrm{LDpdf}: X \times P_v \longrightarrow [0,1]^s \times [0,1]^s, \qquad (x,a) \longmapsto \big(\boldsymbol{\mu}(x,a), \boldsymbol{\alpha}(a)\big),$$

such that, for all $x \in X$ and $a \in P_v$,

$$0 \le \sum_{i=1}^{s} \alpha_i(a)\, \mu_i(x,a) \le C, \qquad 0 \le \sum_{i=1}^{s} \alpha_i(a) \le C. \tag{3.13}$$

Here

$$\boldsymbol{\mu}(x,a) = \big(\mu_1(x,a), \dots, \mu_s(x,a)\big), \qquad \boldsymbol{\alpha}(a) = \big(\alpha_1(a), \dots, \alpha_s(a)\big),$$

are, respectively, plithogenic component degrees of $x$ with respect to the value $a$ and the associated reference parameters for $a$.

For each $(x,a)$ the pair

$$\Lambda(x,a) := \big(\boldsymbol{\mu}(x,a), \boldsymbol{\alpha}(a)\big)$$

is an $s$–dimensional linear Diophantine plithogenic number in the sense of the previous definition. The residual degree at $(x,a)$ is

$$\pi(x,a) := C - \sum_{i=1}^{s} \alpha_i(a)\, \mu_i(x,a).$$

Table 3.26 presents the description of the Linear Diophantine plithogenic set as a common generalization of several existing Linear Diophantine models.

Table 3.26: Linear Diophantine plithogenic set as a common generalization

| Model | Components | Ref. params | Diophantine constraint | Plithogenic part |
|---|---|---|---|---|
| Linear Diophantine fuzzy set [754–756] | $(\mu_D, \nu_D)$ | $(\alpha, \beta)$ | $0 \le \alpha\mu_D + \beta\nu_D \le 1, \quad 0 \le \alpha + \beta \le 1$ | none |
| Linear Diophantine neutrosophic set [757] | $(T_D, I_D, F_D)$ | $(\alpha, \delta, \beta)$ | $0 \le \alpha T_D + \delta I_D + \beta F_D \le 2, \quad 0 \le \alpha + \delta + \beta \le 2$ | none |
| Linear Diophantine spherical fuzzy set [758–760] | $(T_D, U_D, K_D)$ | $(\alpha, \beta, \eta)$ | $0 \le \alpha T_D + \beta U_D + \eta K_D \le 1, \quad 0 \le \alpha + \beta + \eta \le 1$ | none |
| Linear Diophantine hesitant fuzzy set [761] | $(h_D^-, h_D^+)$ | $(\alpha, \beta)$ | $0 \le \alpha h_D^- + \beta h_D^+ \le 1, \quad 0 \le \alpha + \beta \le 1$ | none |
| Linear Diophantine plithogenic set | $(\mu_1, \dots, \mu_s)$ | $(\alpha_1, \dots, \alpha_s)$ | $0 \le \sum_{i=1}^{s} \alpha_i\mu_i \le C, \quad 0 \le \sum_{i=1}^{s} \alpha_i \le C$ | $(v, P_v, pCF)$ |

A concrete example is provided below.

**Example 3.18.3** (Linear Diophantine plithogenic set for smartphone selection)**.** Consider a household choosing among smartphones

$$X = \{x_1, x_2, x_3\} = \{\text{EcoPhone}, \ \text{GamePhone}, \ \text{BudgetPhone}\}.$$

Let $v$ be the plithogenic attribute "evaluation criterion" with value set

$$P_v = \{\text{price}, \ \text{battery}, \ \text{camera}\}.$$

The plithogenic degree of contradiction $pCF : P_v \times P_v \to [0,1]$ is set as

$$pCF(\text{price}, \text{battery}) = 0.30, \quad pCF(\text{price}, \text{camera}) = 0.20, \quad pCF(\text{battery}, \text{camera}) = 0.40,$$

with $pCF(a, a) = 0$ and symmetry $pCF(a, b) = pCF(b, a)$.

Take $s = 3$ components and $C = 2$. For each value $a \in P_v$ let

$$\boldsymbol{\alpha}(a) = (\alpha_1(a), \alpha_2(a), \alpha_3(a)) = (1,\ 0.5,\ 0.5),$$

so that

$$0 \le \sum_{i=1}^{3} \alpha_i(a) = 1 + 0.5 + 0.5 = 2 \le C.$$

Interpret the three membership components for $(x, a)$ as a neutrosophic-type triple

$$\boldsymbol{\mu}(x, a) = (\mu_1(x, a), \mu_2(x, a), \mu_3(x, a)) = (T, I, F) \in [0, 1]^3,$$

where $T$ measures support that $x$ is good on criterion $a$, $I$ expresses indeterminacy, and $F$ expresses counter-evidence.

For the EcoPhone $x_1$ define

$$\boldsymbol{\mu}(x_1, \text{price}) = (0.70,\ 0.10,\ 0.20),$$
$$\boldsymbol{\mu}(x_1, \text{battery}) = (0.90,\ 0.05,\ 0.10),$$
$$\boldsymbol{\mu}(x_1, \text{camera}) = (0.75,\ 0.15,\ 0.20).$$

The Linear Diophantine plithogenic constraints at $(x_1, a)$ read

$$0 \le \sum_{i=1}^{3} \alpha_i(a)\mu_i(x_1, a) \le C, \qquad 0 \le \sum_{i=1}^{3} \alpha_i(a) \le C.$$

Check explicitly for $a = \text{battery}$:

$$\sum_{i=1}^{3} \alpha_i(\text{battery})\mu_i(x_1, \text{battery}) = 1 \cdot 0.90 + 0.5 \cdot 0.05 + 0.5 \cdot 0.10 = 0.90 + 0.025 + 0.05 = 0.975.$$

Thus

$$0 \le 0.975 \le C = 2, \qquad 0 \le 1 + 0.5 + 0.5 = 2 \le 2,$$

so the Linear Diophantine conditions are satisfied at $(x_1, \text{battery})$. The corresponding residual degree is

$$\pi(x_1, \text{battery}) := C - \sum_{i=1}^{3} \alpha_i(\text{battery})\mu_i(x_1, \text{battery}) = 2 - 0.975 = 1.025.$$

Similarly, one defines $\boldsymbol{\mu}(x_2, a)$ and $\boldsymbol{\mu}(x_3, a)$ for $a \in P_v$ (e.g. GamePhone better on camera, BudgetPhone better on price), always ensuring

$$0 \le \sum_{i=1}^{3} \alpha_i(a)\mu_i(x, a) \le 2.$$

The resulting structure

$$\mathsf{PS}_{LD} = \big(X,\ v,\ P_v,\ pCF, \text{LDpdf},\ C\big),$$

with

$$\text{LDpdf}(x, a) = \big(\boldsymbol{\mu}(x, a), \boldsymbol{\alpha}(a)\big), \qquad (x, a) \in X \times P_v,$$

is a Linear Diophantine plithogenic set. It combines: (i) Diophantine-type linear constraints on the weighted memberships, (ii) plithogenic contradictions between criteria via $pCF$, to model a realistic multi-criteria smartphone selection problem.

## 3.19 TreePlithogenic Set

A TreePlithogenic set organizes attributes in a rooted tree and aggregates their plithogenic memberships using contradiction-aware hierarchical weighting across levels [762]. Related concepts such as TreeSoft Set [319–321, 763, 764] and TreeRough Set [762, 765] are also known in the literature.

**Definition 3.19.1** (TreePlithogenic Set)**.** [762] Let $S$ be a universal set and $P \subseteq S$ a nonempty subset. Let Tree($A$) be a finite rooted attribute–tree whose nodes are attributes $a_i$ arranged in levels $1, \ldots, m$. For every node (attribute) $a_i \in \mathrm{Tree}(A)$ let $Pv_i$ be the set of admissible values of $a_i$, and let

$$pdf_i : P \times Pv_i \longrightarrow [0,1]^{s_i}$$

be the (possibly multi–component) plithogenic degree of appurtenance attached to $a_i$. Assume further a global degree of contradiction function

$$pCF : \Big(\bigcup_i Pv_i\Big) \times \Big(\bigcup_i Pv_i\Big) \longrightarrow [0,1]^t$$

satisfying (i) $pCF(a,a) = 0$ for all $a$ (reflexivity), and (ii) $pCF(a,b) = pCF(b,a)$ for all $a, b$ (symmetry).

A *TreePlithogenic Set* is the tuple

$$\mathrm{TPS} \;=\; \big(P,\ \mathrm{Tree}(A),\ \{Pv_i\}_{a_i \in \mathrm{Tree}(A)},$$
$$\{pdf_i\}_{a_i \in \mathrm{Tree}(A)},\ pCF\big),$$

which, for every subset of nodes $X \subseteq \mathrm{Tree}(A)$ and every element $x \in P$, collects all plithogenic memberships $pdf_i(x, \cdot)$ of $x$ with respect to the attributes present in $X$, while modulating/weighting them by $pCF$ according to the mutual contradiction of their value–labels. In particular, if the tree has only one level and one attribute, this reduces to the ordinary plithogenic set.

Table 3.27 presents the description of a TreePlithogenic Set as a unifying model for tree–based fuzzy, intuitionistic fuzzy, and neutrosophic sets.

Table 3.27: TreePlithogenic Set as a unifying model for tree–based fuzzy / intuitionistic fuzzy / neutrosophic sets.

| **Target (tree)** | **Attribute/tree restriction** | **Outcome in TPS** |
|---|---|---|
| TreeFuzzy Set [762] | One rooted attribute–tree; each node has a single scalar membership in $[0, 1]$ | TreePlithogenic Set with $s_i = 1$ for all nodes and $pCF \equiv 0$; fuzzy case recovered |
| Tree-Intuitionistic Fuzzy Set [762] (Tree-Vague Set) | One rooted attribute–tree; each node stores $(\mu, \nu)$ with $\mu + \nu \leq 1$ | TreePlithogenic Set with $s_i = 2$ and $pCF \equiv 0$; intuitionistic tree recovered |
| TreeNeutrosophic Set [762] (Tree-Picture Fuzzy, Tree-Hesitant Fuzzy, and Tree-Spherical Fuzzy Set) | One rooted attribute–tree; each node stores $(T, I, F) \in [0, 1]^3$ | TreePlithogenic Set with $s_i = 3$ and $pCF \equiv 0$; neutrosophic tree recovered |
| Tree-Quadripartitioned Neutrosophic Set | One rooted attribute–tree; | TreePlithogenic Set with $s_i = 4$ and $pCF \equiv 0$; neutrosophic tree recovered |
| Tree-Pentapartitioned Neutrosophic Set | One rooted attribute–tree; | TreePlithogenic Set with $s_i = 5$ and $pCF \equiv 0$; neutrosophic tree recovered |

A concrete example of this concept is provided below.

**Example 3.19.2** (Hiring a data engineer: contradiction-aware hierarchical fit)**.** Let the attribute-tree be root Fit with three children: Technical, Experience, Culture. Each child has two leaf attributes:

$$\text{Technical} = \{\text{Coding},\ \text{DataModeling}\},$$

$$\text{Experience} = \{\text{Projects},\ \text{Domain}\},$$

$$\text{Culture} = \{\text{Communication},\ \text{Teamwork}\}.$$

Each node uses a fuzzy plithogenic membership (scalar in $[0,1]$), with dominant value set to High at every level. Let the linguistic labels be $\{\text{Low},\ \text{Med},\ \text{High}\}$ and the contradiction to High be

$$pCF(\text{High},\text{High}) = 0, \quad pCF(\text{Med},\text{High}) = 0.3, \quad pCF(\text{Low},\text{High}) = 0.8.$$

The compatibility weight is $w(\ell \mid \text{High}) = 1 - pCF(\ell, \text{High})$.

Leaf evaluations for candidate $c$ (observed label; base membership $\mu$):

| Leaf | Label | $\mu$ | $w(\cdot \mid \text{High})$ |
|---|---|---|---|
| Coding | High | 0.82 | 1.0 |
| DataModeling | Med | 0.68 | 0.7 |
| Projects | High | 0.75 | 1.0 |
| Domain | Med | 0.55 | 0.7 |
| Communication | Med | 0.62 | 0.7 |
| Teamwork | High | 0.78 | 1.0 |

Aggregate each internal node by weighted mean $\widehat{\mu} = \frac{\sum w\mu}{\sum w}$:

$$\mu_{\text{Technical}} = \frac{0.82 \cdot 1 + 0.68 \cdot 0.7}{1 + 0.7} = \frac{0.82 + 0.476}{1.7} = \frac{1.296}{1.7} = 0.76235.$$

$$\mu_{\text{Experience}} = \frac{0.75 \cdot 1 + 0.55 \cdot 0.7}{1 + 0.7} = \frac{0.75 + 0.385}{1.7} = \frac{1.135}{1.7} = 0.66765.$$

$$\mu_{\text{Culture}} = \frac{0.62 \cdot 0.7 + 0.78 \cdot 1}{0.7 + 1} = \frac{0.434 + 0.78}{1.7} = \frac{1.214}{1.7} = 0.71412.$$

Intermediate node labels (for root-level contradiction) via thresholding: High if $\mu \geq 0.70$, Med if $0.50 \leq \mu < 0.70$, Low otherwise. Hence

$$\text{Technical} = \text{High}, \quad \text{Experience} = \text{Med}, \quad \text{Culture} = \text{High},$$

so the root weights are $w(\text{High} \mid \text{High}) = 1$, $w(\text{Med} \mid \text{High}) = 0.7$.

Root aggregation (Fit):

$$\mu_{\text{Fit}} = \frac{0.76235 \cdot 1 + 0.66765 \cdot 0.7 + 0.71412 \cdot 1}{1 + 0.7 + 1} = \frac{0.76235 + 0.46736 + 0.71412}{2.7} = \frac{1.94383}{2.7} = 0.72068.$$

Thus, candidate $c$ attains a contradiction-aware hierarchical Fit score $\approx 0.721$.

**Example 3.19.3** (Emergency triage: pneumonia severity under a clinical tree)**.** Tree: root Severity with children Vitals, Imaging, Labs; leaves Vitals=$\{SpO_2$, RespRate$\}$, Imaging=$\{$CXR, CT$\}$, Labs=$\{$CRP, WBC$\}$. Dominant label at all levels is Severe. Labels $\{\text{Mild},\ \text{Moderate},\ \text{Severe}\}$ with contradictions

$$pCF(\text{Severe},\text{Severe}) = 0, \quad pCF(\text{Moderate},\text{Severe}) = 0.35, \quad pCF(\text{Mild},\text{Severe}) = 0.85,$$

and weights $w(\ell \mid \text{Severe}) = 1 - pCF(\ell, \text{Severe})$.

Leaf assessments for patient $p$:

| Leaf | Label | $\mu$ | $w(\cdot \mid \text{Severe})$ |
|---|---|---|---|
| $SpO_2$ | Severe | 0.88 | 1.0 |
| RespRate | Moderate | 0.65 | 0.65 |
| CXR | Moderate | 0.70 | 0.65 |
| CT | Mild | 0.40 | 0.15 |
| CRP | Moderate | 0.60 | 0.65 |
| WBC | Severe | 0.75 | 1.0 |

Node aggregations:

$$\mu_{\text{Vitals}} = \frac{0.88 \cdot 1 + 0.65 \cdot 0.65}{1 + 0.65} = \frac{0.88 + 0.4225}{1.65} = \frac{1.3025}{1.65} = 0.78939.$$

$$\mu_{\text{Imaging}} = \frac{0.70 \cdot 0.65 + 0.40 \cdot 0.15}{0.65 + 0.15} = \frac{0.455 + 0.06}{0.80} = \frac{0.515}{0.80} = 0.64375.$$

$$\mu_{\text{Labs}} = \frac{0.60 \cdot 0.65 + 0.75 \cdot 1}{0.65 + 1} = \frac{0.39 + 0.75}{1.65} = \frac{1.14}{1.65} = 0.69091.$$

Intermediate labels (High/Severe if $\mu \geq 0.70$): Vitals=Severe, Imaging=Moderate, Labs=Moderate. Root weights: $w(\text{Severe} \mid \text{Severe}) = 1$, $w(\text{Moderate} \mid \text{Severe}) = 0.65$.

Root aggregation (Severity):

$$\mu_{\text{Severity}} = \frac{0.78939 \cdot 1 + 0.64375 \cdot 0.65 + 0.69091 \cdot 0.65}{1 + 0.65 + 0.65}$$

$$= \frac{0.78939 + 0.41844 + 0.44909}{2.30} = \frac{1.65692}{2.30} = 0.72040.$$

The patient's contradiction-aware severity score is $\approx 0.720$.

**Example 3.19.4** (Supplier selection: suitability under cost–quality–delivery tree)**.** Tree: root Suitability with children Cost, Quality, Delivery. Dominant label at all levels is Preferred. Labels {Risky, Acceptable, Preferred} with

$$pCF(\text{Preferred}, \text{Preferred}) = 0, \quad pCF(\text{Acceptable}, \text{Preferred}) = 0.4, \quad pCF(\text{Risky}, \text{Preferred}) = 0.85,$$

and $w(\ell \mid \text{Preferred}) = 1 - pCF(\ell, \text{Preferred}) \in \{1, 0.6, 0.15\}$.

Leaves for supplier $s$:

| Leaf | Label | $\mu$ | $w(\cdot \mid \text{Preferred})$ |
|---|---|---|---|
| UnitPrice | Acceptable | 0.62 | 0.6 |
| TotalCost | Preferred | 0.78 | 1.0 |
| DefectRate | Preferred | 0.83 | 1.0 |
| Certifications | Acceptable | 0.58 | 0.6 |
| OnTime | Acceptable | 0.65 | 0.6 |
| LeadTime | Risky | 0.35 | 0.15 |

Node aggregations:

$$\mu_{\text{Cost}} = \frac{0.62 \cdot 0.6 + 0.78 \cdot 1}{0.6 + 1} = \frac{0.372 + 0.78}{1.6} = \frac{1.152}{1.6} = 0.72.$$

$$\mu_{\text{Quality}} = \frac{0.83 \cdot 1 + 0.58 \cdot 0.6}{1 + 0.6} = \frac{0.83 + 0.348}{1.6} = \frac{1.178}{1.6} = 0.73625.$$

$$\mu_{\text{Delivery}} = \frac{0.65 \cdot 0.6 + 0.35 \cdot 0.15}{0.6 + 0.15} = \frac{0.39 + 0.0525}{0.75} = \frac{0.4425}{0.75} = 0.59.$$

Intermediate labels: Cost=Preferred, Quality=Preferred, Delivery=Acceptable. Root weights: $w(\text{Preferred} \mid \text{Preferred}) = 1$, $w(\text{Acceptable} \mid \text{Preferred}) = 0.6$.

Root aggregation (Suitability):

$$\mu_{\text{Suitability}} = \frac{0.72 \cdot 1 + 0.73625 \cdot 1 + 0.59 \cdot 0.6}{1 + 1 + 0.6} = \frac{0.72 + 0.73625 + 0.354}{2.6} = \frac{1.81025}{2.6} = 0.69625.$$

Hence the supplier receives a contradiction-aware suitability of $\approx 0.696$, i.e., borderline "preferred" under the specified hierarchy and contradictions.

## 3.20 ForestPlithogenic Set

A ForestPlithogenic set spans multiple attribute trees and fuses their plithogenic memberships while weighting inter-tree contradictions for consistent evaluation results [762].

**Definition 3.20.1** (ForestPlithogenic Set)**.** [762] Let $\{\, \mathrm{TPS}_t \,\}_{t\in T}$ be a family of TreePlithogenic Sets, where the $t$–th tree has attribute–tree $\mathrm{Tree}(A^{(t)})$, value–sets $\{Pv_i^{(t)}\}$, appurtenance maps $\{pdf_i^{(t)}\}$, and (possibly) its own contradiction map $pCF^{(t)}$. Form the disjoint union (forest) of attribute–trees

$$\mathrm{Forest}\big(\{A^{(t)}\}_{t\in T}\big) \;:=\; \bigsqcup_{t\in T} \mathrm{Tree}(A^{(t)}).$$

A *ForestPlithogenic Set* on the common universe $P$ is the tuple

$$\mathrm{FPS} \;=\; \big(P,\ \mathrm{Forest}(\{A^{(t)}\}),\ \{Pv_i^{(t)}\},\ \{pdf_i^{(t)}\},\ \widetilde{pCF}\big),$$

where $\widetilde{pCF}$ extends all $pCF^{(t)}$'s to the union of value–labels coming from every tree. For any subset of forest–nodes $X$ and any element $x \in P$, the forest–level membership of $x$ is obtained by aggregating the plithogenic memberships provided by every tree whose nodes appear in $X$, under the common contradiction control $\widetilde{pCF}$. If the forest consists of exactly one tree, the above reduces to a TreePlithogenic Set.

Table 3.28 provides an explanation of the ForestPlithogenic Set as a unifying model for forest–based fuzzy, intuitionistic fuzzy, and neutrosophic sets.

A concrete example of this concept is provided below.

**Example 3.20.2** (Smart-city site selection: forest of environmental, infrastructure, and socio-economic trees)**.** City site selection evaluates potential urban locations using criteria like transport, resources, risk, environment, and growth to choose optimal development (cf. [766]).

Consider three TreePlithogenic Sets over parcels $P$: $\mathrm{TPS}_{\mathrm{Env}}$, $\mathrm{TPS}_{\mathrm{Infra}}$, and $\mathrm{TPS}_{\mathrm{Socio}}$. Each node uses a fuzzy membership in $[0, 1]$. Linguistic labels are $\{\mathrm{Low}, \mathrm{Med}, \mathrm{High}\}$ with contradiction to the dominant label High:

$$pCF(\mathrm{High}, \mathrm{High}) = 0, \quad pCF(\mathrm{Med}, \mathrm{High}) = 0.3,$$

Table 3.28: ForestPlithogenic Set as a unifying model for forest–based fuzzy / intuitionistic fuzzy / neutrosophic sets.

| **Target (forest)** | **Forest restriction** | **Outcome in FPS** |
|---|---|---|
| ForestFuzzy Set [762] | Finite disjoint family of attribute–trees; every node fuzzy in $[0,1]$; no cross–tree contradiction | ForestPlithogenic Set with $s_i = 1$ for all nodes and $\widetilde{pCF} \equiv 0$ |
| ForestIntuitionistic Fuzzy Set [762] | Forest of trees; each node has $(\mu, \nu)$; no cross–tree contradiction | ForestPlithogenic Set with $s_i = 2$ and $\widetilde{pCF} \equiv 0$ |
| ForestNeutrosophic Set [762] | Forest of trees; each node has $(T, I, F)$ | ForestPlithogenic Set with $s_i = 3$ and $\widetilde{pCF} \equiv 0$ |
| Forest-Quadripartitioned Neutrosophic Set | Forest of trees | ForestPlithogenic Set with $s_i = 4$ and $\widetilde{pCF} \equiv 0$ |
| Forest-Pentapartitioned Neutrosophic Set | Forest of trees | ForestPlithogenic Set with $s_i = 5$ and $\widetilde{pCF} \equiv 0$ |

$$pCF(\text{Low}, \text{High}) = 0.8,$$

and weights $w(\ell \mid \text{High}) = 1 - pCF(\ell, \text{High}) \in \{1,\ 0.7,\ 0.2\}$.

For a parcel $x \in P$, leaf assessments (label; base membership $\mu$) are:

(Env) AirQuality=(High; 0.80), GreenCover=(Med; 0.65), Noise=(High; 0.75).

$$\mu_{\text{Env}}(x) = \frac{0.80 \cdot 1 + 0.65 \cdot 0.7 + 0.75 \cdot 1}{1 + 0.7 + 1} = \frac{0.80 + 0.455 + 0.75}{2.7} = \frac{2.005}{2.7} = 0.74259.$$

(Infra) TransitAccess=(Med; 0.68), RoadConnectivity=(High; 0.77), Utilities=(Med; 0.60).

$$\mu_{\text{Infra}}(x) = \frac{0.68 \cdot 0.7 + 0.77 \cdot 1 + 0.60 \cdot 0.7}{0.7 + 1 + 0.7} = \frac{0.476 + 0.77 + 0.42}{2.4} = \frac{1.666}{2.4} = 0.69417.$$

(Socio) Safety=(Med; 0.70), CommunitySupport=(High; 0.74), Rent=(Low; 0.45).

$$\mu_{\text{Socio}}(x) = \frac{0.70 \cdot 0.7 + 0.74 \cdot 1 + 0.45 \cdot 0.2}{0.7 + 1 + 0.2} = \frac{0.49 + 0.74 + 0.09}{1.9} = \frac{1.32}{1.9} = 0.69474.$$

Thresholding (High if $\mu \geq 0.70$, Med if $0.50 \leq \mu < 0.70$) gives Env=High, Infra=Med, Socio=Med. The forest–level (dominant High) aggregate is

$$\mu_{\text{Forest}}(x) = \frac{0.74259 \cdot 1 + 0.69417 \cdot 0.7 + 0.69474 \cdot 0.7}{1 + 0.7 + 0.7}$$

$$= \frac{0.74259 + 0.48592 + 0.48632}{2.4} = \frac{1.71483}{2.4} = 0.71451.$$

Thus $x$ attains a contradiction-aware forest score $\approx 0.715$.

**Example 3.20.3** (Hospital readmission risk: forest of ClinicalHistory, CurrentStatus, SocialDeterminants)**.** Hospital readmission risk estimates the probability a discharged patient returns soon, guiding care coordination, follow up planning, and resource allocation (cf. [767]).

Trees: $\mathrm{TPS}_{\mathrm{CH}}$, $\mathrm{TPS}_{\mathrm{CS}}$, $\mathrm{TPS}_{\mathrm{SD}}$ with labels $\{\mathrm{Low}, \mathrm{Moderate}, \mathrm{High}\}$ and dominant High risk.

Contradictions to High: $pCF(\mathrm{High}, \mathrm{High}) = 0$, $pCF(\mathrm{Moderate}, \mathrm{High}) = 0.35$, $pCF(\mathrm{Low}, \mathrm{High}) = 0.85$, so $w(\cdot \mid \mathrm{High}) \in \{1,\, 0.65,\, 0.15\}$.

Patient $p$ leaf evaluations (label; $\mu$):

(ClinicalHistory) Comorbidity=(High; 0.82), PriorAdmissions=(Moderate; 0.66).

$$\mu_{\mathrm{CH}}(p) = \frac{0.82 \cdot 1 + 0.66 \cdot 0.65}{1 + 0.65} = \frac{0.82 + 0.429}{1.65} = \frac{1.249}{1.65} = 0.75697.$$

(CurrentStatus) BPControl=(Moderate; 0.55), Adherence=(Low; 0.40), Mobility=(Moderate; 0.60).

$$\mu_{\mathrm{CS}}(p) = \frac{0.55 \cdot 0.65 + 0.40 \cdot 0.15 + 0.60 \cdot 0.65}{0.65 + 0.15 + 0.65} = \frac{0.3575 + 0.06 + 0.39}{1.45} = \frac{0.8075}{1.45} = 0.55690.$$

(SocialDeterminants) Support=(Low; 0.35), Housing=(Moderate; 0.58), Access=(Moderate; 0.62).

$$\mu_{\mathrm{SD}}(p) = \frac{0.35 \cdot 0.15 + 0.58 \cdot 0.65 + 0.62 \cdot 0.65}{0.15 + 0.65 + 0.65} = \frac{0.0525 + 0.377 + 0.403}{1.45} = \frac{0.8325}{1.45} = 0.57310.$$

Tree labels: CH=High, CS=Moderate, SD=Moderate. Forest aggregation (dominant High):

$$\mu_{\mathrm{Forest}}(p) = \frac{0.75697 \cdot 1 + 0.55690 \cdot 0.65 + 0.57310 \cdot 0.65}{1 + 0.65 + 0.65}$$
$$= \frac{0.75697 + 0.36199 + 0.37252}{2.30} = \frac{1.49148}{2.30} = 0.64847.$$

Hence $p$'s contradiction-aware forest readmission score is $\approx 0.648$ (moderate–high).

**Example 3.20.4** (Portfolio selection: forest of Risk, Return, and Liquidity)**.** Portfolio selection chooses an optimal mix of assets balancing expected return, risk tolerance, diversification, and constraints like liquidity or regulations (cf. [768]).

Trees: $\mathrm{TPS}_{\mathrm{Risk}}$, $\mathrm{TPS}_{\mathrm{Return}}$, $\mathrm{TPS}_{\mathrm{Liq}}$. Labels $\{\mathrm{Undesirable}, \mathrm{Acceptable}, \mathrm{Attractive}\}$ with dominant Attractive.

Contradictions:

$$pCF(\mathrm{Attractive}, \mathrm{Attractive}) = 0$$

,

$$pCF(\mathrm{Acceptable}, \mathrm{Attractive}) = 0.4$$

,

$$pCF(\mathrm{Undesirable}, \mathrm{Attractive}) = 0.85$$

, so $w \in \{1,\, 0.6,\, 0.15\}$.

Asset $a$ leaf evaluations (label; $\mu$):

(Risk) Volatility=(Acceptable; 0.68), Drawdown=(Acceptable; 0.70), Diversification=(Attractive; 0.74).

$$\mu_{\text{Risk}}(a)=\frac{0.68\cdot 0.6+0.70\cdot 0.6+0.74\cdot 1}{0.6+0.6+1}=\frac{0.408+0.42+0.74}{2.2}=\frac{1.568}{2.2}=0.71273.$$

(Return) CAGR=(Attractive; 0.80), Sharpe=(Acceptable; 0.66), Alpha=(Attractive; 0.75).

$$\mu_{\text{Return}}(a)=\frac{0.80\cdot 1+0.66\cdot 0.6+0.75\cdot 1}{1+0.6+1}=\frac{0.80+0.396+0.75}{2.6}=\frac{1.946}{2.6}=0.74846.$$

(Liquidity) Turnover=(Acceptable; 0.64), BidAsk=(Undesirable; 0.40), Depth=(Acceptable; 0.67).

$$\mu_{\text{Liq}}(a)=\frac{0.64\cdot 0.6+0.40\cdot 0.15+0.67\cdot 0.6}{0.6+0.15+0.6}=\frac{0.384+0.06+0.402}{1.35}=\frac{0.846}{1.35}=0.62667.$$

Tree labels: Risk=Attractive, Return=Attractive, Liquidity=Acceptable. Forest aggregation (dominant Attractive):

$$\mu_{\text{Forest}}(a)=\frac{0.71273\cdot 1+0.74846\cdot 1+0.62667\cdot 0.6}{1+1+0.6}=\frac{0.71273+0.74846+0.37600}{2.6}=\frac{1.83719}{2.6}=0.70738.$$

Thus asset $a$ achieves a contradiction-aware forest score $\approx 0.707$, i.e., marginally attractive overall.

## 3.21 Plithogenic Soft Expert Set

Soft expert set extends soft sets by incorporating multiple experts' parameterized opinions for decision-making under uncertainty and differing viewpoints simultaneously [769–771]. A plithogenic soft expert set enriches the classical (fuzzy / intuitionistic / neutrosophic) soft expert set by attaching to each expert–parameter–opinion triple a plithogenic degree of appurtenance together with a contradiction-aware fusion [772]. This allows modeling heterogeneous multi-component memberships while explicitly penalizing conflicting labels.

**Definition 3.21.1** (Plithogenic Soft Expert Set (PSES))**.** [772] Let $U$ be a universe of discourse; $E$ a set of parameters; $X$ a set of experts; and $O=\{1=\text{agree},\,0=\text{disagree}\}$ a set of opinions. Put $Z=E\times X\times O$ and fix a finite index set $A\subseteq Z$ of activated triples.

For each parameter $e\in E$, let $Pv_e$ be its set of admissible value-labels. Define the label universe

$$\mathcal{L}\;:=\;\Big(\bigsqcup_{e\in E}Pv_e\Big)\;\sqcup\;O$$

as a disjoint union of all parameter value-sets and opinions. For every $e\in E$ fix a (possibly vector-valued) plithogenic degree-of-appurtenance (DAF)

$$pdf_e:\;U\times Pv_e\;\longrightarrow\;[0,1]^{s_e},$$

and fix a global degree-of-contradiction function (DCF)

$$pCF:\;\mathcal{L}\times\mathcal{L}\;\longrightarrow\;[0,1]^t,\qquad pCF(\ell,\ell)=\mathbf{0},\quad pCF(\ell_1,\ell_2)=pCF(\ell_2,\ell_1).$$

Let $\Phi : [0,1]^s \times [0,1]^t \to [0,1]$ be an aggregation ($s := \max_e s_e$) which is monotone nondecreasing in its first argument and nonincreasing in its second, and satisfies the neutral/annihilation conditions

$$\Phi(\mathbf{0}, \mathbf{b}) = 0, \qquad \Phi(\mathbf{a}, \mathbf{0}) = \Psi(\mathbf{a}) \in [0,1].$$

A *Plithogenic Soft Expert Set (PSES) on $U$* is a tuple

$$\text{PSES} \;=\; \big(U, E, X, O,\; A,\; \{Pv_e\}_{e\in E},\; \{pdf_e\}_{e\in E},\; pCF,\; \Phi,\; \text{val}\big),$$

where $\text{val} : A \to \bigsqcup_{e\in E} Pv_e$ assigns to each $\alpha = (e, x, o) \in A$ the value-label $\text{val}(\alpha) \in Pv_e$ used by expert $x$ under opinion $o$.

Its *plithogenic soft expert mapping* is

$$F_{\text{PL}} : A \;\longrightarrow\; [0,1]^U, \qquad F_{\text{PL}}(\alpha)(u) \;:=\; \Phi\Big(\, \iota_e\big(pdf_e(u, \text{val}(\alpha))\big),\; pCF\big(\text{val}(\alpha), o\big)\Big),$$

where $\alpha = (e, x, o)$ and $\iota_e : [0,1]^{s_e} \hookrightarrow [0,1]^s$ is the natural padding into the common $s$-dimensional cube.

**Remark 3.21.2** (Well-posedness and reductions)**.** (i) If $t = 0$ (no contradiction channel) then $F_{\text{PL}}(\alpha)(u) = \Psi\big(\iota_e(pdf_e(u, \text{val}(\alpha)))\big)$, recovering multi-component memberships without contradiction penalization.

(ii) Choosing $s = 1$ and $t = 0$ recovers fuzzy soft expert sets; choosing $s = 2$ and $t = 0$ recovers intuitionistic fuzzy soft expert sets; choosing $s = 3$ and $t = 0$ recovers neutrosophic soft expert sets. See Table 3.29.

Describe the content in Table 3.29 as Reductions showing that a Plithogenic Soft Expert Set (PSES) subsumes fuzzy, intuitionistic fuzzy, and neutrosophic soft expert sets.

Table 3.29: Reductions of a Plithogenic Soft Expert Set (PSES) to several soft expert models.

| Target model | Image of $F(\alpha)$ | Contradiction | Reduction from PSES |
|---|---|---|---|
| Fuzzy Soft Expert Set [771, 773, 774] | $[0,1]$ | $t = 0$ | Take $s = 1$; each $pdf_e(u, v) \in [0,1]$; set $pCF \equiv 0$ and $\Phi(a, 0) = a$. |
| Intuitionistic Fuzzy Soft Expert Set [769, 770, 775] | $[0,1]^2$ | $t = 0$ | Take $s = 2$; $pdf_e(u, v) = (\mu, \nu)$ with usual IF constraint; set $pCF \equiv 0$ and use standard IF-operations. |
| Neutrosophic Soft Expert Set [776–778] | $[0,1]^3$ | $t = 0$ | Take $s = 3$; $pdf_e(u, v) = (T, I, F)$; set $pCF \equiv 0$ and apply neutrosophic soft expert operations componentwise. |
| Hesitant Fuzzy Soft Expert Set [779, 780] | $[0,1]^3$ | $t = 0$ | Take $s = 3$; encode hesitant information in $(T, I, F)$-type components; set $pCF \equiv 0$ and use hesitant soft expert aggregation. |
| Picture Fuzzy Soft Expert Set [781] | $[0,1]^3$ | $t = 0$ | Take $s = 3$; interpret $(T, I, F)$ as picture fuzzy acceptance/neutrality/rejection; set $pCF \equiv 0$ and use picture fuzzy rules. |
| Spherical Fuzzy Soft Expert Set [782–784] | $[0,1]^3$ | $t = 0$ | Take $s = 3$; impose spherical fuzzy constraint on $(T, I, F)$; set $pCF \equiv 0$ and adopt spherical soft expert operators. |

A concrete example of this concept is provided below.

**Example 3.21.3** (Sustainable supplier selection as a PSES)**.** Systematic evaluation of suppliers using environmental, social, and economic criteria to reduce impacts and ensure reliable, ethical, long-term sourcing choices (cf. [785]).

Let $U = \{u_1, u_2, u_3\}$ be a set of candidate suppliers. Let

$E = \{e_1 = \text{“cost”},\ e_2 = \text{“sustainability”}\}, \quad X = \{x_1 = \text{procurement manager},\ x_2 = \text{sustainability officer}\},$

and $O = \{1 = \text{agree},\ 0 = \text{disagree}\}$.

For the parameters we take

$$Pv_{e_1} = \{\text{low}, \text{medium}, \text{high}\}, \qquad Pv_{e_2} = \{\text{green}, \text{neutral}, \text{risky}\},$$

and let $\mathcal{L}$ be the disjoint union of all $Pv_e$ together with $O$. Consider the activated triples

$$A = \big\{(e_1, x_1, 1),\ (e_2, x_1, 1),\ (e_2, x_2, 1)\big\}.$$

The value-assignment map $\text{val} : A \to \bigsqcup_{e \in E} Pv_e$ is given by

$$\text{val}(e_1, x_1, 1) = \text{low}, \quad \text{val}(e_2, x_1, 1) = \text{neutral}, \quad \text{val}(e_2, x_2, 1) = \text{green}.$$

Let $s_{e_1} = s_{e_2} = 1$ and define fuzzy DAFs

$$pdf_{e_1}, pdf_{e_2} : U \times Pv_{e_i} \to [0, 1]$$

by, for example,

$$\begin{aligned}
pdf_{e_1}(u_1, \text{low}) &= 0.9, & pdf_{e_1}(u_2, \text{low}) &= 0.6, & pdf_{e_1}(u_3, \text{low}) &= 0.2,\\
pdf_{e_2}(u_1, \text{green}) &= 0.7, & pdf_{e_2}(u_2, \text{green}) &= 0.4, & pdf_{e_2}(u_3, \text{green}) &= 0.1,\\
pdf_{e_2}(u_1, \text{neutral}) &= 0.5, & pdf_{e_2}(u_2, \text{neutral}) &= 0.7, & pdf_{e_2}(u_3, \text{neutral}) &= 0.6.
\end{aligned}$$

To model plithogenic contradiction, let $t = 1$ and define

$$pCF : \mathcal{L} \times \mathcal{L} \to [0, 1]$$

by

$$pCF(\text{low}, 1) = 0.1, \quad pCF(\text{green}, 1) = 0.1, \quad pCF(\text{neutral}, 1) = 0.3,$$

and set $pCF(\ell, \ell) = 0$ and $pCF(\ell_1, \ell_2) = 0$ for all other pairs, so that “neutral” sustainability is slightly more contradictory to a strong positive opinion than “green”.

Take the aggregation

$$\Phi : [0, 1] \times [0, 1] \to [0, 1], \qquad \Phi(a, b) = a(1 - b),$$

which decreases the membership when the contradiction degree $b$ is large.

Then the plithogenic soft expert mapping

$$F_{\text{PL}} : A \to [0, 1]^U$$

is given, for $\alpha = (e, x, o) \in A$ and $u \in U$, by

$$F_{\text{PL}}(\alpha)(u) = \Phi\big(pdf_e(u, \text{val}(\alpha)),\ pCF(\text{val}(\alpha), o)\big).$$

For instance, for $\alpha_1 = (e_1, x_1, 1)$ (cost, manager, agree) we obtain

$$F_{\mathrm{PL}}(\alpha_1)(u_1) = \Phi\big(0.9,\ 0.1\big) = 0.9 \times (1 - 0.1) = 0.81,$$

so supplier $u_1$ is strongly supported as "low cost" by the procurement manager. Similarly, for $\alpha_2 = (e_2, x_1, 1)$ and supplier $u_2$,

$$F_{\mathrm{PL}}(\alpha_2)(u_2) = \Phi\big(pdf_{e_2}(u_2, \text{neutral}),\ pCF(\text{neutral}, 1)\big) = 0.7 \times (1 - 0.3) = 0.49,$$

reflecting that a "neutral" sustainability assessment is partially penalized by the contradiction degree 0.3.

Thus this decision scenario forms a Plithogenic Soft Expert Set on the supplier universe $U$.

**Example 3.21.4** (Course recommendation as a PSES)**.** Let $U = \{c_1, c_2, c_3\}$ be a set of elective courses at a university. Let

$$E = \{e_1 = \text{"difficulty"},\ e_2 = \text{"job-market relevance"}\}, \quad X = \{x_1 = \text{advisor},\ x_2 = \text{industry mentor}\},$$

and again $O = \{1 = \text{agree},\ 0 = \text{disagree}\}$.

For the parameters take

$$Pv_{e_1} = \{\text{easy}, \text{moderate}, \text{hard}\}, \qquad Pv_{e_2} = \{\text{low}, \text{medium}, \text{high}\},$$

and consider the activated triples

$$A = \big\{(e_1, x_1, 1),\ (e_2, x_1, 1),\ (e_2, x_2, 1)\big\}.$$

Suppose

$$\mathrm{val}(e_1, x_1, 1) = \text{moderate}, \quad \mathrm{val}(e_2, x_1, 1) = \text{high}, \quad \mathrm{val}(e_2, x_2, 1) = \text{high},$$

so both experts emphasize high job-market relevance, while the advisor describes difficulty as moderate.

Let $s_{e_1} = s_{e_2} = 1$, and define fuzzy DAFs by

$$pdf_{e_1}(c_1, \text{moderate}) = 0.8, \quad pdf_{e_1}(c_2, \text{moderate}) = 0.5, \quad pdf_{e_1}(c_3, \text{moderate}) = 0.3,$$
$$pdf_{e_2}(c_1, \text{high}) = 0.6, \qquad pdf_{e_2}(c_2, \text{high}) = 0.9, \qquad pdf_{e_2}(c_3, \text{high}) = 0.4.$$

To keep the contradiction structure simple, let $t = 1$ and define

$$pCF(\text{moderate}, 1) = 0.2, \quad pCF(\text{high}, 1) = 0.1,$$

and $pCF(\ell, \ell) = 0$, $pCF(\ell_1, \ell_2) = 0$ for all other pairs. Thus "high job-market relevance" is almost fully compatible with a positive opinion, while "moderate difficulty" carries a slightly higher contradiction penalty.

Using the same aggregation $\Phi(a, b) = a(1 - b)$ as above, we obtain for $\alpha = (e_2, x_1, 1)$ (advisor, high relevance) and course $c_2$:

$$F_{\mathrm{PL}}(\alpha)(c_2) = \Phi\big(pdf_{e_2}(c_2, \text{high}),\ pCF(\text{high}, 1)\big) = 0.9 \times (1 - 0.1) = 0.81,$$

indicating strong plithogenic support for recommending $c_2$ due to high job-market relevance with low contradiction.

In contrast, for $\alpha' = (e_1, x_1, 1)$ (advisor, moderate difficulty) and course $c_3$:

$$F_{\mathrm{PL}}(\alpha')(c_3) = \Phi\big(pdf_{e_1}(c_3, \text{moderate}),\ pCF(\text{moderate}, 1)\big) = 0.3 \times (1 - 0.2) = 0.24,$$

showing weaker support for $c_3$ under the same expert opinion.

This university course recommendation scenario thus provides another concrete real-life instance of a Plithogenic Soft Expert Set on $U$.

## 3.22 Dynamic Plithogenic Set

A Dynamic Plithogenic Set (DPS) makes the plithogenic membership and contradiction depend on time, so each instant yields a (static) plithogenic set snapshot [786].

**Definition 3.22.1** (Dynamic Plithogenic Set)**.** Let $P \neq \varnothing$ be a universe, let $v$ be an attribute with value–set $P_v$, and let $T \subseteq \mathbb{R}$ be a nonempty time domain. Fix integers $s \geq 1$ and $t \geq 0$. A *Dynamic Plithogenic Set (of dimension $(s,t)$)* is a tuple

$$\mathrm{DPS} \;=\; (P,\; v,\; P_v,\; pdf,\; pCF,\; T),$$

where

$$pdf:\; T \times P \times P_v \longrightarrow [0,1]^s, \qquad pCF:\; T \times P_v \times P_v \longrightarrow [0,1]^t$$

are, respectively, the time–dependent Degree of Appurtenance Function (DAF) and Degree of Contradiction Function (DCF). For each fixed $t_0 \in T$, the *snapshot* at time $t_0$ is the (static) plithogenic set

$$\mathrm{PS}^{(t_0)} \;=\; \big(P,\; v,\; P_v,\; pdf^{(t_0)},\; pCF^{(t_0)}\big),$$

with $pdf^{(t_0)}(x,a) := pdf(t_0,x,a)$ and $pCF^{(t_0)}(a,b) := pCF(t_0,a,b)$. We require, for all $t \in T$ and all $a,b \in P_v$,

$$pCF(t,a,a) = \mathbf{0} \in [0,1]^t \quad \text{(reflexivity)}, \qquad pCF(t,a,b) = pCF(t,b,a) \quad \text{(symmetry)}.$$

We provide in Table 3.30 a concise summary of how dynamic fuzzy, dynamic intuitionistic fuzzy, and dynamic neutrosophic models arise as special cases of the Dynamic Plithogenic Set (DPS) framework.

Table 3.30: Dynamic fuzzy/intuitionistic/neutrosophic families as special cases of the Dynamic Plithogenic Set (DPS).

| **Dynamic model** | $s$ | $t$ | **How it is obtained from DPS** |
|---|---|---|---|
| Dynamic Fuzzy Set (DFS) [344, 787] | 1 | 0 | Take $pdf: T \times P \times P_v \to [0,1]$ and $pCF$ absent (or $pCF \equiv \mathbf{0}$). Each snapshot is a fuzzy set; membership varies with $t$. |
| Dynamic Intuitionistic Fuzzy Set (DIFS) [788, 789] | 2 | 0 | Take $pdf(t,x,a) = (\mu,\nu) \in [0,1]^2$ with $\mu + \nu \leq 1$; $pCF$ absent (or $\equiv \mathbf{0}$). |
| Dynamic Neutrosophic Set (DNS) (cf. [786, 790]) | 3 | 0 | Take $pdf(t,x,a) = (T,I,F) \in [0,1]^3$; $pCF$ absent (or $\equiv \mathbf{0}$). |
| Dynamic Hesitant Fuzzy Set (cf. [791, 792]) | 3 | 0 | Take $pdf(t,x,a) \in [0,1]^3$; $pCF$ absent (or $\equiv \mathbf{0}$). |

A concrete example of this concept is provided below.

**Example 3.22.2** (Dynamic plithogenic customer satisfaction in an online platform)**.** Customer satisfaction measures how well products and services meet or exceed customer expectations, positively influencing loyalty, repeat purchases, and reputation (cf. [793]).

Consider an online subscription platform that tracks how customers feel about its *Basic* and *Premium* plans before and after a major interface redesign.

Let

$$P = \{\text{Basic},\ \text{Premium}\}$$

be the universe of plans, and let the attribute be $v =$ "overall satisfaction" with value set

$$P_v = \{\text{low},\ \text{medium},\ \text{high}\}.$$

We take a time domain

$$T = \{t_1, t_2\} \subset \mathbb{R},$$

where $t_1$ is "before redesign" and $t_2$ is "after redesign".

Choose $s = 3$ and $t = 1$. For each $(t, x, a) \in T \times P \times P_v$ the time–dependent DAF

$$pdf : T \times P \times P_v \longrightarrow [0, 1]^3$$

returns a triple $pdf(t, x, a) = (\mu_T, \mu_I, \mu_F)$, interpreted as degrees of *approval* ($\mu_T$), *hesitation* ($\mu_I$), and *rejection* ($\mu_F$) of value $a$ for plan $x$ at time $t$.

For instance, for the *Premium* plan we may set

| | $t_1$ (before redesign) | $t_2$ (after redesign) |
|---|---|---|
| $pdf(t, \text{Premium}, \text{high})$ | $(0.45, 0.30, 0.25)$ | $(0.80, 0.10, 0.10)$ |
| $pdf(t, \text{Premium}, \text{medium})$ | $(0.35, 0.25, 0.40)$ | $(0.15, 0.15, 0.70)$ |
| $pdf(t, \text{Premium}, \text{low})$ | $(0.10, 0.20, 0.70)$ | $(0.05, 0.10, 0.85)$ |

and similarly for the *Basic* plan (with smaller improvements after $t_2$).

The contradiction degree function

$$pCF : T \times P_v \times P_v \longrightarrow [0, 1]$$

is taken time–independent for simplicity:

| $pCF(\cdot, \cdot, \cdot)$ | low | medium | high |
|---|---|---|---|
| low | 0 | 0.3 | 0.9 |
| medium | 0.3 | 0 | 0.5 |
| high | 0.9 | 0.5 | 0 |

and symmetric by definition. Then

$$\text{DPS}_{\text{cust}} = (P,\ v,\ P_v,\ pdf,\ pCF,\ T)$$

is a Dynamic Plithogenic Set. The *dynamic* aspect is captured by the change in the appurtenance triples from $t_1$ to $t_2$: after the redesign, the degree of "high satisfaction" for *Premium* shifts from $(0.45, 0.30, 0.25)$ to $(0.80, 0.10, 0.10)$, while the contradiction structure between satisfaction levels is kept fixed. This allows the analyst to compare customer sentiment snapshots over time within a plithogenic framework.

**Example 3.22.3** (Dynamic plithogenic air–quality risk assessment in a city)**.** Air-quality risk assessment evaluates pollutant exposure, health impacts, and uncertainty to prioritize mitigation policies, regulations, and urban planning decisions effectively (cf. [794]).

A city monitors daily air quality in two districts, $A$ and $B$, and classifies each district into qualitative *health risk levels.*

Let the universe be

$$P = \{\text{District } A,\ \text{District } B\},$$

and let the attribute be $v =$ "air–quality health risk" with value set

$$P_v = \{\text{low},\ \text{moderate},\ \text{high}\}.$$

Take a discrete time domain of three consecutive days,

$$T = \{d_1, d_2, d_3\} \subset \mathbb{R},$$

with $d_1$ = "Monday", $d_2$ = "Tuesday", $d_3$ = "Wednesday".

We again fix $s = 3$, $t = 1$, and interpret $pdf(d, x, a) = (\mu_T, \mu_I, \mu_F) \in [0, 1]^3$ as degrees of *acceptable risk*, *uncertainty*, and *unacceptable risk* associated with label $a$ for district $x$ on day $d$.

Suppose that for District $A$ we obtain the following DAF values:

| | $d_1$ (clear) | $d_2$ (haze) | $d_3$ (pollution episode) |
|---|---|---|---|
| $pdf(d,A,\text{low})$ | $(0.80, 0.10, 0.10)$ | $(0.40, 0.20, 0.40)$ | $(0.10, 0.10, 0.80)$ |
| $pdf(d,A,\text{moderate})$ | $(0.15, 0.20, 0.65)$ | $(0.40, 0.25, 0.35)$ | $(0.30, 0.20, 0.50)$ |
| $pdf(d,A,\text{high})$ | $(0.05, 0.10, 0.85)$ | $(0.20, 0.20, 0.60)$ | $(0.60, 0.15, 0.25)$ |

On clear day $d_1$ the city sees District $A$ mostly as "low risk", while during the pollution episode $d_3$ the weight moves toward "high risk". Similar (possibly different) evolutions can be specified for District $B$.

The time–dependent contradiction degree

$$pCF : T \times P_v \times P_v \longrightarrow [0, 1]$$

can encode changing public–health emphasis. For example, the authorities may become more sensitive to differences between "moderate" and "high" after the pollution episode. A simple specification is

| $pCF(d_1, \cdot, \cdot)$ | low | moderate | high |
|---|---|---|---|
| low | 0 | 0.2 | 0.8 |
| moderate | 0.2 | 0 | 0.4 |
| high | 0.8 | 0.4 | 0 |

| $pCF(d_3, \cdot, \cdot)$ | low | moderate | high |
|---|---|---|---|
| low | 0 | 0.3 | 0.9 |
| moderate | 0.3 | 0 | 0.6 |
| high | 0.9 | 0.6 | 0 |

with $pCF(d_2, \cdot, \cdot)$ chosen in between and symmetry enforced. Here, the contradiction between "moderate" and "high" risk increases from 0.4 on $d_1$ to 0.6 on $d_3$, reflecting stricter health–policy thresholds after observing a serious episode.

Then

$$\text{DPS}_{\text{air}} = (P,\ v,\ P_v,\ pdf,\ pCF,\ T)$$

is a Dynamic Plithogenic Set whose snapshots $\text{PS}^{(d_1)}$, $\text{PS}^{(d_2)}$, and $\text{PS}^{(d_3)}$ represent, respectively, the city's plithogenic risk assessments on each day, taking into account both time-varying memberships and time-varying contradiction between risk labels.

## 3.23 Probabilistic Plithogenic Set

A probabilistic plithogenic set attaches to each element a *random* (possibly multi–component) plithogenic degree, and fuses such degrees across attributes by a contradiction–aware aggregator. This subsumes probabilistic fuzzy, probabilistic intuitionistic fuzzy, and probabilistic neutrosophic sets.

**Definition 3.23.1** (Probabilistic Plithogenic Degree (PPD))**.** Let $(\Omega, \mathcal{A}, \mathbb{P})$ be a probability space. Fix a finite attribute system

$$\text{Att} = \{a_1, \dots, a_m\}, \qquad Pv_i \text{ (admissible values of } a_i\text{)}.$$

For an underlying universe $P$, an $s_i$–component *probabilistic plithogenic degree* at node $a_i$ is a measurable map

$$\mu_i :\ P \times Pv_i \times \Omega \longrightarrow [0, 1]^{s_i}, \qquad (x, \alpha, \omega) \longmapsto \mu_i(x, \alpha; \omega),$$

meaning that, for each fixed $(x, \alpha)$, the random vector $\mu_i(x, \alpha; \cdot)$ describes the stochastic membership (possibly multi–valued such as $(\mu, \nu)$ or $(T, I, F)$).

**Definition 3.23.2** (Probabilistic Plithogenic Set (PPS))**.** Let $pCF : \left(\bigcup_i Pv_i\right) \times \left(\bigcup_i Pv_i\right) \to [0,1]$ be a symmetric contradiction degree with $pCF(a,a) = 0$. Let $\mathsf{Agg}_{pCF}$ be a measurable *contradiction–aware aggregator* that, for any finite family of components $\{z_j\}_j$ in $[0,1]$ and their pairwise contradictions $\{c_{jk}\}$, returns a value in $[0,1]$; a typical choice is the contradiction–weighted $t$–norm/$t$–conorm blend

$$\mathsf{Agg}_{pCF}(u,v;c) \;:=\; (1-c)\,T(u,v) + c\,S(u,v), \quad c \in [0,1],$$

extended iteratively to $n \geq 2$ inputs, where $T$ is a $t$–norm and $S$ a $t$–conorm.

A *Probabilistic Plithogenic Set* is the tuple

$$\mathrm{PPS} = \Big(P,\; \{(a_i, Pv_i, \mu_i)\}_{i=1}^{m},\; pCF,\; \mathsf{Agg}_{pCF}\Big).$$

For any selection of attribute–values $\gamma = (\alpha_1, \ldots, \alpha_m) \in \prod_i Pv_i$ and $x \in P$, define the *aggregated random degree*

$$\overline{\mu}(x,\gamma;\omega) \;:=\; \underset{1\leq i\leq m}{\mathsf{Agg}_{\mathrm{pCF}}}\big(\mu_i(x,\alpha_i;\omega)\big) \;\in [0,1]^{\,s},$$

where the aggregation is taken componentwise when $s := \max_i s_i > 1$ and the pairwise contradictions used inside $\mathsf{Agg}_{pCF}$ are $pCF(\alpha_i, \alpha_j)$.

Representative crisp summaries can be extracted, e.g. $\mathbb{E}[\overline{\mu}(x,\gamma;\omega)]$, or quantile profiles $q_p(x,\gamma)$ defined componentwise by

$$q_p(x,\gamma) \;:=\; \inf\big\{t \in [0,1] \;:\; \mathbb{P}\big(\overline{\mu}(x,\gamma;\omega) \leq t\big) \geq p\big\}.$$

**Remark 3.23.3** (Set operations (defined $\omega$–wise))**.** For two PPSs on the same $(P, \{Pv_i\}, pCF, \mathsf{Agg}_{pCF})$ with degrees $\overline{\mu}_A, \overline{\mu}_B$:

$$\begin{aligned}
&\text{Union:} && \overline{\mu}_{A\cup B}(x,\gamma;\omega) := \mathsf{Agg}^{(\cup)}_{pCF}\big(\overline{\mu}_A(x,\gamma;\omega), \overline{\mu}_B(x,\gamma;\omega)\big),\\
&\text{Intersection:} && \overline{\mu}_{A\cap B}(x,\gamma;\omega) := \mathsf{Agg}^{(\cap)}_{pCF}\big(\overline{\mu}_A(x,\gamma;\omega), \overline{\mu}_B(x,\gamma;\omega)\big),\\
&\text{Complement:} && \overline{\mu}_{A^c}(x,\gamma;\omega) := \mathsf{Comp}\big(\overline{\mu}_A(x,\gamma;\omega)\big),
\end{aligned}$$

where $\mathsf{Agg}^{(\cup)}_{pCF}$ (resp. $\mathsf{Agg}^{(\cap)}_{pCF}$) is typically obtained by biasing toward $S$ (resp. $T$) via the local contradiction degree(s), and $\mathsf{Comp}$ is the plithogenic complement (for $s = 1$, $\mathsf{Comp}(u) = 1 - u$; for $(\mu, \nu)$ or $(T, I, F)$, apply the standard dualities componentwise). A crisp output can be taken as $\mathbb{E}[\cdot]$ of the above random outputs when needed.

**Remark 3.23.4** (Connection to probabilistic fuzzy sets)**.** In a probabilistic fuzzy set (PFS), the membership grade is a random variable on a probability space, often formalized as a measurable map $\mu_A : T \times \Omega \to [0,1]$ (measurable in $\omega$ for fixed $t$, and a fuzzy membership in $t$ for fixed $\omega$). This yields a "3D" view where the grade has a probability distribution. The PPS reduces to a PFS when $m = 1$, $s = 1$, and $pCF \equiv 0$ (so $\mathsf{Agg}_{pCF}$ is the identity). See the measurable definition and random–grade viewpoint of PFS in the literature.

Table 3.31 summarizes the reductions of the Probabilistic Plithogenic Set into its corresponding fuzzy, intuitionistic fuzzy, neutrosophic, and classical plithogenic special cases.

A concrete example of this concept is provided below.

**Example 3.23.5** (Probabilistic plithogenic supplier evaluation in sustainable procurement)**.** Consider a manufacturing company that needs to select a long–term raw–material supplier under uncertain future demand and regulatory conditions.

Table 3.31: Reductions of the Probabilistic Plithogenic Set (PPS). A PPS collapses to PFS/PIFS/PNS by fixing one attribute, taking $pCF \equiv 0$, and choosing $s = 1, 2, 3$ respectively.

| Target model | Components (codomain) | Constraints | Reduction from PPS |
|---|---|---|---|
| Probabilistic Fuzzy Set (PFS) [795–798] | scalar grade $u \in [0,1]$ (random) | none beyond $u \in [0,1]$ | $m = 1$, $s = 1$, $pCF \equiv 0$; $\mathsf{Agg}_{pCF}$ identity; $\overline{\mu}(x, \alpha; \cdot) \in [0,1]$ |
| Probabilistic Intuitionistic Fuzzy Set (PIFS) [180, 799, 800] | pair $(\mu, \nu) \in [0,1]^2$ (random) | $\mu + \nu \le 1$ a.s. | $m = 1$, $s = 2$, $pCF \equiv 0$; $\overline{\mu}(x, \alpha; \cdot) \in [0,1]^2$ with a.s. $\mu + \nu \le 1$ |
| Probabilistic Neutrosophic Set (PNS) [801–803] | triple $(T, I, F) \in [0,1]^3$ (random) | typically $0 \le T, I, F \le 1$ | $m = 1$, $s = 3$, $pCF \equiv 0$; $\overline{\mu}(x, \alpha; \cdot) \in [0,1]^3$ |
| Probabilistic Hesitant Fuzzy Set [804, 805, 805] | hesitant fuzzy triple (random) | – | $m = 1$, $s = 3$, $pCF \equiv 0$; $\overline{\mu}(x, \alpha; \cdot) \in [0,1]^3$ |
| Probabilistic Quadripartitioned Neutrosophic Set | Quadripartitioned Neutrosophic Quadruple (random) | – | $m = 1$, $s = 4$, $pCF \equiv 0$; $\overline{\mu}(x, \alpha; \cdot) \in [0,1]^4$ |
| Probabilistic Pentapartitioned Neutrosophic Set | Pentapartitioned Neutrosophic QuinTuple (random) | – | $m = 1$, $s = 5$, $pCF \equiv 0$; $\overline{\mu}(x, \alpha; \cdot) \in [0,1]^5$ |

**Universe and attributes.** Let the universe of alternatives be

$$P = \{S_1, S_2, S_3\},$$

where $S_1$ is the incumbent supplier and $S_2, S_3$ are new candidates. The company evaluates suppliers according to the finite attribute system

$$\mathrm{Att} = \{a_1, a_2, a_3\},$$

where

$$a_1 = \text{“unit cost”}, \quad a_2 = \text{“quality stability”}, \quad a_3 = \text{“environmental performance”}.$$

For each $a_i$ we use the same linguistic value set

$$Pv_i = \{\text{low, medium, high}\}$$

interpreted as low/medium/high *favorability* for the company.

**Uncertain scenarios and probabilistic degrees.** Future conditions are uncertain, so the company models three equally likely scenarios:

$$\Omega = \{\omega_1, \omega_2, \omega_3\}, \qquad \mathbb{P}(\{\omega_k\}) = \tfrac{1}{3} \quad (k = 1, 2, 3),$$

representing

$$\omega_1 = \text{“demand boom”}, \ \omega_2 = \text{“baseline”}, \ \omega_3 = \text{“strict carbon regulation”}.$$

We choose $s_i = 1$ for all $i$, so each probabilistic plithogenic degree $\mu_i(x, \alpha; \omega) \in [0,1]$ is an ordinary random fuzzy membership.

For illustration, consider attribute $a_3$ (environmental performance) and supplier $S_1$. The random membership $\mu_3(S_1, \alpha; \omega)$ for $\alpha \in Pv_3$ is specified as follows:

| $\mu_3(S_1, \alpha; \omega)$ | $\omega_1$ | $\omega_2$ | $\omega_3$ |
|---|---|---|---|
| $\alpha$ = low | 0.20 | 0.10 | 0.05 |
| $\alpha$ = medium | 0.50 | 0.40 | 0.30 |
| $\alpha$ = high | 0.30 | 0.50 | 0.65 |

Before strict regulation $(\omega_1, \omega_2)$, $S_1$ is mostly "medium" to "high" in environmental performance, but under strict regulation $\omega_3$, the company expects $S_1$ to invest more in green technology, increasing the membership of "high" to 0.65.

Similarly, we define $\mu_1$ and $\mu_2$ for $a_1$ (cost) and $a_2$ (quality) for each supplier, attribute–value, and scenario. For example, for $S_2$ on cost:

| $\mu_1(S_2, \alpha; \omega)$ | $\omega_1$ | $\omega_2$ | $\omega_3$ |
|---|---|---|---|
| $\alpha = \text{low}$ | 0.80 | 0.75 | 0.70 |
| $\alpha = \text{medium}$ | 0.15 | 0.20 | 0.25 |
| $\alpha = \text{high}$ | 0.05 | 0.05 | 0.05 |

indicating that $S_2$ is expected to be consistently low–cost, even with regulatory changes.

**Contradiction degrees and aggregation.** We define a simple contradiction degree on the union of all value sets by

$$pCF(\text{low}, \text{high}) = pCF(\text{high}, \text{low}) = 0.9, \quad pCF(\text{low}, \text{medium}) = pCF(\text{medium}, \text{low}) = 0.4,$$

$$pCF(\text{medium}, \text{high}) = pCF(\text{high}, \text{medium}) = 0.5, \quad pCF(\alpha, \alpha) = 0,$$

and extend $pCF$ trivially across attributes (values from different attributes are considered moderately contradictory, if needed).

As contradiction–aware aggregator, the company chooses

$$\mathsf{Agg}_{pCF}(u, v; c) = (1 - c) \min\{u, v\} + c \max\{u, v\},$$

which biases toward the $t$–norm min when $c$ is small (compatible values) and toward the $t$–conorm max when $c$ is large (strong contradiction).

For a fixed supplier $S_2$ and a particular value profile

$$\gamma = (\alpha_1, \alpha_2, \alpha_3) = (\text{low cost},\ \text{high quality},\ \text{high environment}),$$

the aggregated random degree

$$\overline{\mu}(S_2, \gamma; \omega) = \mathsf{Agg}_{pCF}\big(\mu_1(S_2, \text{low}; \omega),\ \mu_2(S_2, \text{high}; \omega),\ \mu_3(S_2, \text{high}; \omega)\big)$$

is a scenario–dependent random membership in $[0, 1]$. A crisp summary such as

$$\mathbb{E}\big[\overline{\mu}(S_2, \gamma; \omega)\big] = \sum_{k=1}^{3} \overline{\mu}(S_2, \gamma; \omega_k)\, \mathbb{P}(\{\omega_k\})$$

gives the expected plithogenic suitability of $S_2$ under profile $\gamma$, while quantiles capture risk–averse views. Altogether,

$$\mathrm{PPS}_{\text{supply}} = \Big(P,\ \{(a_i, Pv_i, \mu_i)\}_{i=1}^{3},\ pCF,\ \mathsf{Agg}_{pCF}\Big)$$

is a Probabilistic Plithogenic Set modeling sustainable supplier evaluation under uncertain future conditions.

**Example 3.23.6** (Probabilistic plithogenic disease–risk assessment in preventive medicine)**.** Disease–risk assessment is systematic evaluation of likelihood and impact of developing specific diseases using medical data, biomarkers, lifestyle, genetics, and environment over time (cf. [806]).

A public–health agency wants to assess an individual's risk of developing a chronic disease (e.g. type–2 diabetes) using several uncertain indicators.

**Universe and attributes.** Let $P$ be the population of patients enrolled in a screening program. For a fixed patient $x \in P$ we evaluate the following attribute system:

$$\mathrm{Att} = \{a_1, a_2, a_3\},$$

with

$$a_1 = \text{“body–mass index (BMI)”}, \quad a_2 = \text{“fasting blood glucose”}, \quad a_3 = \text{“physical activity level”}.$$

Each attribute is described by a qualitative value set:

$$Pv_1 = \{\text{normal}, \ \text{overweight}, \ \text{obese}\},$$

$$Pv_2 = \{\text{normal}, \ \text{impaired}, \ \text{high}\},$$

$$Pv_3 = \{\text{low}, \ \text{moderate}, \ \text{high}\}.$$

**Measurement uncertainty as probability space.** Measurements are noisy and may vary across visits, so the agency considers three equiprobable “data states”:

$$\Omega = \{\omega_1, \omega_2, \omega_3\}, \qquad \mathbb{P}(\{\omega_k\}) = \tfrac{1}{3},$$

where, for a given patient $x$,

$$\omega_1 = \text{“optimistic lab readings”}, \quad \omega_2 = \text{“typical readings”}, \quad \omega_3 = \text{“pessimistic readings”}.$$

We choose $s_1 = s_2 = s_3 = 3$ and regard

$$\mu_i(x, \alpha; \omega) = \big(T_i(x, \alpha; \omega), \ I_i(x, \alpha; \omega), \ F_i(x, \alpha; \omega)\big) \in [0,1]^3$$

as a neutrosophic–type triple of degrees of *support* ($T$), *indeterminacy* ($I$), and *counter–evidence* ($F$) for value $\alpha$ of attribute $a_i$ under data state $\omega$.

For example, for patient $x$ with moderately high BMI, the random degrees for $a_1$ may be

| $\mu_1(x, \alpha; \omega)$ | $\omega_1$ | $\omega_2$ | $\omega_3$ |
|---|---|---|---|
| $\alpha =$ normal | $(0.30, 0.20, 0.50)$ | $(0.20, 0.20, 0.60)$ | $(0.10, 0.15, 0.75)$ |
| $\alpha =$ overweight | $(0.50, 0.20, 0.30)$ | $(0.55, 0.20, 0.25)$ | $(0.45, 0.20, 0.35)$ |
| $\alpha =$ obese | $(0.20, 0.20, 0.60)$ | $(0.25, 0.25, 0.50)$ | $(0.45, 0.20, 0.35)$ |

indicating that, under pessimistic readings $\omega_3$, the evidence for $\alpha =$ obese increases.

Similarly, for fasting glucose $a_2$ we might have, for the same patient $x$,

| $\mu_2(x, \alpha; \omega)$ | $\omega_1$ | $\omega_2$ | $\omega_3$ |
|---|---|---|---|
| $\alpha =$ normal | $(0.70, 0.10, 0.20)$ | $(0.55, 0.15, 0.30)$ | $(0.40, 0.15, 0.45)$ |
| $\alpha =$ impaired | $(0.20, 0.15, 0.65)$ | $(0.30, 0.20, 0.50)$ | $(0.35, 0.20, 0.45)$ |
| $\alpha =$ high | $(0.10, 0.20, 0.70)$ | $(0.15, 0.20, 0.65)$ | $(0.25, 0.20, 0.55)$ |

and analogous tables for $a_3$ describing physical activity.

**Contradiction structure and risk aggregation.** Some attribute values jointly *increase* disease risk (e.g. “obese” and “high glucose”), while others are protective or partially compensating (e.g.

“obese” but “high activity”). This is captured by a plithogenic contradiction degree $pCF$ on the union of all $Pv_i$. For instance,

$$pCF(\text{normal BMI}, \text{normal glucose}) = 0.1, \qquad pCF(\text{obese}, \text{high glucose}) = 0.95,$$

$$pCF(\text{obese}, \text{high activity}) = 0.6,$$

and $pCF(\alpha, \alpha) = 0$ with symmetry.

The contradiction–aware aggregator

$$\mathsf{Agg}_{pCF} : ([0,1]^3)^n \times [0,1]^{n\times n} \longrightarrow [0,1]^3$$

combines the neutrosophic triples across attributes. For a chosen risk–oriented profile

$$\gamma = (\alpha_1, \alpha_2, \alpha_3) = (\text{obese},\ \text{high glucose},\ \text{low activity}),$$

the aggregated random triple

$$\overline{\mu}(x, \gamma; \omega) \in [0,1]^3$$

represents the probabilistic plithogenic degree of being a “high–risk patient” for $x$ under data state $\omega$.

Taking the expectation

$$\mathbb{E}\big[\overline{\mu}(x, \gamma; \omega)\big]$$

yields an average neutrosophic risk triple, while quantiles of $\overline{\mu}(x, \gamma; \omega)$ provide conservative risk bounds. Thus

$$\mathrm{PPS}_{\mathrm{med}} = \Big(P,\ \{(a_i, Pv_i, \mu_i)\}_{i=1}^{3},\ pCF,\ \mathsf{Agg}_{pCF}\Big)$$

is a Probabilistic Plithogenic Set that models disease–risk assessment under measurement noise and uncertain future health states.

## 3.24 Triangular Plithogenic Set

A triangular plithogenic set models each element’s appurtenance by one or more triangular fuzzy numbers on $[0, 1]$, while modulating these degrees through an attribute–value contradiction function.

**Definition 3.24.1** (Triangular Fuzzy Number (TFN))**.** (cf. [807–809]) A TFN on $[0, 1]$ is a triple $\tau = (\ell, m, u)$ with $0 \le \ell \le m \le u \le 1$. Its membership function $\mu_\tau : [0, 1] \to [0, 1]$ is

$$\mu_\tau(z) = \begin{cases} 0, & z \le \ell, \\ \dfrac{z-\ell}{m-\ell}, & \ell < z \le m, \\ \dfrac{u-z}{u-m}, & m < z < u, \\ 0, & z \ge u. \end{cases}$$

**Definition 3.24.2** (Triangular Plithogenic Set (TPS))**.** Let $U$ be a universe and $v$ an attribute with value set $P_v$. Fix integers $p, q, r \ge 0$ (a *triangular refinement signature*) and set $s := p+q+r \in \{1, 2, 3\}$. Define the *triangular plithogenic appurtenance* map

$$\mathrm{tPDF} :\ U \times P_v \longrightarrow \big([0,1]^3\big)^{s}, \qquad (x, a) \ \longmapsto\ \Big(T_i(x,a)\Big)_{i=1}^{p} \,\Big|\, \Big(I_j(x,a)\Big)_{j=1}^{q} \,\Big|\, \Big(F_k(x,a)\Big)_{k=1}^{r},$$

where each component $T_i(x, a)$, $I_j(x, a)$, $F_k(x, a)$ is a TFN $T_i(x, a) = (\ell^{T_i}, m^{T_i}, u^{T_i})$, etc. Let the *plithogenic contradiction* be

$$pCF :\ P_v \times P_v \to [0,1]^t, \qquad pCF(a, b) = pCF(b, a), \quad pCF(a, a) = 0.$$

A *Triangular Plithogenic Set* is the tuple

$$\mathsf{TPS} = \big(U,\ v,\ P_v,\ (p,q,r),\ \text{tPDF},\ pCF,\ \Phi\big),$$

together with a fixed monotone aggregator

$$\Phi:\ \big([0,1]^3\big)^s \times [0,1]^t \longrightarrow [0,1]^3$$

that returns a TFN and satisfies $\Phi(\cdot,\mathbf{0})$ = componentwise TFN aggregation without contradiction. For any $x \in U$ and *dominant* value $a^* \in P_v$, the effective triangular degree of $x$ is the TFN

$$\tau_{\mathsf{TPS}}(x \mid a^*) \ :=\ \Phi\Big(\text{tPDF}(x,a^*),\ pCF\big(a^*,\cdot\big)\Big) \in [0,1]^3.$$

Decision scores (if needed) can be scalarized by a TFN defuzzifier, e.g. the centroid $\text{cen}(\ell,m,u) = \frac{\ell+m+u}{3}$, applied to $\tau_{\mathsf{TPS}}(x \mid a^*)$.

Table 3.32 explains that the classical triangular fuzzy, intuitionistic, and neutrosophic sets arise as exact special cases of the Triangular Plithogenic Set.

Table 3.32: Classical triangular fuzzy/intuitionistic/neutrosophic sets as exact special cases of the Triangular Plithogenic Set by signature choice and $pCF \equiv 0$.

| Target triangular model | Signature $(p,q,r)$ | Recovery inside TPS (set $pCF \equiv 0$) |
|---|---|---|
| Triangular Fuzzy Set (TFS) [810–812] | $(1,0,0)$ | One TFN $T_1(x,a)$ per $x$ and $a$; the effective TFN is the aggregated $T$ (no contradiction). |
| Triangular Intuitionistic Fuzzy Set (TIFS) [813–815] | $(1,1,0)$ | A pair of TFNs $(T_1,I_1)$ per $(x,a)$ with a consistency condition (e.g. peak values $m^{T_1} + m^{I_1} \le 1$); operations act componentwise on $(T_1,I_1)$. |
| Triangular Neutrosophic Set (TNS) [816–818] | $(1,1,1)$ | A triple of TFNs $(T_1,I_1,F_1)$ per $(x,a)$; no coupling required beyond range constraints (optional bounds may be imposed per application). |
| Triangular Hesitant Fuzzy Sets [819–821] | $(1,1,1)$ | A triple per $(x,a)$; no coupling required beyond range constraints (optional bounds may be imposed per application). |

A brief concrete example of this concept is provided below.

**Example 3.24.3** (Triangular plithogenic assessment of laptop options in online shopping)**.** A customer wants to choose a new laptop from three alternatives

$$U = \{L_1, L_2, L_3\}.$$

They evaluate each laptop with respect to the attribute

$$v = \text{“overall perceived utility”}$$

and a set of linguistic attribute values

$$P_v = \{\text{low},\ \text{medium},\ \text{high}\}.$$

We consider a purely triangular fuzzy–type refinement with

$$(p,q,r) = (1,0,0), \qquad s = p+q+r = 1,$$

so that for every $(x, a) \in U \times P_v$ there is a *single* triangular fuzzy number $T_1(x, a)$ describing the truth–membership of "laptop $x$ has utility level $a$". Thus

$$\text{tPDF}(x, a) = \big(T_1(x, a)\big), \qquad T_1(x, a) = (\ell^{T_1}, m^{T_1}, u^{T_1}) \in [0, 1]^3.$$

Suppose the customer's approximate judgements (already normalized to $[0, 1]$) are as follows.

**Triangular degrees for $L_1$.** Laptop $L_1$ is a budget option: mostly between "medium" and "high", but with some uncertainty. The triangular truth–memberships are

$$T_1(L_1, \text{low}) = (0.0, 0.1, 0.3), \qquad T_1(L_1, \text{medium}) = (0.2, 0.5, 0.8), \qquad T_1(L_1, \text{high}) = (0.4, 0.7, 1.0).$$

**Triangular degrees for $L_2$.** Laptop $L_2$ is a mid–range balanced model:

$$T_1(L_2, \text{low}) = (0.0, 0.0, 0.2), \qquad T_1(L_2, \text{medium}) = (0.3, 0.6, 0.9), \qquad T_1(L_2, \text{high}) = (0.5, 0.8, 1.0).$$

**Triangular degrees for $L_3$.** Laptop $L_3$ is a premium model; the user believes it is rarely "low" utility but sometimes only "medium", depending on personal taste:

$$T_1(L_3, \text{low}) = (0.0, 0.0, 0.1), \qquad T_1(L_3, \text{medium}) = (0.2, 0.4, 0.7), \qquad T_1(L_3, \text{high}) = (0.6, 0.9, 1.0).$$

**Plithogenic contradiction on utility levels.** We impose a contradiction degree on $P_v$:

$$pCF(\text{low}, \text{high}) = pCF(\text{high}, \text{low}) = 0.9, \qquad pCF(\text{low}, \text{medium}) = pCF(\text{medium}, \text{low}) = 0.4,$$

$$pCF(\text{medium}, \text{high}) = pCF(\text{high}, \text{medium}) = 0.5, \qquad pCF(a, a) = 0 \quad \text{for } a \in P_v.$$

This reflects that "low" and "high" perceived utility are strongly contradictory, while "medium" is moderately contradictory to them.

**Contradiction–aware aggregation toward a dominant value.** Assume the customer desires *high* utility as the dominant value $a^* = \text{high}$. Then the triangular plithogenic degree

$$\tau_{\mathsf{TPS}}(x \mid a^*) = \Phi\Big(\text{tPDF}(x, \text{high}),\ pCF(\text{high}, \cdot)\Big)$$

is obtained via a chosen TFN aggregator $\Phi$ which biases the triangle according to how strongly "high" contradicts the other utility levels. For example, if $\Phi$ puts more weight on optimistic parts of the triangle whenever $pCF(\text{high}, \text{low})$ is large, then laptops which rarely have "low" utility (such as $L_3$) will obtain a more favorable $\tau_{\mathsf{TPS}}(L_3 \mid \text{high})$ than those with sizable "low" components.

Defuzzifying via the centroid

$$\text{cen}(\ell, m, u) = \frac{\ell + m + u}{3}$$

yields a scalar plithogenic score for each $L_i$, thus providing a concrete Triangular Plithogenic Set

$$\mathsf{TPS}_{\text{laptop}} = \big(U,\ v,\ P_v,\ (1, 0, 0),\ \text{tPDF},\ pCF,\ \Phi\big)$$

for real–life online laptop selection under contradictory impressions of utility levels.

**Example 3.24.4** (Triangular plithogenic evaluation of renewable–energy projects)**.** Renewable energy refers to power from naturally replenished sources like sun, wind, water, and biomass, reducing emissions and dependence globally (cf. [822]).

A city council considers three renewable–energy projects

$$U = \{P_1, P_2, P_3\},$$

where $P_1$ is a solar farm, $P_2$ an onshore wind park, and $P_3$ a biomass plant. The main attribute is

$$v = \text{“overall sustainability impact”},$$

with attribute values

$$P_v = \{\text{poor}, \ \text{acceptable}, \ \text{good}, \ \text{excellent}\}.$$

Because sustainability must balance environmental, social, and economic dimensions, the council wants to model not only truth but also indeterminacy and falsity degrees, each in triangular fuzzy form. We therefore choose

$$(p, q, r) = (1, 1, 1), \qquad s = p + q + r = 3,$$

so that

$$\text{tPDF}(x, a) = \Big(T_1(x, a) \mid I_1(x, a) \mid F_1(x, a)\Big),$$

where each of

$$T_1(x, a), \ I_1(x, a), \ F_1(x, a)$$

is a triangular fuzzy number in $[0, 1]^3$.

**Sustainability profile for the solar farm $P_1$.** For $P_1$, suppose the council's expert panel provides the following triangular neutrosophic–type assessments for the value “good”:

$$T_1(P_1, \text{good}) = (0.5, 0.7, 0.9),$$

indicating that the truth of “$P_1$ is good” is most plausible around 0.7, with minimum support 0.5 and maximum 0.9. Due to uncertainty in long–term lifecycle data, the indeterminacy is

$$I_1(P_1, \text{good}) = (0.1, 0.2, 0.4),$$

while falsity (counter–evidence, e.g. land–use concerns) is

$$F_1(P_1, \text{good}) = (0.0, 0.1, 0.3).$$

For a more ambitious value “excellent”, they specify

$$T_1(P_1, \text{excellent}) = (0.2, 0.4, 0.7), \quad I_1(P_1, \text{excellent}) = (0.2, 0.3, 0.5), \quad F_1(P_1, \text{excellent}) = (0.1, 0.2, 0.5).$$

**Profiles for wind park $P_2$ and biomass plant $P_3$.** For $P_2$ (wind park), visual impact and noise cause more conflict with stakeholders, but carbon savings are high. For the value “good” one may have

$$T_1(P_2, \text{good}) = (0.4, 0.6, 0.8), \quad I_1(P_2, \text{good}) = (0.2, 0.3, 0.4), \quad F_1(P_2, \text{good}) = (0.1, 0.2, 0.4).$$

For $P_3$ (biomass), fuel supply chain and emissions create further uncertainty:

$$T_1(P_3,\text{good}) = (0.3, 0.5, 0.7), \quad I_1(P_3,\text{good}) = (0.3, 0.4, 0.6), \quad F_1(P_3,\text{good}) = (0.1, 0.3, 0.5).$$

**Plithogenic contradiction between sustainability levels.** We define a contradiction degree on $P_v$ by viewing it as an ordered scale:

$$\text{poor} \prec \text{acceptable} \prec \text{good} \prec \text{excellent}.$$

Set

$$pCF(a,b) := \frac{|\iota(a) - \iota(b)|}{3}, \qquad a,b \in P_v,$$

where $\iota(\text{poor}) = 0$, $\iota(\text{acceptable}) = 1$, $\iota(\text{good}) = 2$, $\iota(\text{excellent}) = 3$. Thus

$$pCF(\text{poor},\text{excellent}) = 1, \quad pCF(\text{poor},\text{good}) = \tfrac{2}{3}, \quad pCF(\text{acceptable},\text{good}) = \tfrac{1}{3},$$

and $pCF(a,a) = 0$.

**Contradiction–aware triangular sustainability degree.** Assume the council aims at *excellent* sustainability as dominant value $a^* = \text{excellent}$. For each project $x \in \{P_1, P_2, P_3\}$ the Triangular Plithogenic Set returns a neutrosophic–type TFN

$$\tau_{\mathsf{TPS}}(x \mid a^*) = \Phi\Big(\text{tPDF}(x,\text{excellent}),\ pCF(\text{excellent},\cdot)\Big) \in [0,1]^3,$$

where $\Phi$ aggregates the $(T, I, F)$ triangles while taking into account how much "excellent" contradicts lower levels such as "poor" or "acceptable".

Defuzzifying $\tau_{\mathsf{TPS}}(x \mid a^*)$ by the centroid componentwise,

$$\text{cen}(T) = \frac{\ell^T + m^T + u^T}{3}, \quad \text{cen}(I) = \frac{\ell^I + m^I + u^I}{3}, \quad \text{cen}(F) = \frac{\ell^F + m^F + u^F}{3},$$

gives a scalar summary of truth, indeterminacy, and falsity for the statement "project $x$ is *excellent* in sustainability". In this way

$$\mathsf{TPS}_{\text{sust}} = \big(U,\ v,\ P_v,\ (1,1,1),\ \text{tPDF},\ pCF,\ \Phi\big)$$

is a concrete Triangular Plithogenic Set for real–life evaluation of renewable–energy projects with contradictory sustainability judgements.

## 3.25 Trapezoidal Plithogenic Sets

A Trapezoidal Plithogenic Set models elements using trapezoidal truth, indeterminacy, falsity memberships, weighted by attribute-wise contradiction degrees for uncertain information.

**Definition 3.25.1** (Trapezoidal fuzzy number)**.** (cf. [823–825]) Let $(a, b, c, d)$ with $a \le b \le c \le d$. The trapezoidal membership profile $\text{Trap}\,abcd : [0, 1] \to [0, 1]$ is

$$\text{Trap}\,abcd = \begin{cases} 0, & x \le a, \\ \dfrac{x-a}{b-a}, & a < x < b, \\ 1, & b \le x \le c, \\ \dfrac{d-x}{d-c}, & c < x < d, \\ 0, & x \ge d. \end{cases}$$

We abbreviate a trapezoidal fuzzy number by $(a, b, c, d)$.

**Definition 3.25.2** (Trapezoidal Plithogenic Set (TPS))**.** Let $U$ be a universe, $v$ an attribute with value-set $P_v$, and $pCF : P_v \times P_v \to [0,1]$ a plithogenic contradiction function with $pCF(a,a) = 0$ and $pCF(a,b) = pCF(b,a)$. For each $a \in P_v$, a TPS on $U$ assigns to every $x \in U$ three trapezoidal fuzzy numbers on $[0,1]$:

$$T(x,a) = (t_1,t_2,t_3,t_4), \qquad I(x,a) = (i_1,i_2,i_3,i_4), \qquad F(x,a) = (f_1,f_2,f_3,f_4),$$

interpreted as truth-, indeterminacy-, and falsity-membership profiles, respectively. The data

$$\mathsf{TPS} = \big(U,\ v,\ P_v;\ T,I,F;\ pCF\big)$$

is called a Trapezoidal Plithogenic Set.

A scalar *plithogenic inclusion grade* at a chosen dominant value $a^* \in P_v$ is obtained by first defuzzifying each trapezoid via the arithmetic mean $S(a,b,c,d) := \frac{a+b+c+d}{4}$ and then contradiction-weighting:

$$\lambda(a^*) := \frac{1}{|P_v|}\sum_{b\in P_v} pCF(a^*,b), \qquad \mu_{\mathrm{pl}}(x \mid a^*) := \big(1-\lambda(a^*)\big)\, S\big(T(x,a^*)\big) + \lambda(a^*)\big(1-S(F(x,a^*))\big) - \beta\, S\big(I(x,a^*)\big),$$

with a design parameter $\beta \in [0,1]$. One finally clips $\mu_{\mathrm{pl}}$ to $[0,1]$ if needed. The set-operations (union/intersection/complement) are defined at the $(T,I,F)$-level by a chosen t-conorm/t-norm/negator and transported through the defuzzification.

Table 3.33 explains how the classical trapezoidal fuzzy, intuitionistic fuzzy, and neutrosophic families arise as specializations of the Trapezoidal Plithogenic Set.

Table 3.33: Classical trapezoidal fuzzy/intuitionistic/neutrosophic families as specializations of the Trapezoidal Plithogenic Set.

| Target model (recovered) | Constraints inside TPS | Recovery of the classical grade |
|---|---|---|
| Trapezoidal Fuzzy Set (TFS) [826–828] | Set $I(x,a) \equiv (0,0,0,0)$, $F(x,a) \equiv (0,0,0,0)$ and $pCF \equiv 0$. | $\mu(x) = S\big(T(x,a^*)\big)$ gives the usual trapezoidal membership. |
| Trapezoidal Intuitionistic Fuzzy Set (TIFS) [829–831] | Use $(T,F)$ only with the intuitionistic constraint $t_4(x,a)+f_4(x,a) \leq 1$, and $pCF \equiv 0$. | $\mu(x) = S(T(x,a^*))$, $\nu(x) = S(F(x,a^*))$, with $\mu+\nu \leq 1$. |
| Trapezoidal Vague Set | Use two value only with the vague constraint $t_4(x,a) + f_4(x,a) \leq 1$, and $pCF \equiv 0$. | Please refer to the references. |
| Trapezoidal Neutrosophic Set (TrNS) [832–834] | Keep $(T,I,F)$ with $t_4(x,a) + i_4(x,a)+f_4(x,a) \leq 3$ and $pCF \equiv 0$. | $T(x), I(x), F(x)$ are trapezoids; classical TrNS operations apply componentwise. |
| Trapezoidal Hesitant Fuzzy Sets [835, 836] | $pCF \equiv 0$ and Hesitant Fuzzy Constraints | Triple Values are trapezoids; classical operations apply componentwise. |
| Trapezoidal Picture Fuzzy Sets [837, 838] | $pCF \equiv 0$. Picture Fuzzy Constraints | Triple Values are trapezoids; classical operations apply componentwise. |
| Trapezoidal Spherical Fuzzy Sets [839] | $pCF \equiv 0$. Spherical Fuzzy Constraints | Triple Values are trapezoids; classical operations apply componentwise. |
| Trapezoidal Quadripartitioned Neutrosophic Sets | $pCF \equiv 0$. Quadripartitioned Neutrosophic Constraints | Quadruple Values are trapezoids; classical operations apply componentwise. |
| Trapezoidal Pentapartitioned Neutrosophic Sets | $pCF \equiv 0$. Pentapartitioned Neutrosophic Constraints | Quinruple Values are trapezoids; classical operations apply componentwise. |

A concrete example of this concept is provided below.

**Example 3.25.3** (Supplier sustainability evaluation as a Trapezoidal Plithogenic Set)**.** Consider a procurement department that must evaluate two candidate suppliers

$$U = \{S_A, S_B\}$$

with respect to an attribute

$$v = \text{“sustainability level”}$$

whose plithogenic value-set is

$$P_v = \{\text{low}, \text{medium}, \text{high}\}.$$

The contradiction degrees between attribute values are chosen as

| $pCF(\cdot,\cdot)$ | low | medium | high |
|---|---|---|---|
| low | 0 | 0.2 | 0.7 |
| medium | 0.2 | 0 | 0.3 |
| high | 0.7 | 0.3 | 0 |

so that $pCF(a, a) = 0$ and $pCF(a, b) = pCF(b, a)$ for all $a, b \in P_v$. Suppose that the decision maker takes the *dominant* value $a^* = \text{high}$.

For each supplier $x \in U$ and the dominant value $a^*$, the Trapezoidal Plithogenic Set specifies trapezoidal truth-, indeterminacy-, and falsity-membership profiles $(T, I, F)$ on $[0, 1]$ as follows:

| | $T(x, \text{high})$ | $I(x, \text{high})$ | $F(x, \text{high})$ |
|---|---|---|---|
| $S_A$ | $(0.6, 0.7, 0.9, 1.0)$ | $(0.1, 0.2, 0.2, 0.3)$ | $(0.0, 0.0, 0.1, 0.2)$ |
| $S_B$ | $(0.3, 0.4, 0.5, 0.6)$ | $(0.2, 0.3, 0.3, 0.4)$ | $(0.3, 0.4, 0.5, 0.6)$ |

Interpreted linguistically, $S_A$ has a trapezoidal profile that is mostly in the “highly sustainable” region with small indeterminacy and low falsity, while $S_B$ is only moderately sustainable with larger falsity and indeterminacy.

Using the arithmetic mean

$$S(a, b, c, d) := \frac{a + b + c + d}{4},$$

we obtain for $S_A$ at $a^* = \text{high}$:

$$S\big(T(S_A, \text{high})\big) = \frac{0.6 + 0.7 + 0.9 + 1.0}{4} = \frac{3.2}{4} = \frac{4}{5},$$

$$S\big(I(S_A, \text{high})\big) = \frac{0.1 + 0.2 + 0.2 + 0.3}{4} = \frac{0.8}{4} = \frac{1}{5},$$

$$S\big(F(S_A, \text{high})\big) = \frac{0 + 0 + 0.1 + 0.2}{4} = \frac{0.3}{4} = \frac{3}{40}.$$

The plithogenic global contradiction factor at the dominant value $a^* = \text{high}$ is

$$\lambda(\text{high}) = \frac{1}{|P_v|} \sum_{b \in P_v} pCF(\text{high}, b) = \frac{1}{3}\big(0 + 0.3 + 0.7\big) = \frac{11}{30}.$$

For a design parameter $\beta \in [0, 1]$, the plithogenic inclusion grade is

$$\mu_{\text{pl}}(x \mid a^*) := \big(1 - \lambda(a^*)\big)\, S\big(T(x, a^*)\big) + \lambda(a^*)\big(1 - S(F(x, a^*))\big) - \beta\, S\big(I(x, a^*)\big),$$

so, taking for instance $\beta = \frac{1}{2}$, we get

$$\mu_{\text{pl}}(S_A \mid \text{high}) = \Big(1 - \frac{11}{30}\Big) \cdot \frac{4}{5} + \frac{11}{30}\Big(1 - \frac{3}{40}\Big) - \frac{1}{2} \cdot \frac{1}{5}.$$

We now simplify each term:

$$1 - \frac{11}{30} = \frac{19}{30}, \qquad 1 - \frac{3}{40} = \frac{37}{40},$$

$$\left(1 - \frac{11}{30}\right) \cdot \frac{4}{5} = \frac{19}{30} \cdot \frac{4}{5} = \frac{76}{150} = \frac{38}{75},$$
$$\frac{11}{30}\left(1 - \frac{3}{40}\right) = \frac{11}{30} \cdot \frac{37}{40} = \frac{407}{1200},$$
$$\frac{1}{2} \cdot \frac{1}{5} = \frac{1}{10}.$$

Hence

$$\mu_{\mathrm{pl}}(S_A \mid \mathrm{high}) = \frac{38}{75} + \frac{407}{1200} - \frac{1}{10} = \frac{608}{1200} + \frac{407}{1200} - \frac{120}{1200} = \frac{895}{1200} = \frac{179}{240} \approx 0.746.$$

Thus, under the trapezoidal plithogenic model, supplier $S_A$ attains a high inclusion grade ($\approx 0.75$) in the plithogenic sustainability set, quantitatively reflecting high truth, low falsity, limited indeterminacy, and the contradiction of the chosen attribute value "high" with other possible values. The data

$$\mathsf{TPS}_1 = \big(U,\ v,\ P_v;\ T, I, F;\ pCF\big)$$

constitute a concrete Trapezoidal Plithogenic Set used for sustainable supplier evaluation.

**Example 3.25.4** (Chronic disease risk assessment as a Trapezoidal Plithogenic Set)**.** Chronic disease risk assessment estimates an individual's long-term likelihood of developing conditions using medical history, lifestyle, biomarkers, and demographics data [840].

Consider a hospital that stratifies patients with a chronic condition into risk categories based on multiple clinical indicators. Let

$$U = \{P_1, P_2\}$$

represent two patients, and let

$$v = \text{"cardiovascular risk level"}$$

with plithogenic value-set

$$P_v = \{\mathrm{low}, \mathrm{moderate}, \mathrm{high}\}.$$

The contradiction degrees capture how far each category is from the desired "low risk" state:

| $pCF(\cdot,\cdot)$ | low | moderate | high |
|---|---|---|---|
| low | 0 | 0.5 | 0.9 |
| moderate | 0.5 | 0 | 0.4 |
| high | 0.9 | 0.4 | 0 |

and the dominant value is chosen as $a^* = \mathrm{low}$, reflecting the clinical goal of keeping patients at low risk.

For each patient $x \in U$ and $a^* = \mathrm{low}$, the Trapezoidal Plithogenic Set prescribes trapezoidal truth-, indeterminacy-, and falsity-membership profiles for the statement "$x$ is of low cardiovascular risk":

| | $T(x, \mathrm{low})$ | $I(x, \mathrm{low})$ | $F(x, \mathrm{low})$ |
|---|---|---|---|
| $P_1$ | $(0.5, 0.6, 0.8, 0.9)$ | $(0.1, 0.2, 0.2, 0.3)$ | $(0.1, 0.2, 0.3, 0.4)$ |
| $P_2$ | $(0.1, 0.2, 0.3, 0.4)$ | $(0.2, 0.3, 0.4, 0.5)$ | $(0.5, 0.6, 0.7, 0.8)$ |

Patient $P_1$ has relatively high truth-membership in "low risk" and moderate falsity-membership, whereas $P_2$ has low truth-membership and high falsity-membership, reflecting a clinically higher risk.

Using the same mean operator $S(a, b, c, d) = \frac{a+b+c+d}{4}$, we compute for $P_1$:

$$S\big(T(P_1, \mathrm{low})\big) = \frac{0.5 + 0.6 + 0.8 + 0.9}{4} = \frac{2.8}{4} = \frac{7}{10},$$

$$S\big(I(P_1, \mathrm{low})\big) = \frac{0.1 + 0.2 + 0.2 + 0.3}{4} = \frac{0.8}{4} = \frac{1}{5},$$

$$S\big(F(P_1,\text{low})\big) = \frac{0.1+0.2+0.3+0.4}{4} = \frac{1.0}{4} = \frac{1}{4}.$$

The contradiction factor at $a^* = \text{low}$ is

$$\lambda(\text{low}) = \frac{1}{|P_v|}\sum_{b\in P_v} pCF(\text{low}, b) = \frac{1}{3}\big(0+0.5+0.9\big) = \frac{1.4}{3} = \frac{14}{30} = \frac{7}{15}.$$

Fixing again $\beta = \frac{1}{2}$, the plithogenic inclusion grade of $P_1$ in the low-risk plithogenic set is

$$\mu_{\text{pl}}(P_1 \mid \text{low}) = \Big(1-\frac{7}{15}\Big)\cdot\frac{7}{10} + \frac{7}{15}\Big(1-\frac{1}{4}\Big) - \frac{1}{2}\cdot\frac{1}{5}.$$

We simplify term by term:

$$1-\frac{7}{15} = \frac{8}{15}, \qquad 1-\frac{1}{4} = \frac{3}{4},$$

$$\Big(1-\frac{7}{15}\Big)\cdot\frac{7}{10} = \frac{8}{15}\cdot\frac{7}{10} = \frac{56}{150} = \frac{28}{75},$$

$$\frac{7}{15}\Big(1-\frac{1}{4}\Big) = \frac{7}{15}\cdot\frac{3}{4} = \frac{21}{60} = \frac{7}{20},$$

$$\frac{1}{2}\cdot\frac{1}{5} = \frac{1}{10}.$$

Therefore

$$\mu_{\text{pl}}(P_1 \mid \text{low}) = \frac{28}{75} + \frac{7}{20} - \frac{1}{10} = \frac{112}{300} + \frac{105}{300} - \frac{30}{300} = \frac{187}{300} \approx 0.623.$$

Thus patient $P_1$ belongs to the plithogenic low-risk set with a moderate inclusion degree (about 0.62), determined jointly by the trapezoidal truth/indeterminacy/falsity profiles and the contradiction between the desired state “low” and the other risk categories.

Collecting the ingredients

$$\mathsf{TPS}_2 = \big(U,\ v,\ P_v;\ T, I, F;\ pCF\big),$$

the hospital obtains a concrete Trapezoidal Plithogenic Set that can be used to rank patients, design follow-up intervals, or allocate monitoring resources under cardiovascular risk uncertainty.

## 3.26 Nonstandard Plithogenic Sets

A nonstandard fuzzy set assigns each element a hyperreal membership near $[0,1]$, allowing infinitesimal underset/overset deviations and analysis via monads. A nonstandard neutrosophic set assigns hyperreal truth, indeterminacy, falsity degrees near $[0,1]$, permitting infinitesimal inconsistencies and refined uncertainty modeling variability [841]. A nonstandard plithogenic set maps items and attribute values to hyperreal membership vectors, aggregating evidence using contradiction-weighted t-norm/t-conorm operators internally.

**Definition 3.26.1** (Nonstandard primitives)**.** [841] Let ${}^*\mathbb{R}$ be a hyperreal field extending $\mathbb{R}$. An element $\varepsilon \in {}^*\mathbb{R}$ is *infinitesimal* if $|\varepsilon| < \frac{1}{n}$ for all $n \in \mathbb{N}$. Write the *halo* of $[0,1]$ as

$$[0,1]_{\text{ns}} \;:=\; \{\, x \in {}^*\mathbb{R} \mid \exists y \in [0,1] \text{ with } x \approx y \,\},$$

and, more liberally, for a fixed positive infinitesimal $\delta$, set the *nonstandard band*

$$[0,1]^{(\delta)} \;:=\; [-\delta,\ 1+\delta] \;\subset {}^*\mathbb{R}.$$

The *standard part* map st sends near–standard $x \in {}^*\mathbb{R}$ to the unique $y \in \mathbb{R}$ with $x \approx y$.

**Definition 3.26.2** (Nonstandard Neutrosophic Set (NSN))**.** [841] A *nonstandard neutrosophic set* on $U$ is a triple of maps

$$T_A, I_A, F_A : U \longrightarrow [0,1]^{(\delta)} \subset {}^*\mathbb{R},$$

optionally subject to the near–standard neutrosophic bound

$$T_A(x) + I_A(x) + F_A(x) \ \le\ 3 + \delta \qquad (x \in U).$$

Set operations act componentwise via an internal $t$-norm $T_*$ and $t$-conorm $S_*$:

$$\begin{aligned}(T,I,F)_{A\cup B}(x) &= \big(S_*(T_A,T_B),\ S_*(I_A,I_B),\ T_*(F_A,F_B)\big),\\ (T,I,F)_{A\cap B}(x) &= \big(T_*(T_A,T_B),\ T_*(I_A,I_B),\ S_*(F_A,F_B)\big),\\ (T,I,F)_{A^{\complement}}(x) &= \big(F_A(x),\ 1-I_A(x),\ T_A(x)\big).\end{aligned}$$

**Definition 3.26.3** (Nonstandard Plithogenic Set (NSP))**.** Fix a plithogenic context

$$PS = \big(P,\ v,\ Pv,\ pdf,\ pCF\big),$$

where $P$ is a universe, $v$ is an attribute with value set $Pv$, the *degree of appurtenance* is

$$pdf:\ P \times Pv \ \longrightarrow\ \big([0,1]^{(\delta)}\big)^s \subset ({}^*\mathbb{R})^s,$$

and the *degree of contradiction* is

$$pCF:\ Pv \times Pv \ \longrightarrow\ [0,1]^{(\delta)}\ \subset {}^*\mathbb{R} \quad \text{with} \quad pCF(a,a)=0,\ \ pCF(a,b)=pCF(b,a).$$

A *nonstandard plithogenic set* is a selection of $(P, v, Pv, pdf, pCF)$ together with fixed internal $t$-norm/$t$-conorm $(T_*, S_*)$ used to aggregate across contradictory values. For $a, b \in [0,1]^{(\delta)}$ and $c \in [0,1]^{(\delta)}$, define the DCF–weighted binary aggregator

$$a \mathbin{\widetilde{\wedge}_c} b \ :=\ (1-c)\,T_*(a,b) \ +\ c\,S_*(a,b),$$

and extend componentwise to vectors in $\big([0,1]^{(\delta)}\big)^s$. Given a finite multiset of attribute values $\{u_1,\ldots,u_m\} \subset Pv$, the *aggregated membership* of $x \in P$ is

$$\mu_{\mathrm{NSP}}(x;\{u_j\}) \ :=\ pdf(x;u_1) \mathbin{\widetilde{\wedge}_{c_{12}}} pdf(x;u_2) \mathbin{\widetilde{\wedge}_{c_{13}}} \cdots \mathbin{\widetilde{\wedge}_{c_{1m}}} pdf(x;u_m), \quad c_{1j} := pCF(u_1,u_j).$$

When only one value $u$ is considered, $\mu_{\mathrm{NSP}}(x;\{u\}) = pdf(x;u)$.

Table 3.34 presents the description of the nonstandard plithogenic set as a common generalization.

Table 3.34: Nonstandard plithogenic set as a common generalization

| Model | Membership form | Realization as NSP |
|---|---|---|
| Nonstandard Fuzzy Set | $\mu : U \to [0,1]^{(\delta)}$ | Take $Pv = \{u_0\}$, $s = 1$, define $pdf(x;u_0) := \mu(x)$, $pCF(u_0,u_0) = 0$. |
| Nonstandard Intuitionistic Fuzzy Set | $(\mu,\nu) : U \to \big([0,1]^{(\delta)}\big)^2$ | Take $Pv = \{u_0\}$, $s = 2$, set $pdf(x;u_0) := (\mu(x),\nu(x))$, $pCF(u_0,u_0) = 0$. |
| Nonstandard Neutrosophic Set [842–844] | $(T,I,F) : U \to \big([0,1]^{(\delta)}\big)^3$ | Take $Pv = \{u_0\}$, $s = 3$, set $pdf(x;u_0) := (T(x),I(x),F(x))$, $pCF(u_0,u_0) = 0$. |
| Nonstandard Quadripartitioned Neutrosophic Set | $q : U \to \big([0,1]^{(\delta)}\big)^4$ | Take $Pv = \{u_0\}$, $s = 4$, set $pdf(x;u_0) := q(x)$, $pCF(u_0,u_0) = 0$. |
| Nonstandard Peta-partitioned Neutrosophic Set | $p : U \to \big([0,1]^{(\delta)}\big)^5$ | Take $Pv = \{u_0\}$, $s = 5$, set $pdf(x;u_0) := p(x)$, $pCF(u_0,u_0) = 0$. |

A concrete example of this concept is provided below.

**Example 3.26.4** (Nonstandard plithogenic evaluation of renewable–energy projects)**.** Consider two renewable projects

$$P = \{x_{\text{wind}}, x_{\text{solar}}\},$$

and a plithogenic attribute $v =$ "sustainability criterion" with value set

$$Pv = \{\text{emissions}, \ \text{cost}\}.$$

Work in a hyperreal field ${}^*\mathbb{R}$ and fix a positive infinitesimal $\delta > 0$; write $[0,1]^{(\delta)} = [-\delta, 1+\delta]$. Set $s = 1$, so that

$$pdf : \ P \times Pv \ \longrightarrow \ [0,1]^{(\delta)}$$

returns a single (hyperreal) membership degree, and let $pCF : Pv \times Pv \to [0,1]^{(\delta)}$.

Interpretation: $pdf(x; u)$ is the (nonstandard) membership of project $x$ in the plithogenic property "sustainable" under criterion $u$; $pCF(u_1, u_2)$ measures the contradiction between two criteria.

Assume, for small infinitesimal $\varepsilon$ with $0 < |\varepsilon| \ll \delta$,

$$\begin{aligned} pdf(x_{\text{wind}}; \text{emissions}) &= 0.94 + \varepsilon, \\ pdf(x_{\text{wind}}; \text{cost}) &= 0.68 - \varepsilon, \\ pdf(x_{\text{solar}}; \text{emissions}) &= 0.90 - \varepsilon, \\ pdf(x_{\text{solar}}; \text{cost}) &= 0.72 + \varepsilon, \end{aligned}$$

so each value lies in $[0,1]^{(\delta)}$ and is "infinitesimally close" to a classical membership.

Let the contradiction degree between criteria be

$$c \ := \ pCF(\text{emissions}, \text{cost}) = 0.30 \in [0,1]^{(\delta)},$$

and choose internal $t$–norm/$t$–conorm

$$T_*(a, b) = \min(a, b), \qquad S_*(a, b) = \max(a, b)$$

(on hyperreals). The DCF–weighted aggregator of Definition of NSP is

$$a \ \widetilde{\wedge}_c \ b := (1 - c)\, T_*(a, b) + c\, S_*(a, b),$$

so for $x_{\text{wind}}$ with both criteria we obtain

$$\mu_{\text{NSP}}\big(x_{\text{wind}}; \{\text{emissions}, \text{cost}\}\big) = (0.94 + \varepsilon) \ \widetilde{\wedge}_{0.30} \ (0.68 - \varepsilon).$$

Here

$$T_* = \min(0.94 + \varepsilon, \, 0.68 - \varepsilon) = 0.68 - \varepsilon, \qquad S_* = \max(0.94 + \varepsilon, \, 0.68 - \varepsilon) = 0.94 + \varepsilon,$$

so

$$\begin{aligned} \mu_{\text{NSP}}\big(x_{\text{wind}}; \{\text{emissions}, \text{cost}\}\big) &= (1 - 0.30)(0.68 - \varepsilon) + 0.30(0.94 + \varepsilon) \\ &= 0.7 \cdot 0.68 + 0.3 \cdot 0.94 + (-0.7 + 0.3)\varepsilon \\ &= 0.758 - 0.4\,\varepsilon \ \in \ [0,1]^{(\delta)}. \end{aligned}$$

Its standard part is

$$\text{st}\big(\mu_{\text{NSP}}(x_{\text{wind}}; \{\text{emissions}, \text{cost}\})\big) = 0.758,$$

interpreted as the classical aggregate sustainability score. The infinitesimal correction $-0.4\varepsilon$ preserves additional hyperreal resolution (e.g. micro–uncertainty in expert judgments) that can be exploited by nonstandard analysis tools, while remaining *plithogenic* through the contradiction–weighted blend of criteria.

## 3.27 Refined Plithogenic Set

A Refined Plithogenic Set represents elements by multi-component truth, indeterminacy, falsity degrees with contradiction-aware weighting across attribute values and contexts. Refined Fuzzy Sets [403, 845, 846], Refined Intuitionistic Fuzzy Sets [844, 847–849], and Refined Neutrosophic Sets [119, 404, 850, 851] are also well-known in the existing literature.

**Definition 3.27.1** (Refined Plithogenic Set)**.** Let $U$ be a universe, $v$ an attribute with value-set $P_v$ (finite or countable), and let

$$p, q, r \in \mathbb{N}_{\geq 0}, \qquad s := p + q + r \geq 1.$$

Fix index sets $I_T = \{1, \dots, p\}$, $I_I = \{1, \dots, q\}$, $I_F = \{1, \dots, r\}$. A *refinement signature* is the triple $(p, q, r)$.

A *refined plithogenic appurtenance (RPA)* attached to $v$ is a map

$$\mathrm{pdf}: \ U \times P_v \longrightarrow [0,1]^s, \qquad \mathrm{pdf}(x, a) = \big(T_i(x, a)_{i \in I_T}, \ I_j(x, a)_{j \in I_I}, \ F_k(x, a)_{k \in I_F}\big),$$

where each listed component lies in $[0, 1]$. A *plithogenic contradiction function* is

$$pCF: \ P_v \times P_v \longrightarrow [0,1]^t, \qquad t \in \mathbb{N}_{\geq 0},$$

satisfying reflexivity $pCF(a, a) = 0$ and symmetry $pCF(a, b) = pCF(b, a)$.

The *Refined Plithogenic Set (RPS)* determined by $(U, v, P_v; \mathrm{pdf}, pCF)$ is the data

$$\mathsf{RPS} = \big(U, \ v, \ P_v, \ (p, q, r), \ \mathrm{pdf}, \ pCF\big).$$

Semantics: for $x \in U$ and $a \in P_v$, $T_i(x, a)$ are refined truth-memberships, $I_j(x, a)$ refined indeterminacy-memberships, and $F_k(x, a)$ refined falsity-memberships (when present), all evaluated for attribute value $a$. No further coupling is imposed at this level, except the usual range constraints $[0, 1]$; model-specific constraints (e.g. intuitionistic or neutrosophic) appear as specializations below.

**Remark 3.27.2** (Typical scalarization/aggregation (optional))**.** In decision tasks, one often fixes a dominant value $a^* \in P_v$ and an aggregation $\Phi : [0,1]^s \times [0,1]^t \to [0,1]$ that is monotone in each argument and $\Phi(\cdot, \mathbf{0})$ reduces to the underlying refined score. A *plithogenic inclusion grade* of $x$ at $a^*$ is

$$\mu_{\mathrm{pl}}(x \mid a^*) := \Phi\big(\mathrm{pdf}(x, a^*), \ pCF(a^*, \cdot)\big),$$

with unions/intersections realized via a chosen pair of t-conorm/t-norm at the aggregated level.

The overview of refined fuzzy, intuitionistic fuzzy, neutrosophic, quadripartitioned neutrosophic, and pentapartitioned neutrosophic sets is presented in Table 3.35. Write $\sup T := \sup_{i \in I_T} T_i$, etc. By choosing $(p, q, r)$ and constraints as follows, the refined fuzzy, refined intuitionistic fuzzy, and refined neutrosophic families are obtained as exact special cases (take $t = 0$ or $pCF \equiv 0$ when "no contradiction").

A concrete example of this concept is provided below.

**Example 3.27.3** (Hospital triage with refined evidence channels)**.** Consider emergency-room pneumonia triage. Let the universe be $U = \{p_A, p_B\}$ (two patients). Let the attribute be $v =$ "evidence source" with value–set $P_v = \{\mathrm{Img}, \mathrm{Lab}, \mathrm{Sym}\} =$ (chest imaging, laboratory markers, symptoms).

Refinement signature $(p, q, r) = (2, 1, 1)$ so that each refined appurtenance vector has $(T_1, T_2; I_1; F_1) \in [0, 1]^4$. For $x \in U$ and $a \in P_v$, the refined plithogenic degree $\mathrm{pdf}(x, a)$ is:

Table 3.35: Refined fuzzy, intuitionistic fuzzy, neutrosophic, quadripartitioned neutrosophic, and pentapartitioned neutrosophic sets as special cases of the Refined Plithogenic Set.

| Target refined model | Constraints and recovery inside RPS |
|---|---|
| Refined Fuzzy Set (RFS) [403, 845, 852] | Only truth components $T_i \in [0,1]$ with $pCF \equiv 0$. Any monotone aggregation of $(T_i)$ yields the usual refined fuzzy grade. |
| Refined Intuitionistic Fuzzy Set (RIFS) [847–849] | Paired components $(T_i, I_i)$ satisfying $T_i + I_i \leq 1$ for each $i$, with $pCF \equiv 0$. Classical refined intuitionistic fuzzy sets are recovered by componentwise operations on $(T_i, I_i)$. |
| Refined Neutrosophic Set (RNS) [404, 853–855] | Refined truth, indeterminacy, and falsity components $T_i, I_j, F_k \in [0,1]$ with $\sup T + \sup I + \sup F \leq 3$. Setting $pCF \equiv 0$ recovers the standard refined neutrosophic framework (single–valued case when there is exactly one $T$, $I$, and $F$). |
| Refined Quadripartitioned Neutrosophic Set (RQNS) | The families $T, I, F$ are grouped into four sub–blocks (quadripartitions), each block obeying the usual neutrosophic bounds (for example, blockwise sums $\leq 1$). With $pCF \equiv 0$, one obtains refined quadripartitioned neutrosophic sets. |
| Refined Pentapartitioned Neutrosophic Set (RPNS) | Truth, indeterminacy, and falsity components are organized into five sub–blocks (pentapartitions) satisfying the corresponding neutrosophic constraints; taking $pCF \equiv 0$ yields refined pentapartitioned neutrosophic sets as a special case. |

| | Img | Lab | Sym |
|---|---|---|---|
| $p_A$ | $(0.80, 0.75;\ 0.10;\ 0.05)$ | $(0.60, 0.55;\ 0.20;\ 0.10)$ | $(0.70, 0.65;\ 0.15;\ 0.12)$ |
| $p_B$ | $(0.45, 0.40;\ 0.25;\ 0.30)$ | $(0.50, 0.48;\ 0.22;\ 0.28)$ | $(0.60, 0.50;\ 0.25;\ 0.20)$ |

A symmetric contradiction map $pCF : P_v \times P_v \to [0,1]$ is chosen as

$$pCF = \begin{pmatrix} 0 & 0.2 & 0.3 \\ 0.2 & 0 & 0.1 \\ 0.3 & 0.1 & 0 \end{pmatrix} \quad \text{(rows/cols ordered as Img, Lab, Sym).}$$

Fix the dominant value $\delta = \text{Img}$ and use contradiction–aware weights $w(a \mid \delta) := 1 - pCF(a, \delta)$, giving

$$w(\text{Img}) = 1, \quad w(\text{Lab}) = 0.8, \quad w(\text{Sym}) = 0.7, \quad \sum_a w(a \mid \delta) = 2.5.$$

Define the $\delta$–relative refined degree of $x$ componentwise by the weighted mean

$$(T_i, I_1, F_1)^{(\delta)}(x) \;=\; \frac{\sum_{a \in P_v} w(a \mid \delta)\,(T_i, I_1, F_1)(x, a)}{\sum_{a \in P_v} w(a \mid \delta)} \quad (i = 1, 2).$$

Numerical aggregation (to three decimals):

| | $T_1^{(\delta)}$ | $T_2^{(\delta)}$ | $I_1^{(\delta)}$ | $F_1^{(\delta)}$ |
|---|---|---|---|---|
| $p_A$ | $\frac{0.80+0.48+0.49}{2.5} = 0.708$ | $\frac{0.75+0.44+0.455}{2.5} = 0.658$ | $\frac{0.10+0.16+0.105}{2.5} = 0.146$ | $\frac{0.05+0.08+0.084}{2.5} = 0.086$ |
| $p_B$ | $\frac{0.45+0.40+0.42}{2.5} = 0.508$ | $\frac{0.40+0.384+0.35}{2.5} = 0.454$ | $\frac{0.25+0.176+0.175}{2.5} = 0.240$ | $\frac{0.30+0.224+0.14}{2.5} = 0.266$ |

For a concrete scalarization, take

$$\Phi(T_1, T_2, I, F) := 0.5 \cdot \tfrac{T_1+T_2}{2} + 0.2 \cdot (1 - I) + 0.3 \cdot (1 - F).$$

Then

$$\Phi(p_A) = 0.5{\cdot}0.683{+}0.2{\cdot}0.854{+}0.3{\cdot}0.914 = 0.787, \qquad \Phi(p_B) = 0.5{\cdot}0.481{+}0.2{\cdot}0.760{+}0.3{\cdot}0.734 = 0.613.$$

Hence, under imaging–dominant context, $p_A$ has a higher refined plithogenic support.

**Example 3.27.4** (Supplier selection with ESG–dominant policy)**.** A retailer evaluates two suppliers $U = \{s_A, s_B\}$. Attribute $v =$ "criterion" with $P_v = \{\text{ESG}, \text{Qual}, \text{Deliv}\}$ (environmental–social–governance, product quality, delivery reliability). Use refinement signature $(p, q, r) = (2, 1, 1)$ and record refined appurtenances:

| | ESG | Qual | Deliv |
|---|---|---|---|
| $s_A$ | $(0.85, 0.80;\ 0.10;\ 0.05)$ | $(0.75, 0.70;\ 0.12;\ 0.08)$ | $(0.60, 0.55;\ 0.15;\ 0.12)$ |
| $s_B$ | $(0.60, 0.55;\ 0.20;\ 0.20)$ | $(0.80, 0.78;\ 0.10;\ 0.10)$ | $(0.75, 0.70;\ 0.12;\ 0.12)$ |

Contradiction matrix (symmetric):

$$pCF = \begin{pmatrix} 0 & 0.2 & 0.3 \\ 0.2 & 0 & 0.1 \\ 0.3 & 0.1 & 0 \end{pmatrix} \quad \text{(rows/cols ordered as ESG, Qual, Deliv).}$$

With dominant value $\delta = \text{ESG}$, weights are

$$w(\text{ESG}) = 1, \quad w(\text{Qual}) = 0.8, \quad w(\text{Deliv}) = 0.7, \quad \sum_a w(a \mid \delta) = 2.5.$$

Aggregate componentwise:

| | $T_1^{(\delta)}$ | $T_2^{(\delta)}$ | $I_1^{(\delta)}$ | $F_1^{(\delta)}$ |
|---|---|---|---|---|
| $s_A$ | $\frac{0.85+0.60+0.42}{2.5} = 0.748$ | $\frac{0.80+0.56+0.385}{2.5} = 0.698$ | $\frac{0.10+0.096+0.105}{2.5} = 0.120$ | $\frac{0.05+0.064+0.084}{2.5} = 0.079$ |
| $s_B$ | $\frac{0.60+0.64+0.525}{2.5} = 0.706$ | $\frac{0.55+0.624+0.490}{2.5} = 0.666$ | $\frac{0.20+0.080+0.084}{2.5} = 0.146$ | $\frac{0.20+0.080+0.084}{2.5} = 0.146$ |

Using the same scalarization $\Phi(T_1, T_2, I, F) = 0.5 \cdot \frac{T_1+T_2}{2} + 0.2(1 - I) + 0.3(1 - F)$, we obtain

$$\Phi(s_A) = 0.5{\cdot}0.723{+}0.2{\cdot}0.880{+}0.3{\cdot}0.921 = 0.814, \qquad \Phi(s_B) = 0.5{\cdot}0.686{+}0.2{\cdot}0.854{+}0.3{\cdot}0.854 = 0.770.$$

Under an ESG–dominant policy with contradiction–aware weighting, $s_A$ is preferred.

## 3.28 Subset–Valued Plithogenic Sets

A subset–valued plithogenic set maps each item and attribute value to subsets of membership vectors, aggregating contradictions via degree-of-contradiction weights.

**Definition 3.28.1** (Subset–Valued Fuzzy Set (SVFS))**.** Let $X$ be a nonempty universe and $\mathcal{P}([0, 1])$ the powerset of $[0, 1]$. A *subset–valued fuzzy set* on $X$ is a map

$$A : X \to \mathcal{P}([0, 1]) \setminus \{\varnothing\}, \qquad x \longmapsto A(x),$$

where $A(x) \subseteq [0, 1]$ is the (possibly non-singleton) set of admissible membership degrees of $x$. When each $A(x)$ is a singleton, one recovers an ordinary fuzzy set.

**Definition 3.28.2** (Subset–Valued Neutrosophic Set (SVNS))**.** [841] Let $X$ be a nonempty universe. A *subset–valued neutrosophic set* on $X$ assigns to each $x \in X$ a triple

$$\mathcal{A}(x) \;=\; \bigl(T_{\mathcal{A}}(x),\, I_{\mathcal{A}}(x),\, F_{\mathcal{A}}(x)\bigr),$$

with $T_{\mathcal{A}}(x), I_{\mathcal{A}}(x), F_{\mathcal{A}}(x) \in \mathcal{P}([0,1]) \setminus \{\varnothing\}$. Writing inf and sup for the usual bounds (with the convention on closedness as needed), one requires

$$0 \;\le\; \inf T_{\mathcal{A}}(x) + \inf I_{\mathcal{A}}(x) + \inf F_{\mathcal{A}}(x) \;\le\; \sup T_{\mathcal{A}}(x) + \sup I_{\mathcal{A}}(x) + \sup F_{\mathcal{A}}(x) \;\le\; 3 \qquad (\forall x \in X).$$

When each of $T_{\mathcal{A}}(x), I_{\mathcal{A}}(x), F_{\mathcal{A}}(x)$ is a singleton, this reduces to a single–valued neutrosophic set.

**Definition 3.28.3** (Subset-Valued Plithogenic Set (SVPS))**.** A *plithogenic context* is a tuple

$$PS = \bigl(P,\; v,\; Pv,\; pdf,\; pCF\bigr),$$

where

- $P$ is a universe of items;
- $v$ is a fixed attribute with value set $Pv$;
- $pdf : P \times Pv \to \mathcal{P}([0,1]^s) \setminus \{\varnothing\}$ (for some fixed $s \in \mathbb{N}$) is a degree–of–appurtenance mapping that assigns to each $(x,u) \in P \times Pv$ a *subset* $pdf(x;u) \subseteq [0,1]^s$ of admissible membership vectors;
- $pCF : Pv \times Pv \to [0,1]$ is the degree of contradiction, satisfying $pCF(a,a) = 0$ and $pCF(a,b) = pCF(b,a)$ for all $a,b \in Pv$.

A *subset–valued plithogenic set* over $PS$ is given by the data

$$(P,\; v,\; Pv,\; pdf,\; pCF).$$

For each item $x \in P$ and each attribute value $u \in Pv$, the subset $pdf(x;u) \subseteq [0,1]^s$ collects all admissible membership vectors of $x$ under $u$.

[3]

**Example 3.28.4** (Subset–valued plithogenic set for medical treatment side–effect risk)**.** Let the universe of items be

$$P = \{\mathrm{DrugA}, \mathrm{DrugB}\},$$

and consider a plithogenic attribute

$$v = \text{“evidence type”}, \qquad P_v = \{\mathrm{Trial}, \mathrm{Post}\},$$

where Trial = controlled clinical trials, Post = post–marketing surveillance reports.

We take $s = 2$, with components

$$\mu_1 = \text{“severe side–effect risk”}, \qquad \mu_2 = \text{“mild side–effect risk”},$$

so that each membership vector lies in $[0,1]^2$. For each $(x,u) \in P \times P_v$, the map

$$pdf : P \times P_v \longrightarrow \mathcal{P}([0,1]^2) \setminus \{\varnothing\}$$

[3]In applications, plithogenic operations (e.g. conjunction or aggregation) are often defined by combining elements of $pdf(x;u)$ via a fixed $t$–norm / $t$–conorm, weighted by the contradiction degree $pCF$. For the representation and reduction results in this section, only the subset–valued mapping $pdf$ and the function $pCF$ are needed.

Table 3.36: Subset–valued plithogenic set as a common generalization.

| Model | Image of $pdf(x; u)$ | $s$ | Specialization of SVPS |
|---|---|---|---|
| Subset-valued fuzzy set (SVFS) | $\mathcal{P}([0,1]) \setminus \{\varnothing\}$ | 1 | Single attribute value $u_0$; set $pdf(x; u_0) = A(x)$; take $pCF \equiv 0$. |
| Subset-valued intuitionistic fuzzy set (SVIFS) | $\mathcal{P}([0,1]^2) \setminus \{\varnothing\}$ | 2 | Single value $u_0$; $pdf(x; u_0)$ is a subset of IF pairs $(\mu, \nu)$ satisfying the usual intuitionistic bounds; $pCF \equiv 0$. |
| Subset-valued neutrosophic set (SVNS) [841] | $\mathcal{P}([0,1]^3) \setminus \{\varnothing\}$ | 3 | Single value $u_0$; $pdf(x; u_0)$ is a subset of neutrosophic triples $(T, I, F)$ obeying neutrosophic constraints; $pCF \equiv 0$. |
| Subset-valued plithogenic set (SVPS) | $\mathcal{P}([0,1]^s) \setminus \{\varnothing\}$ | $s \geq 1$ | General case: arbitrary $s$, multiple attribute values in $P_v$, and nontrivial contradiction function $pCF : P_v \times P_v \to [0,1]$. |

assigns a *subset* of possible pairs $(\mu_1, \mu_2)$ obtained from different studies or expert groups.

A possible specification is:

$$\begin{aligned} pdf(\text{DrugA}; \text{Trial}) &= \{(0.10, 0.30),\ (0.15, 0.35)\}, \\ pdf(\text{DrugA}; \text{Post}) &= \{(0.20, 0.40),\ (0.25, 0.50)\}, \\ pdf(\text{DrugB}; \text{Trial}) &= \{(0.05, 0.20),\ (0.08, 0.25)\}, \\ pdf(\text{DrugB}; \text{Post}) &= \{(0.12, 0.28),\ (0.18, 0.35)\}. \end{aligned}$$

The plithogenic degree of contradiction between evidence types is modeled by

$$pCF : P_v \times P_v \to [0,1], \qquad pCF(u,u) = 0, \quad pCF(u_1, u_2) = pCF(u_2, u_1),$$

for example

$$pCF(\text{Trial}, \text{Post}) = 0.4, \quad pCF(\text{Trial}, \text{Trial}) = pCF(\text{Post}, \text{Post}) = 0.$$

Interpretation. For each drug $x$ and evidence type $u$, the subset $pdf(x; u)$ collects several plausible risk profiles $(\mu_1, \mu_2)$ coming from different datasets or statistical models, rather than a single fixed pair. The plithogenic mechanism then aggregates these subset–valued degrees across Trial and Post using the contradiction weight $pCF(\text{Trial}, \text{Post})$: high disagreement between evidence sources (large $pCF$) yields more cautious or conservative combined risk assessments. Thus $(P, v, P_v, pdf, pCF)$ forms a subset–valued plithogenic set describing real–world medical side–effect uncertainty.

## 3.29 Picture Plithogenic Set

A Picture Plithogenic Set models neutral and other degrees with attribute-based contradiction–weighted aggregation, capturing complex opinions in uncertain environments.

**Definition 3.29.1** (Picture Fuzzy Set (PFS))**.** [7] Let $U$ be a nonempty universe. A *picture fuzzy set* $A$ on $U$ is a family

$$A = \Big\{ \langle x, \mu_A(x), \eta_A(x), \nu_A(x) \rangle \ : \ x \in U \Big\},$$

where, for every $x \in U$,

$$\mu_A(x),\ \eta_A(x),\ \nu_A(x) \in [0,1]$$

denote respectively the *positive*, *neutral*, and *negative* degrees of $x$ with respect to $A$, subject to

$$0 \ \leq \ \mu_A(x) + \eta_A(x) + \nu_A(x) \ \leq \ 1.$$

The remaining quantity

$$\pi_A(x) \ := \ 1 - \mu_A(x) - \eta_A(x) - \nu_A(x)$$

is called the *refusal degree* of $x$ in $A$.

**Definition 3.29.2** (Picture Neutrosophic Set (PNS))**.** [445] Let $U$ be a nonempty universe. A *picture neutrosophic set* $B$ on $U$ is a family

$$B = \Big\{ \langle x, T_B(x), I_B(x), F_B(x) \rangle \ : \ x \in U \Big\},$$

where, for each $x \in U$,

$$T_B(x),\ I_B(x),\ F_B(x) \in [0,1]$$

denote respectively the *truth*, *indeterminacy*, and *falsity* degrees of $x$ with respect to $B$, and

$$0 \ \leq \ T_B(x) + I_B(x) + F_B(x) \ \leq \ 3.$$

The term "picture" emphasizes that the triple $(T_B(x), I_B(x), F_B(x))$ gives a three–fold view (positive / neutral / negative) of the evaluation, while mathematically $B$ is a single–valued neutrosophic set.

**Definition 3.29.3** (Picture Plithogenic Set (PPlS))**.** Let $P$ be a nonempty universe, $v$ an attribute, and $P_v$ a nonempty set of attribute values of $v$. Fix integers $t \geq 1$ and $r \geq 0$. A *picture plithogenic set* on $(P, v, P_v)$ is a tuple

$$PS_{\text{pic}} \ = \ (P,\ v,\ P_v,\ pdf,\ pCF),$$

where:

- the *picture plithogenic degree of appurtenance function*

  $$pdf:\ P \times P_v \longrightarrow [0,1]^{t+1}$$

  is written as

  $$pdf(x,a) \ = \ \big(\alpha_1(x,a), \dots, \alpha_t(x,a), \eta(x,a)\big),$$

  with each component in $[0,1]$; here

  $$\alpha(x,a) := \big(\alpha_1(x,a), \dots, \alpha_t(x,a)\big)$$

  is the *underlying plithogenic membership vector*, and

  $$\eta(x,a) := \alpha_{t+1}(x,a)$$

  is the additional *picture–neutral degree* of $x$ with respect to $v = a$;
- the *degree of contradiction function* (DCF)

  $$pCF:\ P_v \times P_v \longrightarrow [0,1]^r$$

  satisfies, for all $a, b \in P_v$,

  $$pCF(a,a) = \mathbf{0} \in [0,1]^r, \qquad pCF(a,b) = pCF(b,a).$$

Table 3.37: Reductions of picture fuzzy, picture neutrosophic, and plithogenic sets to the picture plithogenic framework.

| Source model | Representation as a picture plithogenic set |
|---|---|
| Picture fuzzy set on $P$ [7, 365] | Choose $t = 2$, $r = 0$, $P_v = \{a^*\}$ and set $pdf(x, a^*) = (\mu(x), \nu(x), \eta(x))$ (positive, negative, neutral). |
| Picture neutrosophic set on $P$ [445] | Choose $t = 3$, $r = 0$, $P_v = \{b^*\}$ and set $pdf(x, b^*) = (T(x), I(x), F(x), 0)$; the first three coordinates give the neutrosophic triple and the neutral channel is fixed to 0. |
| Plithogenic set on $P$ | Given $pdf_0 : P \times P_v \to [0,1]^t$, define $pdf(x, a) = (pdf_0(x, a), 0)$; the extra neutral coordinate is always 0. |

Thus a picture plithogenic set is a plithogenic set whose membership vectors have $t$ "ordinary" components together with one extra *neutral* component, stored in the $(t + 1)$–st coordinate.

Table 3.37 presents a concise overview of how picture fuzzy sets, picture neutrosophic sets, and plithogenic sets can each be represented within the unified picture plithogenic framework.

**Theorem 3.29.4** (Picture Plithogenic Set as a common generalization)**.** *The class of picture plithogenic sets strictly generalizes the classes of picture fuzzy sets, picture neutrosophic sets, and plithogenic sets. More precisely:*

1. *Every picture fuzzy set on $P$ can be represented as a picture plithogenic set with suitable $(t, r)$, $P_v$, $pdf$, and $pCF$.*
2. *Every picture neutrosophic set on $P$ can be represented as a picture plithogenic set with suitable $(t, r)$, $P_v$, $pdf$, and $pCF$.*
3. *Every plithogenic set on $P$ can be embedded into a picture plithogenic set by appending a neutral coordinate equal to zero.*

*Proof.* (1) Recovery of picture fuzzy sets.

Let $A$ be a picture fuzzy set on $P$:

$$A = \Big\{ \langle x, \mu_A(x), \eta_A(x), \nu_A(x) \rangle : x \in P \Big\},$$

with

$$\mu_A(x), \eta_A(x), \nu_A(x) \in [0, 1], \qquad 0 \le \mu_A(x) + \eta_A(x) + \nu_A(x) \le 1.$$

Choose

$$t := 2, \qquad r := 0, \qquad P_v := \{a^*\}$$

and define $pCF(a^*, a^*) := 0$. Define $pdf : P \times P_v \to [0, 1]^{t+1} = [0, 1]^3$ by

$$pdf(x, a^*) := \big(\alpha_1(x, a^*), \alpha_2(x, a^*), \eta(x, a^*)\big) := \big(\mu_A(x), \nu_A(x), \eta_A(x)\big).$$

Then for each $x \in P$, we have

$$\alpha_1(x, a^*) = \mu_A(x), \quad \alpha_2(x, a^*) = \nu_A(x), \quad \eta(x, a^*) = \eta_A(x),$$

and the normalization

$$0 \le \alpha_1(x,a^*) + \eta(x,a^*) + \alpha_2(x,a^*) \le 1$$

is exactly the constraint of a picture fuzzy triple $(\mu_A(x), \eta_A(x), \nu_A(x))$. Thus

$$PS^A_{\text{pic}} := (P,\ v,\ P_v,\ pdf,\ pCF)$$

is a picture plithogenic set whose coordinates recover $A$.

Conversely, suppose we are given a picture plithogenic set $(P, v, P_v, pdf, pCF)$ with

$$t = 2, \quad r = 0, \quad P_v = \{a^*\},$$

and such that, for each $x \in P$,

$$pdf(x,a^*) = \big(\alpha_1(x,a^*), \alpha_2(x,a^*), \eta(x,a^*)\big) \in [0,1]^3$$

satisfies

$$0 \le \alpha_1(x,a^*) + \eta(x,a^*) + \alpha_2(x,a^*) \le 1.$$

Define

$$\mu_A(x) := \alpha_1(x,a^*), \quad \nu_A(x) := \alpha_2(x,a^*), \quad \eta_A(x) := \eta(x,a^*).$$

Then

$$0 \le \mu_A(x) + \eta_A(x) + \nu_A(x) \le 1,$$

so

$$A := \Big\{ \langle x, \mu_A(x), \eta_A(x), \nu_A(x)\rangle : x \in P \Big\}$$

is a picture fuzzy set. Hence picture fuzzy sets are exactly those picture plithogenic sets with $t = 2$, $r = 0$, a singleton $P_v$, and the constraint $\alpha_1 + \eta + \alpha_2 \le 1$; this proves (1).

(2) Recovery of picture neutrosophic sets.

Let $B$ be a picture neutrosophic set on $P$:

$$B = \Big\{ \langle x, T_B(x), I_B(x), F_B(x)\rangle : x \in P \Big\},$$

with

$$T_B(x), I_B(x), F_B(x) \in [0,1], \qquad 0 \le T_B(x) + I_B(x) + F_B(x) \le 3.$$

Choose

$$t := 3, \qquad r := 0, \qquad P_v := \{b^*\},$$

and $pCF(b^*, b^*) := 0$. Define $pdf : P \times P_v \to [0,1]^{t+1} = [0,1]^4$ by

$$pdf(x,b^*) := \big(\alpha_1(x,b^*), \alpha_2(x,b^*), \alpha_3(x,b^*), \eta(x,b^*)\big) := \big(T_B(x), I_B(x), F_B(x), 0\big).$$

Then, for each $x \in P$,

$$\alpha_1(x,b^*) = T_B(x), \quad \alpha_2(x,b^*) = I_B(x), \quad \alpha_3(x,b^*) = F_B(x), \quad \eta(x,b^*) = 0.$$

Thus the first $t = 3$ components of $pdf(x,b^*)$ reproduce the neutrosophic triple $(T_B, I_B, F_B)$, and the extra neutral coordinate is identically 0.

Hence

$$PS^B_{\text{pic}} := (P,\ v,\ P_v,\ pdf,\ pCF)$$

is a picture plithogenic set that encodes $B$.

Conversely, suppose we are given a picture plithogenic set with

$$t = 3, \quad r = 0, \quad P_v = \{b^*\},$$

and such that

$$pdf(x, b^*) = \big(\alpha_1(x, b^*), \alpha_2(x, b^*), \alpha_3(x, b^*), \eta(x, b^*)\big) \in [0, 1]^4,$$

with the additional constraint

$$0 \le \alpha_1(x, b^*) + \alpha_2(x, b^*) + \alpha_3(x, b^*) \le 3,$$

and where, for example, $\eta(x, b^*) = 0$ for all $x$. Define

$$T_B(x) := \alpha_1(x, b^*), \quad I_B(x) := \alpha_2(x, b^*), \quad F_B(x) := \alpha_3(x, b^*).$$

Then $B$ as above is a picture neutrosophic set. Therefore picture neutrosophic sets correspond to picture plithogenic sets with $t = 3$, $r = 0$, singleton $P_v$, and neutral coordinate identically zero (or otherwise separated from the neutrosophic triple); this proves (2).

(3) Embedding plithogenic sets.

Let

$$PS = (P,\ v,\ P_v,\ pdf_0,\ pCF)$$

be any plithogenic set with

$$pdf_0 :\ P \times P_v \to [0, 1]^t, \qquad pdf_0(x, a) = \big(\alpha_1(x, a), \dots, \alpha_t(x, a)\big),$$

and

$$pCF :\ P_v \times P_v \to [0, 1]^r$$

reflexive and symmetric.

Define

$$pdf :\ P \times P_v \to [0, 1]^{t+1}$$

by appending a neutral coordinate equal to zero:

$$pdf(x, a) := \big(\alpha_1(x, a), \dots, \alpha_t(x, a), \eta(x, a)\big), \qquad \eta(x, a) := 0.$$

Then $(P, v, P_v, pdf, pCF)$ is a picture plithogenic set with the same attribute system and contradiction structure as $PS$, and with the underlying plithogenic part unchanged:

$$\alpha_i(x, a) \text{ in } PS \quad \longleftrightarrow \quad \alpha_i(x, a) \text{ in } PS_{\text{pic}} \text{ for } i = 1, \dots, t.$$

Conversely, any picture plithogenic set $(P, v, P_v, pdf, pCF)$ for which the neutral coordinate satisfies

$$\eta(x, a) \equiv 0 \quad \text{for all } (x, a) \in P \times P_v$$

determines a plithogenic set by simply discarding the last coordinate:

$$pdf_0(x, a) := \big(\alpha_1(x, a), \dots, \alpha_t(x, a)\big).$$

Thus plithogenic sets are exactly those picture plithogenic sets whose neutral coordinate is identically zero, proving (3).

Combining (1), (2), and (3), we conclude that picture fuzzy sets, picture neutrosophic sets, and plithogenic sets appear as particular, well–defined subclasses of picture plithogenic sets obtained by suitable choices of the parameters $(t, r)$, the value set $P_v$, the contradiction function $pCF$, and algebraic constraints on the coordinates of $pdf$. Therefore the picture plithogenic framework strictly generalizes all three structures. □

A concrete example of this concept is provided below.

**Example 3.29.5** (Public transport policy evaluation)**.** Consider a city that evaluates two candidate public transport policies

$$P = \{p_1 = \text{“new bus lanes”},\ p_2 = \text{“fare discount”}\}.$$

Let $v$ be the attribute *citizen attitude*, with value set

$$P_v = \{a^+ = \text{support},\ a^- = \text{oppose}\}.$$

We take $t = 2$, $r = 1$ and interpret, for each $(x, a) \in P \times P_v$,

$$pdf(x, a) = \big(\alpha_1(x, a), \alpha_2(x, a), \eta(x, a)\big),$$

where $\alpha_1$ is the positive degree, $\alpha_2$ the negative degree, and $\eta$ a picture–neutral degree (“undecided / mixed”).

For instance, based on a survey we may set

$$\begin{aligned} pdf(p_1, a^+) &= (0.7, 0.1, 0.1), \quad pdf(p_1, a^-) = (0.2, 0.6, 0.1), \\ pdf(p_2, a^+) &= (0.6, 0.2, 0.1), \quad pdf(p_2, a^-) = (0.1, 0.5, 0.2). \end{aligned}$$

To encode plithogenic contradiction between *support* and *oppose* we define

$$pCF(a^+, a^+) = pCF(a^-, a^-) = 0, \qquad pCF(a^+, a^-) = pCF(a^-, a^+) = 0.9.$$

Thus the picture plithogenic set

$$PS^{\text{policy}}_{\text{pic}} = \big(P,\ v,\ P_v,\ pdf,\ pCF\big)$$

models, in a single framework, positive / negative / neutral citizen opinions together with their strong contradiction between “support” and “oppose”.

**Example 3.29.6** (Online product recommendation)**.** Let

$$P = \{x_1, x_2, x_3\}$$

be three candidate smartphones on an e–commerce platform. Let $v$ be the attribute *customer sentiment category* with

$$P_v = \{a_1 = \text{battery},\ a_2 = \text{camera}\}.$$

We again choose $t = 2$, $r = 1$ and write

$$pdf(x, a) = (\alpha_1(x, a), \alpha_2(x, a), \eta(x, a)),$$

where $\alpha_1$ is the positive sentiment degree, $\alpha_2$ the negative sentiment degree, and $\eta$ the neutral (“mixed / no clear opinion”) degree for product $x$ on aspect $a$.

Example membership assessments:

$$\begin{aligned} pdf(x_1, a_1) &= (0.8, 0.1, 0.05), \quad pdf(x_1, a_2) = (0.4, 0.3, 0.2), \\ pdf(x_2, a_1) &= (0.5, 0.3, 0.1), \quad\ \ pdf(x_2, a_2) = (0.7, 0.1, 0.1), \\ pdf(x_3, a_1) &= (0.3, 0.5, 0.1), \quad\ \ pdf(x_3, a_2) = (0.4, 0.4, 0.1). \end{aligned}$$

To model that "battery" and "camera" are only mildly contradictory as aspects, we define

$$pCF(a_1, a_1) = pCF(a_2, a_2) = 0, \qquad pCF(a_1, a_2) = pCF(a_2, a_1) = 0.3.$$

Then the picture plithogenic set

$$PS_{\text{pic}}^{\text{phone}} = \big(P,\ v,\ P_v,\ pdf,\ pCF\big)$$

captures, for each phone and each aspect, the positive / negative / neutral picture of customer opinions together with a contradiction–aware interaction between the aspects "battery" and "camera" when aggregating overall recommendations.

# Chapter 4

# Unifying Framework of Fuzzy, Intuitionistic, Neutrosophic, Plithogenic, and Other Set

In this chapter, we present a unifying framework for Fuzzy, Intuitionistic, Neutrosophic, Plithogenic, and other uncertainty-handling sets.

## 4.1 Uncertain Set

An Uncertain Set is any set-theoretic model assigning graded, possibly multi-component membership degrees to elements, generalizing fuzzy, intuitionistic, neutrosophic, plithogenic and related uncertainty frameworks unified [856].

**Definition 4.1.1** (Uncertain Set)**.** [856] Let $\mathbb{U}$ be the collection of all Uncertain Models as in Definition **??**. Fix

$$U \in \mathbb{U}, \quad \mathrm{Dom}(U) \subseteq [0,1]^r$$

for some integer $r \geq 1$, and let $X$ be a nonempty base set (universe of discourse).

An *Uncertain Set of type* $U$ *on* $X$ is a pair

$$\mathsf{A}^U = (X, \mu),$$

where

$$\mu \colon X \longrightarrow \mathrm{Dom}(U)$$

assigns to each element $x \in X$ a $U$–membership degree

$$\mu(x) \in \mathrm{Dom}(U).$$

Equivalently, once the base set $X$ and the Uncertain Model $U$ are fixed, we may identify the Uncertain Set with its membership function and simply write

$$\mathsf{A}^U : X \longrightarrow \mathrm{Dom}(U), \quad x \longmapsto \mu(x),$$

and view the collection of all Uncertain Sets of type $U$ on $X$ as the function space

$$\big(\mathrm{Dom}(U)\big)^X = \{\, \mu \mid \mu : X \to \mathrm{Dom}(U) \,\}.$$

In this sense, an Uncertain Set is a $U$–labeling of the base set $X$ by membership–degree tuples taken from $\mathrm{Dom}(U)$.

**Remark 4.1.2** (Recovery of classical fuzzy–type sets)**.** Let $X$ be a nonempty set and let $\mathsf{A}^U = (X, \mu)$ be an Uncertain Set of type $U$.

1. (*Fuzzy Set*) Take $U$ = Fuzzy with
$$\mathrm{Dom}(U) = [0,1] = [0,1]^1.$$
Then an Uncertain Set of type $U$ is exactly a fuzzy set in the sense of Zadeh, since
$$\mu : X \to [0,1]$$
is the usual fuzzy membership function.
2. (*Intuitionistic Fuzzy Set*) Take $U$ = Intuitionistic Fuzzy with
$$\mathrm{Dom}(U) = \{(\mu, \nu) \in [0,1]^2 \mid \mu + \nu \leq 1\} \subseteq [0,1]^2.$$
Then $\mathsf{A}^U = (X, \mu)$ coincides with an intuitionistic fuzzy set, because for each $x \in X$,
$$\mu(x) = (\mu_A(x), \nu_A(x)) \in [0,1]^2$$
satisfies $\mu_A(x) + \nu_A(x) \leq 1$.
3. (*Neutrosophic Set*) Take $U$ = Neutrosophic with
$$\mathrm{Dom}(U) = \{(T, I, F) \in [0,1]^3 \mid 0 \leq T + I + F \leq 3\} \subseteq [0,1]^3.$$
Then $\mathsf{A}^U = (X, \mu)$ is exactly a single–valued neutrosophic set, since
$$\mu(x) = (T_A(x), I_A(x), F_A(x)) \in [0,1]^3$$
with $0 \leq T_A(x) + I_A(x) + F_A(x) \leq 3$ for all $x \in X$.
4. (*Plithogenic Set*) For a Plithogenic Model $U$ = Plithogenic with degree–domain
$$\mathrm{Dom}(U) = \Big\{ (v, \mathrm{pdf}(x,v), \mathrm{pCF}(v_1,v_2)) \Bigm| v \in P_v,\ \mathrm{pdf}(x,v) \in [0,1]^s,\ \mathrm{pCF}(v_1,v_2) \in [0,1]^t \Big\} \subseteq [0,1]^{s+t+\ell},$$
an Uncertain Set of type $U$ on $X$ reproduces a Plithogenic Set on $X$, since each
$$\mu(x) \in \mathrm{Dom}(U)$$
encodes the Plithogenic degrees associated with $x \in X$.

Thus, by choosing different Uncertain Models $U \in \mathbb{U}$ and their corresponding domains $\mathrm{Dom}(U) \subseteq [0,1]^r$, the general notion of an Uncertain Set in Definition 4.1.1 unifies fuzzy sets, intuitionistic fuzzy sets, neutrosophic sets, plithogenic sets, and many other existing uncertainty–set frameworks.

## 4.2 Functional Set

A functorial set assigns to each object of a category a corresponding set and transports its elements along morphisms via the functor [856]. These constructions not only encompass fuzzy, intuitionistic, neutrosophic, plithogenic, and related uncertainty frameworks, but also generalize set concepts that possess features beyond uncertainty.

**Definition 4.2.1** (Functorial Set)**.** [856] Let $\mathcal{C}$ be a category and

$$F\colon \mathcal{C} \longrightarrow \mathbf{Set}$$

be a (covariant) endofunctor. For any object $X \in \mathrm{Ob}(\mathcal{C})$, an *$F$-set over $X$* is an element

$$s \in F(X).$$

We denote the collection of all $F$-sets over $X$ simply by $F(X)$. A morphism $f\colon X \to Y$ in $\mathcal{C}$ induces a *pushforward*

$$F(f)\colon F(X) \longrightarrow F(Y), \qquad s \mapsto F(f)(s).$$

A concrete example of this concept is provided below.

**Example 4.2.2** (City logistics: purchase orders functor)**.** Let $\mathcal{C}$ be the (small) category generated by the cities $T =$ Tokyo, $O =$ Osaka, $F =$ Fukuoka and route–morphisms

$$f : T \to O, \qquad g : O \to F, \qquad g \circ f : T \to F,$$

together with identities $\mathrm{id}_T, \mathrm{id}_O, \mathrm{id}_F$. Define a functor $F : \mathcal{C} \to \mathbf{Set}$ by:

$$F(T) = \{\text{PO } T{-}101 : (50),\ \ T{-}102 : (30)\},$$

$$F(O) = \{\text{PO } O{-}201 : (40)\}, \quad F(F) = \varnothing,$$

where "PO $X - \cdot\, : (q)$" denotes a purchase order for destination $X$ with quantity $q$. For a route $h : X \to Y$, define $F(h) : F(X) \to F(Y)$ by *retagging the destination* while preserving the item and quantity:

$$F(h) : \text{PO } X{-}n : (q) \longmapsto \text{PO } Y{-}n : (q).$$

Then functoriality holds:

$$F(\mathrm{id}_X) = \mathrm{id}_{F(X)} \quad \text{(trivial retagging)}, \qquad F(g \circ f) = F(g) \circ F(f) \ \ \text{(same two-step retagging)}.$$

Concrete check on PO $T{-}101 : (50)$:

$$F(f) :\ \text{PO } T{-}101 : (50) \mapsto \text{PO } O{-}101 : (50), \qquad F(g) :\ \text{PO } O{-}101 : (50) \mapsto \text{PO } F{-}101 : (50),$$

so $(F(g) \circ F(f))(\text{PO } T{-}101 : (50)) = \text{PO } F{-}101 : (50)$, which equals $F(g \circ f)(\text{PO } T{-}101 : (50))$ by the definition of $g \circ f : T \to F$. Thus $F$ models *functorial transport* of orders along the logistics network.

**Example 4.2.3** (Cross–timezone calendars: meeting slots functor)**.** Let $\mathcal{T}$ be the category whose objects are fixed UTC offsets $\{0, +9, -5\}$ (UTC, JST, EST), and whose morphisms are offset–changes given by

$$h_{\alpha\to\beta} : \alpha \to \beta,$$

$$h_{\beta\to\gamma} \circ h_{\alpha\to\beta} = h_{\alpha\to\gamma},$$

$$h_{\alpha\to\alpha} = \mathrm{id}_\alpha.$$

Define $G : \mathcal{T} \to \mathbf{Set}$ by letting $G(\alpha)$ be the set of *local* meeting timestamps at offset $\alpha$ (as naive datetimes). For $h_{\alpha\to\beta} : \alpha \to \beta$, define $G(h_{\alpha\to\beta})$ to *convert* local times:

$$G(h_{\alpha\to\beta})(\text{YYYY-MM-DD hh:mm at } \alpha) \ := \ \text{YYYY-MM-DD hh:mm} + (\beta - \alpha) \text{ hours, now at } \beta.$$

Functoriality is immediate:

$$G(\mathrm{id}_\alpha) = \mathrm{id}_{G(\alpha)},$$

$$G(h_{\beta\to\gamma} \circ h_{\alpha\to\beta}) = G(h_{\beta\to\gamma}) \circ G(h_{\alpha\to\beta})$$

because time–shift addition is associative. Numerical check:

$$\text{Slot } s = \text{2025-11-08 10:00 at UTC} \in G(0).$$

Then

$$G(h_{0\to+9})(s) = \text{2025-11-08 19:00 at JST}, \quad G(h_{+9\to-5}) = \text{shift by } -14\text{ h},$$

so

$$\big(G(h_{+9\to-5}) \circ G(h_{0\to+9})\big)(s) = \text{2025-11-08 05:00 at EST}.$$

Directly,

$$G(h_{0\to-5})(s) = \text{2025-11-08 05:00 at EST},$$

hence $G(h_{+9\to-5} \circ h_{0\to+9}) = G(h_{+9\to-5}) \circ G(h_{0\to+9})$ on $s$. Thus $G$ functorially transports meeting slots across timezones.

**Example 4.2.4** (Geospatial containment: points–of–interest functor)**.** Let $(\mathcal{R}, \subseteq)$ be a poset of regions with inclusions as morphisms, forming a category: objects are regions $X$, morphisms are $i_{X\subseteq Y} : X \to Y$ when $X \subseteq Y$. Consider the chain

$$\text{Ward} = \text{Shinjuku} \ \subseteq \ \text{City} = \text{Tokyo} \ \subseteq \ \text{Country} = \text{Japan}.$$

Define $H : \mathcal{R} \to \textbf{Set}$ by

$$H(X) = \{\text{POIs located in } X\}.$$

For $i_{X\subseteq Y} : X \to Y$, define $H(i_{X\subseteq Y}) : H(X) \to H(Y)$ as the *inclusion on elements* (same POI, now regarded inside the larger region). Concrete data:

$$H(\text{Shinjuku}) = \{\text{POI1: park, POI2: museum, POI3: station}\},$$

$$H(\text{Tokyo}) = \{\text{POI1}, \text{POI2}, \text{POI3}, \text{POI4: gallery}\}.$$

Then

$$H(i_{\text{Shinjuku}\subseteq\text{Tokyo}})(\text{POI2}) = \text{POI2} \in H(\text{Tokyo}).$$

Functoriality:

$$H(\text{id}_X) = \text{id}_{H(X)} \quad \text{(identity inclusion)}, \qquad H(i_{X\subseteq Z}) = H(i_{Y\subseteq Z}) \circ H(i_{X\subseteq Y})$$

whenever $X \subseteq Y \subseteq Z$. Numerically,

$$H(i_{\text{Shinjuku}\subseteq\text{Japan}})(\text{POI3}) = \text{POI3} = \Big(H(i_{\text{Tokyo}\subseteq\text{Japan}}) \circ H(i_{\text{Shinjuku}\subseteq\text{Tokyo}})\Big)(\text{POI3}).$$

Thus $H$ functorially embeds POIs along region inclusions, preserving composition.

## 4.3 Other Uncertain Sets and Future Works

This section presents additional uncertainty–oriented frameworks that are frequently studied alongside Fuzzy Sets, Intuitionistic Fuzzy Sets, Neutrosophic Sets, and Plithogenic Sets. Looking ahead, it is expected that further extensions and new theoretical developments will emerge by combining these auxiliary concepts with the Fuzzy, Intuitionistic Fuzzy, Neutrosophic, and Plithogenic models examined in this paper. Note that these can also be generalized within the Functorial Set framework.

- Near Set [857–859]: A near set groups objects that are perceptually or descriptively similar under tolerance relations, supporting approximate classification, retrieval, and matching tasks. Related concepts such as Fuzzy Near Sets [860–863] are also known in the literature.
- Weighted Set [864, 865]: A weighted set equips each element with a positive importance coefficient, enabling prioritized aggregation, ranking, or decision analysis over items. Related concepts such as Weighted Fuzzy Sets [716, 866, 867], Weighted Intuitionistic Fuzzy Sets [868, 869], Weighted Rough Sets [870–872], and Weighted Neutrosophic Sets [873, 874] are also known in the literature.

- Z-Number [875–877]: A Z-number describes uncertain information by pairing a fuzzy restriction with a reliability measure, capturing both value and confidence simultaneously. Related concepts such as Intuitionistic Fuzzy Z-Numbers [878–881] and Neutrosophic Z-Numbers [882–886] are also known in the literature.
- D-Number [887–889]: A D-number generalizes Dempster–Shafer evidence by allowing incomplete or non-exclusive masses, improving knowledge fusion under open or dynamic frames situations [890]. Neutrosophic D-Numbers are also known as a related concept in the literature.
- Multiple Sets [891–893]: A Multiple Set assigns several membership grades, arranged in an $n \times k$ matrix, to each element for modeling vagueness and multiplicity. Neutrosophic Multiple Sets are also known as a related concept [12, 894].

# Disclaimer

## Funding

This study was conducted without any financial support from external organizations or grants.

## Acknowledgments

We would like to express our sincere gratitude to everyone who provided valuable insights, support, and encouragement throughout this research. We also extend our thanks to the readers for their interest and to the authors of the referenced works, whose scholarly contributions have greatly influenced this study. Lastly, we are deeply grateful to the publishers and reviewers who facilitated the dissemination of this work.

## Data Availability

Since this research is purely theoretical and mathematical, no empirical data or computational analysis was utilized. Researchers are encouraged to expand upon these findings with data-oriented or experimental approaches in future studies.

## Ethical Statement

As this study does not involve experiments with human participants or animals, no ethical approval was required.

## Conflicts of Interest

The authors declare that they have no conflicts of interest related to the content or publication of this paper.

## Code Availability

No code or software was developed for this study.

## Clinical Trial

This study did not involve any clinical trials.

## Consent to Participate

Not applicable.

## Use of Generative AI and AI-Assisted Tools

I use generative AI and AI-assisted tools for tasks such as English grammar checking, and I do not employ them in any way that violates ethical standards.

## Disclaimer (Others)

This work presents theoretical ideas and frameworks that have not yet been empirically validated. Readers are encouraged to explore practical applications and further refine these concepts. Although care has been taken to ensure accuracy and appropriate citations, any errors or oversights are unintentional. The perspectives and interpretations expressed herein are solely those of the authors and do not necessarily reflect the viewpoints of their affiliated institutions.

# Appendix (List of Tables)

*

This dynamic survey comprehensively examines numerous generalized set-theoretic frameworks designed to model and capture real-world uncertainty, which frequently involves vagueness, partial truth, and incomplete information. The work traces the evolution of these mathematical concepts, starting with classical models like Fuzzy Sets and Intuitionistic Fuzzy Sets. The survey then extends to more sophisticated and recent models, including Neutrosophic Sets, Plithogenic Sets, and Extensional Sets. Additionally, it covers related concepts such as Vague Sets, Hesitant Fuzzy Sets, Picture Fuzzy Sets, and Quadripartitioned Neutrosophic Sets. The aim is to provide a unified and up-to-date analysis of these diverse theories, highlighting their foundational principles and their capacity for rigorous mathematical modeling of complex uncertainty in various scientific domains.